\documentclass[lettersize,journal]{IEEEtran}

\usepackage{standalone}

\usepackage{tikz} 
\usetikzlibrary{arrows.meta, positioning} 

\usepackage{amsmath,amsfonts} 
\usepackage{algorithmic}      
\usepackage{algorithm}        

\usepackage{amsmath}  
\usepackage{amssymb}
\usepackage{amsthm}

\usepackage[table]{xcolor}

\definecolor{stepgray}{gray}{0.25}
\definecolor{bgblue}{RGB}{245,248,255}
\definecolor{metagreen}{RGB}{0,130,0}

\usepackage{array}            
\usepackage{textcomp}         
\usepackage{stfloats}         
\usepackage{url}              
\usepackage{verbatim}         
\usepackage{graphicx}         
\usepackage{booktabs}         
\usepackage{subcaption}       
\usepackage{cite}             
\usepackage{dsfont}           
\usepackage{bm}               
\usepackage{multirow}         

\usepackage{siunitx}          
\usepackage[pagebackref=true,breaklinks=true,colorlinks,bookmarks=false,citecolor=blue]{hyperref}     
\usepackage{listings}         
\usepackage{multirow}         

\usepackage{mathtools}  

\usepackage{amsmath,booktabs,pifont,makecell}
\newcommand{\cmark}{\ding{51}}  
\newcommand{\xmark}{\ding{55}}

\newtheorem{theorem}{Theorem}[section]
\newtheorem{proposition}{Proposition}[section]

\newtheorem{assumption}{Assumption}[section]

\newcommand{\PARFOR}[1]{\STATE \textbf{parfor} #1 \textbf{do}\begin{ALC@g}}
	\newcommand{\ENDPARFOR}{\end{ALC@g}\STATE \textbf{end parfor}}

\definecolor{mygreen}{rgb}{0,0.6,0}  
\definecolor{mygray}{rgb}{0.5,0.5,0.5}
\definecolor{mymauve}{rgb}{0.58,0,0.82}
\definecolor{myblue}{rgb}{0,0,1}

\lstdefinestyle{mystyle}{
	backgroundcolor=\color{white},      
	commentstyle=\color{mygreen},       
	keywordstyle=\color{myblue},        
	identifierstyle=\color{black},      
	numberstyle=\tiny\color{mygray},    
	stringstyle=\color{mymauve},        
	basicstyle=\ttfamily\footnotesize,  
	breakatwhitespace=false,            
	breaklines=true,                    
	captionpos=b,                       
	keepspaces=true,                    
	numbers=left,                       
	numbersep=2pt,                      
	showspaces=false,                   
	showstringspaces=false,             
	showtabs=false,                     
	tabsize=2,                          
	language=Python,                    
	literate={all}{{\textcolor{black}{all}}}3 
	{any}{{\textcolor{black}{any}}}3
	{sum}{{\textcolor{black}{sum}}}3,
	xleftmargin=0pt,                    
}

\usepackage[linesnumbered,ruled,vlined,algo2e]{algorithm2e}

\definecolor{cvprblue}{rgb}{0.0, 0.0, 1.0} 

\newcommand{\FigTeaser}{%
	\begin{figure}[!tbp]
		\centering
		\includegraphics[scale=0.58]{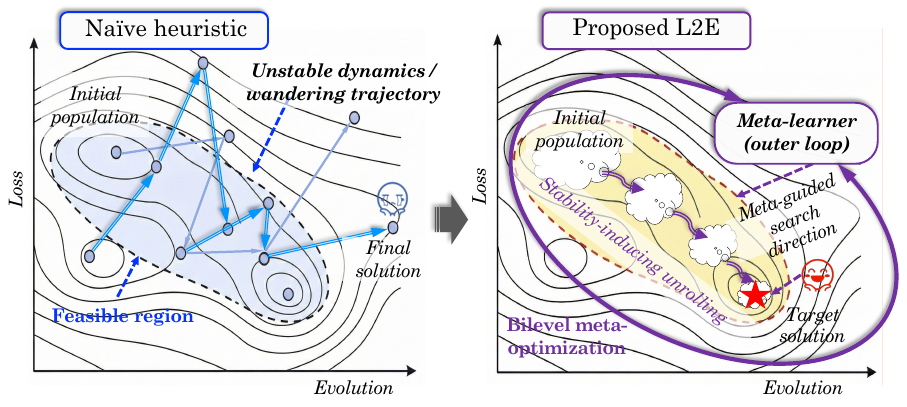} 
\caption{
	Conceptual contrast. \textit{Left:} Unconstrained learned heuristics may overfit training landscapes, resulting in unstable trajectories on unseen problems. 
	\textit{Right:} L2E learns a  {stability-inducing} unrolled evolutionary dynamics with bilevel meta-training, shaping the search trajectory toward controlled progress and improved robustness under distribution shifts.}
		\label{fig:fig0}
	\end{figure}
}

\newcommand{\FigPipeline}{%
	\begin{figure*}[!tbp]
		\centering
		\includegraphics[scale=0.7]{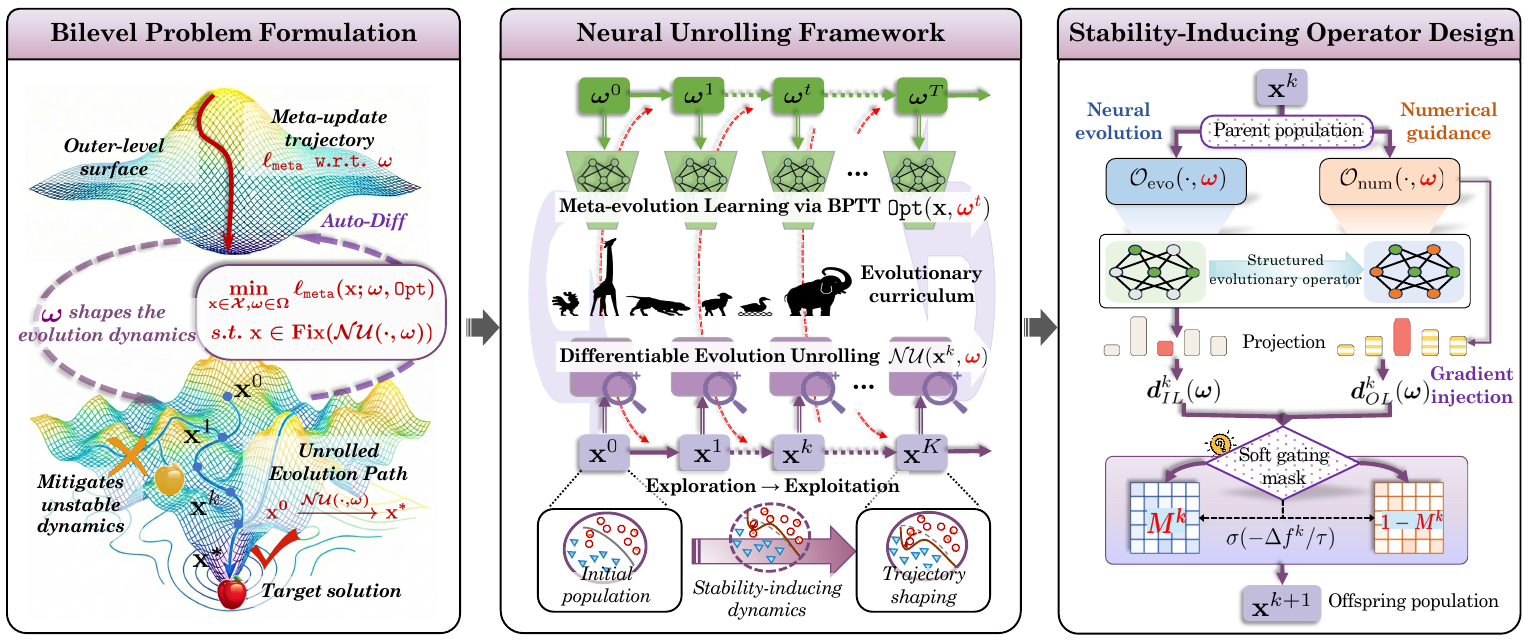} 
\caption{
	 {Overall framework of L2E.} 
	\textit{Left:} L2E is formulated as a bilevel meta-optimization problem where the outer loop updates meta-parameters to shape the induced evolution dynamics.
	\textit{Middle:} The evolutionary search is unrolled into a differentiable pipeline via AutoDiff to learn stable and effective search trajectories.
	\textit{Right:} L2E implements a stability-inducing operator design that combines a learned evolutionary operator with numerical guidance, promoting controlled population updates.
}
		\label{fig:pipeline}
	\end{figure*}
}

\newcommand{\AlgLTwoE}{%
	\begin{algorithm2e}[t]
		\small
		\DontPrintSemicolon 
		\SetKwInOut{Input}{\textbf{Input}}
		\SetKwInOut{Output}{\textbf{Output}}
		\SetKw{Return}{\textbf{Return}}
		\SetKwFunction{SortPop}{\textsc{SortPop}}
		\SetKwFunction{Proj}{\textsc{Proj}}
		\SetKwFunction{AutoDiff}{\texttt{AutoDiff}}
		
		\Input{Initial population $\mathbf{x}_0 \in \mathcal{X}$, 
			Meta-parameters $\bm{\omega}$, 
			Learning rate $\gamma$,
			Unroll horizon $K$, 
			Outer iterations $T$,
			Relaxation factor $\alpha \in (0,1]$,
			Differentiable operators $\mathcal{O}_{\mathrm{evo}}$ (Mamba) / $\mathcal{O}_{\mathrm{num}}$ (Proxy)
		}
		\Output{Optimized population $\mathbf{x}^K$, meta-parameters $\bm{\omega}^T$}
		
		\vspace{1mm}
		\textbf{Initialization:} $\bm{\omega}^0 \leftarrow \bm{\omega}_{\text{init}}$; \
		
		\For{$t = 0$ \KwTo $T-1$}{
			$\mathbf{x}^0 \leftarrow \mathbf{x}_0$ \tcp*{\textbf{Reset Trajectory}}
			
			\vspace{1mm}
			\tcc{\textcolor{black}{\textbf{Differentiable Evolution Unrolling (Inner-Level)}}}
			\For{$k = 1$ \KwTo $K$}{
				\tcc{Step 1: Stability-inducing Neural Evolution (KM Iteration)}
				$\mathbf{d}_{IL}^{k} \gets (1-\alpha)\mathbf{x}^{k-1} + \alpha \cdot \mathcal{O}_{\mathrm{evo}}(\mathbf{x}^{k-1}, \bm{\omega})$\;
				
				\tcc{Step 2: Numerical Guidance}
				\eIf{\textnormal{\textbf{is training}}}{
					$\mathbf{g}_{\text{proxy}} \leftarrow \bm{P}_{\bm{\omega}}^{-1} \nabla_{\mathbf{x}} \ell_{\mathtt{meta}}(\mathbf{x}^{k-1}; \bm{\omega}^t)$ \tcp*{\textbf{Gradient Injection}}
					$\mathbf{d}_{OL}^k \leftarrow \mathbf{x}^{k-1} - s_k \cdot \mathbf{g}_{\text{proxy}}$ \;
				}{
					$\mathbf{d}_{OL}^k \leftarrow \mathbf{x}^{k-1}$ \tcp*{\textbf{Evaluation Mode}}
				}
				
				\tcc{Step 3:  Soft-gated Fusion}
				$\Delta f^k \leftarrow f(\mathbf{d}_{OL}^k) - f(\mathbf{d}_{IL}^k)$; \quad
				$\bm{M}^k \leftarrow \sigma(-\Delta f^k / \tau)$  
				$\mathbf{x}^k \leftarrow\text{Proj}_{\mathcal{X}, \bm{P}_{\bm{\omega}}} \left( \bm{M}^k \odot \bm{d}_{OL}^k + (1 - \bm{M}^k) \odot \bm{d}_{IL}^k \right)$
			}
			
			\vspace{1mm}
			\tcc{\textcolor{black}{\textbf{Meta-Update via AutoDiff (Outer-Level)}}}
			$\ell_{\mathtt{meta}}(\bm{\omega}^t) \leftarrow - \mathbb{E}_{f \sim p(f)} \left[ \frac{\bar{f}(\mathbf{x}^0) - \bar{f}(\mathbf{x}^K(\bm{\omega}^t))}{|\bar{f}(\mathbf{x}^0)| + \epsilon} \right]$ \tcp*{Normalized Improvement} 
			
			$\nabla_{\bm{\omega}} \ell_{\mathtt{meta}} \leftarrow \AutoDiff(\ell_{\mathtt{meta}}, \{\mathbf{x}^k\}_{k=0}^K)$ \tcp*{BPTT}
			$\bm{\omega}^{t+1} \leftarrow \bm{\omega}^t - \gamma \cdot \nabla_{\bm{\omega}}\ell_{\mathtt{meta}}$
		}
		
		\Return $\mathbf{x}^K$, $\bm{\omega}^T$
		
		\caption{\textbf{L2E: Differentiable Bilevel Evolution Strategy.} The algorithm couples a KM-style averaged neural update with  {dual-mode guidance}: proxy gradients are injected during training, while inference remains evaluation-only.}
		\label{alg:l2e_advanced_dual}
	\end{algorithm2e}
}

\newcommand{\FigMamba}{%
	\begin{figure}[!tbp]
		\centering
		\includegraphics[scale=0.68]{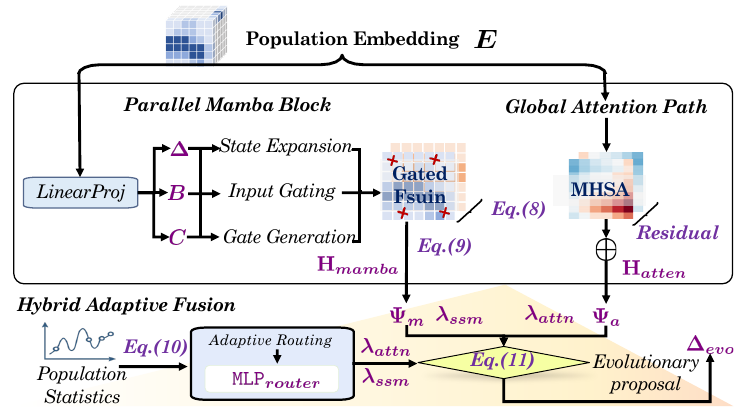} 
		\caption{
			Architecture of the Structured Mamba Neural Operator.
		}
		\label{fig:fig3}
	\end{figure}
}

\newcommand{\FigbbobTen}{%
\begin{table*}[!tbp]
	\centering
	\scriptsize
	\setlength{\tabcolsep}{2pt}
	\caption{Numerical performance evaluation (mean $\pm$ std) of representative algorithms on the \emph{BBOB-10D} benchmark. Each entry is calculated from 10 independent test runs. The best and second-best results are highlighted using \underline{\textbf{bold}} and \textbf{bold}, respectively.}
	\label{tab:bbob-10d}
	\begin{tabular}{lcccccccc}
		\toprule
		\textbf{Method} & \texttt{Buche\_Ras} & \texttt{Attractive\_Sec} & \texttt{Step\_Ell} & \texttt{Rosenbrock\_ori} & \texttt{Rosenbrock\_rot} & \texttt{Ellipsoidal\_hc} & \texttt{Discus} & \texttt{Bent\_Cig} \\
		\midrule 
		\multirow{2}{*}{PSO} & 3.429E+02 & 1.109E+03 & 8.667E+01 & 4.003E+03 & 1.798E+03 & 3.288E+05 & 7.338E+01 & 1.768E+07 \\
		& ($\pm$6.591E+01) & ($\pm$7.005E+02) & ($\pm$1.083E+01) & ($\pm$1.471E+03) & ($\pm$7.772E+02) & ($\pm$2.522E+05) & ($\pm$3.266E+01) & ($\pm$1.077E+07) \\
		\multirow{2}{*}{DE} & \textbf{1.012E+02} & 4.934E+01 & 3.099E+01 & 3.527E+02 & 7.201E+02 & 2.859E+04 & 4.860E+01 & 2.070E+06 \\
		& ($\pm$2.988E+01) & ($\pm$6.491E+00) & ($\pm$7.657E+00) & ($\pm$1.685E+02) & ($\pm$4.851E+02) & ($\pm$2.196E+04) & ($\pm$6.461E+00) & ($\pm$5.398E+05) \\
		\multirow{2}{*}{SAHLPSO} & 1.424E+02 & 1.167E+02 & 2.402E+01 & 9.622E+02 & 1.070E+03 & 4.344E+04 & 1.141E+02 & 7.618E+06 \\
		& ($\pm$6.008E+00) & ($\pm$8.459E+01) & ($\pm$1.786E+01) & ($\pm$6.640E+02) & ($\pm$6.981E+02) & ($\pm$2.739E+04) & ($\pm$2.008E+01) & ($\pm$2.039E+06) \\
		\midrule 
		\multirow{2}{*}{RNNOPT} & 6.616E+03 & 6.720E+04 & 7.170E+01 & 1.930E+04 & 6.619E+01 & 3.805E+06 & 1.740E+05 & 6.700E+07 \\
		& ($\pm$0.000E+00) & ($\pm$0.000E+00) & ($\pm$0.000E+00) & ($\pm$0.000E+00) & ($\pm$0.000E+00) & ($\pm$0.000E+00) & ($\pm$0.000E+00) & ($\pm$0.000E+00) \\
		\multirow{2}{*}{DEDQN} & 3.627E+02 & 9.099E+03 & 1.018E+02 & 1.369E+04 & 1.021E+04 & 5.756E+05 & 2.154E+02 & 3.578E+07 \\
		& ($\pm$9.568E+01) & ($\pm$1.020E+04) & ($\pm$2.562E+01) & ($\pm$3.021E+03) & ($\pm$3.073E+03) & ($\pm$1.608E+05) & ($\pm$5.994E+01) & ($\pm$1.490E+07) \\
		\multirow{2}{*}{LES} & 1.473E+03 & 1.816E+03 & 3.076E+02 & 2.697E+03 & 1.334E+03 & 3.207E+05 & 9.363E+02 & 1.308E+07 \\
		& ($\pm$4.452E+01) & ($\pm$1.624E+03) & ($\pm$7.422E-01) & ($\pm$2.109E+02) & ($\pm$4.173E+00) & ($\pm$6.588E+04) & ($\pm$8.504E-01) & ($\pm$1.848E+06) \\
		\multirow{2}{*}{LGA} & 2.462E+02 & 4.033E+02 & 1.794E+02 & 5.188E+03 & 3.062E+04 & 2.578E+05 & 2.265E+02 & 5.390E+07 \\
		& ($\pm$5.814E+01) & ($\pm$0.000E+00) & ($\pm$9.467E+00) & ($\pm$2.615E+03) & ($\pm$1.402E+04) & ($\pm$8.073E+04) & ($\pm$7.462E+01) & ($\pm$1.930E+07) \\
		\multirow{2}{*}{GLHF} & 4.513E+02 & 1.689E+04 & 9.959E+01 & 1.324E+04 & 1.156E+04 & 4.462E+05 & 7.396E+02 & 3.449E+07 \\
		& ($\pm$1.747E+02) & ($\pm$1.254E+04) & ($\pm$1.196E+01) & ($\pm$3.514E+03) & ($\pm$2.001E+03) & ($\pm$1.668E+05) & ($\pm$5.473E+02) & ($\pm$3.621E+06) \\
		\multirow{2}{*}{B2OPT} & 2.734E+02 & 1.116E+02 & 1.061E+01 & 3.067E+02 & \textbf{2.422E+01} & 3.050E+04 & 4.083E+01 & 7.585E+06 \\
		& ($\pm$4.075E+01) & ($\pm$4.417E+00) & ($\pm$3.472E+00) & ($\pm$2.971E+02) & ($\pm$3.558E+00) & ($\pm$1.391E+04) & ($\pm$3.942E+00) & ($\pm$1.256E+06) \\
		\midrule
		\rowcolor{blue!2} \multirow{1}{*}{Ours} & 1.079E+02 & \textbf{4.523E+01} & \textbf{5.048E+00} & \textbf{5.148E+01} & \underline{\textbf{8.213E+00}} & \textbf{1.798E+04} & \underline{\textbf{3.001E+01}} & \textbf{4.299E+05} \\
		\rowcolor{blue!2} & ($\pm$1.640E+01) & ($\pm$1.305E+01) & ($\pm$8.704E$-$01) & ($\pm$1.390E+01) & ($\pm$1.041E+00) & ($\pm$4.549E+03) & ($\pm$3.792E+00) & ($\pm$9.796E+04) \\
		\rowcolor{blue!5} \multirow{1}{*}{Ours$^{\ddag}$} & \underline{\textbf{8.209E+01}} & \underline{\textbf{5.386E+00}} & \underline{\textbf{3.714E+00}} & \underline{\textbf{3.111E+01}} & 2.764E+01 & \underline{\textbf{5.438E+03}} & \textbf{3.869E+01} & \underline{\textbf{1.774E+05}} \\
		\rowcolor{blue!5} & ($\pm$1.394E+01) & ($\pm$1.164E+00) & ($\pm$1.130E+00) & ($\pm$5.541E+00) & ($\pm$3.051E+00) & ($\pm$8.623E+02) & ($\pm$8.844E+00) & ($\pm$2.297E+04) \\
		\bottomrule
	\end{tabular}
	\vspace{0.3em}
	\begin{minipage}{\textwidth}
		\scriptsize
		\raggedright
		~~~~$^{\ddag}$ indicates ours variant enhanced with the proposed Mamba-based neural operator.  {Due to space limitations, only a subset is shown here; the full table is provided in the Supplementary Material (Table~\ref{tab:supp-bbob-10d}).}
	\end{minipage}
\end{table*}
}

\newcommand{\FigbbobThirty}{%
	\begin{figure*}[!tbp]
	\centering
	
	\begin{subfigure}[b]{0.24\textwidth}
		\includegraphics[width=\linewidth]{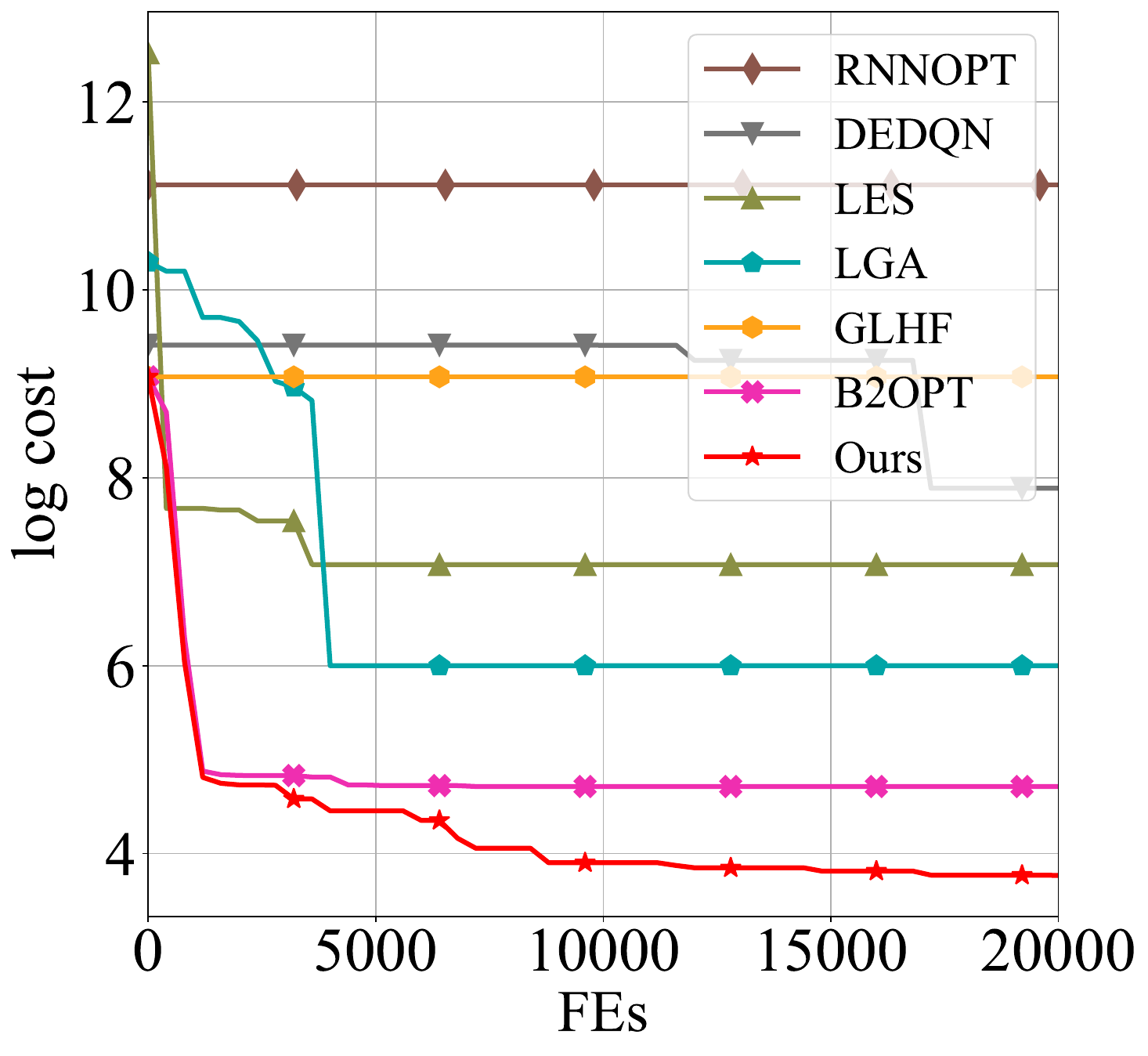}
		\caption*{Attractive Sector}
	\end{subfigure}
	\begin{subfigure}[b]{0.24\textwidth}
		\includegraphics[width=\linewidth]{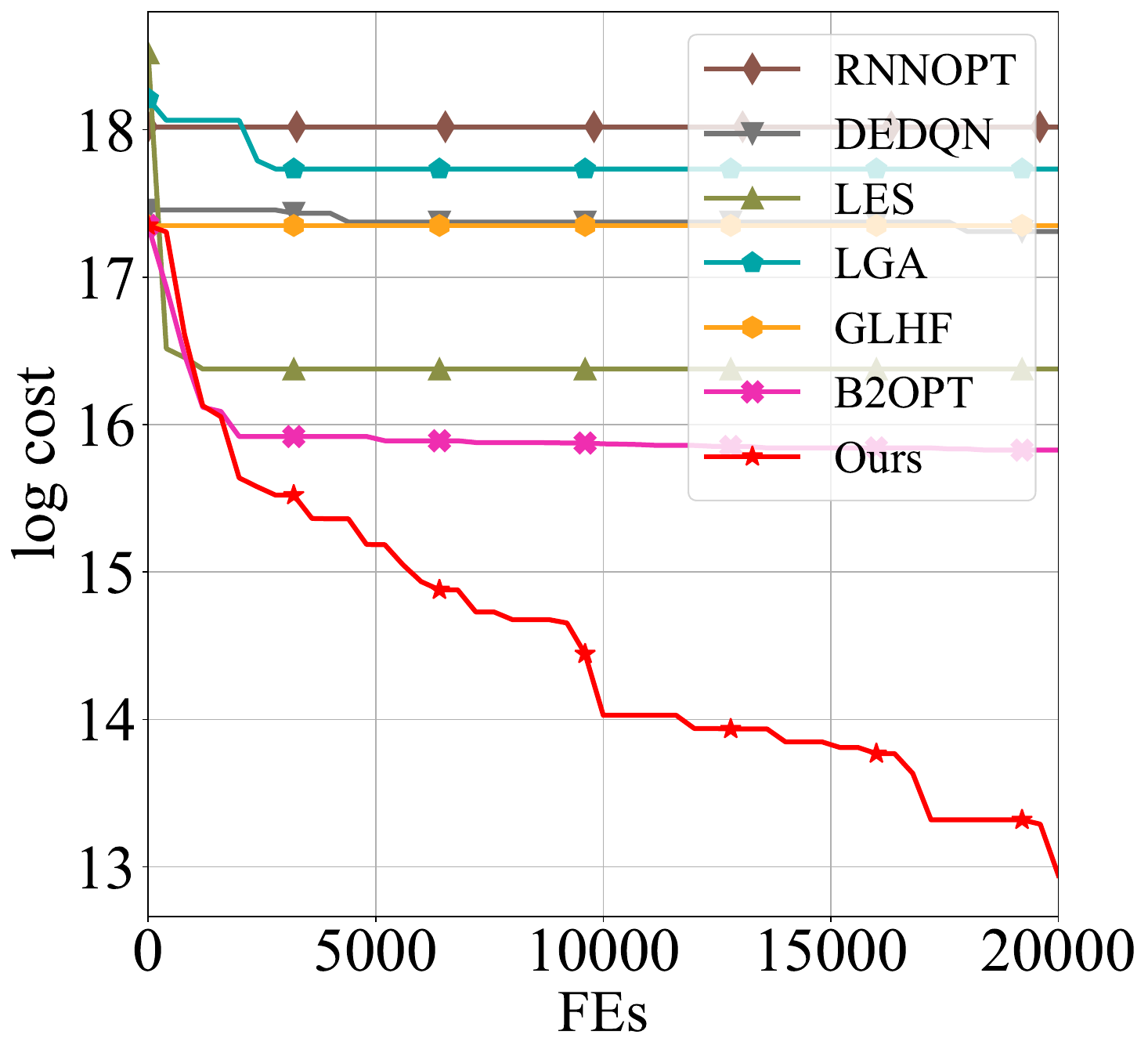}
		\caption*{Bent Cigar}
	\end{subfigure}
	\begin{subfigure}[b]{0.24\textwidth}
		\includegraphics[width=\linewidth]{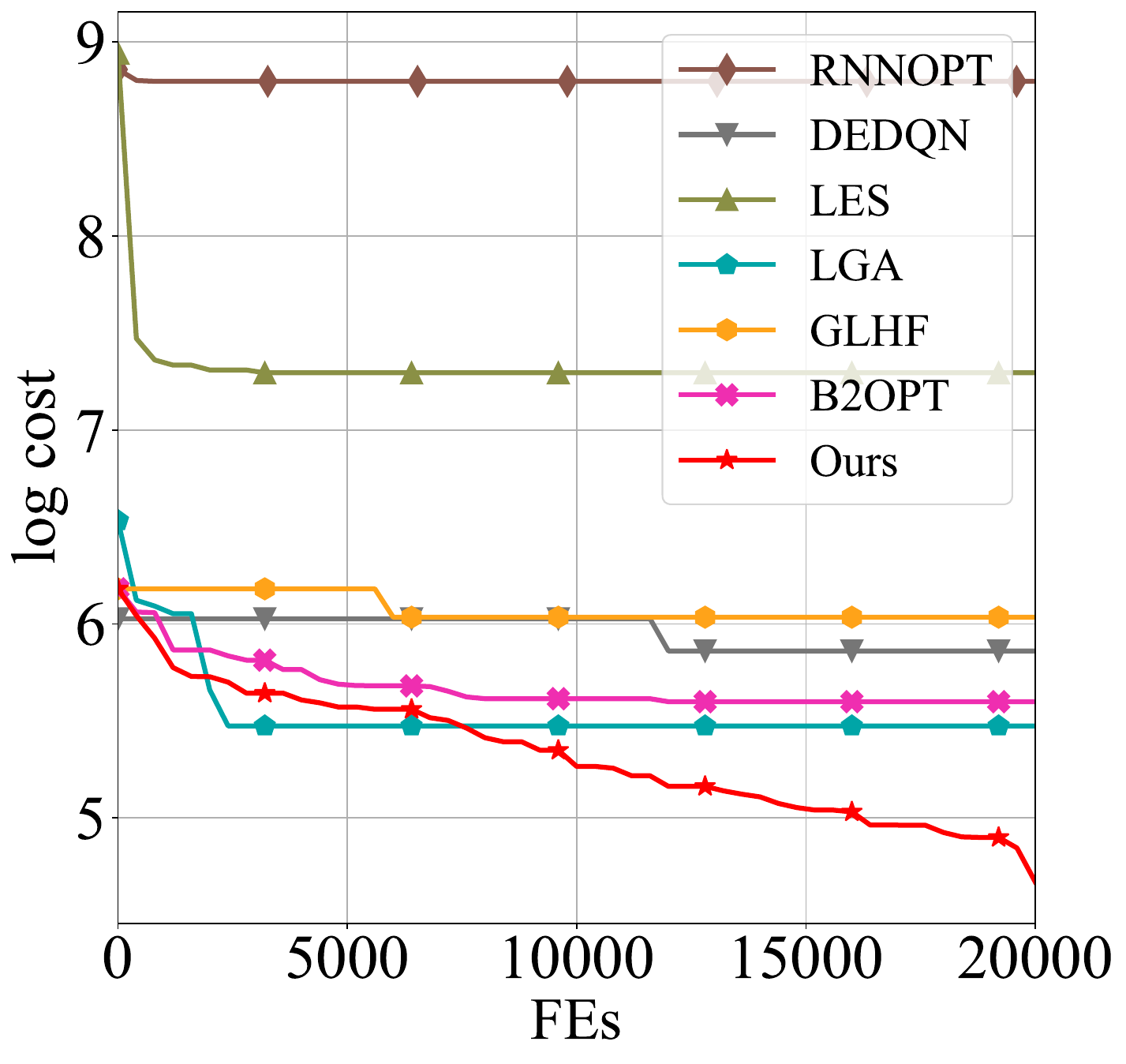}
		\caption*{Buche Rastrigin}
	\end{subfigure}
	\begin{subfigure}[b]{0.24\textwidth}
		\includegraphics[width=\linewidth]{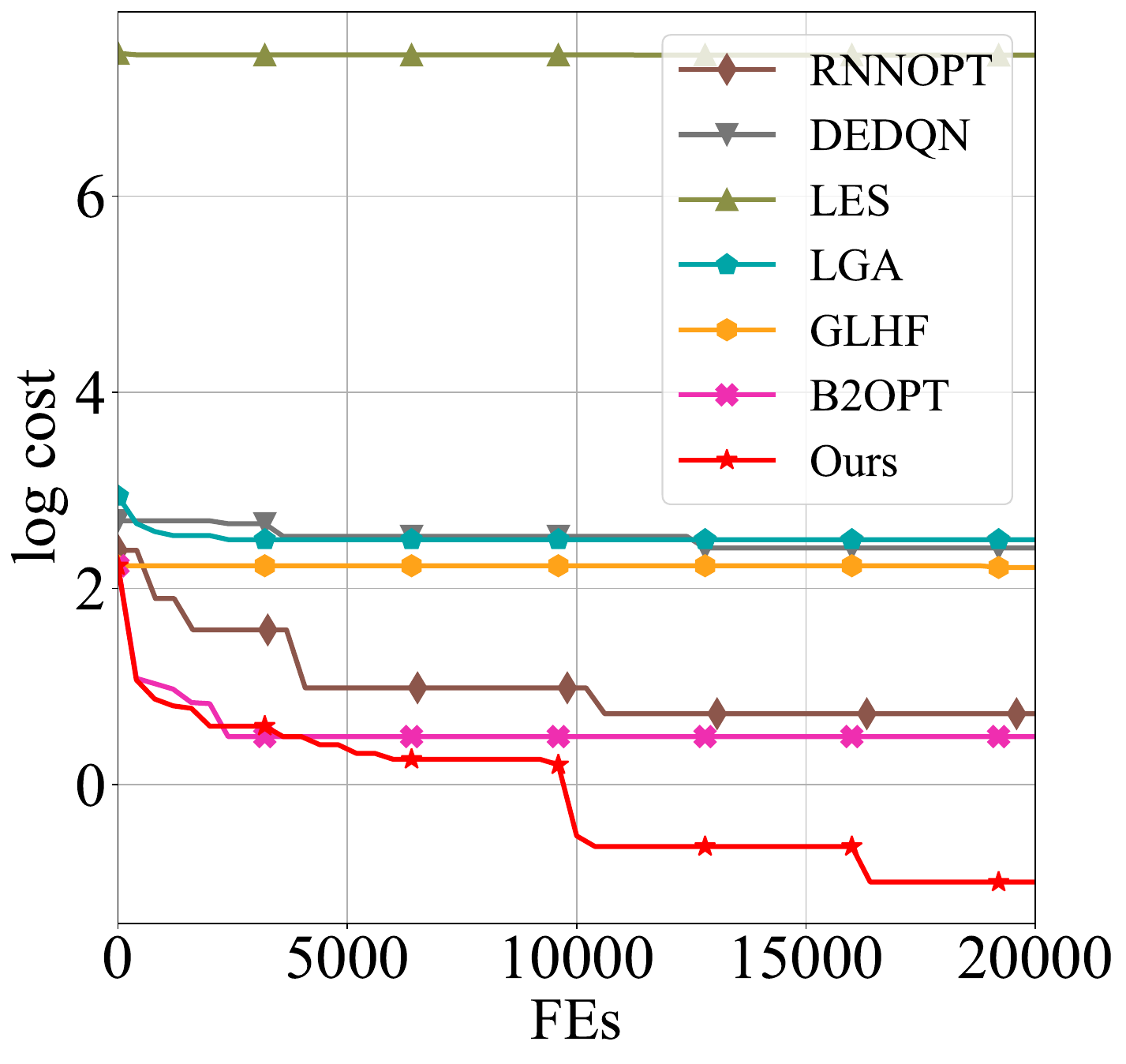}
		\caption*{Composite Grie-Rosen}
	\end{subfigure}
	
	\begin{subfigure}[b]{0.24\textwidth}
		\includegraphics[width=\linewidth]{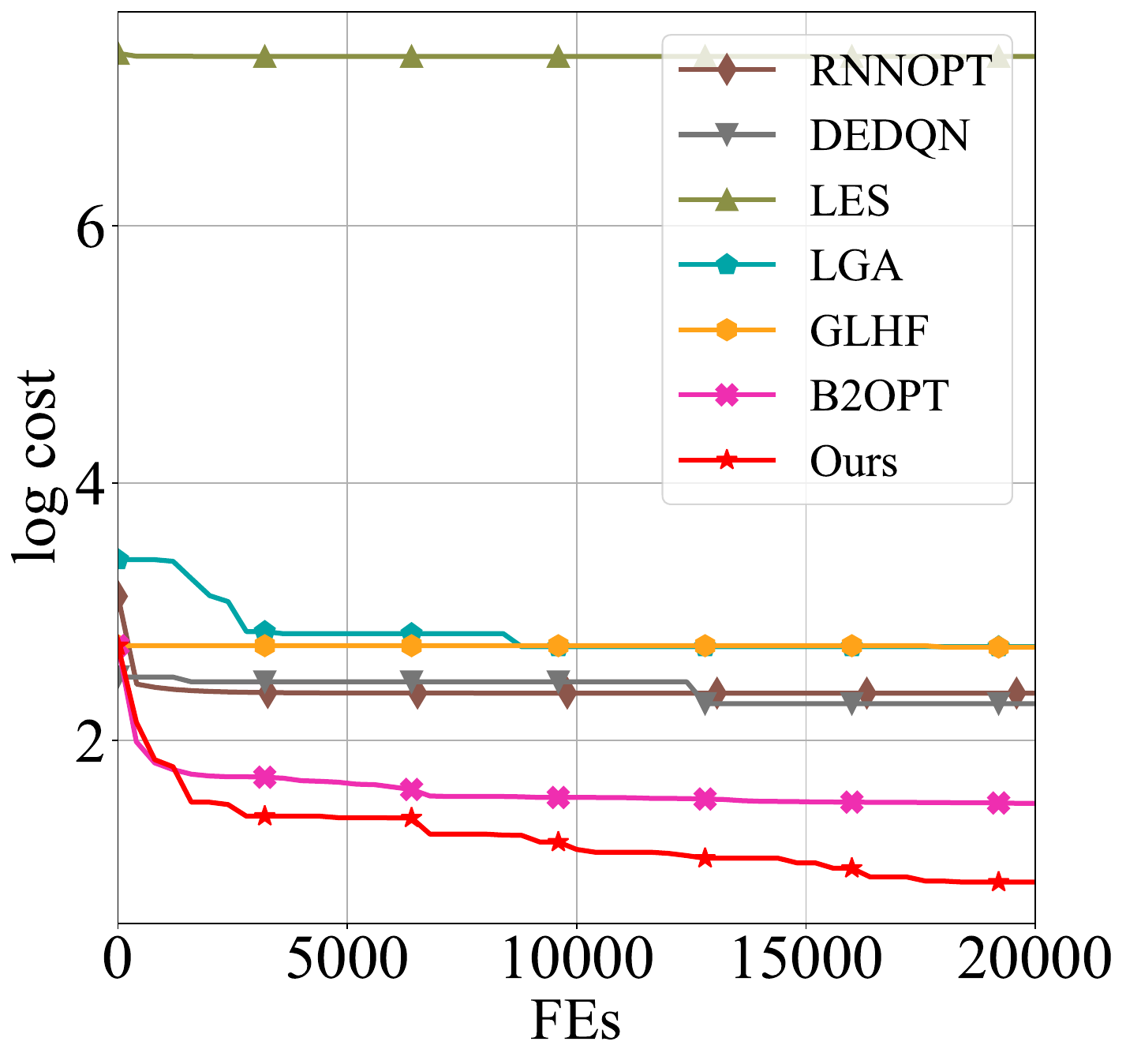}
		\caption*{Different Powers}
	\end{subfigure}
	\begin{subfigure}[b]{0.24\textwidth}
		\includegraphics[width=\linewidth]{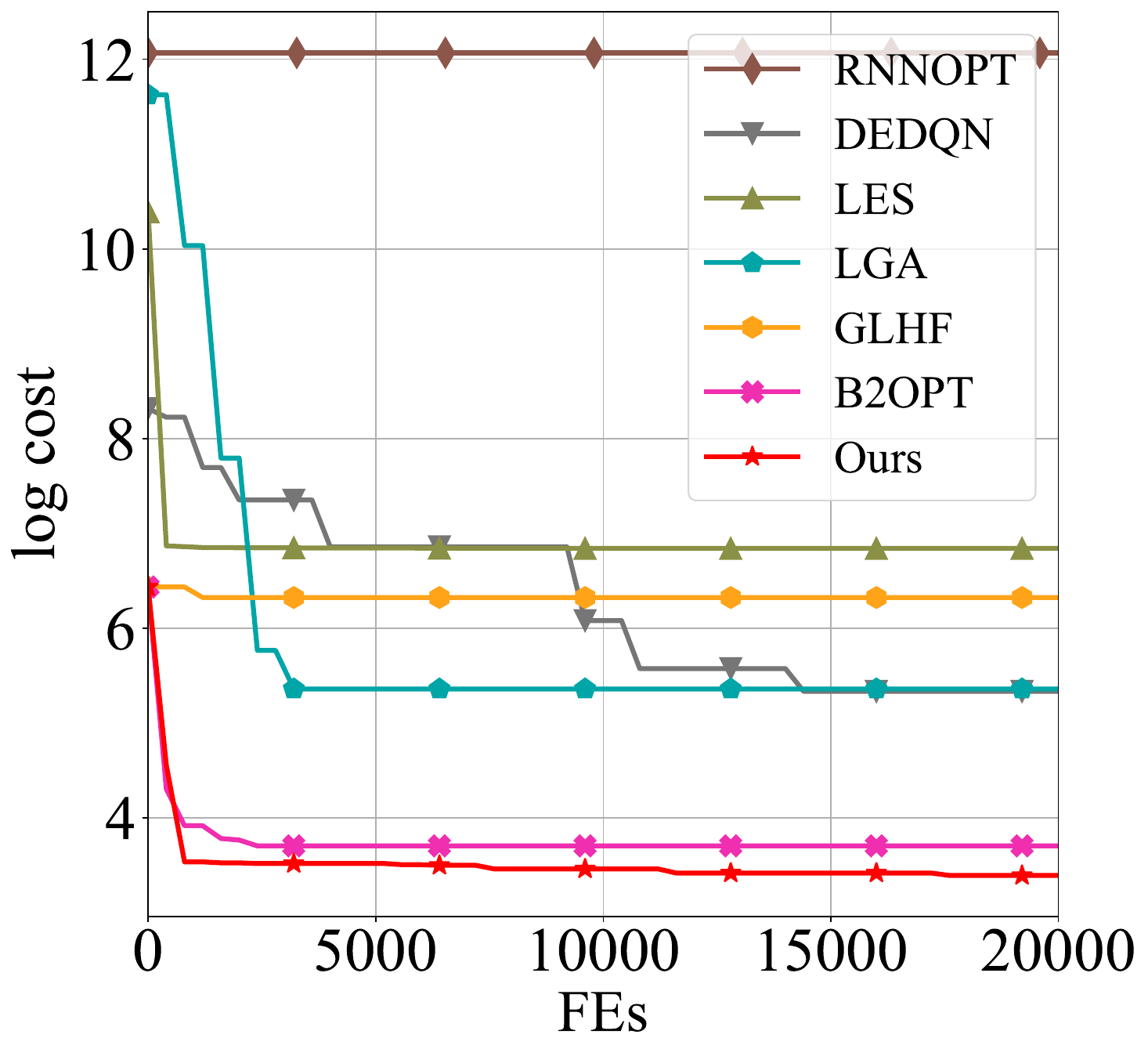}
		\caption*{Discus}
	\end{subfigure}
	\begin{subfigure}[b]{0.24\textwidth}
		\includegraphics[width=\linewidth]{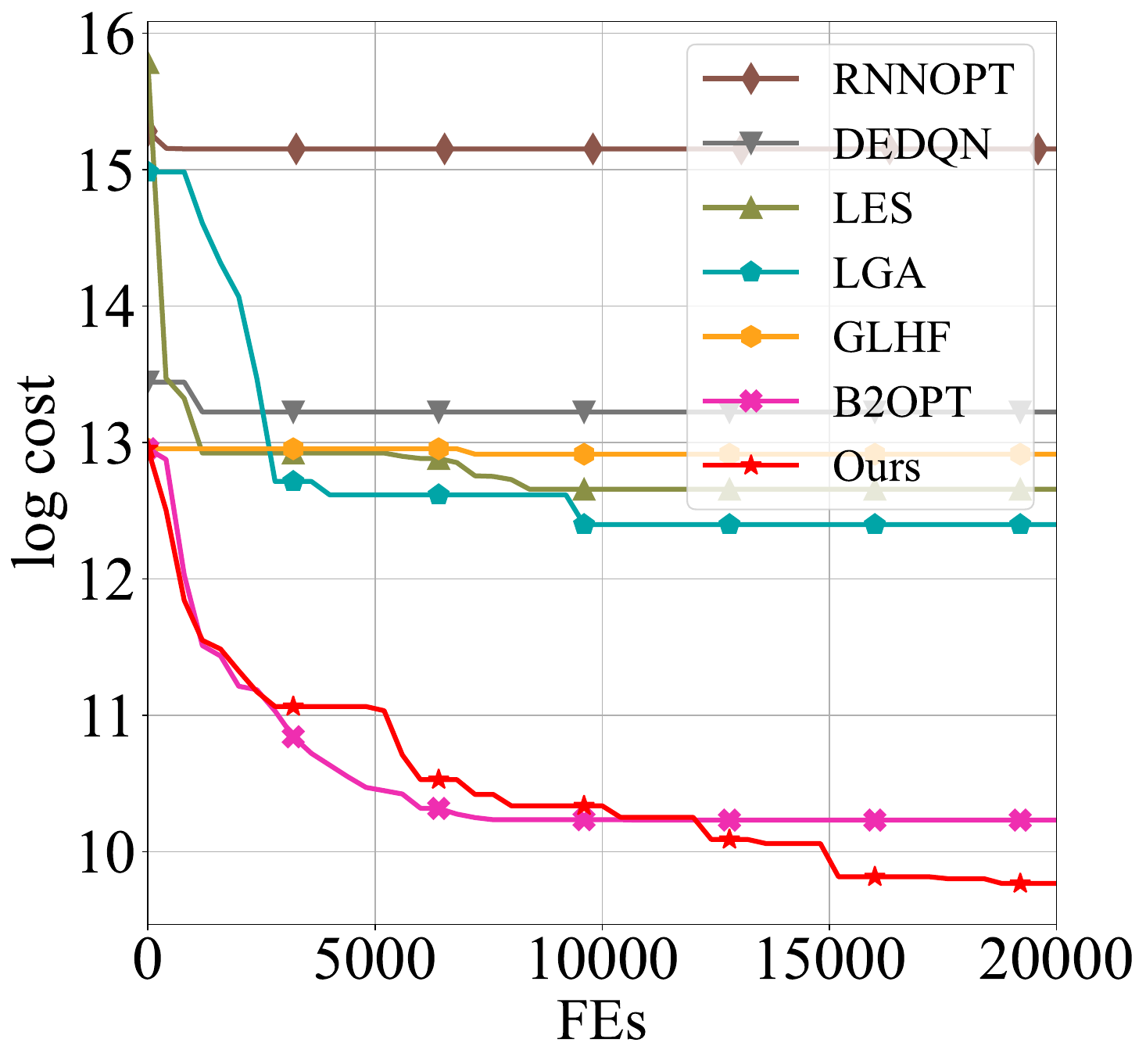}
		\caption*{Ellipsoidal}
	\end{subfigure}
	\begin{subfigure}[b]{0.24\textwidth}
		\includegraphics[width=\linewidth]{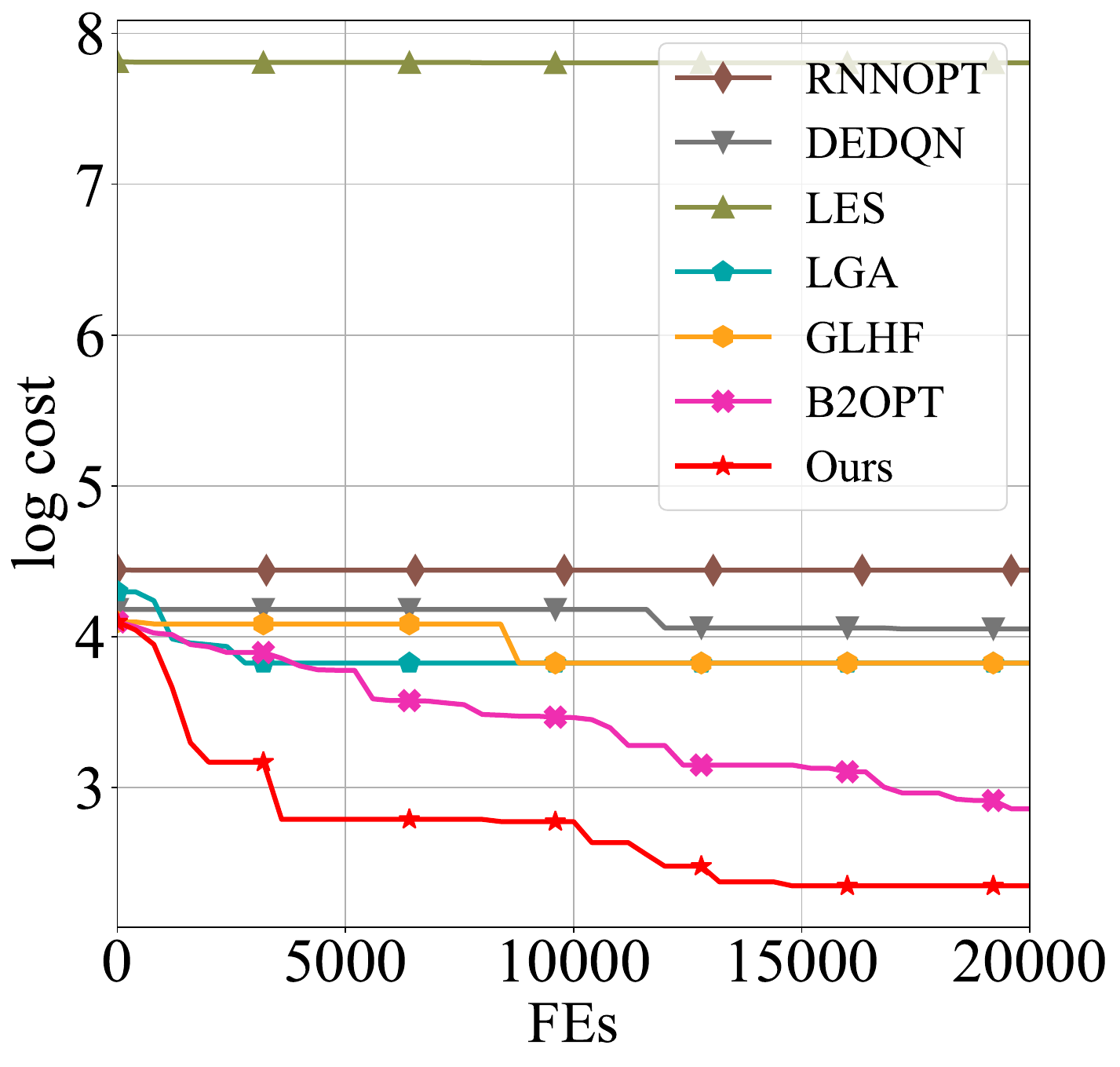}
		\caption*{Gallagher 21Peaks}
	\end{subfigure}
	
	\caption{Log-scaled convergence trajectories of our method and representative methods on the first 8 benchmark tasks from the BBOB-10D suite~\cite{hansen2021coco}. 
		Complete results on all 16 functions are provided in the \textit{supplementary material} (i.e., Fig~\ref{fig:supp-bbob-10d-diff-16}).}
	\label{fig:bbob-10d-diff-16}
\end{figure*}
}

\newcommand{\TabbbobTen}{%
	\begin{table*}[!tbp]
	\centering
	\scriptsize
	\setlength{\tabcolsep}{2pt}
	\caption{Numerical performance evaluation (mean $\pm$ std) of representative algorithms on the \emph{BBOB-30D} benchmark.. Each entry is calculated from 10 independent test runs. The best and second-best results are highlighted using \underline{\textbf{bold}} and \textbf{bold}, respectively.}
	\label{tab:bbob-30d}
	\begin{tabular}{lcccccccc}
		\toprule
		\textbf{Method} & \texttt{Buche\_Ras} & \texttt{Attractive\_Sec} & \texttt{Step\_Ell} & \texttt{Rosenbrock\_ori} & \texttt{Rosenbrock\_rot} & \texttt{Ellipsoidal\_hc} & \texttt{Discus} & \texttt{Bent\_Cig} \\
		\midrule 
		\multirow{2}{*}{PSO} & 2.169E+03 & 1.224E+05 & 7.002E+02 & 1.109E+04 & 5.884E+02 & 1.110E+06 & \textbf{1.633E+02} & 1.159E+08 \\
		& ($\pm$5.998E+02) & ($\pm$5.208E+04) & ($\pm$4.272E+01) & ($\pm$2.664E+03) & ($\pm$2.849E+02) & ($\pm$2.838E+05) & ($\pm$2.059E+01) & ($\pm$1.495E+07) \\
		\multirow{2}{*}{DE} &  {\textbf{4.996E+02}} & \textbf{8.765E+02} & \textbf{1.224E+02} & 2.471E+03 & 2.814E+03 & \textbf{3.085E+05} & 1.820E+02 & 2.201E+07 \\
		& ($\pm$1.715E+01) & ($\pm$7.413E+02) & ($\pm$3.713E+01) & ($\pm$7.355E+02) & ($\pm$5.186E+02) & ($\pm$1.613E+05) & ($\pm$1.608E+01) & ($\pm$3.474E+06) \\
		\multirow{2}{*}{SAHLPSO} & 9.574E+02 & 5.158E+04 & 4.115E+02 & 2.687E+04 & 9.397E+03 & 1.296E+06 & 2.760E+02 & 1.124E+08 \\
		& ($\pm$6.404E+01) & ($\pm$4.482E+04) & ($\pm$1.473E+02) & ($\pm$1.173E+04) & ($\pm$2.132E+03) & ($\pm$5.424E+05) & ($\pm$7.851E+01) & ($\pm$1.062E+08) \\
		\midrule 
		\multirow{2}{*}{RNNOPT} & 1.603E+04 & 5.768E+05 & 1.511E+03 & 9.192E+04 & 1.975E+02 & 1.431E+07 & 4.950E+06 & 4.499E+08 \\
		& ($\pm$0.000E+00) & ($\pm$0.000E+00) & ($\pm$0.000E+00) & ($\pm$1.455E-11) & ($\pm$0.000E+00) & ($\pm$0.000E+00) & ($\pm$0.000E+00) & ($\pm$0.000E+00) \\
		\multirow{2}{*}{DEDQN} & 2.061E+03 & 2.325E+05 & 9.007E+02 & 1.712E+05 & 1.518E+05 & 3.772E+06 & 5.374E+02 & 2.796E+08 \\
		& ($\pm$3.744E+02) & ($\pm$4.669E+04) & ($\pm$4.917E+01) & ($\pm$1.802E+04) & ($\pm$9.146E+03) & ($\pm$4.637E+05) & ($\pm$2.222E+02) & ($\pm$3.171E+07) \\
		\multirow{2}{*}{LES} & 5.822E+03 & 3.620E+05 & 3.044E+03 & 5.335E+04 & 1.485E+03 & 3.974E+06 & 1.953E+03 & 2.468E+08 \\
		& ($\pm$3.715E+02) & ($\pm$2.230E+04) & ($\pm$1.385E+02) & ($\pm$9.407E+02) & ($\pm$5.223E-01) & ($\pm$2.115E+05) & ($\pm$3.210E+01) & ($\pm$1.006E+07) \\
		\multirow{2}{*}{LGA} & 1.082E+04 & 6.709E+05 & 1.266E+03 & 2.607E+05 & 2.086E+05 & 6.214E+06 & 1.187E+05 & 4.517E+08 \\
		& ($\pm$4.568E+03) & ($\pm$2.485E+05) & ($\pm$1.630E+02) & ($\pm$6.975E+04) & ($\pm$6.213E+04) & ($\pm$2.719E+06) & ($\pm$1.669E+05) & ($\pm$9.124E+07) \\
		\multirow{2}{*}{GLHF} & 4.510E+03 & 3.470E+05 & 1.275E+03 & 3.373E+05 & 1.993E+05 & 7.354E+06 & 1.892E+03 & 5.111E+08 \\
		& ($\pm$5.042E+02) & ($\pm$2.271E+05) & ($\pm$3.484E+02) & ($\pm$2.066E+04) & ($\pm$2.834E+04) & ($\pm$1.035E+06) & ($\pm$1.603E+03) & ($\pm$6.690E+07) \\
		\multirow{2}{*}{B2OPT} & 1.173E+03 & 1.243E+05 & 4.753E+02 & 2.698E+04 & 1.863E+02 & 1.270E+06 & 1.809E+02 & 1.656E+08 \\
		& ($\pm$1.705E+02) & ($\pm$4.855E+04) & ($\pm$5.596E+01) & ($\pm$1.223E+04) & ($\pm$9.315E-01) & ($\pm$3.203E+05) & ($\pm$1.733E+01) & ($\pm$1.558E+07) \\
		\midrule
		\rowcolor{blue!2} \multirow{1}{*}{Ours} &\underline{\textbf{4.914E+02}} & 2.497E+03 & 1.342E+02 & \textbf{1.753E+03} & \underline{\textbf{1.638E+02}} & 6.490E+05 & \underline{\textbf{1.518E+02}} & \textbf{2.096E+07} \\
		\rowcolor{blue!2} & ($\pm$3.327E+01) & ($\pm$2.761E+03) & ($\pm$1.739E+01) & ($\pm$2.956E+02) & ($\pm$9.913E-01) & ($\pm$1.787E+05) & ($\pm$2.712E+01) & ($\pm$7.699E+05) \\
		\rowcolor{blue!5} \multirow{1}{*}{Ours$^{\ddag}$} & 5.452E+02 & \underline{\textbf{1.585E+02}} & \underline{\textbf{6.737E+01}} & \underline{\textbf{4.076E+02}} & 2.378E+02 & \underline{\textbf{9.960E+04}} & 1.861E+02 & \underline{\textbf{2.684E+06}} \\
		\rowcolor{blue!5} & ($\pm$1.656E+01) & ($\pm$3.687E+01) & ($\pm$4.068E+00) & ($\pm$1.098E+02) & ($\pm$1.392E+01) & ($\pm$1.120E+04) & ($\pm$3.168E+01) & ($\pm$5.046E+05) \\
		\bottomrule
	\end{tabular}
	\vspace{0.3em}
	\begin{minipage}{\textwidth}
		\scriptsize
		\raggedright
		~~~~$^{\ddag}$ indicates ours variant enhanced with the proposed Mamba-based neural operator. {Due to space limitations, only a subset is shown here; the full table is provided in the Supplementary Material (Table~\ref{tab:supp-bbob-30d}).}
	\end{minipage}
\end{table*}
}

\newcommand{\Figlsgo}{%
\begin{figure*}[!tbp]
	\centering
	
	\begin{subfigure}[b]{0.24\textwidth}
		\includegraphics[width=\linewidth]{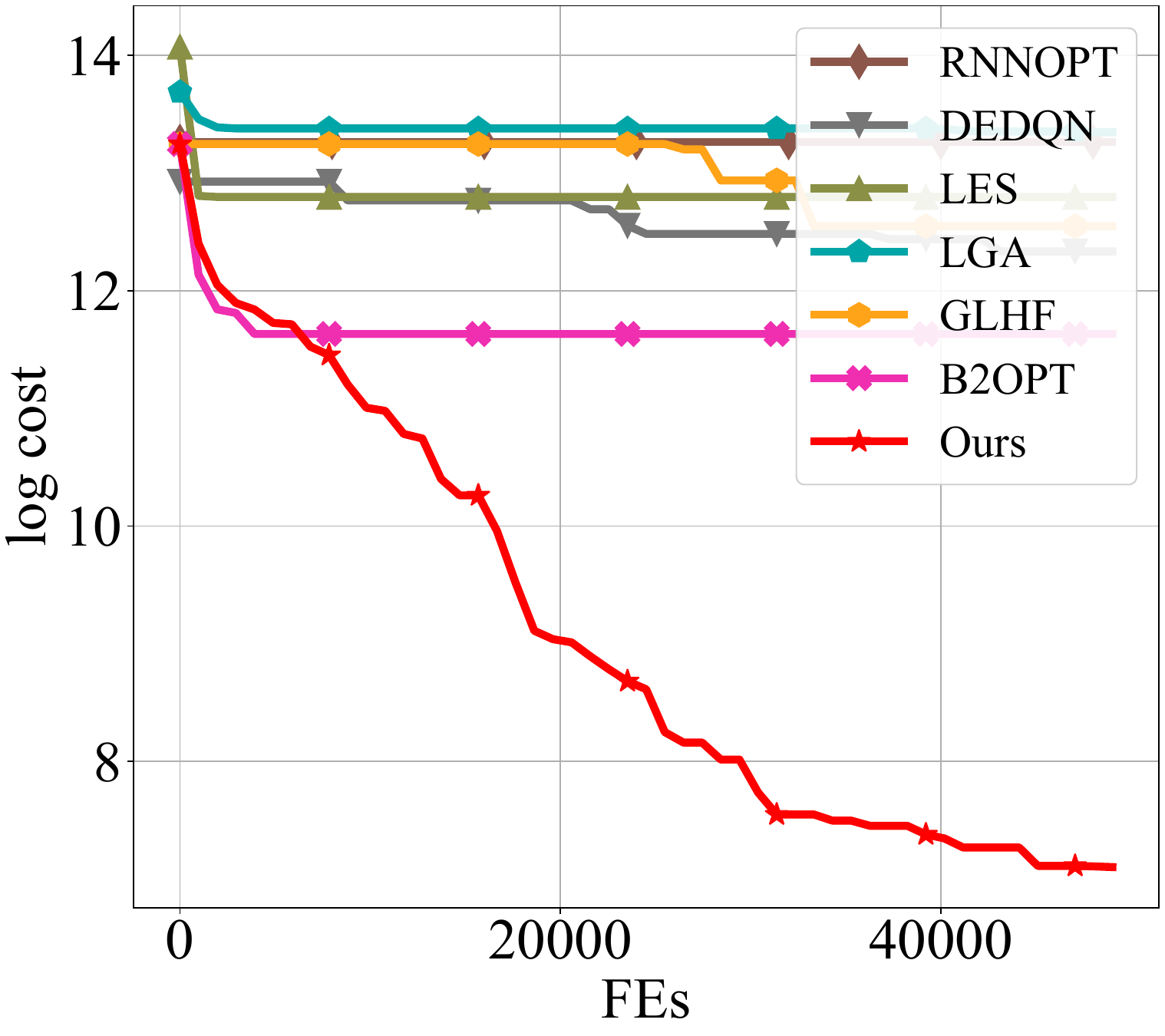}
		\caption*{Attractive Sector}
	\end{subfigure}
	\begin{subfigure}[b]{0.24\textwidth}
		\includegraphics[width=\linewidth]{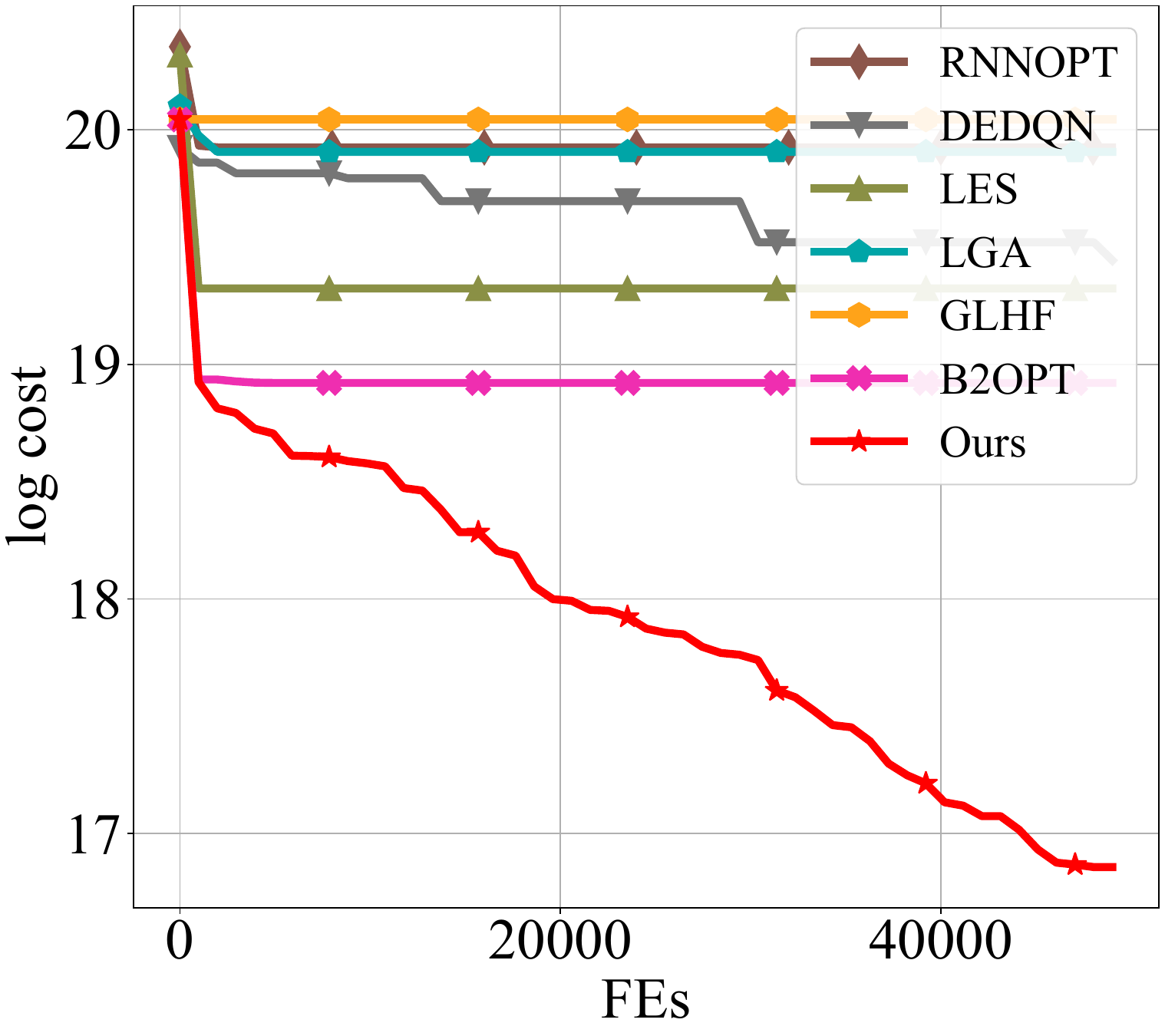}
		\caption*{Bent Cigar}
	\end{subfigure}
	\begin{subfigure}[b]{0.24\textwidth}
		\includegraphics[width=\linewidth]{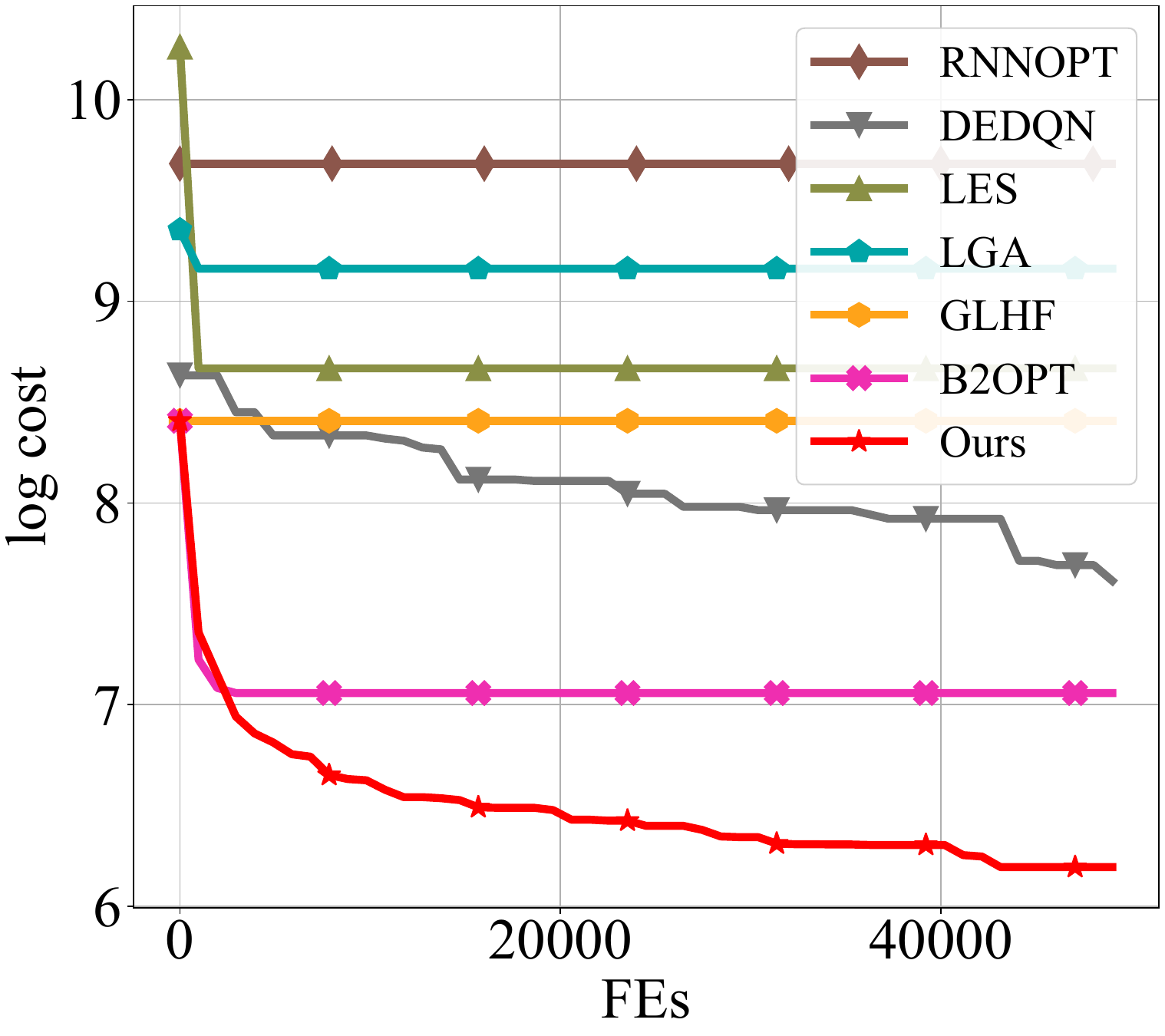}
		\caption*{Buche Rastrigin}
	\end{subfigure}
	\begin{subfigure}[b]{0.24\textwidth}
		\includegraphics[width=\linewidth]{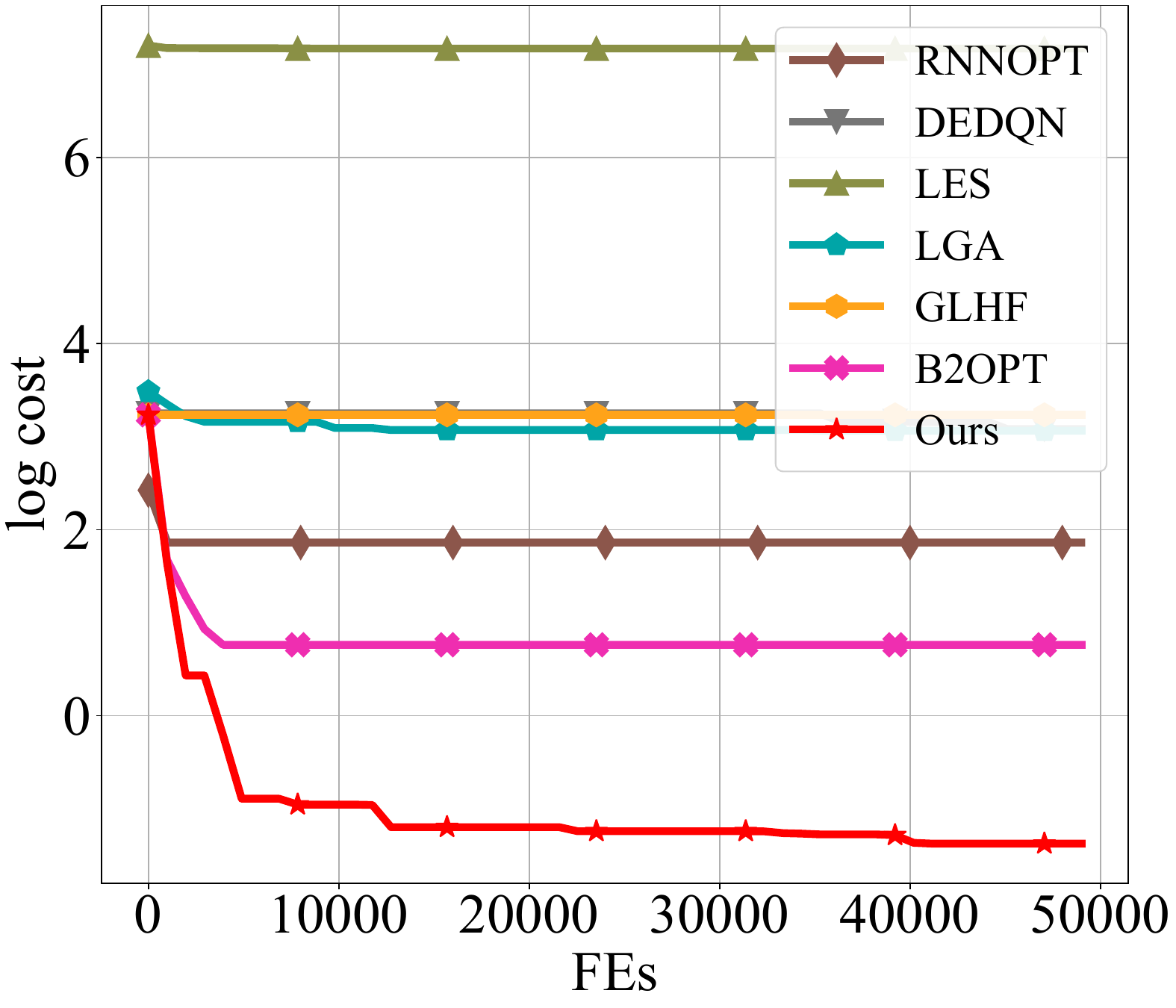}
		\caption*{Composite Grie-Rosen}
	\end{subfigure}
	
	\begin{subfigure}[b]{0.24\textwidth}
		\includegraphics[width=\linewidth]{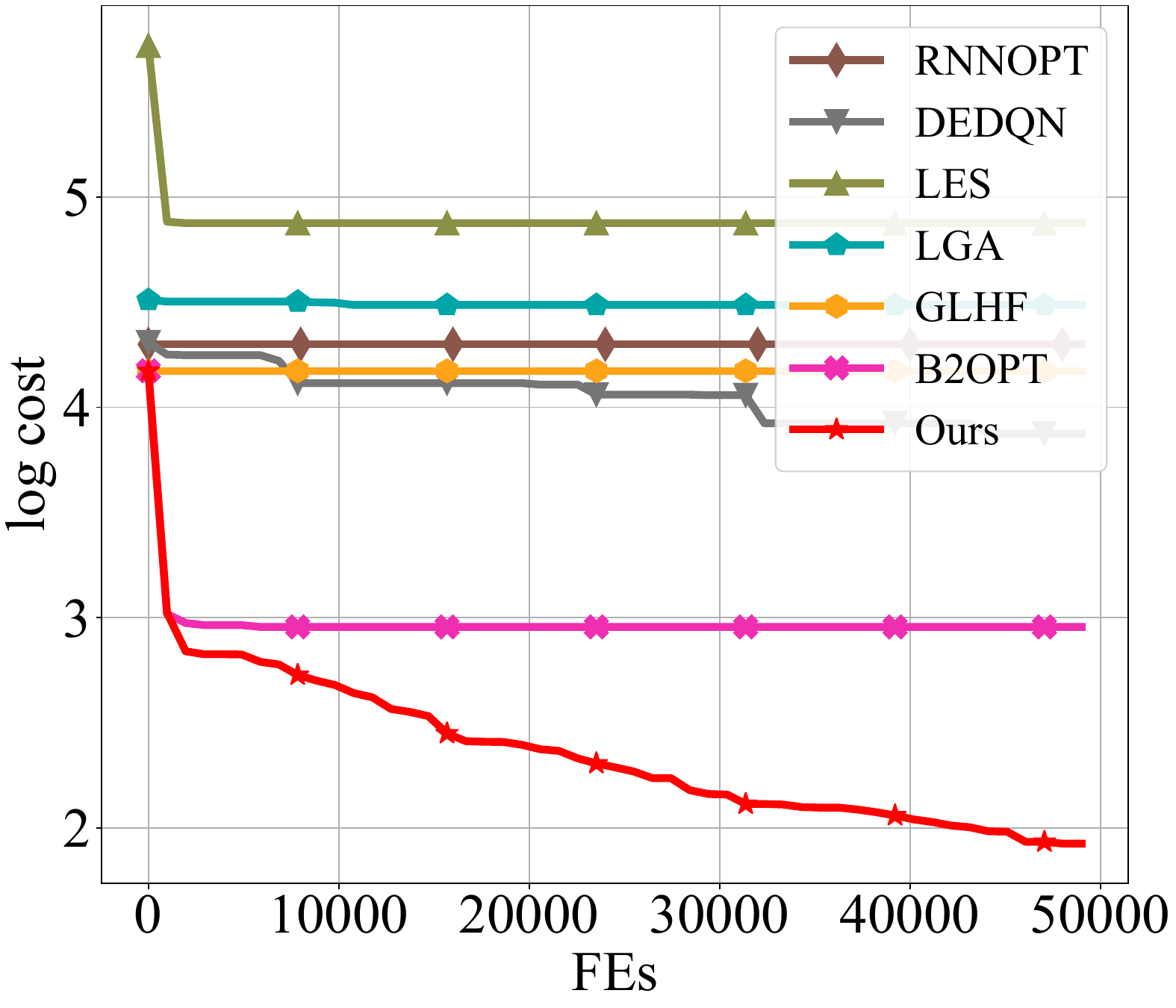}
		\caption*{Different Powers}
	\end{subfigure}
	\begin{subfigure}[b]{0.24\textwidth}
		\includegraphics[width=\linewidth]{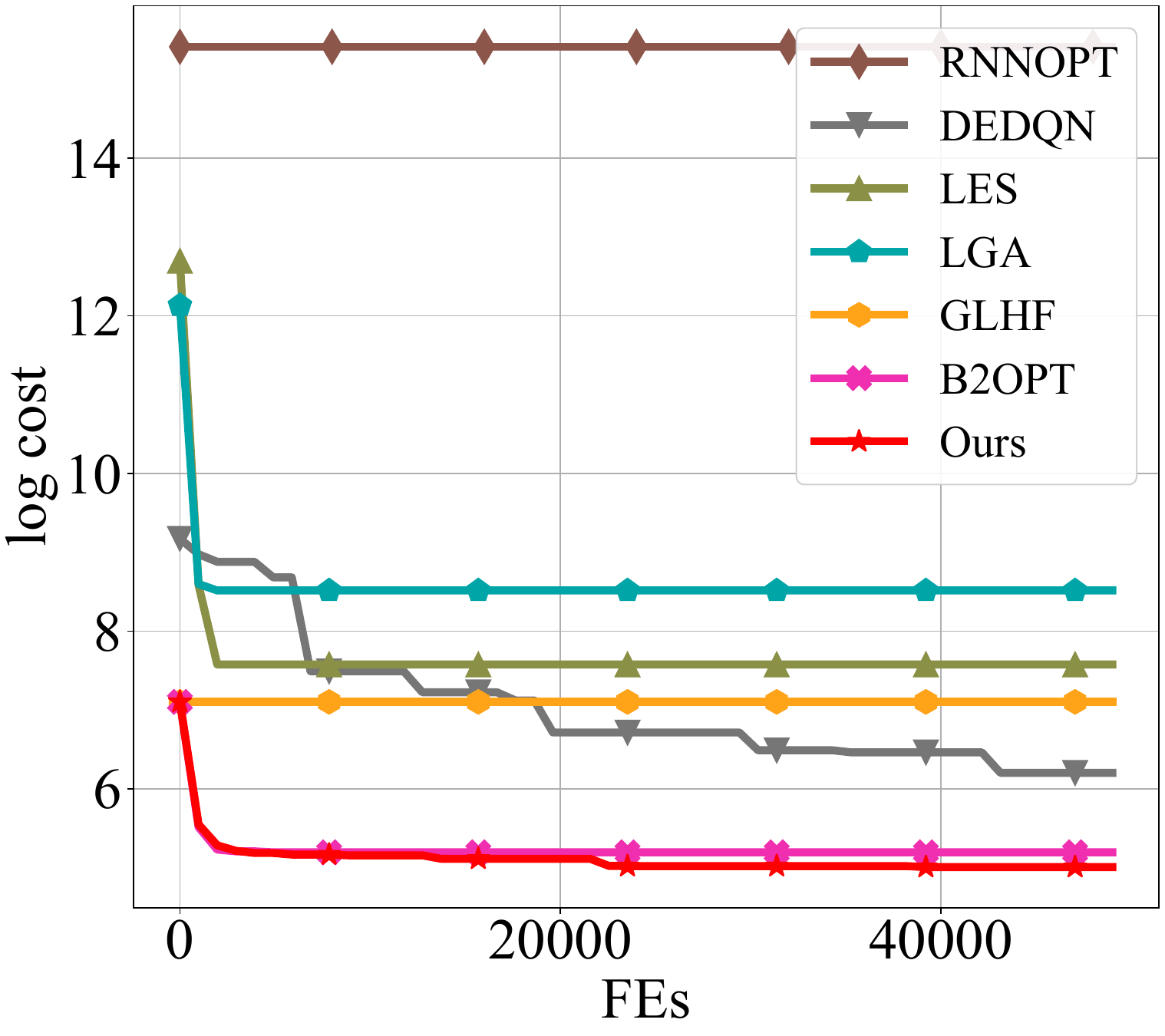}
		\caption*{Discus}
	\end{subfigure}
	\begin{subfigure}[b]{0.24\textwidth}
		\includegraphics[width=\linewidth]{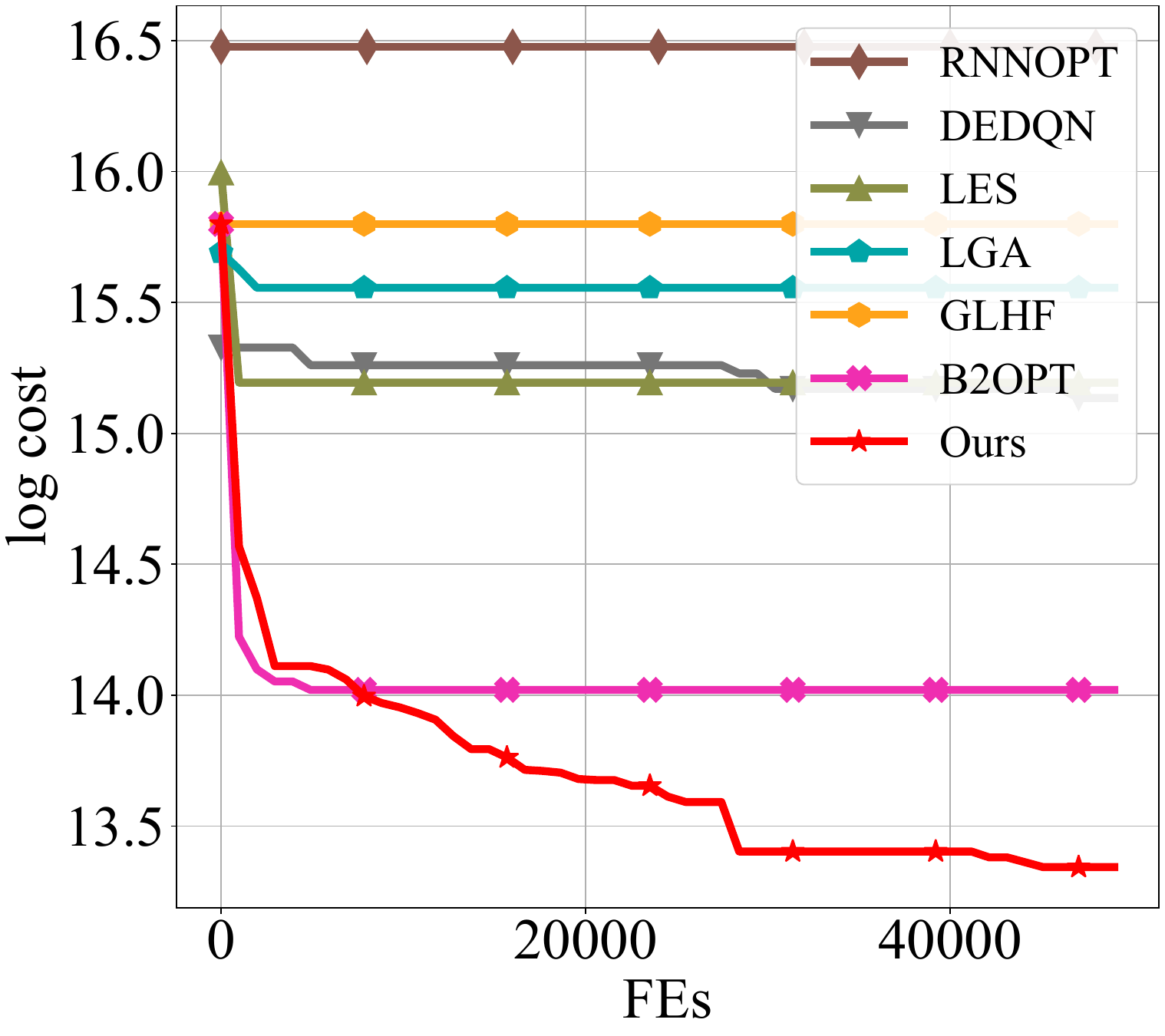}
		\caption*{Ellipsoidal}
	\end{subfigure}
	\begin{subfigure}[b]{0.24\textwidth}
		\includegraphics[width=\linewidth]{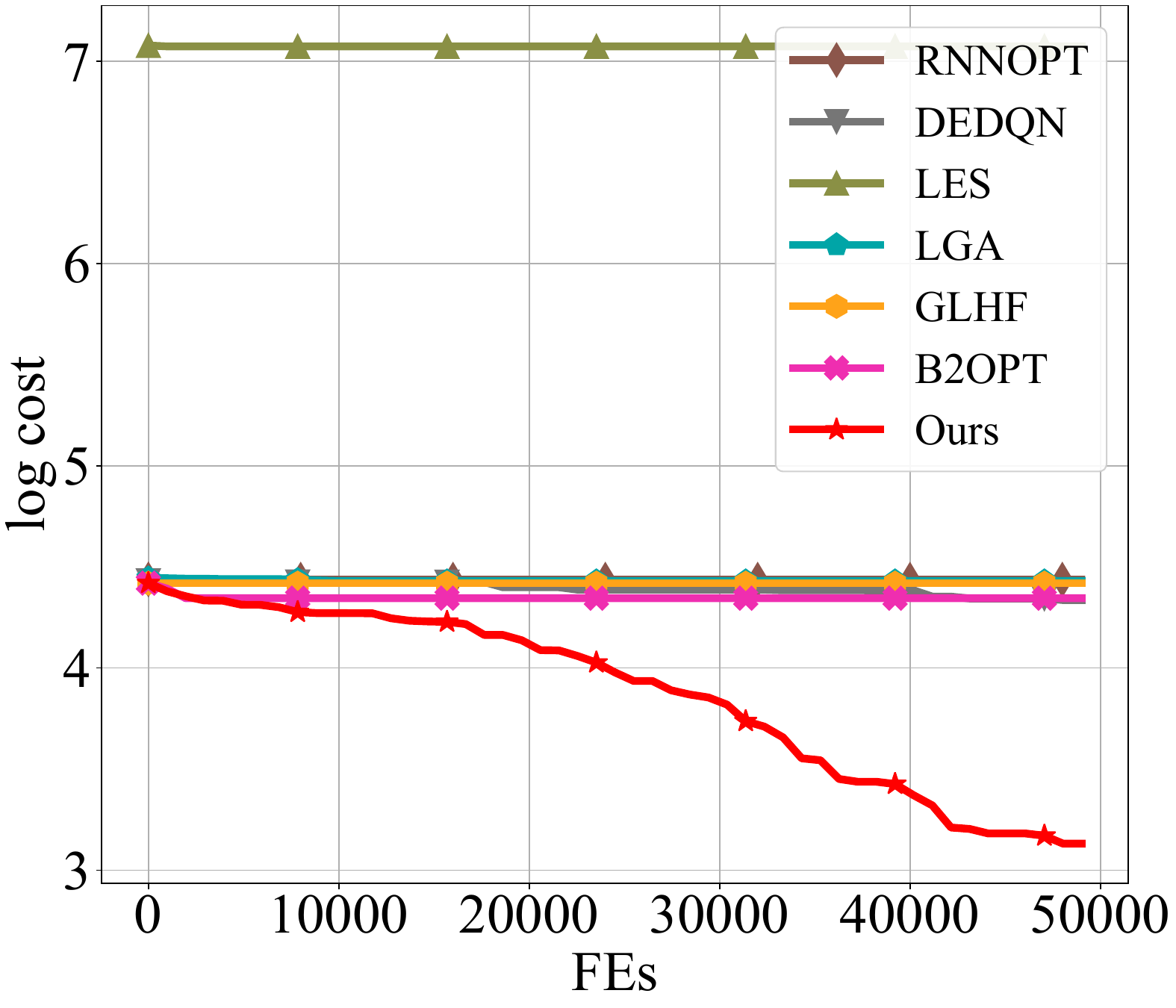}
		\caption*{Gallagher 21Peaks}
	\end{subfigure}
	
	\caption{Log-scaled convergence trajectories of our method and representative methods on the first 8 benchmark tasks from the BBOB-30D suite~\cite{hansen2021coco}. 
		Complete results on all 16 functions are provided in the \textit{supplementary material} (i.e., Fig~\ref{fig:supp-bbob-30d-diff-16}).}
	\label{fig:bbob-30d-diff-16}
\end{figure*}
}

\newcommand{\TabbbobThirty}{%
	\begin{table*}[!tbp]
		\centering
		\scriptsize
		\setlength{\tabcolsep}{4pt}
		\caption{High-Dimensional numerical evaluation (mean $\pm$ std) of representative algorithms on the \emph{LSGO-1000D}. Each entry is calculated from 10 independent test runs. The best and second-best results are highlighted using \underline{\textbf{bold}} and \textbf{bold}, respectively.}
		\label{tab:lsgo}
		\begin{tabular}{lcccccc}
			\toprule
			\textbf{Method} & \texttt{Shifted\_Elliptic} & \texttt{Shifted\_Rastrigin} & \texttt{Shifted\_Ackley} & \texttt{7n1s\_Shifted\_Ell} & \texttt{7n1s\_Shifted\_Ras} & \texttt{7n1s\_Shifted\_Ack} \\
			\midrule 
			\multirow{2}{*}{PSO} & 3.689E+11 & 1.158E+05 & 2.164E+01 & 5.376E+13 & 7.802E+07 & \underline{\textbf{1.067E+06}} \\
			& ($\pm$1.040E+10) & ($\pm$2.302E+03) & ($\pm$7.262E-03) & ($\pm$1.431E+13) & ($\pm$2.797E+06) & ($\pm$1.919E+03) \\
			\multirow{2}{*}{DE} & 3.673E+11 & 1.236E+05 & 2.167E+01 & 3.722E+13 & 7.243E+07 & 1.071E+06 \\
			& ($\pm$5.347E+09) & ($\pm$2.036E+03) & ($\pm$4.901E-03) & ($\pm$9.326E+12) & ($\pm$2.896E+06) & ($\pm$5.270E+02) \\
			\multirow{2}{*}{CMAES} & 2.098E+11 & 6.919E+04 & \underline{\textbf{2.162E+01}} & 1.068E+14 & 5.580E+07 & 1.073E+06 \\
			& ($\pm$7.548E+05) & ($\pm$1.941E+01) & ($\pm$2.688E-03) & ($\pm$3.869E+09) & ($\pm$3.958E+04) & ($\pm$2.152E+02) \\
			\multirow{2}{*}{SAHLPSO} &{\textbf{1.877E+11}} & \textbf{5.983E+04} & \underline{\textbf{2.162E+01}} & \textbf{1.511E+13} & \underline{\textbf{3.510E+07}} & \textbf{1.068E+06} \\
			& ($\pm$1.578E+09) & ($\pm$1.677E+03) & ($\pm$3.670E-02) & ($\pm$9.400E+12) & ($\pm$4.426E+06) & ($\pm$1.274E+03) \\
			\midrule 
			
			\multirow{2}{*}{RNNOPT} & 2.099E+11 & 4.764E+04 & 2.169E+01 & 1.111E+14 & 4.870E+07 & 1.076E+06 \\
			& ($\pm$0.000E+00) & ($\pm$7.276E-12) & ($\pm$0.000E+00) & ($\pm$0.000E+00) & ($\pm$0.000E+00) & ($\pm$0.000E+00) \\
			\multirow{2}{*}{DEDQN} & 3.585E+11 & 1.182E+05 & 2.166E+01 & 3.071E+13 & 6.578E+07 & 1.072E+06 \\
			& ($\pm$1.908E+10) & ($\pm$3.626E+03) & ($\pm$9.260E-03) & ($\pm$1.101E+13) & ($\pm$6.791E+06) & ($\pm$1.731E+03) \\
			\multirow{2}{*}{LES} & 1.961E+11 & 4.707E+04 & 2.164E+01 & 2.778E+13 & 4.179E+07 & 1.069E+06 \\
			& ($\pm$1.948E+09) & ($\pm$2.559E+02) & ($\pm$7.111E-03) & ($\pm$9.982E+12) & ($\pm$1.164E+06) & ($\pm$7.265E+02) \\
			\multirow{2}{*}{LGA} & 4.042E+11 & 1.262E+05 & 2.164E+01 & 5.189E+13 & 8.025E+07 & \textbf{1.068E+06} \\
			& ($\pm$2.897E+10) & ($\pm$5.863E+03) & ($\pm$5.471E-03) & ($\pm$2.901E+13) & ($\pm$9.417E+06) & ($\pm$1.616E+03) \\
			\multirow{2}{*}{GLHF} & 3.645E+11 & 1.176E+05 & 2.166E+01 & 3.329E+13 & 6.923E+07 & 1.072E+06 \\
			& ($\pm$1.521E+10) & ($\pm$3.311E+03) & ($\pm$9.850E-03) & ($\pm$1.083E+13) & ($\pm$6.324E+06) & ($\pm$1.332E+03) \\
			\multirow{2}{*}{B2OPT} & 3.564E+11 & 1.152E+05 & 2.166E+01 & 3.589E+13 & 7.028E+07 & 1.071E+06 \\
			& ($\pm$1.644E+10) & ($\pm$4.849E+03) & ($\pm$6.181E-03) & ($\pm$1.583E+13) & ($\pm$4.294E+06) & ($\pm$2.221E+03) \\
			\rowcolor{blue!5} \multirow{1}{*}{Ours} & \underline{\textbf{1.812E+11}} & \underline{\textbf{4.576E+04}} & \textbf{2.163E+01} & \underline{\textbf{8.577E+12}} & \textbf{3.544E+07} & \underline{\textbf{1.067E+06}} \\
			\rowcolor{blue!5} & ($\pm$6.097E+09) & ($\pm$3.792E+02) & ($\pm$1.583E-02) & ($\pm$3.489E+12) & ($\pm$2.315E+06) & ($\pm$1.841E+03) \\
			\bottomrule
		\end{tabular}
	\end{table*}
}

\newcommand{\Figuav}{%
\begin{figure*}[!tbp]
	\centering
	
	\begin{subfigure}[b]{0.24\textwidth}
		\includegraphics[width=\linewidth]{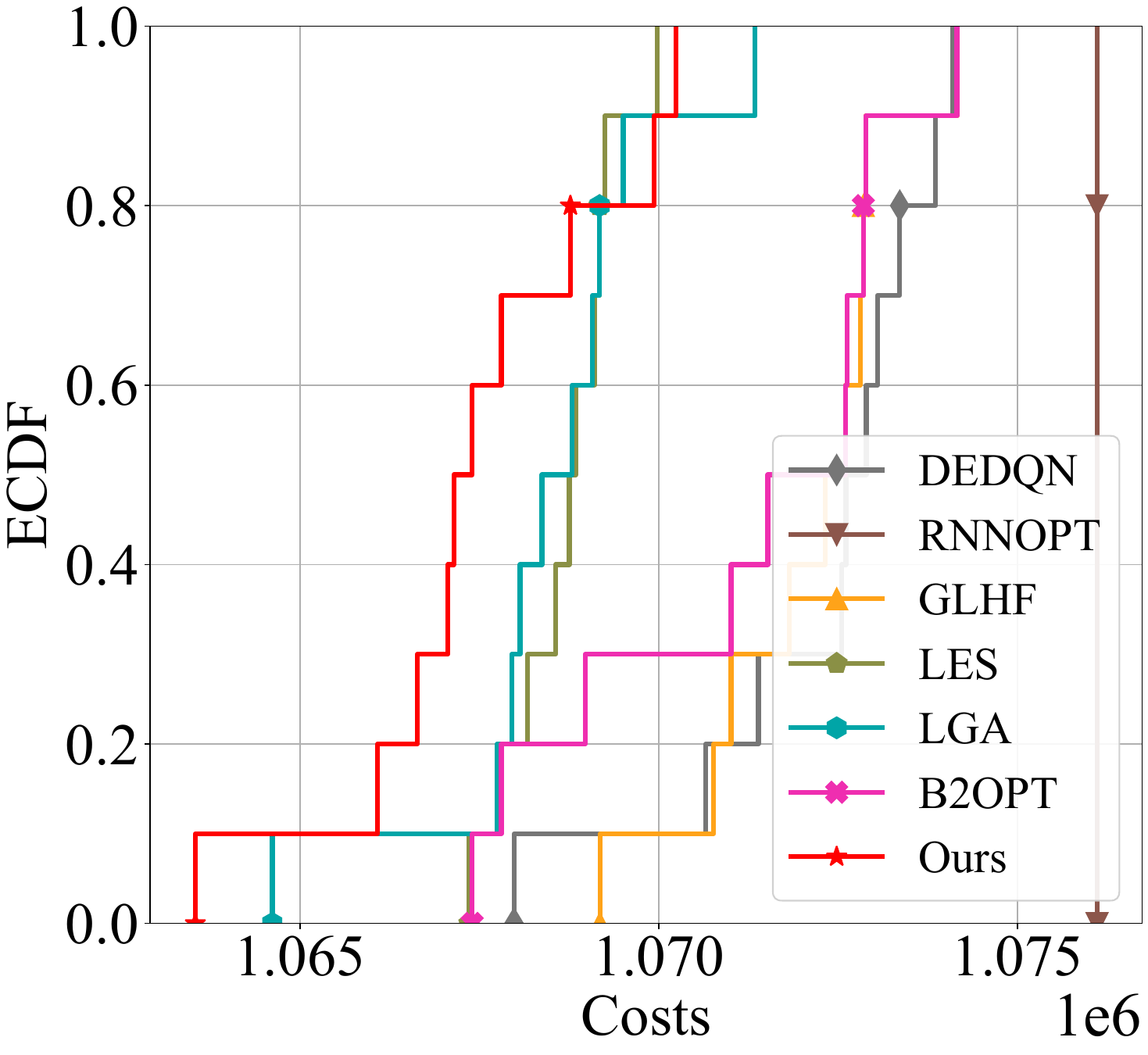}
		\caption*{Rot. Ackley}
	\end{subfigure}
	\begin{subfigure}[b]{0.24\textwidth}
		\includegraphics[width=\linewidth]{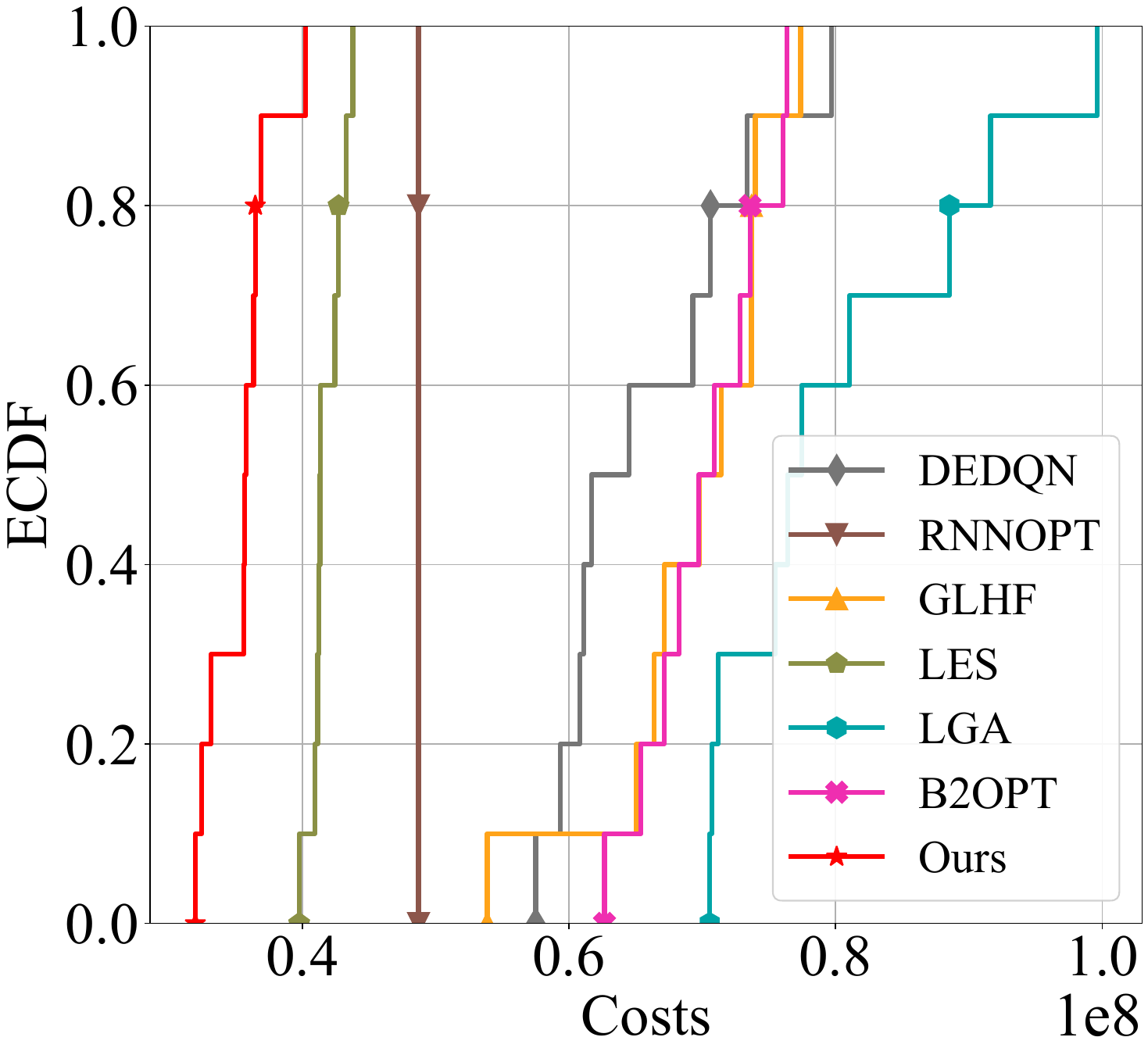}
		\caption*{Rot. Rastrigin}
	\end{subfigure}
	\begin{subfigure}[b]{0.24\textwidth}
		\includegraphics[width=\linewidth]{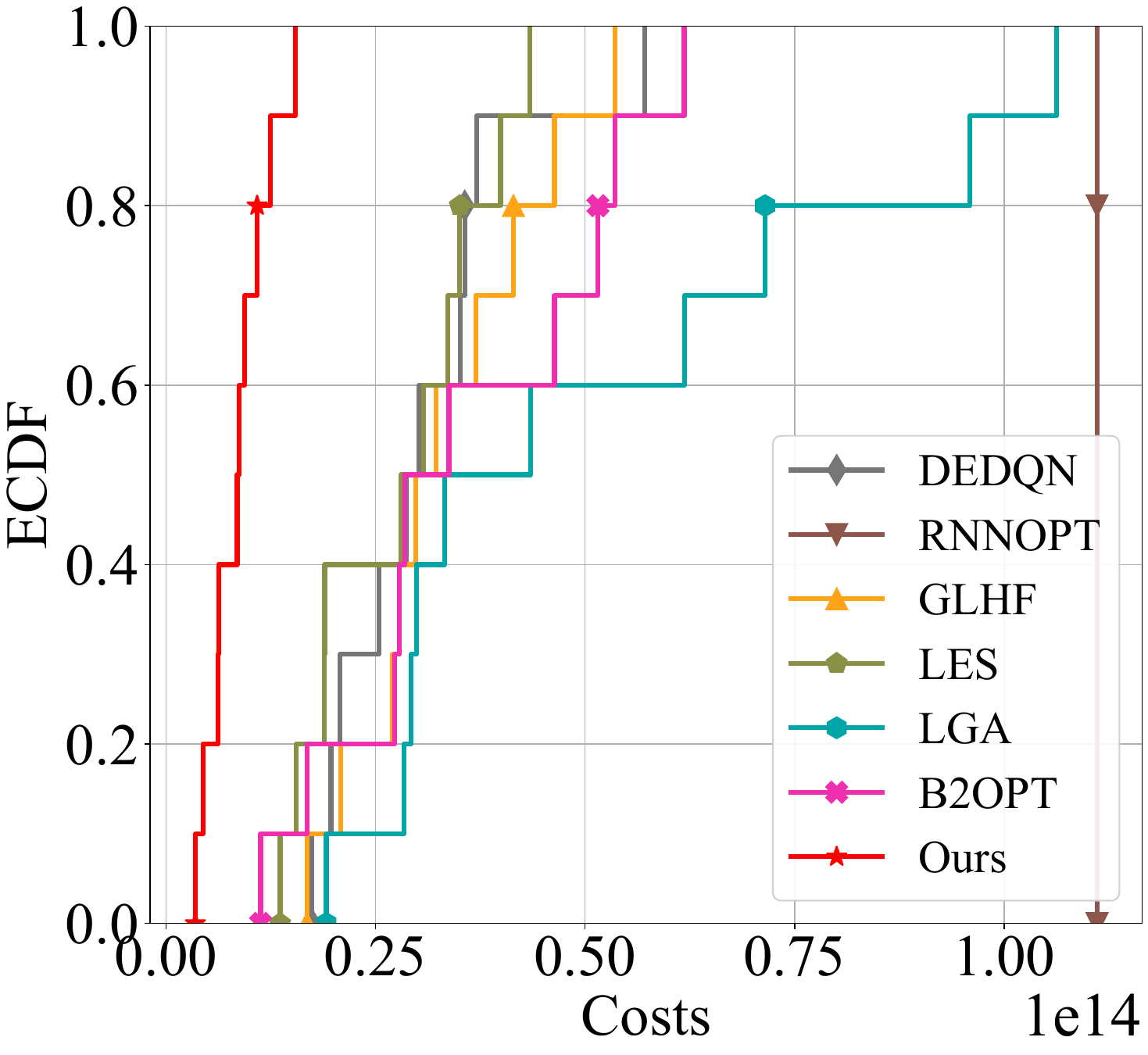}
		\caption*{Rot. Elliptic}
	\end{subfigure}
	\begin{subfigure}[b]{0.24\textwidth}
		\includegraphics[width=\linewidth]{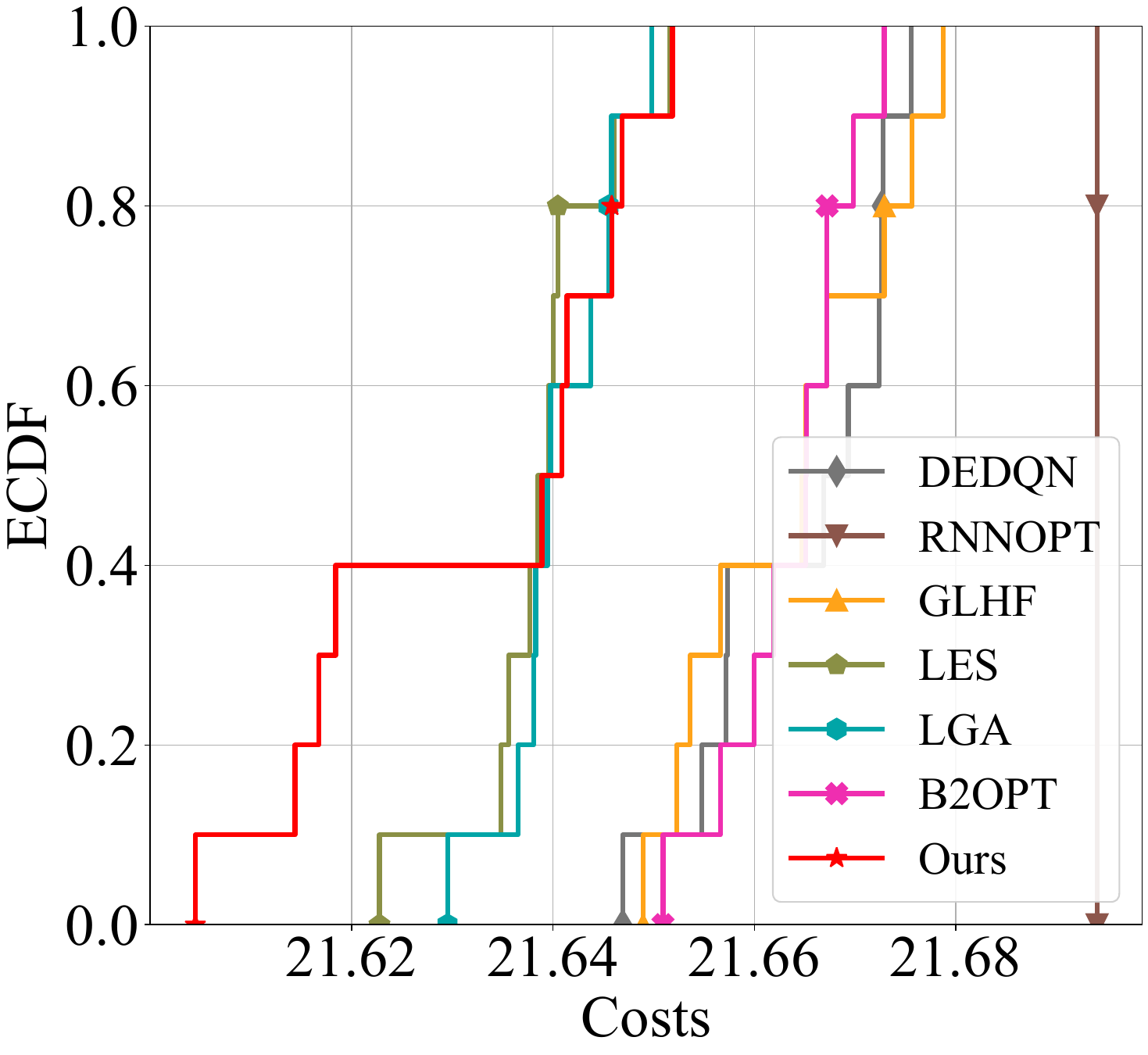}
		\caption*{Shifted Ackley}
	\end{subfigure}
	\caption{Empirical Cumulative Distribution Functions (ECDFs) of different methods on four representative \textit{LSGO-1000D}~\cite{li2013benchmark}.
		The x-axis indicates the number of function evaluations  required to reach a predefined target cost, while the y-axis reflects the cumulative proportion of runs that achieved this target.  Each curve illustrates the probability that a given method solves a task within a certain cost budget.  
		Complete results on all 6 functions are provided in the \textit{supplementary material} (i.e., Fig~\ref{fig:supp-ecdf-lsgo-1000D}).
	}
	\label{fig:ecdf-lsgo-1000D}
\end{figure*}
}

\newcommand{\FigNE}{%
\begin{figure*}[!tbp]
	\centering
	
	\begin{subfigure}[b]{0.24\textwidth}
		\includegraphics[width=\linewidth]{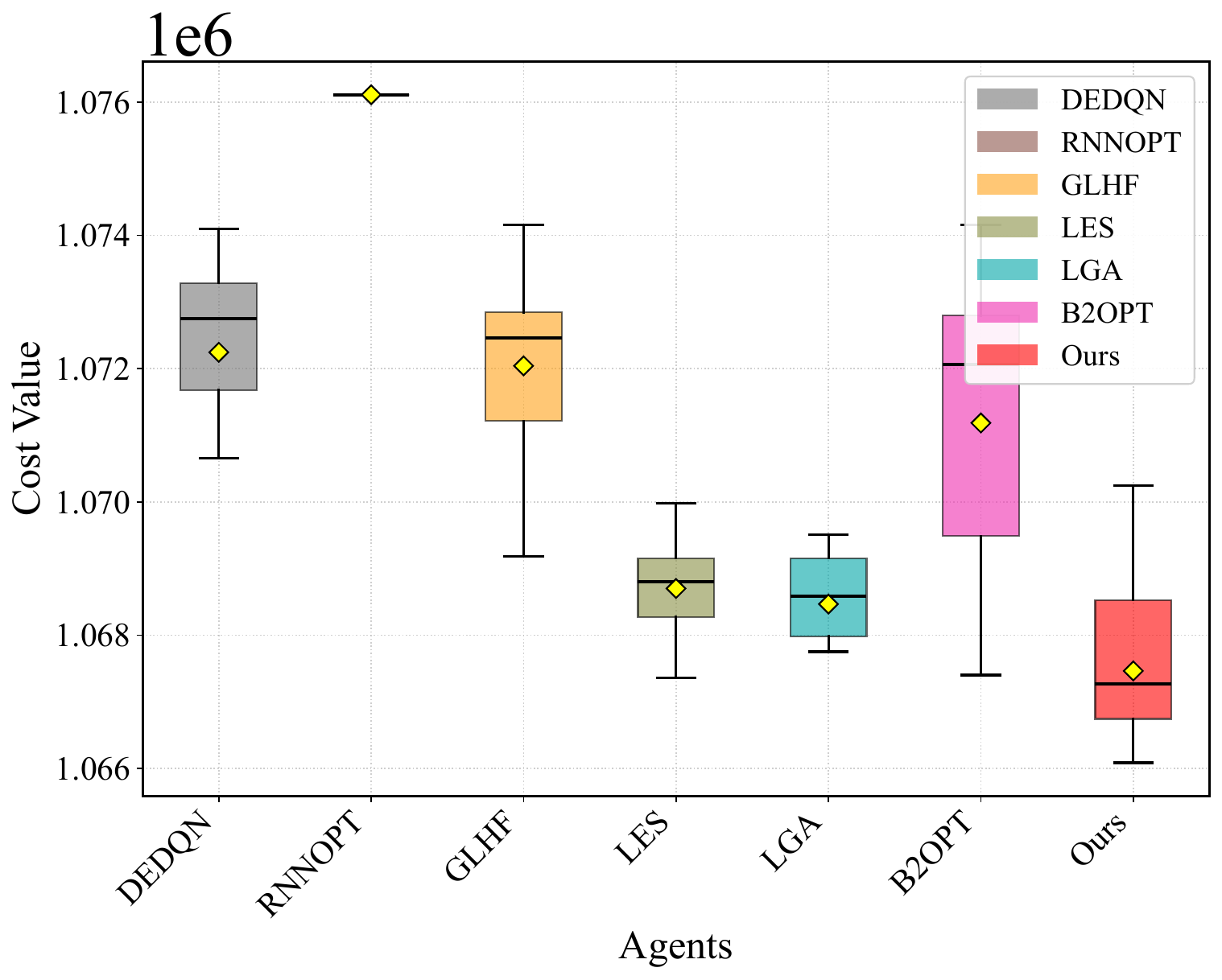}
		\caption*{Rot. Ackley}
	\end{subfigure}
	\begin{subfigure}[b]{0.24\textwidth}
		\includegraphics[width=\linewidth]{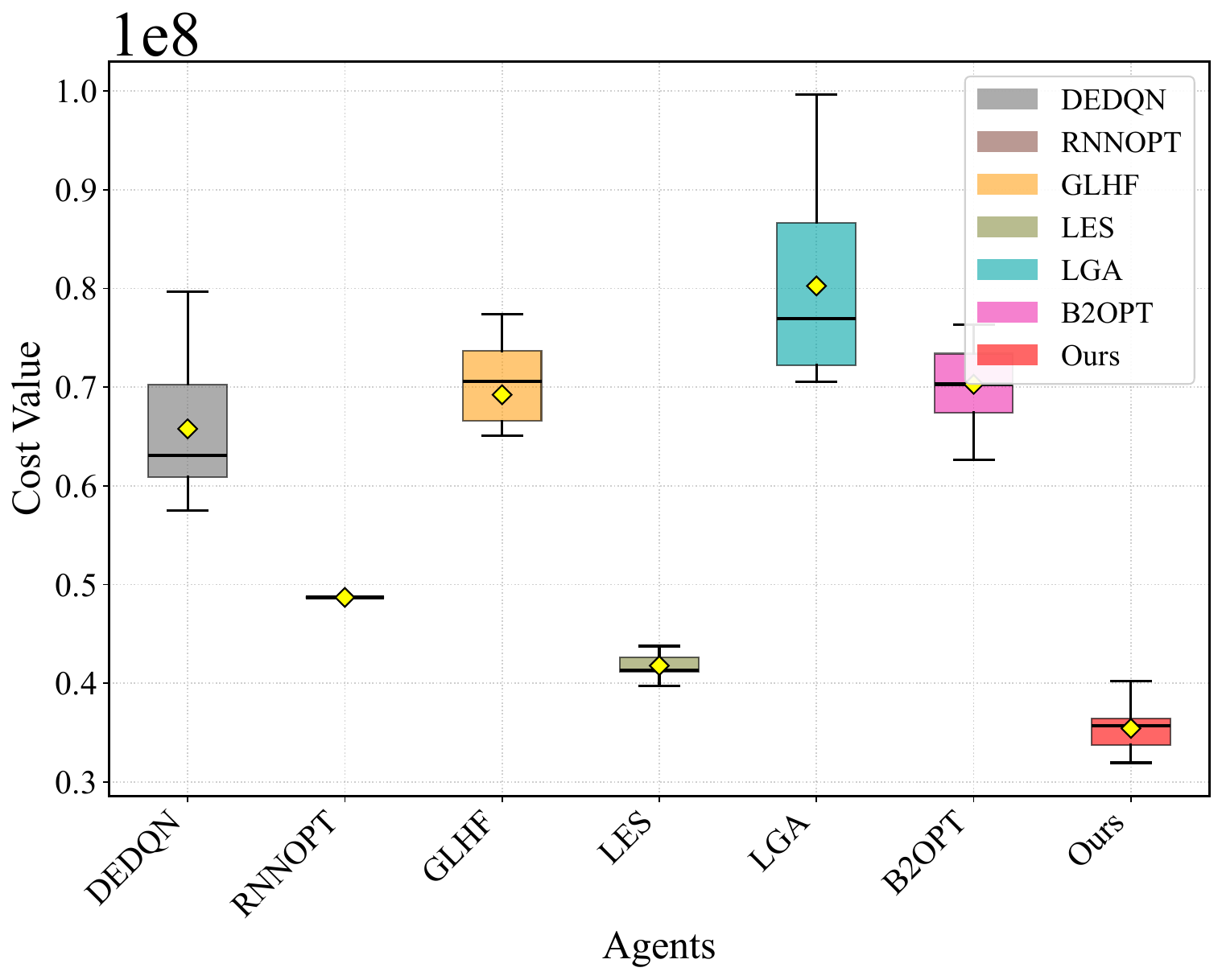}
		\caption*{Rot. Rastrigin}
	\end{subfigure}
	\begin{subfigure}[b]{0.24\textwidth}
		\includegraphics[width=\linewidth]{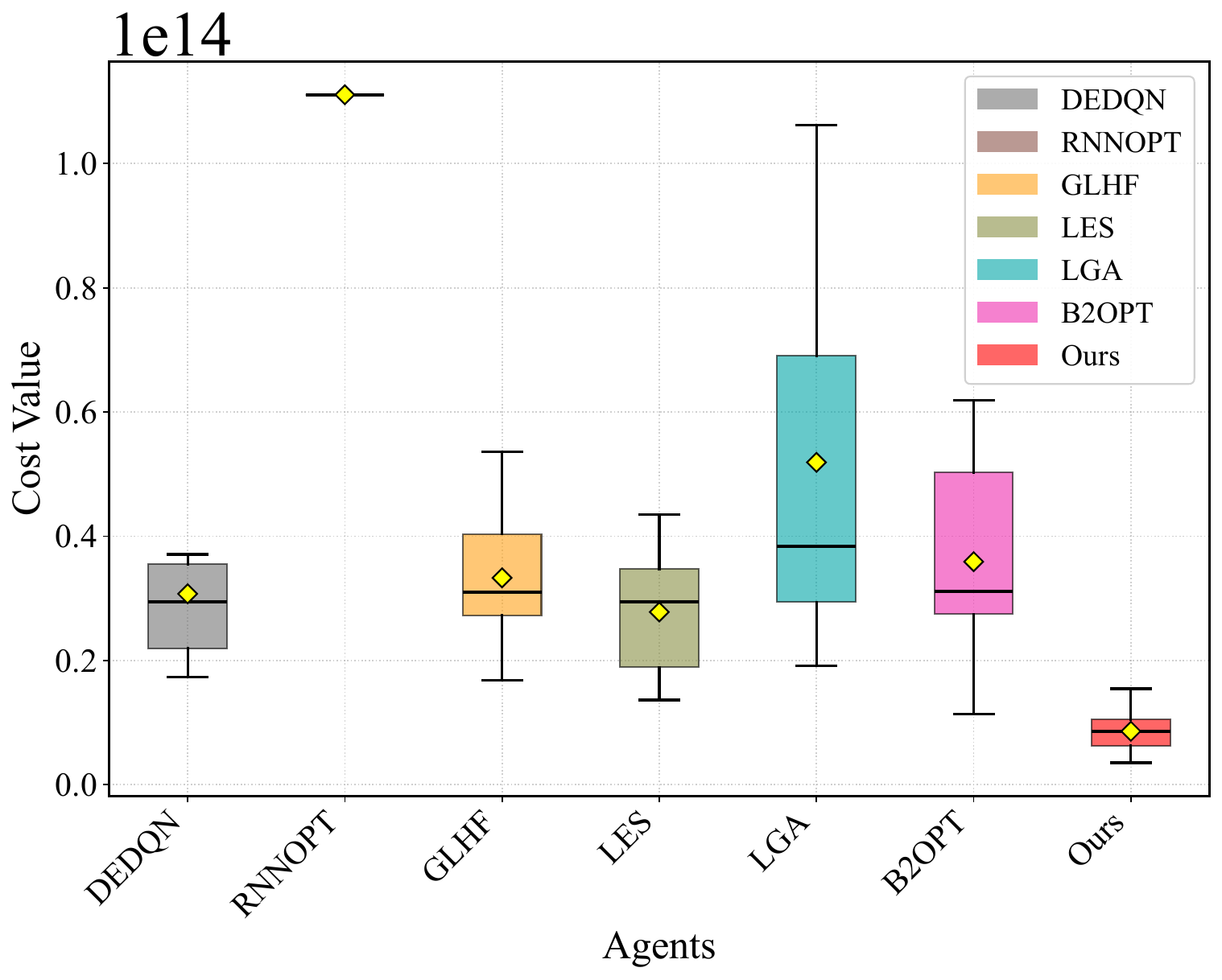}
		\caption*{Rot. Elliptic}
	\end{subfigure}
	\begin{subfigure}[b]{0.24\textwidth}
		\includegraphics[width=\linewidth]{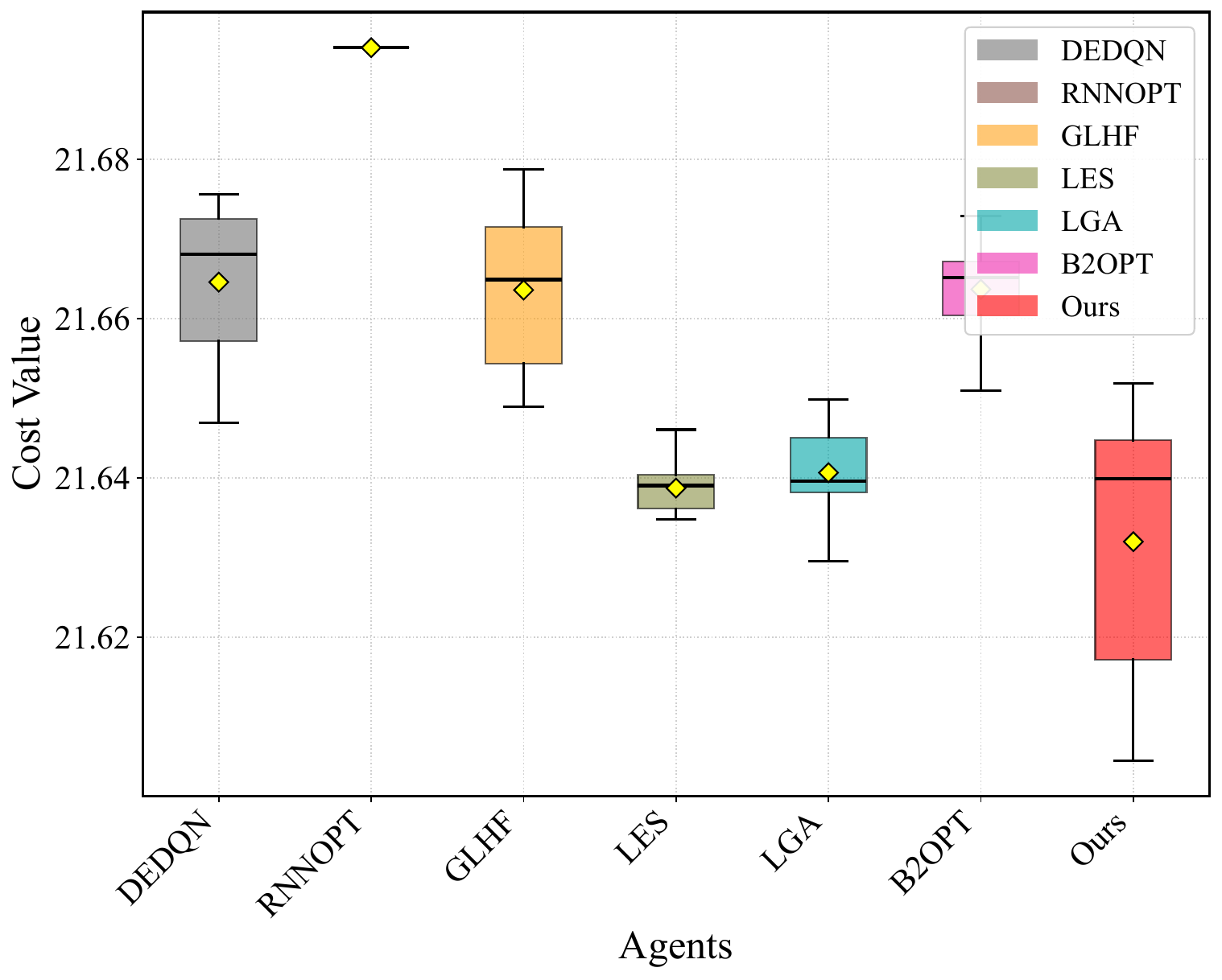}
		\caption*{Shifted Ackley}
	\end{subfigure}
	
	\caption{Boxplots of representative methods on \textit{LSGO-1000D}~\cite{li2013benchmark}.
		Each box shows the distribution of results for one method, where the central line denotes the median, the box boundaries indicate the interquartile range, and whiskers represent the variability across trials. 
		Complete results on all 6 functions are provided in the \textit{supplementary material} (i.e., Fig~\ref{fig:supp-boxplot-lsgo-1000D}).
	}
	\label{fig:boxplot-lsgo-1000D}
\end{figure*}
}

\newcommand{\Figablshare}{%
\begin{figure*}[!tbp]
	\centering
	
	\begin{subfigure}[b]{0.24\textwidth}
		\includegraphics[width=\linewidth]{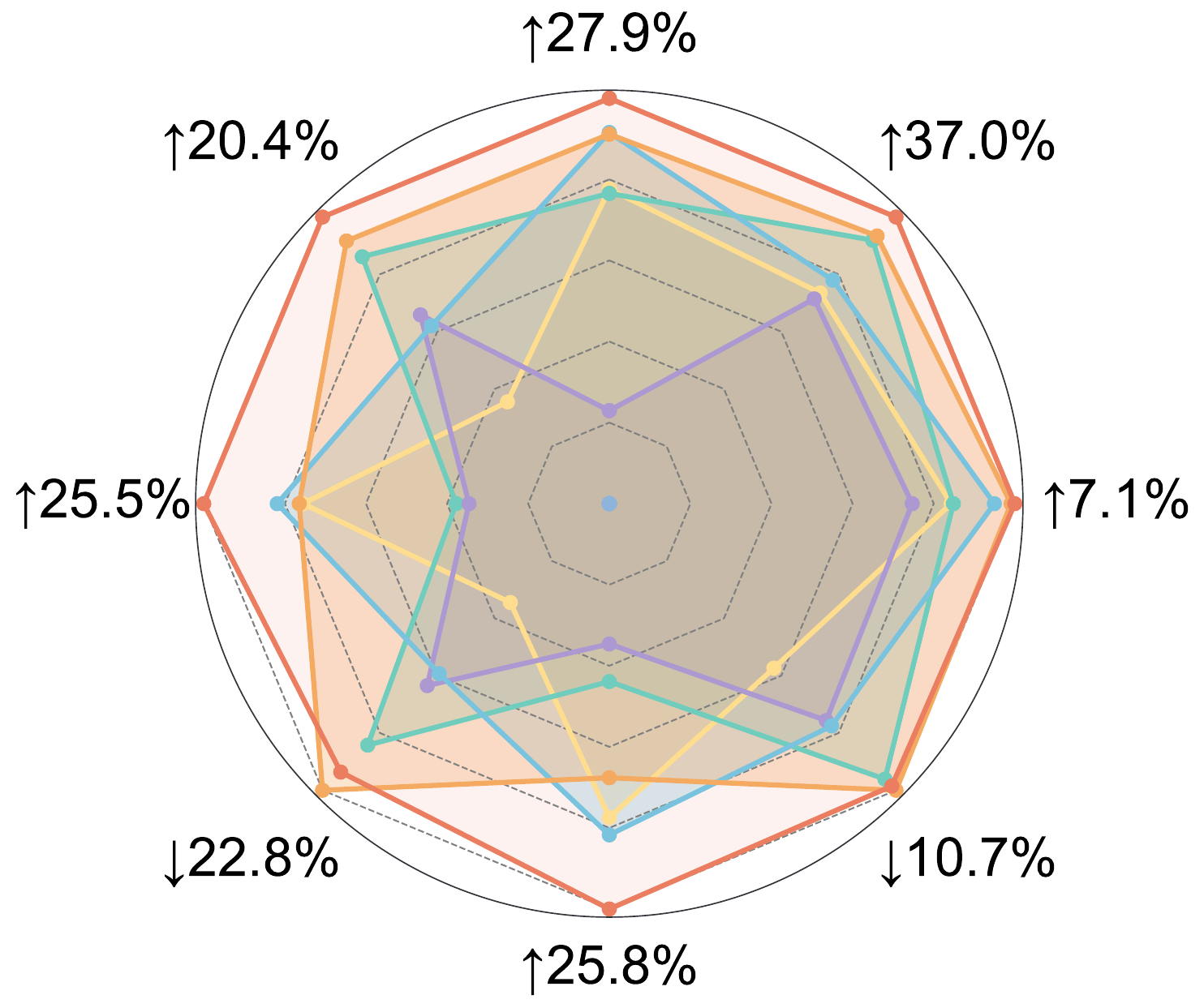}
		\caption*{Terrain 1--8}
	\end{subfigure}
	\begin{subfigure}[b]{0.24\textwidth}
		\includegraphics[width=\linewidth]{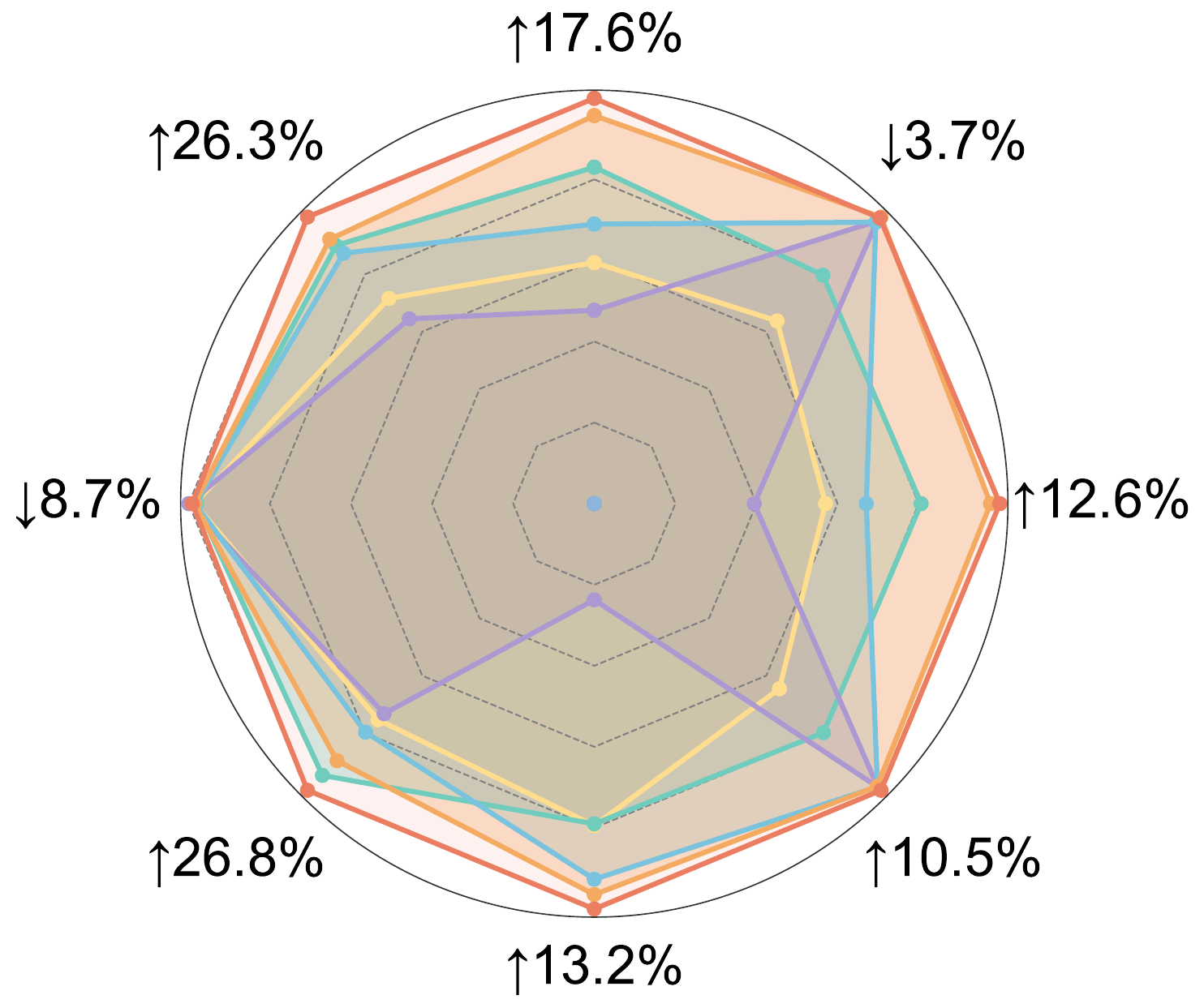}
		\caption*{Terrain 9--16}
	\end{subfigure}
	\begin{subfigure}[b]{0.24\textwidth}
		\includegraphics[width=\linewidth]{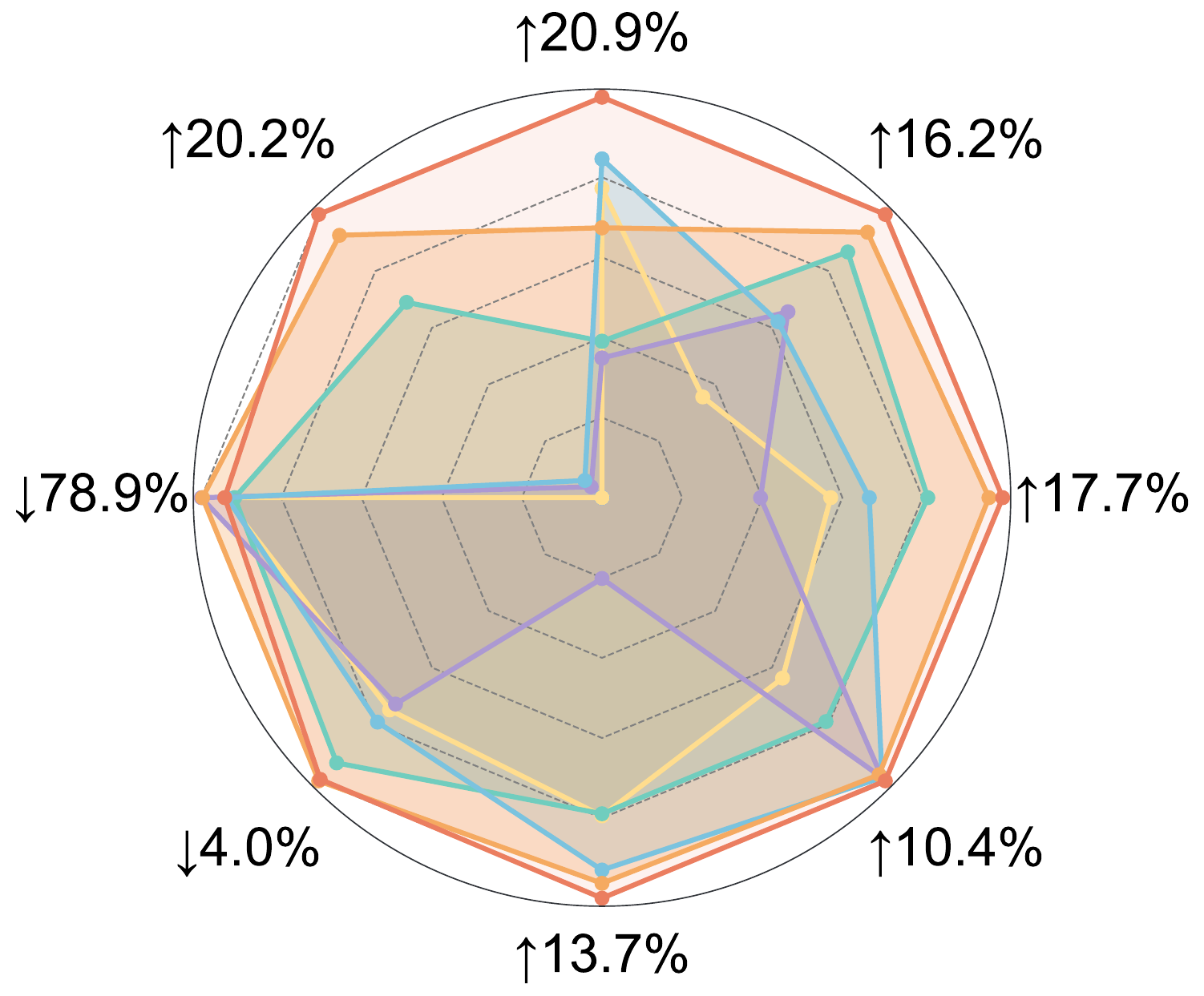}
		\caption*{Terrain 17--24}
	\end{subfigure}
	\begin{subfigure}[b]{0.24\textwidth}
		\includegraphics[width=\linewidth]{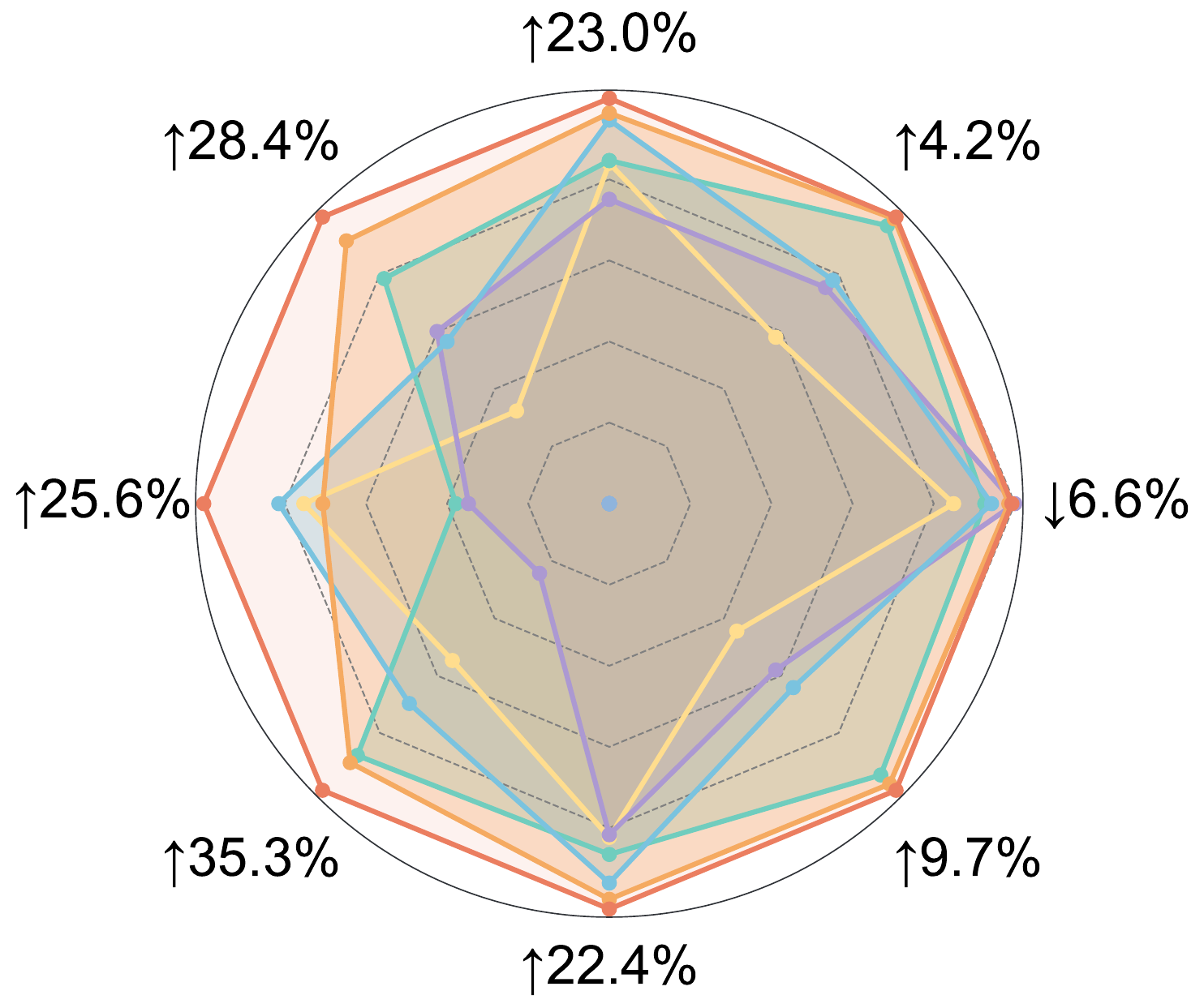}
		\caption*{Terrain 25--32}
	\end{subfigure}
	
	\begin{subfigure}[b]{0.25\textwidth}
		\includegraphics[width=\linewidth]{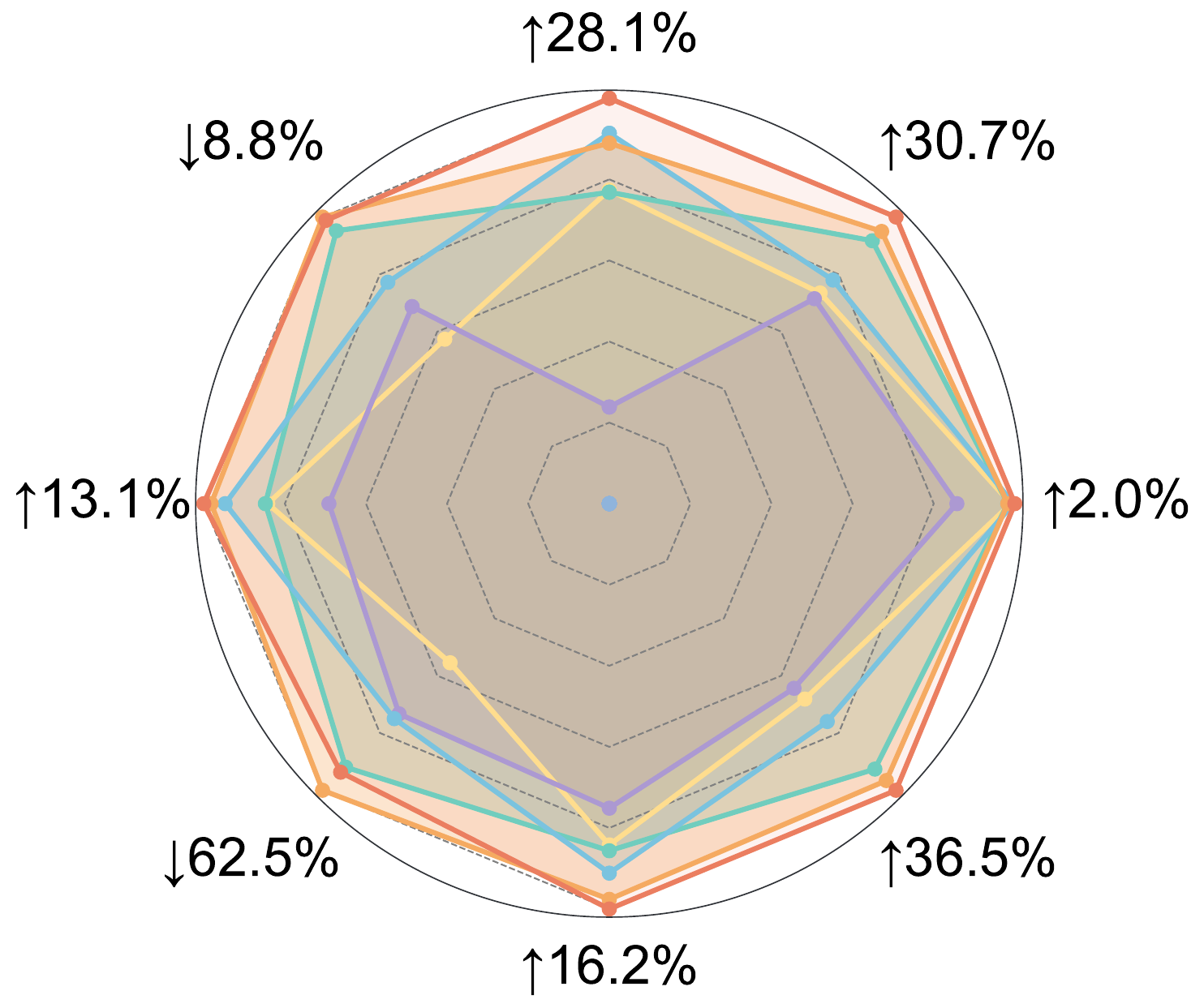}
		\caption*{Terrain 33--40}
	\end{subfigure}
	\begin{subfigure}[b]{0.25\textwidth}
		\includegraphics[width=\linewidth]{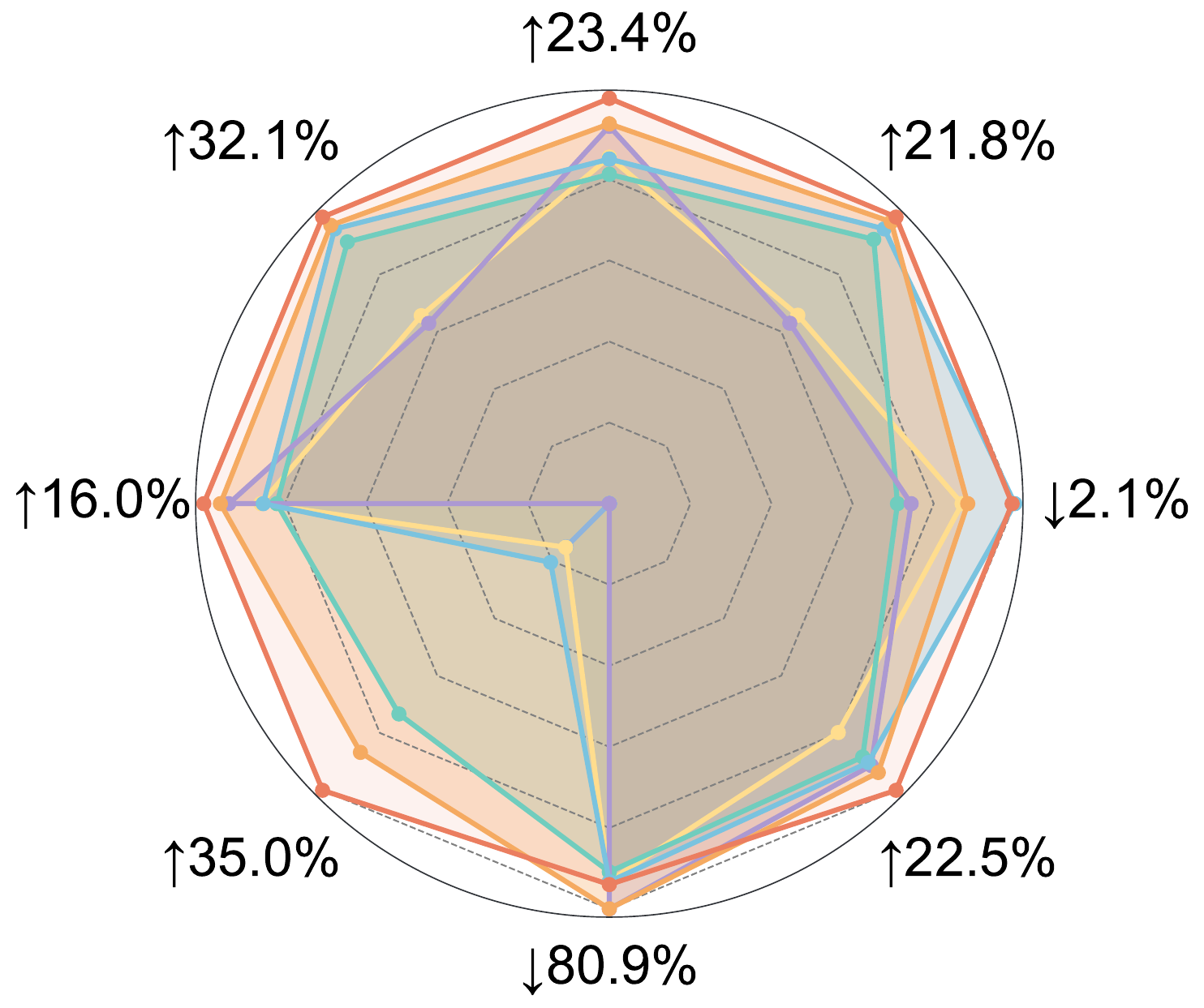}
		\caption*{Terrain 41--48}
	\end{subfigure}
	\begin{subfigure}[b]{0.25\textwidth}
		\includegraphics[width=\linewidth]{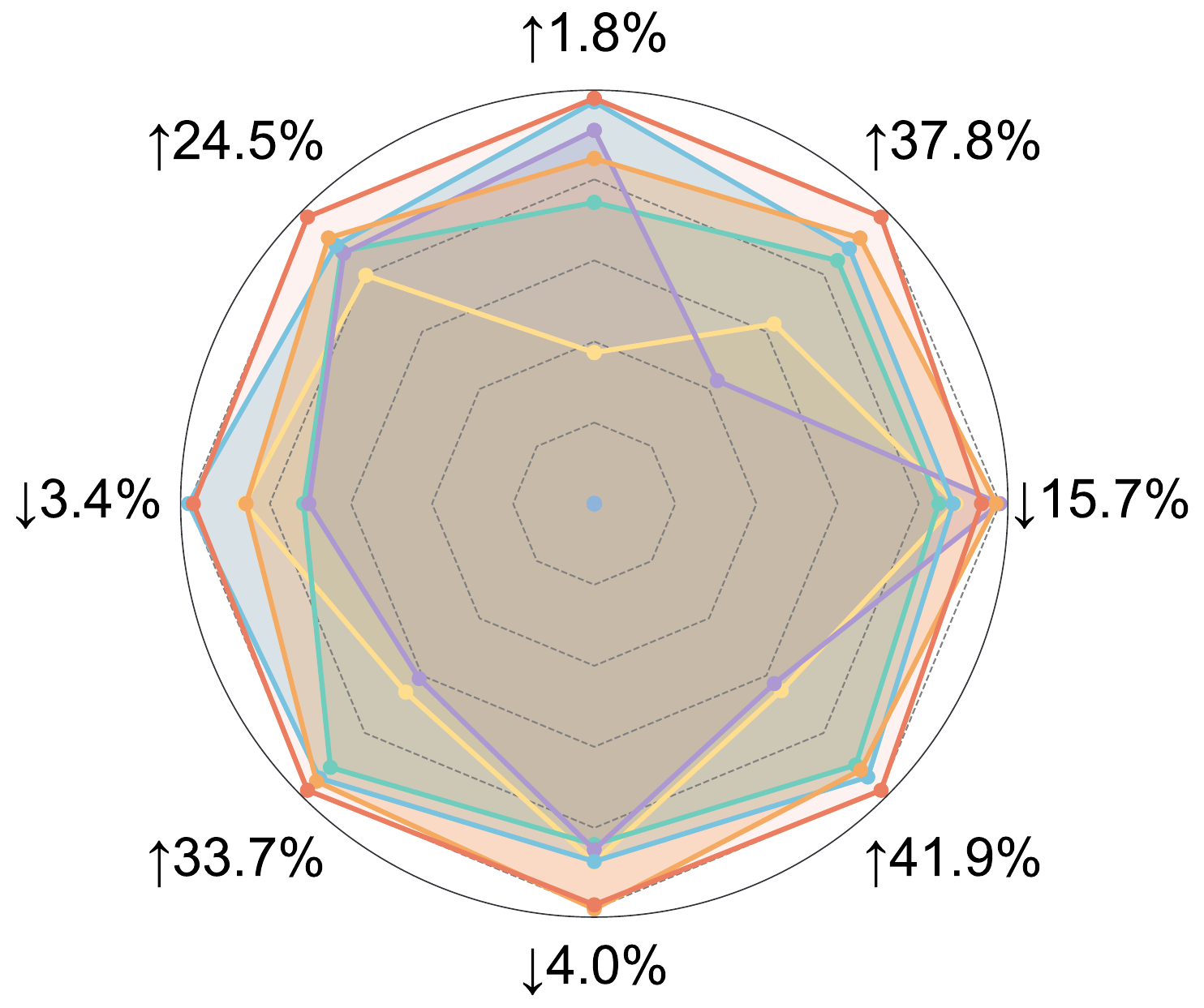}
		\caption*{Terrain 49--56}
	\end{subfigure}
	\begin{subfigure}[b]{0.1\textwidth}
		\includegraphics[width=\linewidth]{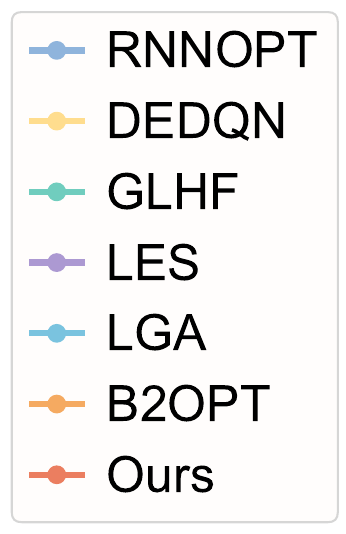}
		\caption*{~}
	\end{subfigure}
	
	\caption{
		Normalized radar plots of representative methods on the  {UAV Path Planning Benchmark}~\cite{shehadeh2025benchmarking}, which consists of 56 realistic planning tasks. The models are trained on \texttt{BBOB-30D} and directly tested in the real-world UAV scenario.  The percentage value marked in each subplot denotes our method's improvement margin over the second-best baseline.
	}
	\label{fig:terrain-radar-all}
\end{figure*}
}

\newcommand{\Figablvariant}{%
\begin{figure*}[!tbp]
	\centering
	\includegraphics[scale=0.87]{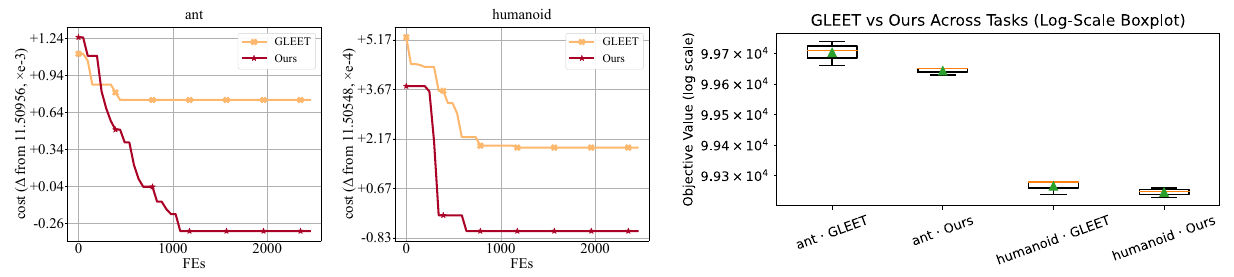} 
	\caption{ 
		Comparison results on the  {Neuroevolution for Robot Control} benchmark~\cite{huang2024evox}, comprising six high-dimensional robot control tasks from two domains: \textit{ant} (ant-3, ant-4, ant-6) and \textit{humanoid} (humanoid-3, humanoid-4, humanoid-5). 
		The left and center plots depict convergence curves on two representative mid-difficulty tasks, where the $y$-axis represents normalized deviations from a task-specific mean, scaled by a shared exponential factor (e.g., $(\Delta\text{cost from }1.15096 \times 10^3$, scaled by $10^{-3})$).
		The right plot shows boxplots of final objective values (log scale) aggregated across all six tasks. 
	}
	\label{fig:ne}
\end{figure*} 
}

\newcommand{\Tabablvariant}{%
\begin{table*}[!tbp]
	\centering
	\scriptsize 
	\caption{Ablation on Evolution Neural Block weight sharing on {BBOB-10D}.  
		Each side aggregates over 16 functions using only the \textbf{Obj} metric (lower is better). Ranks are computed per function across the 8 configurations with average-tie handling. \textit{Std is computed across the 16 functions per configuration using population statistics.} 
		\textbf{EMS} denotes the number of epochs/main steps, computed as $\big\lceil \mathrm{MaxFEs}/\mathrm{NP} \big\rceil$ with $\mathrm{NP}=100$ (i.e., \texttt{ems = (MaxFEs + NP ) // NP - 1}).   Best overall is \textbf{bold}.} 
	\label{tab:ablation-bbob10d-obj-agg-std-14x7}
	\renewcommand{\arraystretch}{1.2}
	\setlength{\tabcolsep}{3pt}  
	\begin{tabular}{rrrrrrr|rrrrrrr}
		\toprule
		\multicolumn{7}{c|}{\textbf{Shared = True}: $\mathcal{O}_{\mathrm{evo}}$ uses \emph{shared meta-parameters} $\bm{\omega}$} & \multicolumn{7}{c}{\textbf{Shared = False}: $\mathcal{O}_{\mathrm{evo}}$ uses \emph{separate per-block meta-parameters} $\bm{\omega}$} \\
		\cmidrule(r){1-7} \cmidrule(l){8-14}
		\textbf{MaxFEs} & \textbf{EMS} & \textbf{Mean $\downarrow$} & \textbf{Median $\downarrow$} & \textbf{GeoMean $\downarrow$} & \textbf{Std Dev $\downarrow$} & \textbf{Avg. Rank $\downarrow$} &
		\textbf{MaxFEs} & \textbf{EMS} & \textbf{Mean $\downarrow$} & \textbf{Median $\downarrow$} & \textbf{GeoMean $\downarrow$} & \textbf{Std Dev $\downarrow$} & \textbf{Avg. Rank $\downarrow$} \\
		\midrule
		2{,}000  & 20  & 2.131E+06 & 1.375E+02 & 4.281E+02 & 8.162E+06 & 8.00 &
		2{,}000  & 20  & 5.442E+05 & 3.706E+01 & 1.106E+02 & 2.088E+06 & 5.06 \\
		5{,}000  & 50  & 5.294E+05 & 4.777E+01 & 1.226E+02 & 2.000E+06 & 5.62 &
		5{,}000  & 50  & 2.800E+05 & 3.091E+01 & 7.729E+01 & 1.070E+06 & 3.31 \\
		10{,}000 & 100 & 8.150E+05 & 4.469E+01 & 1.088E+02 & 3.126E+06 & 5.00 &
		10{,}000 & 100 & 1.350E+05 & 2.768E+01 & 5.521E+01 & 5.139E+05 & 2.00 \\
		20{,}000 & 200 & 9.302E+05 & 4.850E+01 & 1.169E+02 & 3.581E+06 & 6.00 &
		20{,}000 & 200 & \textbf{3.506E+04} & \textbf{2.239E+01} & \textbf{3.879E+01} & \textbf{1.312E+05} & \textbf{1.00} \\
		\bottomrule
	\end{tabular}
\end{table*}
}

\newcommand{\Figablcurve}{%
\begin{figure*}[!tbp]
	\centering
	\begin{subfigure}[b]{0.24\textwidth}
		\includegraphics[width=\linewidth,
		trim=0 0 0 0, clip]{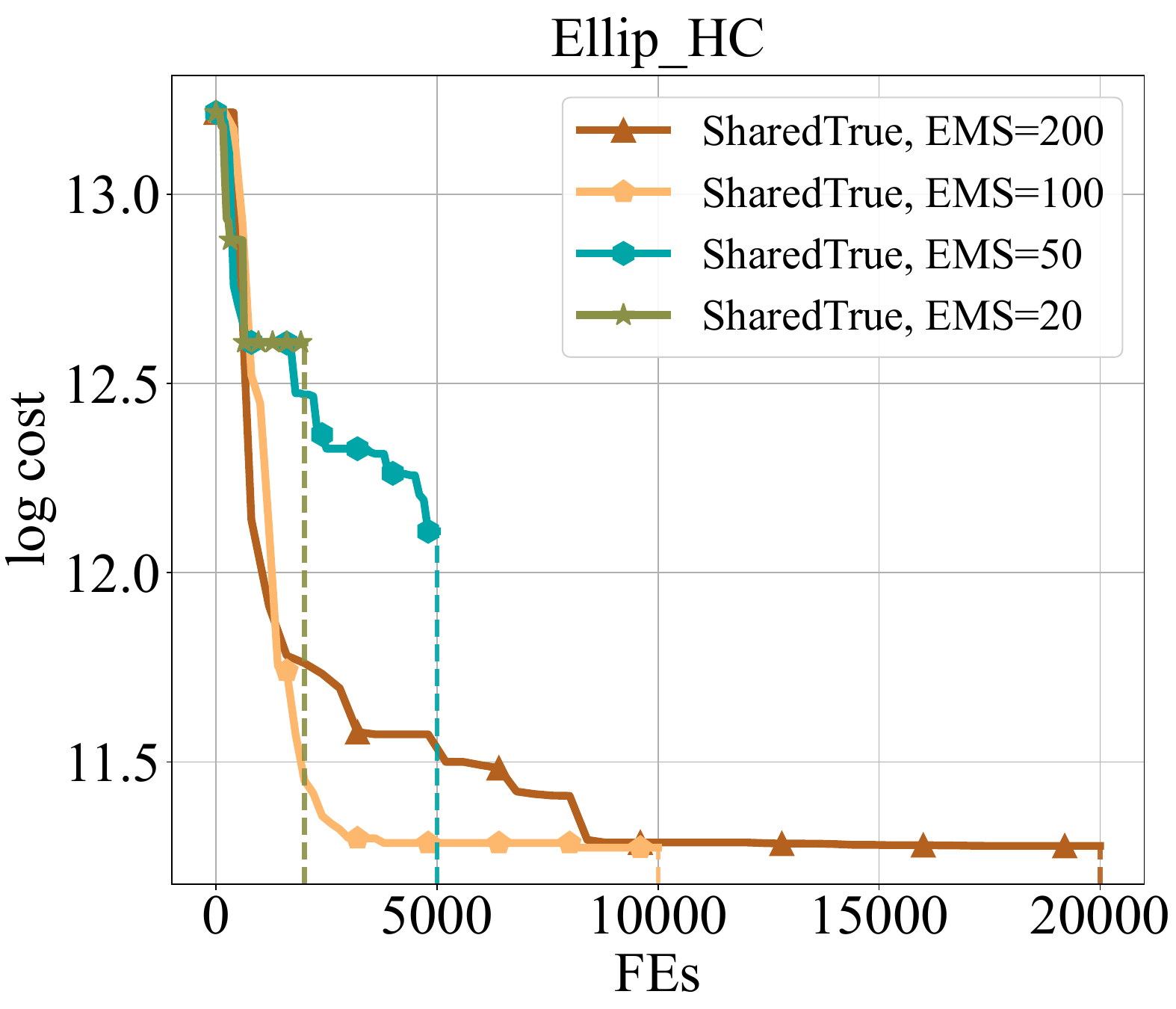}
	\end{subfigure}
	\begin{subfigure}[b]{0.24\textwidth}
		\includegraphics[width=\linewidth,
		trim=0 0 0 0, clip]{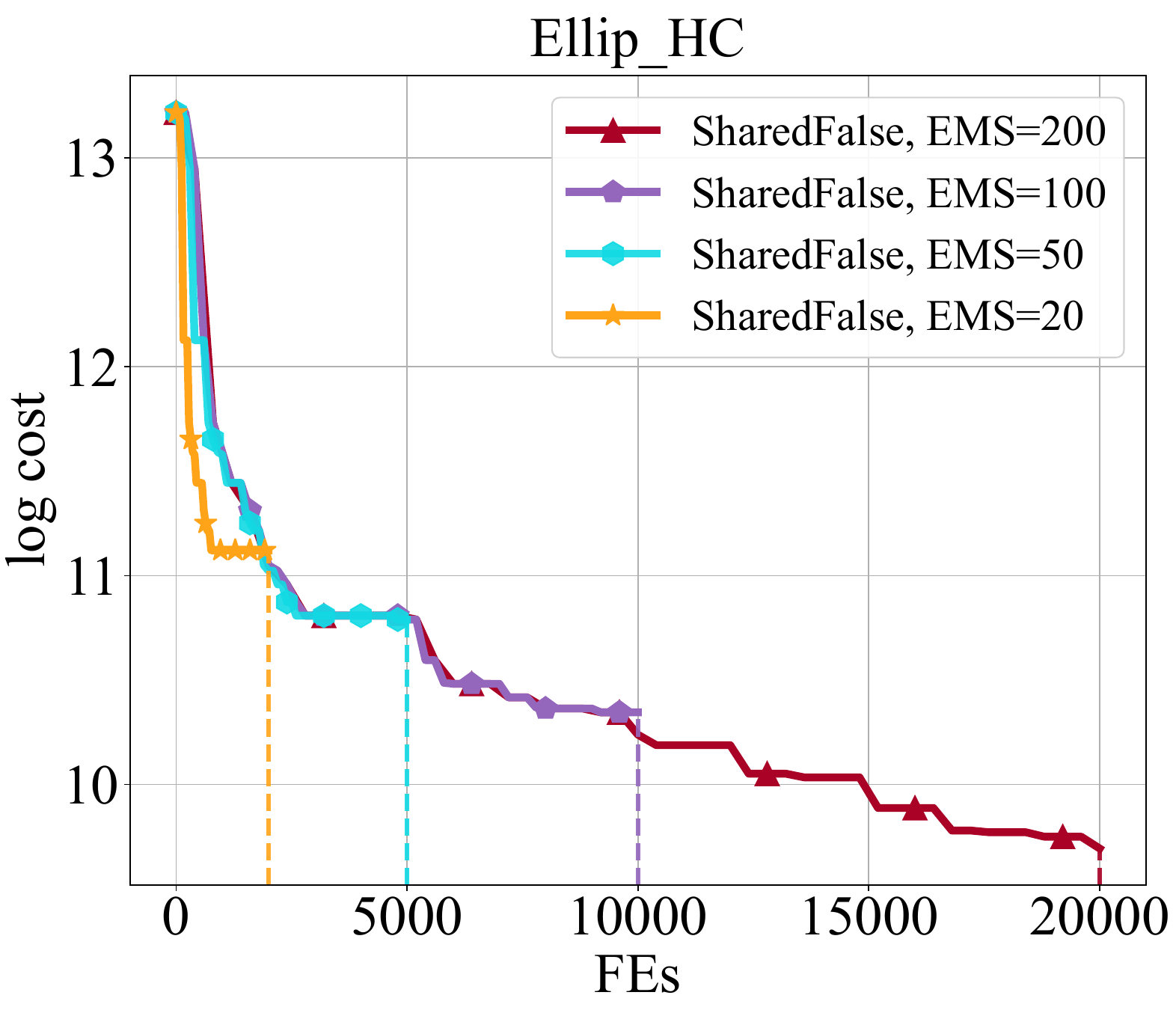}
	\end{subfigure} 
	\begin{subfigure}[b]{0.24\textwidth}
		\includegraphics[width=\linewidth,
		trim=0 0 0 0, clip]{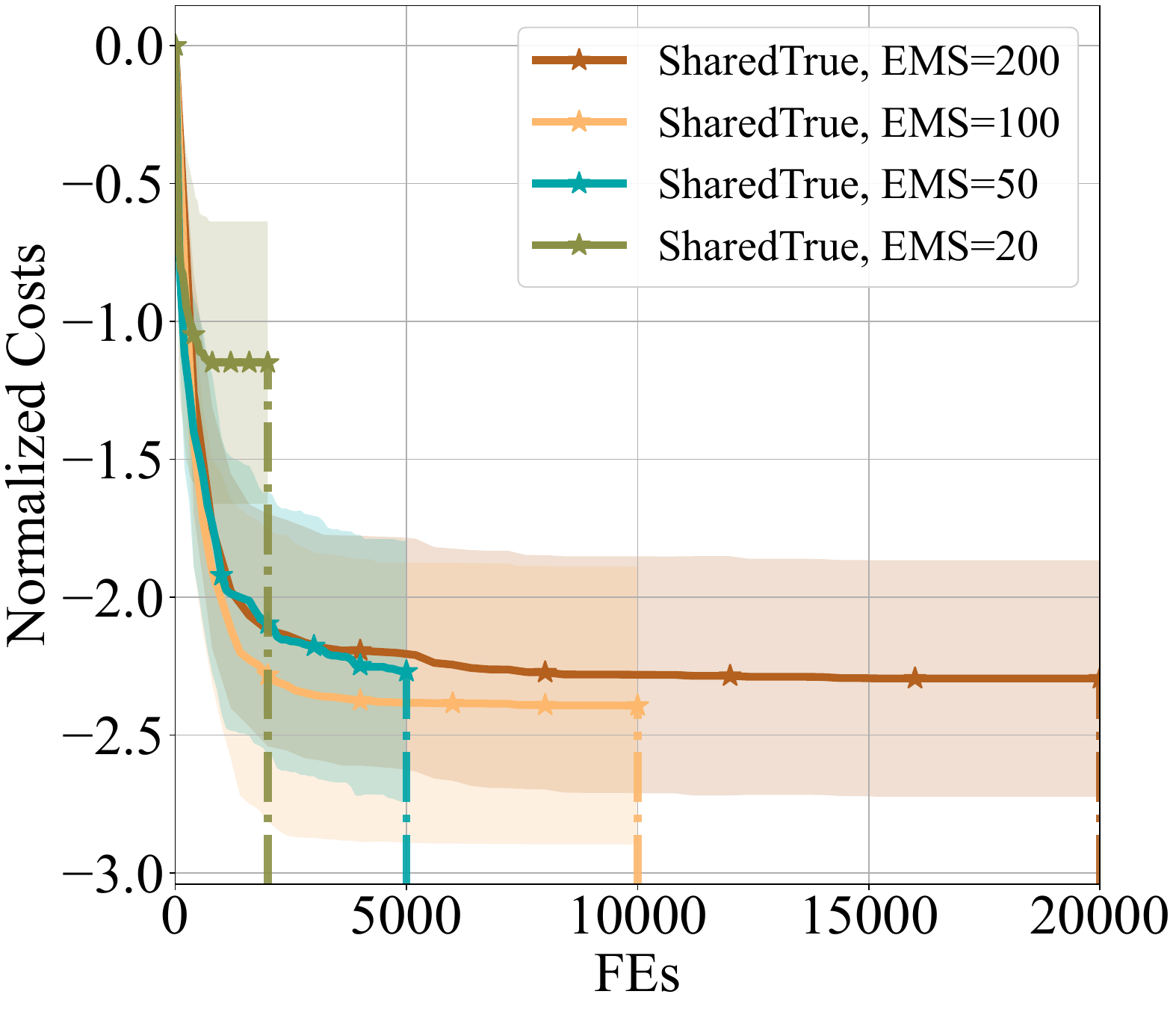}
	\end{subfigure}
	\begin{subfigure}[b]{0.24\textwidth}
		\includegraphics[width=\linewidth,
		trim=0 0 0 0, clip]{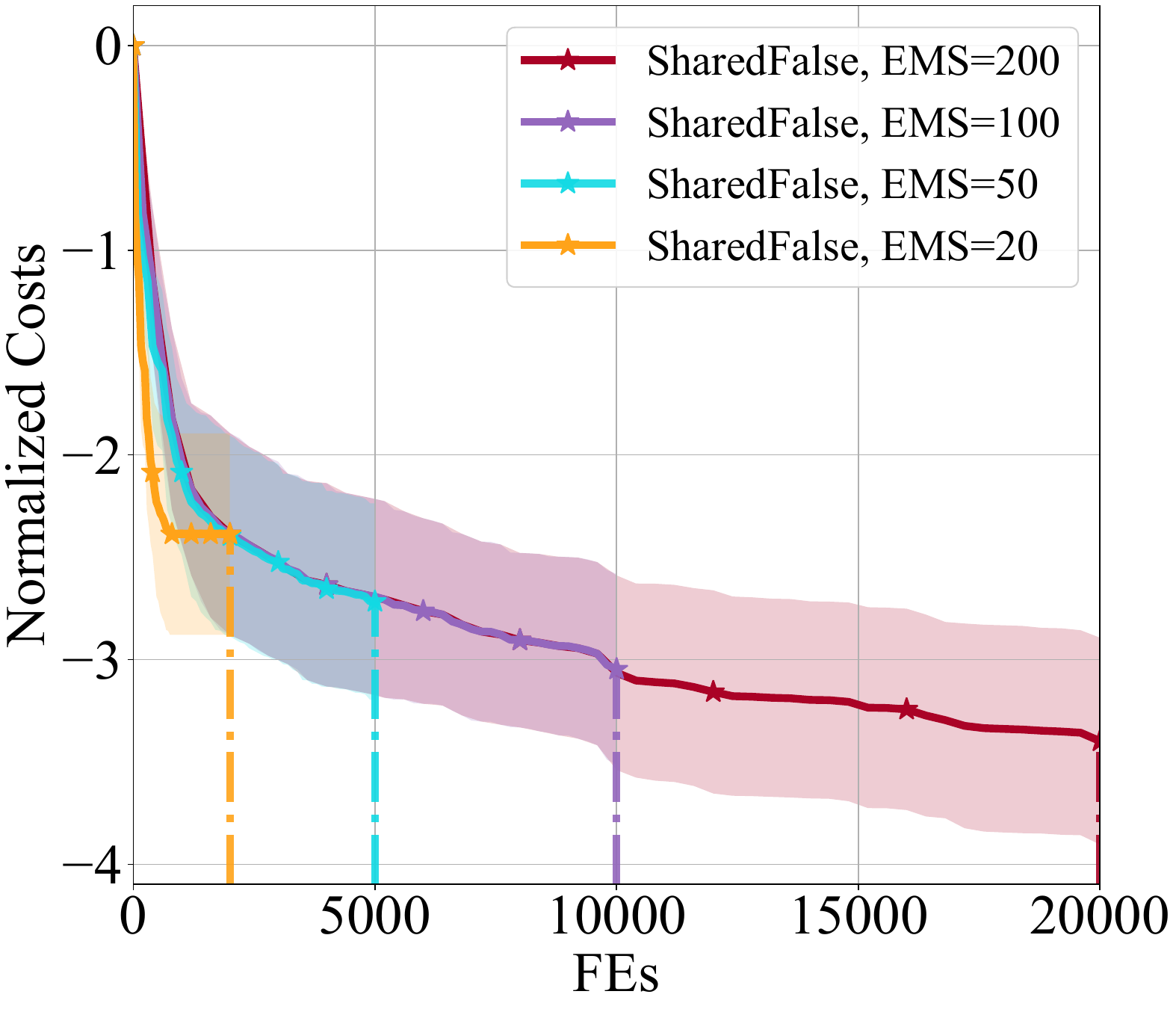}
	\end{subfigure} 
	\caption{
		Convergence behaviors under the \emph{shared} vs.\ \emph{unshared} parameterization of the 
		neural evolution operator~$\mathcal{O}_{\mathrm{evo}}$. 
		Left two:  log-cost convergence on the \textit{Ellip-HC} problem; Right two: aggregated normalized cost curves over all 16 BBOB-like test problems.   
		Additional convergence curves for all problems are provided in the \textit{supplementary material} (i.e., Figs. \ref{fig:abl_shared} and \ref{fig:abl_unshared}).
	}
	\label{fig:abl3}
\end{figure*}
}

\newcommand{\Tabablationobj}{%
	\begin{table}[!tbp]
	\scriptsize
	\centering
	\caption{Overall objective on  {16 BBOB problems}. The first three columns indicate whether each component is enabled (\cmark) or disabled (\xmark). In the \textit{Mean} ($\downarrow$) column, the parentheses report the reduction of  {Ours (Full), Item 4} relative to each ablated variant, computed as $(x-\text{Full})/x$. Best is marked with {\textbf{bold}}.}
	\label{tab:ablation-obj-16}
	\renewcommand{\arraystretch}{1.2}
	\setlength{\tabcolsep}{4pt}
	\begin{tabular}{cccccc}
		\toprule
		\textbf{Item} & \textbf{ProxyGrad} & \textbf{SoftGate} & \textbf{Mamba} & \textbf{Mean ($\downarrow$)} & \textbf{Std Dev} \\
		\midrule
		1       & \xmark & \cmark & \cmark & 5.294E+05 \,\textit{($\downarrow$99.04\%)} & 2.007E+06 \\
		2         & \cmark & \xmark & \cmark & 8.213E+05 \,\textit{($\downarrow$99.38\%)} & 3.150E+06 \\
		3   & \cmark & \cmark & \xmark &  {3.506E+04} \, {\textit{($\downarrow$85.46\%)}} &  {1.312E+05} \\
		4  & \cmark & \cmark & \cmark &  {\textbf{5.098E+03}} \, {\textit{\textbf{(ref.)}}} & {\textbf{1.845E+04}} \\
		\bottomrule
	\end{tabular}
\end{table}
}

\newcommand{\Figablcurvefirsteight}{%
\begin{figure}[!tbp]
	\centering
	\includegraphics[scale=0.41]{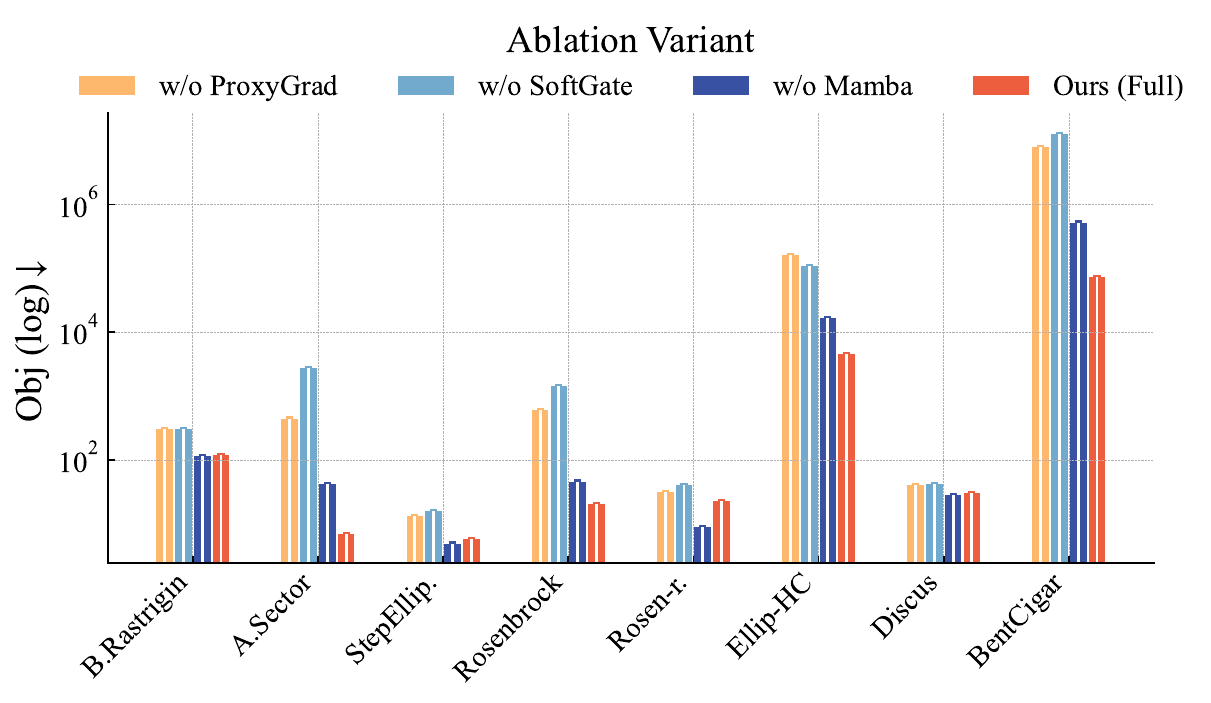}  
	\caption{Log-scale objective values (lower is better) on the first eight BBOB problems under four ablation configurations.
	}
	\label{fig:abl}
\end{figure} 
}

\newcommand{\FigablcurveEllipsoidal}{%
\begin{figure}[!tbp]
	\centering
	\begin{subfigure}[b]{0.24\textwidth}
		\includegraphics[width=\linewidth,
		trim=0 0 0 40, clip]{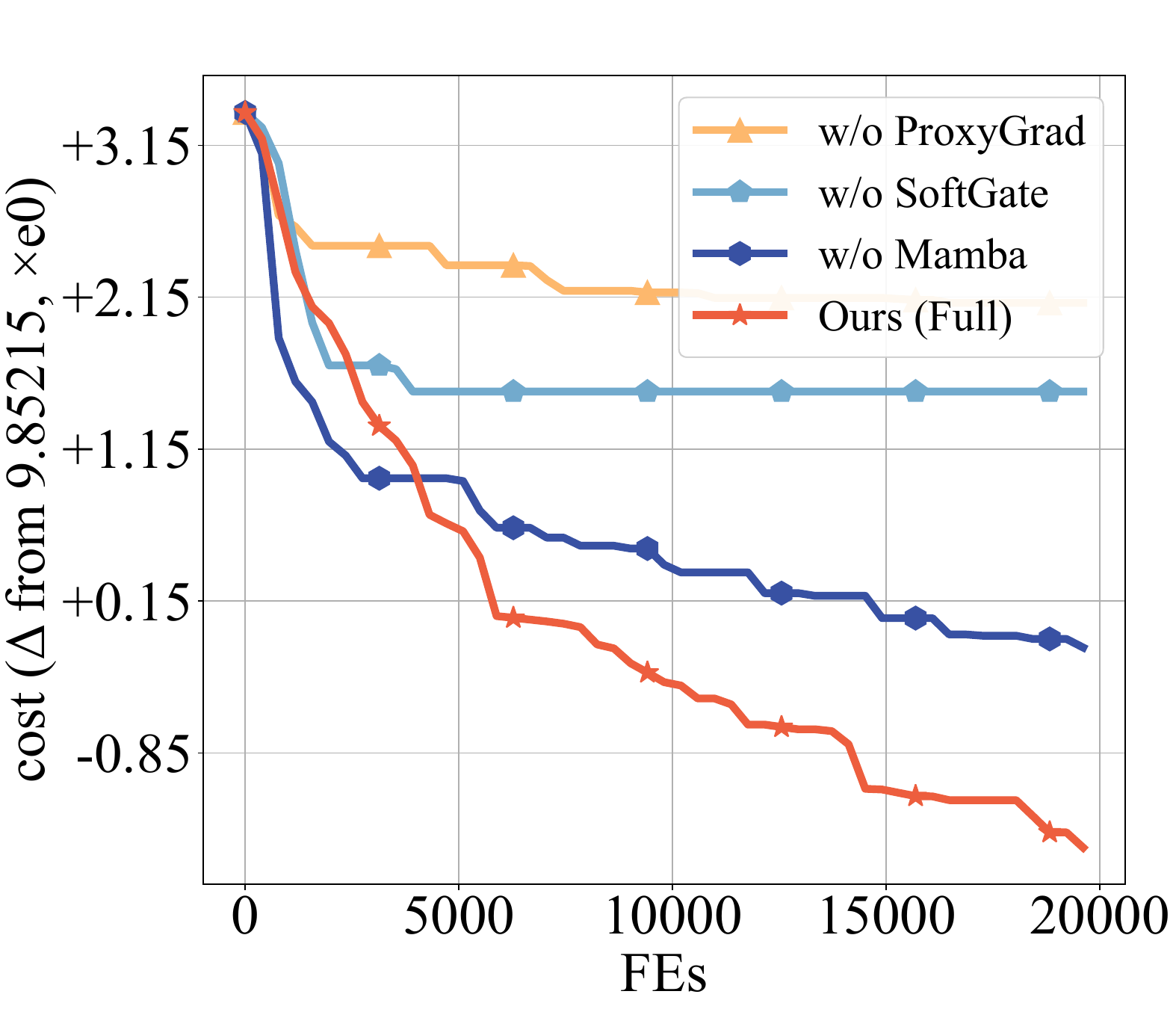}
	\end{subfigure}
	\begin{subfigure}[b]{0.24\textwidth}
		\includegraphics[width=\linewidth,
		trim=0 0 0 0, clip]{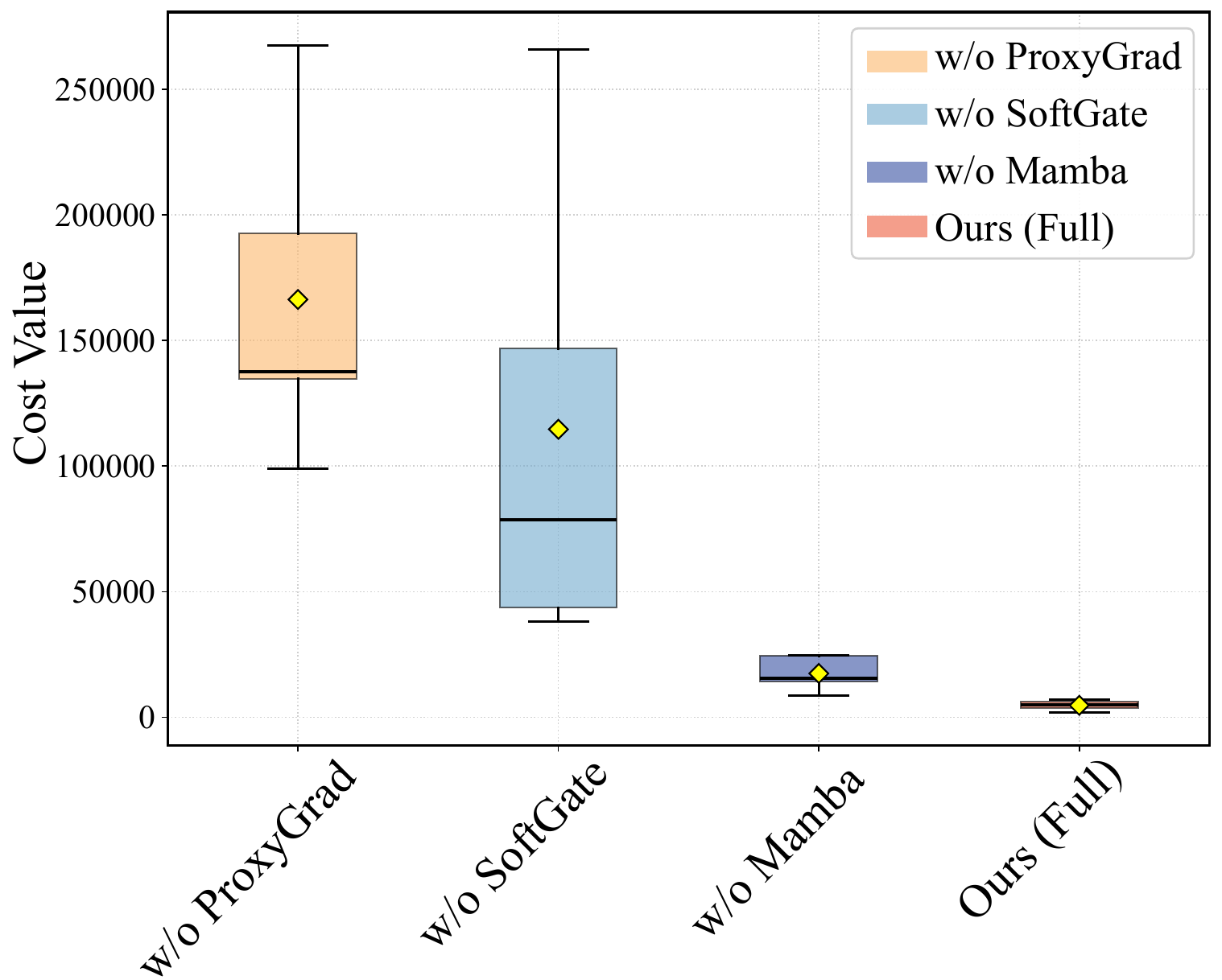}
\end{subfigure} 
\caption{Ablation on convergence and terminal cost across four variants. 
	\textit{Left:} log objective gap vs. function evaluations (FEs) on two representative problems. \textit{Right:} box-whisker statistics of terminal cost across runs. 
	}
\label{fig:abl2}
\end{figure}
}

\newcommand{\FigSupbbobTencurve}{%
\begin{figure*}[htbp]
	\centering
	
	\begin{subfigure}[b]{0.22\textwidth}
		\includegraphics[width=\linewidth]{figures/bbob-10-diff/pics/Attractive_Sector_log_cost_curve.pdf}
		\caption*{Attractive Sector}
	\end{subfigure}
	\begin{subfigure}[b]{0.22\textwidth}
		\includegraphics[width=\linewidth]{figures/bbob-10-diff/pics/Bent_Cigar_log_cost_curve.pdf}
		\caption*{Bent Cigar}
	\end{subfigure}
	\begin{subfigure}[b]{0.22\textwidth}
		\includegraphics[width=\linewidth]{figures/bbob-10-diff/pics/Buche_Rastrigin_log_cost_curve.pdf}
		\caption*{Buche Rastrigin}
	\end{subfigure}
	\begin{subfigure}[b]{0.22\textwidth}
		\includegraphics[width=\linewidth]{figures/bbob-10-diff/pics/Composite_Grie_rosen_log_cost_curve.pdf}
		\caption*{Composite Grie-Rosen}
	\end{subfigure}
	
	\begin{subfigure}[b]{0.22\textwidth}
		\includegraphics[width=\linewidth]{figures/bbob-10-diff/pics/Different_Powers_log_cost_curve.pdf}
		\caption*{Different Powers}
	\end{subfigure}
	\begin{subfigure}[b]{0.22\textwidth}
		\includegraphics[width=\linewidth]{figures/bbob-10-diff/pics/Discus_log_cost_curve.pdf}
		\caption*{Discus}
	\end{subfigure}
	\begin{subfigure}[b]{0.22\textwidth}
		\includegraphics[width=\linewidth]{figures/bbob-10-diff/pics/Ellipsoidal_high_cond_log_cost_curve.pdf}
		\caption*{Ellipsoidal}
	\end{subfigure}
	\begin{subfigure}[b]{0.22\textwidth}
		\includegraphics[width=\linewidth]{figures/bbob-10-diff/pics/Gallagher_21Peaks_log_cost_curve.pdf}
		\caption*{Gallagher 21Peaks}
	\end{subfigure}
	
	\begin{subfigure}[b]{0.22\textwidth}
		\includegraphics[width=\linewidth]{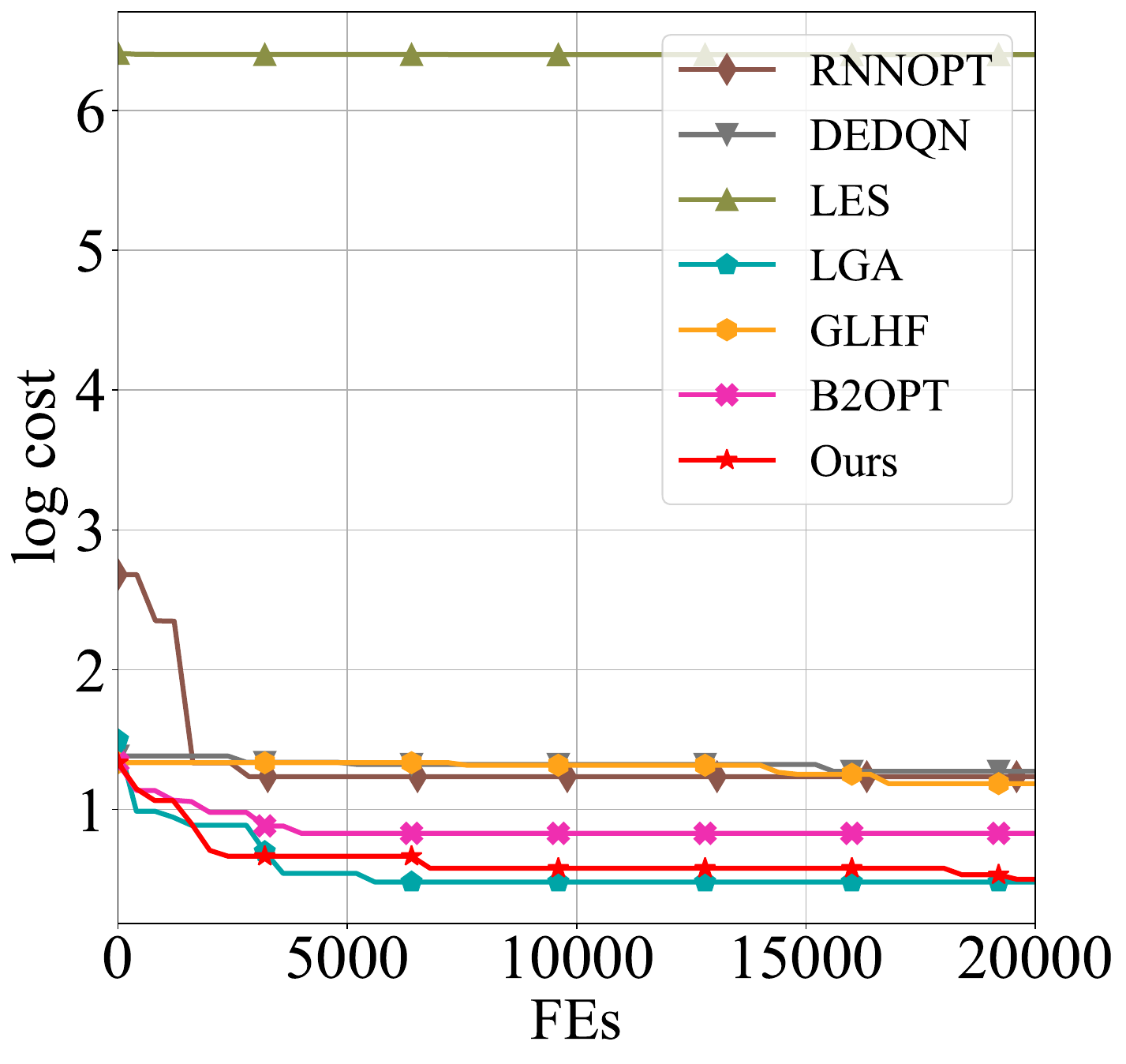}
		\caption*{Katsuura}
	\end{subfigure}
	\begin{subfigure}[b]{0.22\textwidth}
		\includegraphics[width=\linewidth]{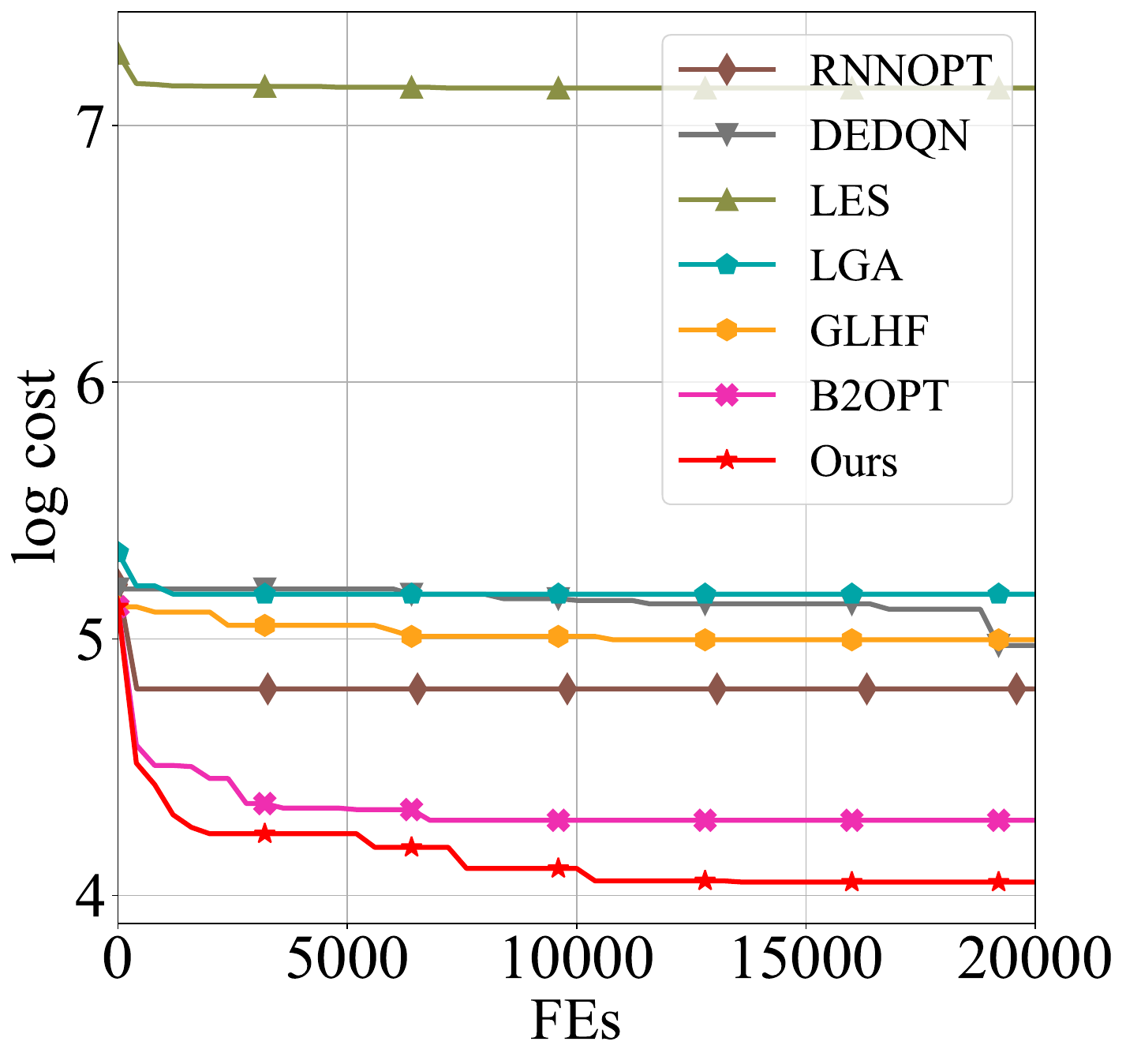}
		\caption*{Lunacek bi-Rastrigin}
	\end{subfigure}
	\begin{subfigure}[b]{0.22\textwidth}
		\includegraphics[width=\linewidth]{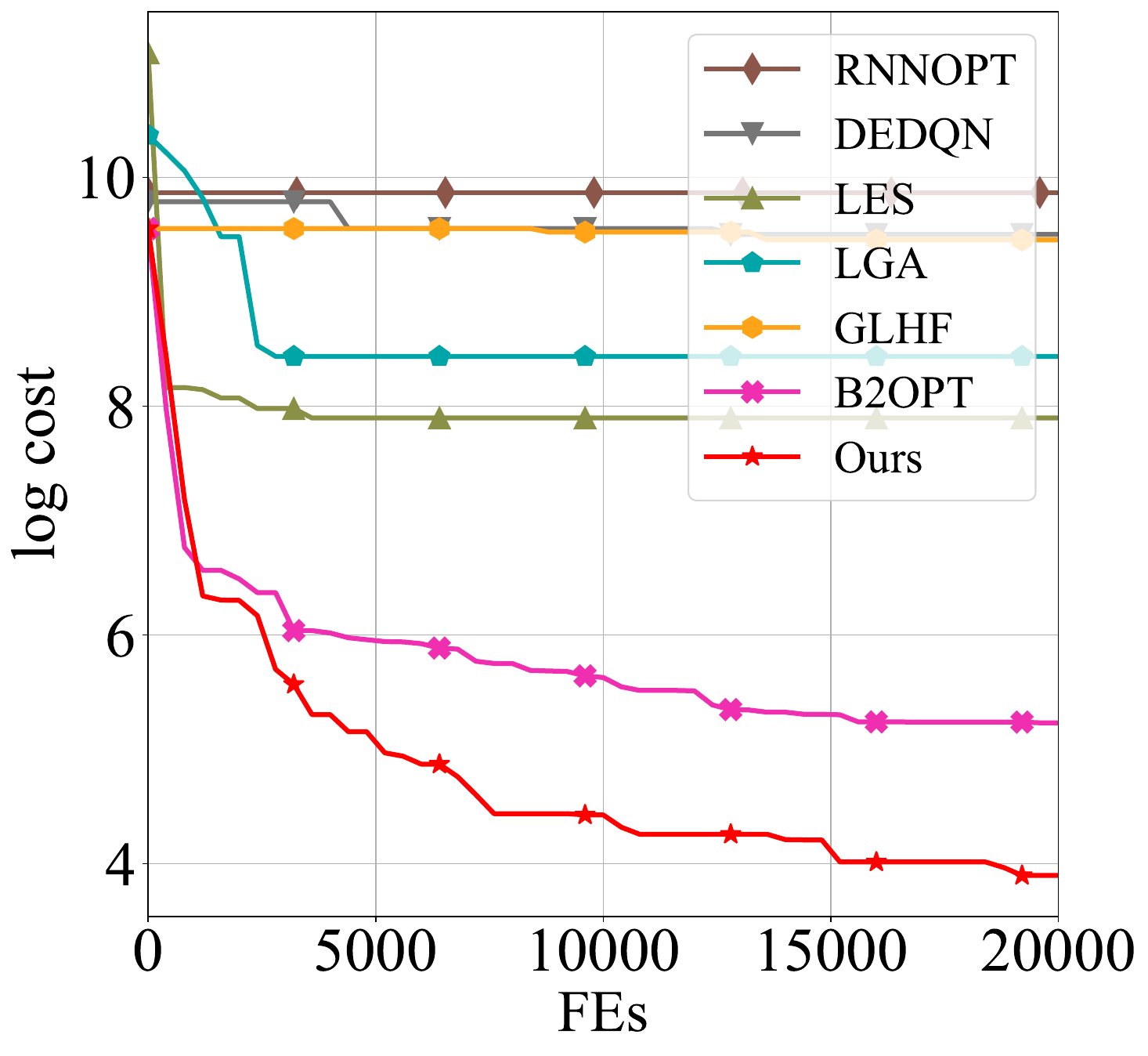}
		\caption*{Rosenbrock Original}
	\end{subfigure}
	\begin{subfigure}[b]{0.22\textwidth}
		\includegraphics[width=\linewidth]{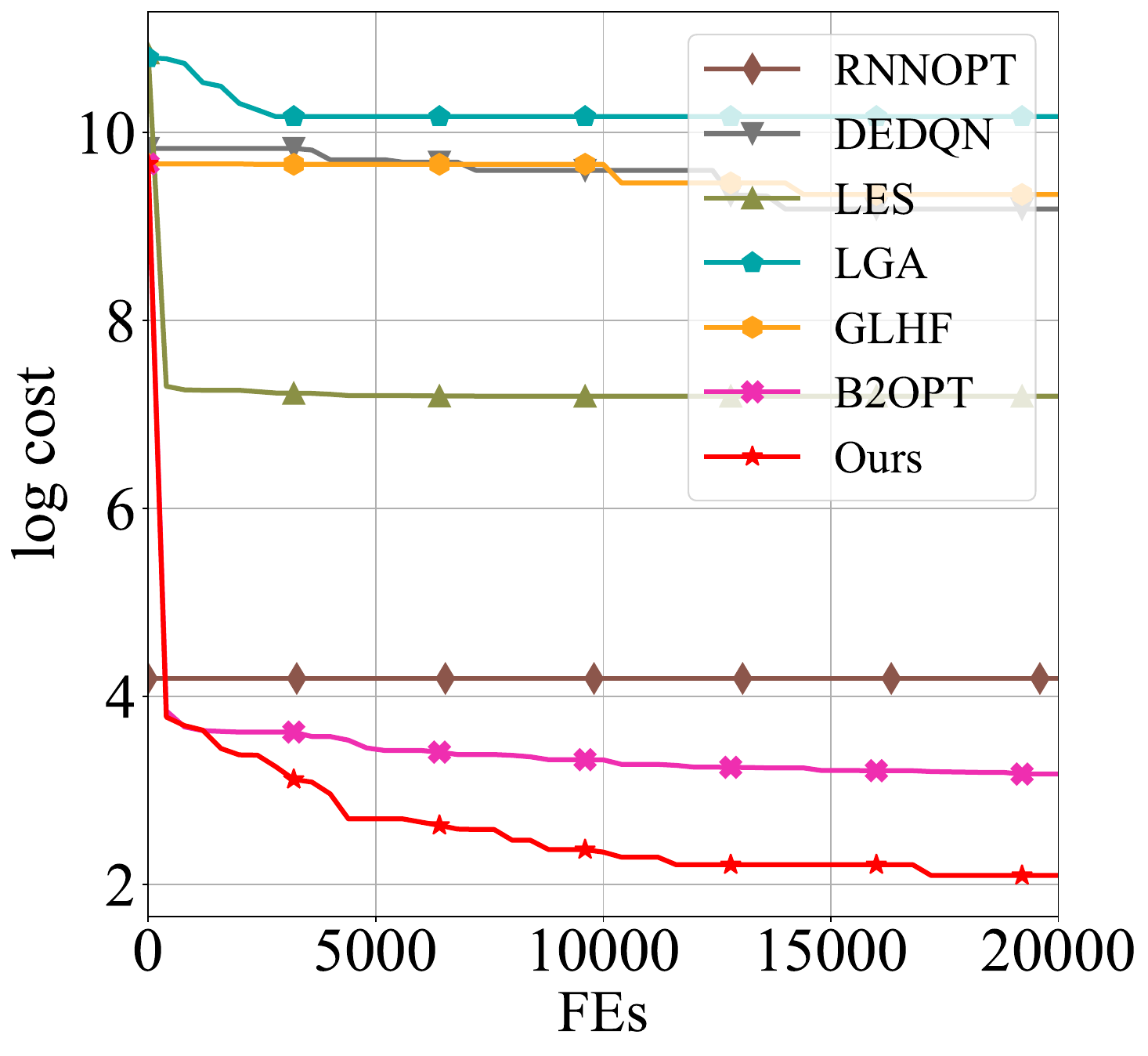}
		\caption*{Rosenbrock Rotated}
	\end{subfigure}
	
	\begin{subfigure}[b]{0.22\textwidth}
		\includegraphics[width=\linewidth]{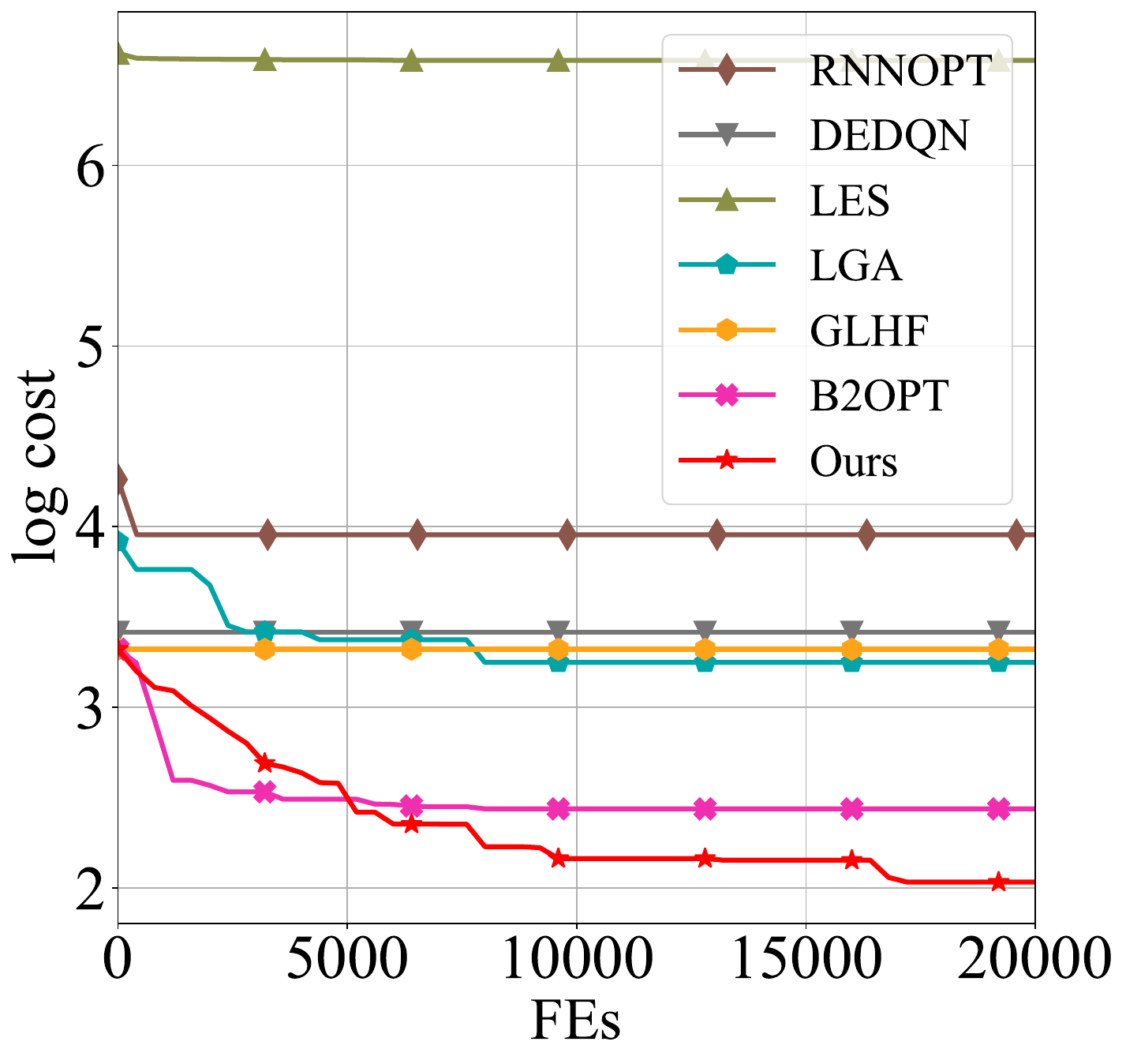}
		\caption*{Schaffers}
	\end{subfigure}
	\begin{subfigure}[b]{0.22\textwidth}
		\includegraphics[width=\linewidth]{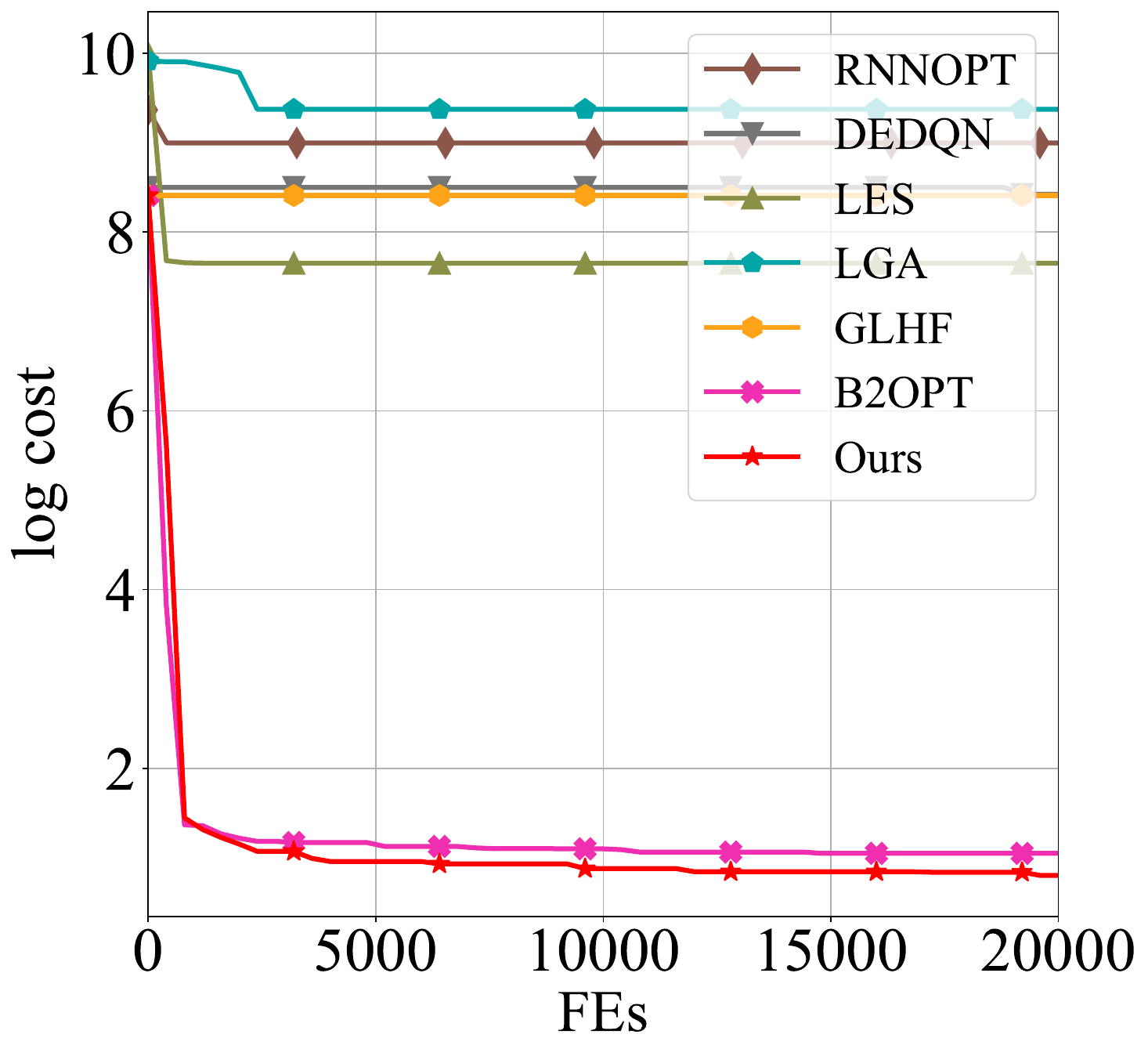}
		\caption*{Schwefel}
	\end{subfigure}
	\begin{subfigure}[b]{0.22\textwidth}
		\includegraphics[width=\linewidth]{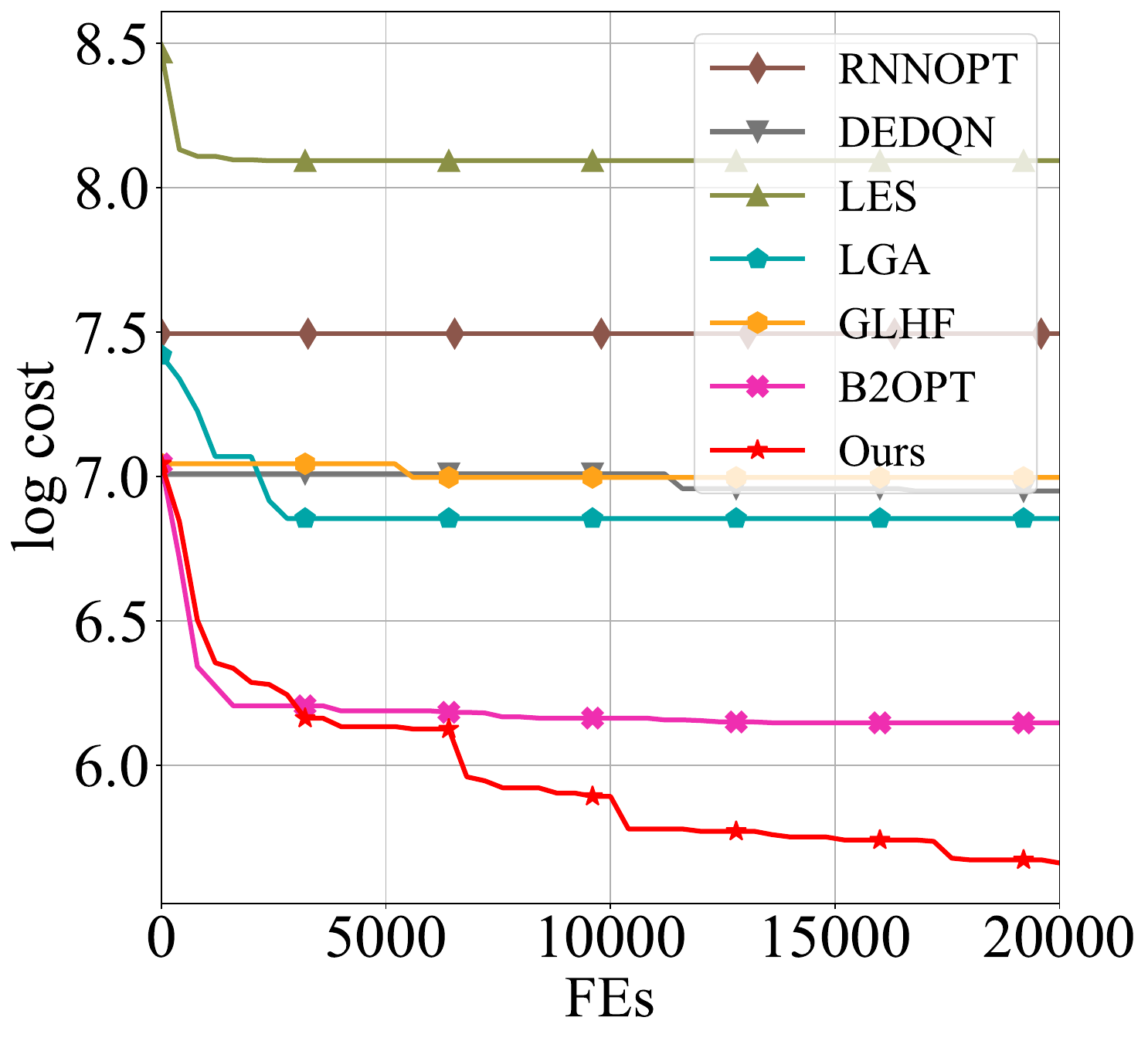}
		\caption*{Sharp Ridge}
	\end{subfigure}
	\begin{subfigure}[b]{0.22\textwidth}
		\includegraphics[width=\linewidth]{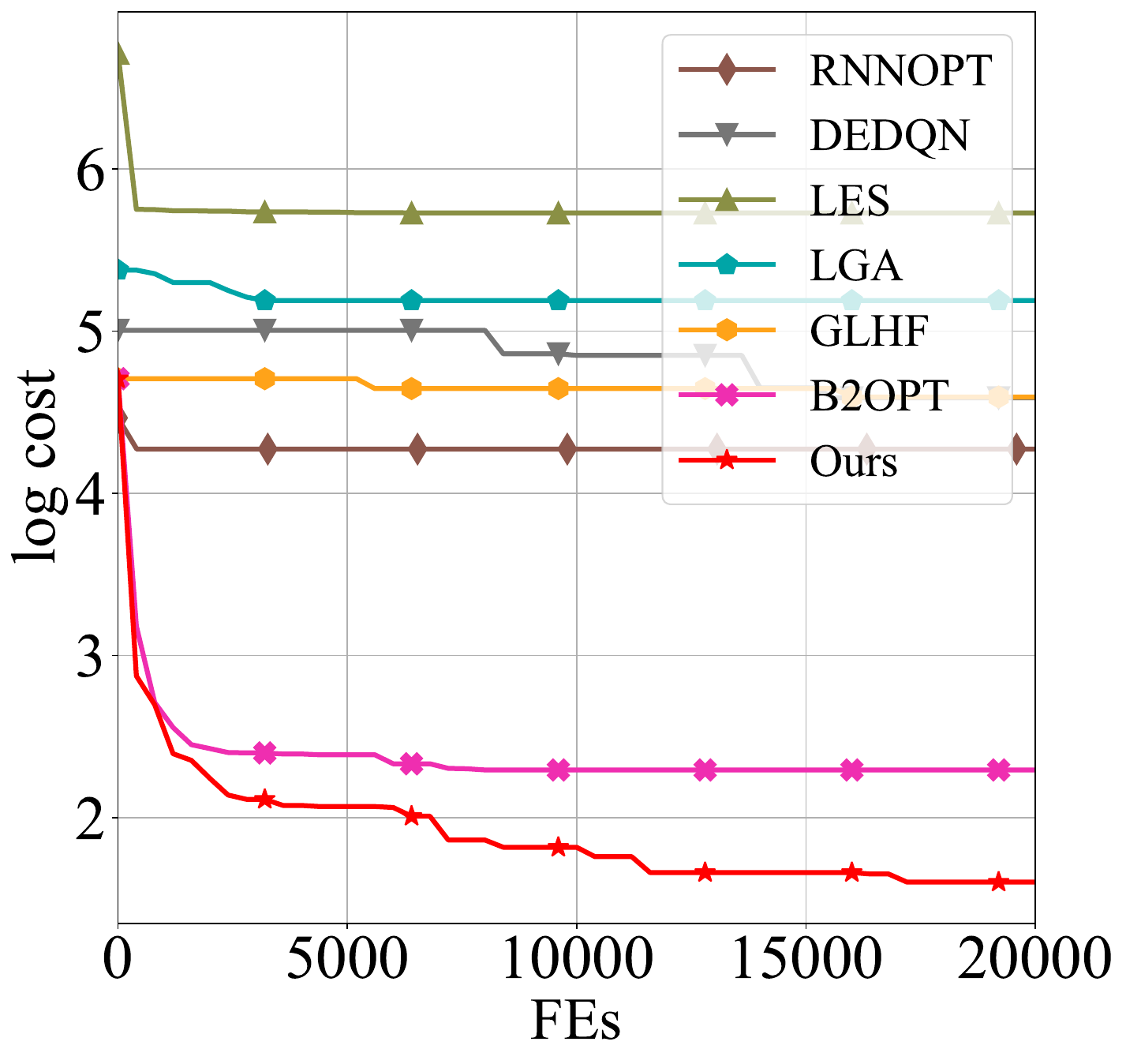}
		\caption*{Step Ellipsoidal}
	\end{subfigure}
	
	\caption{In-distribution log-scaled convergence curves of various representative methods on \emph{BBOB-10D}~\cite{hansen2021coco}.}
	\label{fig:supp-bbob-10d-diff-16}
\end{figure*}
}

\newcommand{\FigSupbbobTenTraditinalEAcurve}{%
\begin{figure*}[htbp]
	\centering
	
	\begin{subfigure}[b]{0.22\textwidth}
		\includegraphics[width=\linewidth]{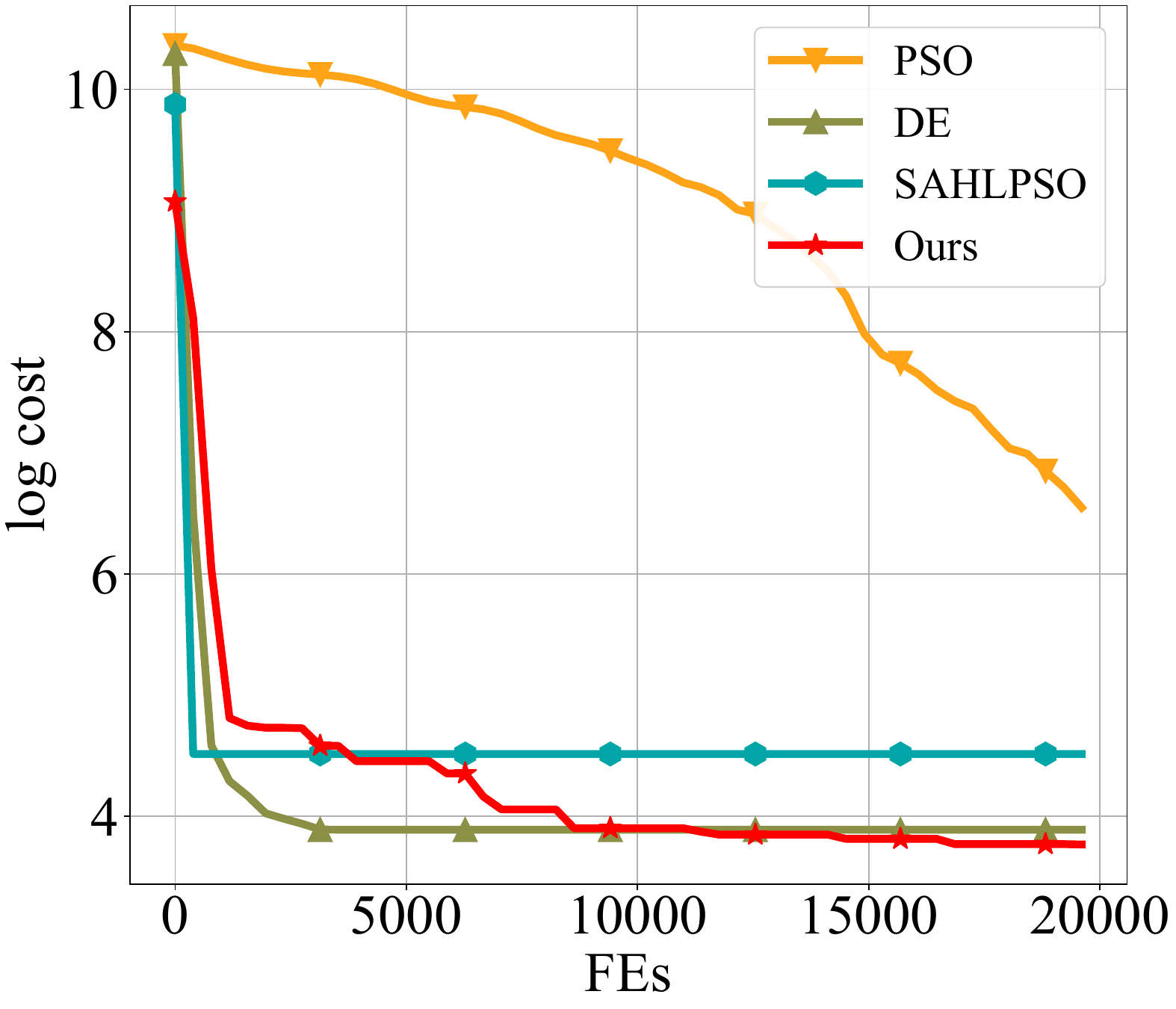}
		\caption*{Attractive Sector}
	\end{subfigure}
	\begin{subfigure}[b]{0.22\textwidth}
		\includegraphics[width=\linewidth]{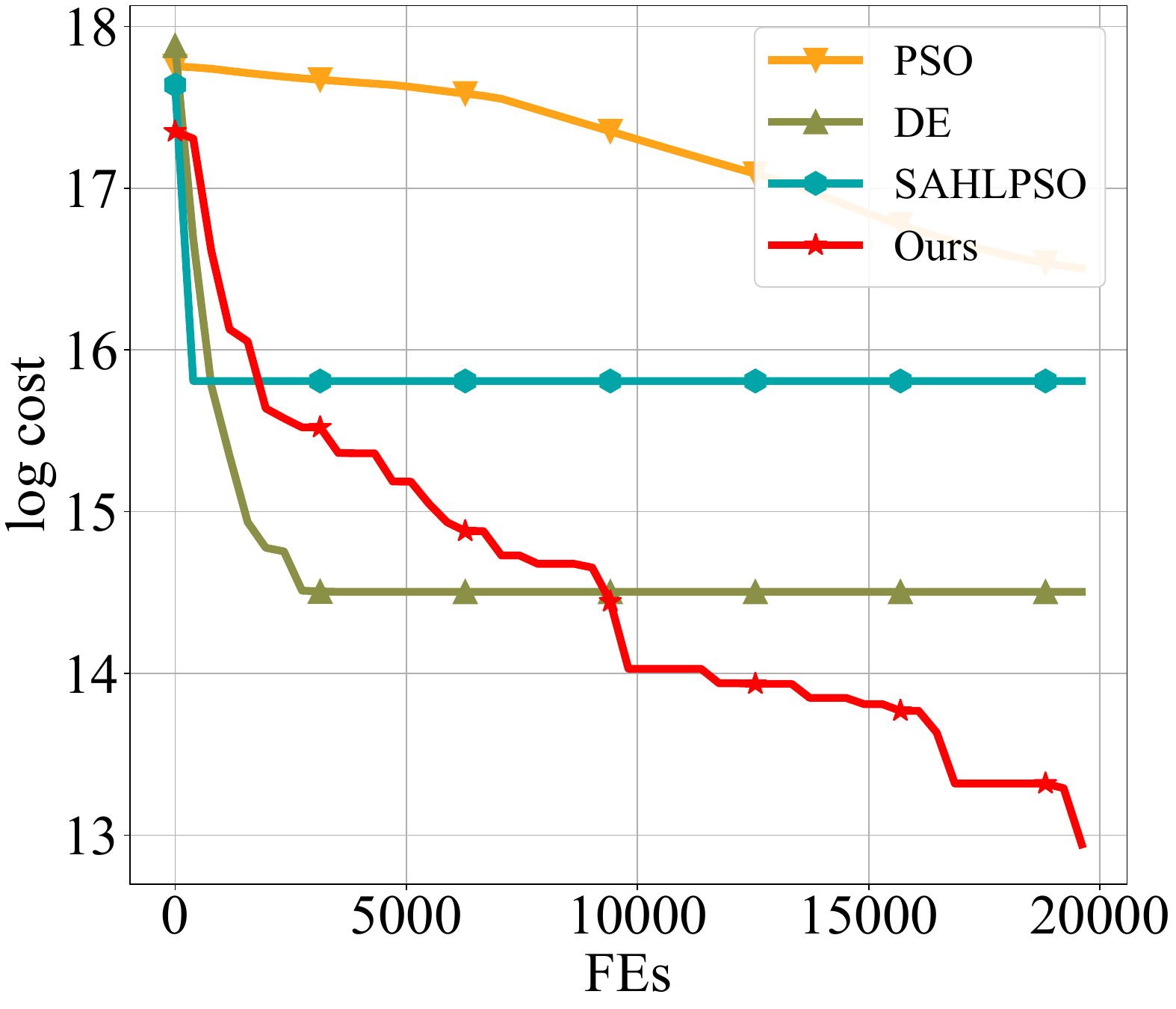}
		\caption*{Bent Cigar}
	\end{subfigure}
	\begin{subfigure}[b]{0.22\textwidth}
		\includegraphics[width=\linewidth]{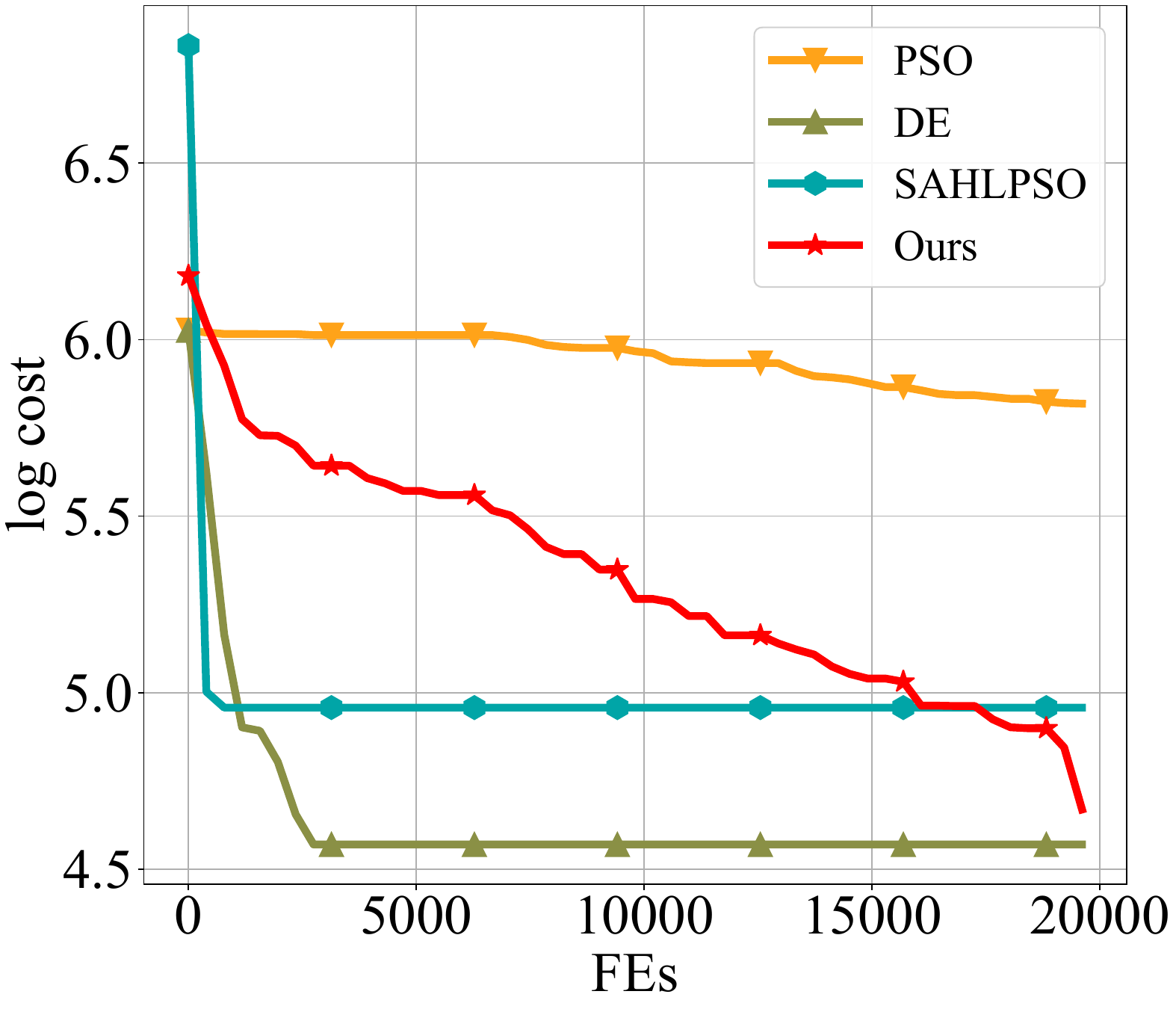}
		\caption*{Buche Rastrigin}
	\end{subfigure}
	\begin{subfigure}[b]{0.22\textwidth}
		\includegraphics[width=\linewidth]{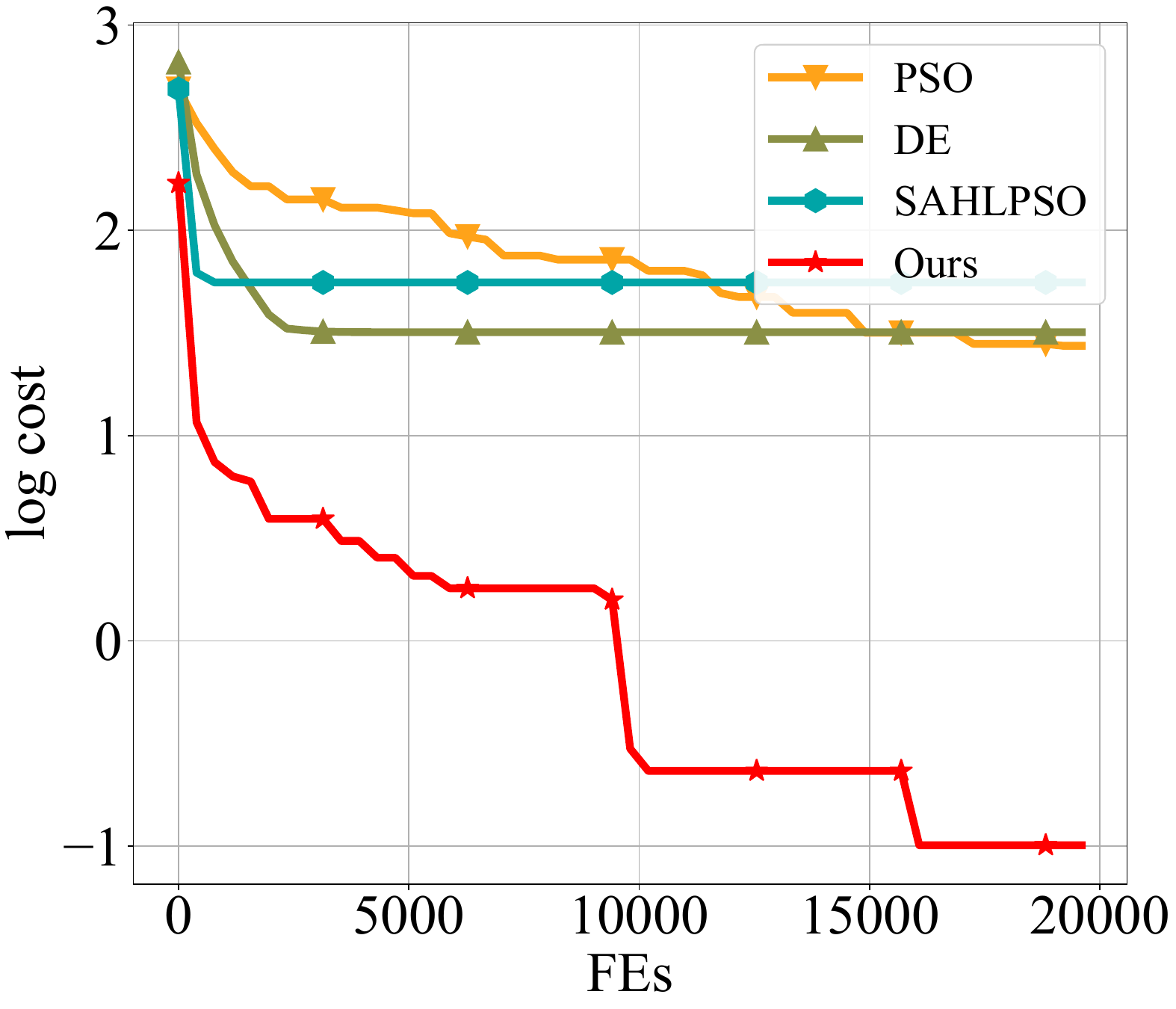}
		\caption*{Composite Grie-Rosen}
	\end{subfigure}
	
	\begin{subfigure}[b]{0.22\textwidth}
		\includegraphics[width=\linewidth]{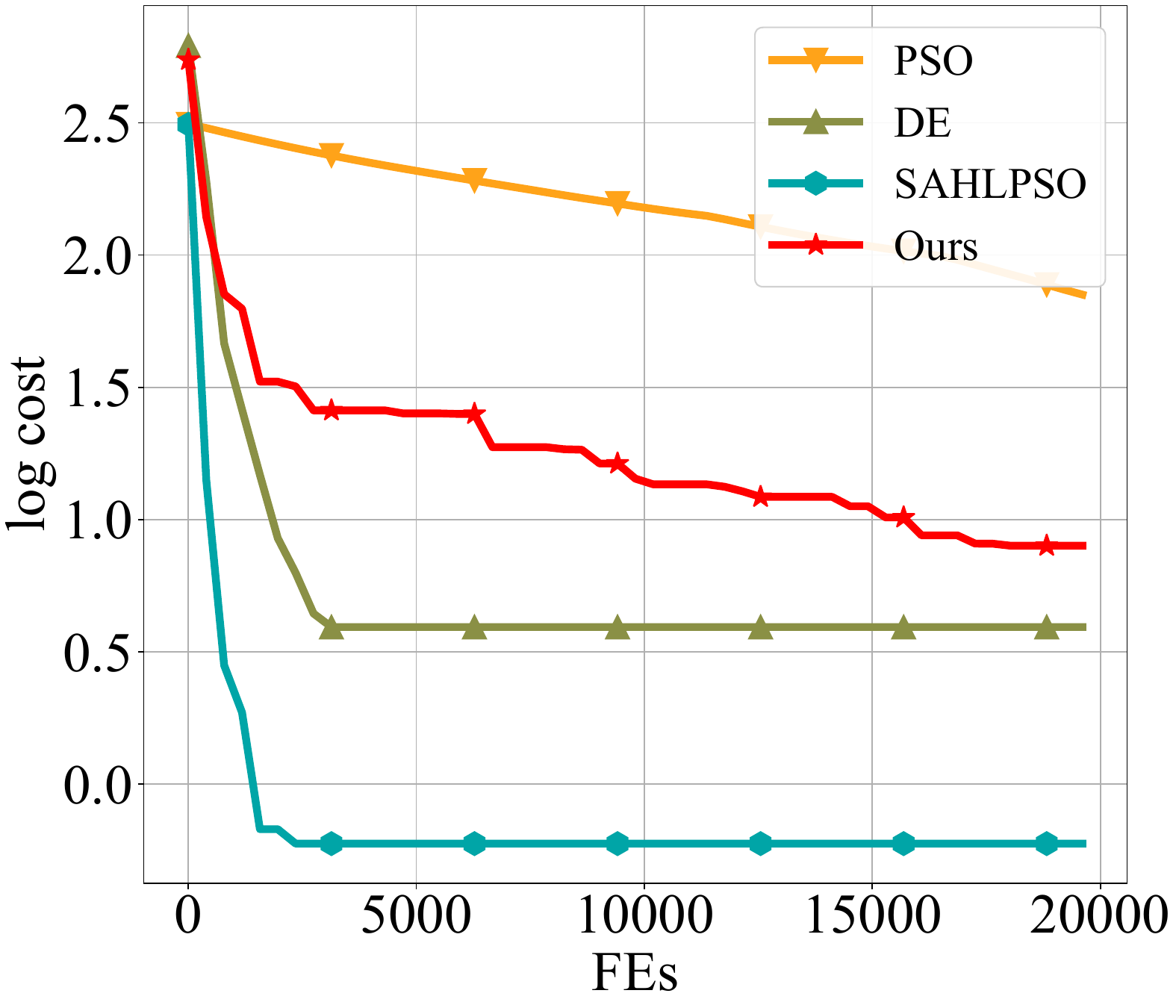}
		\caption*{Different Powers}
	\end{subfigure}
	\begin{subfigure}[b]{0.22\textwidth}
		\includegraphics[width=\linewidth]{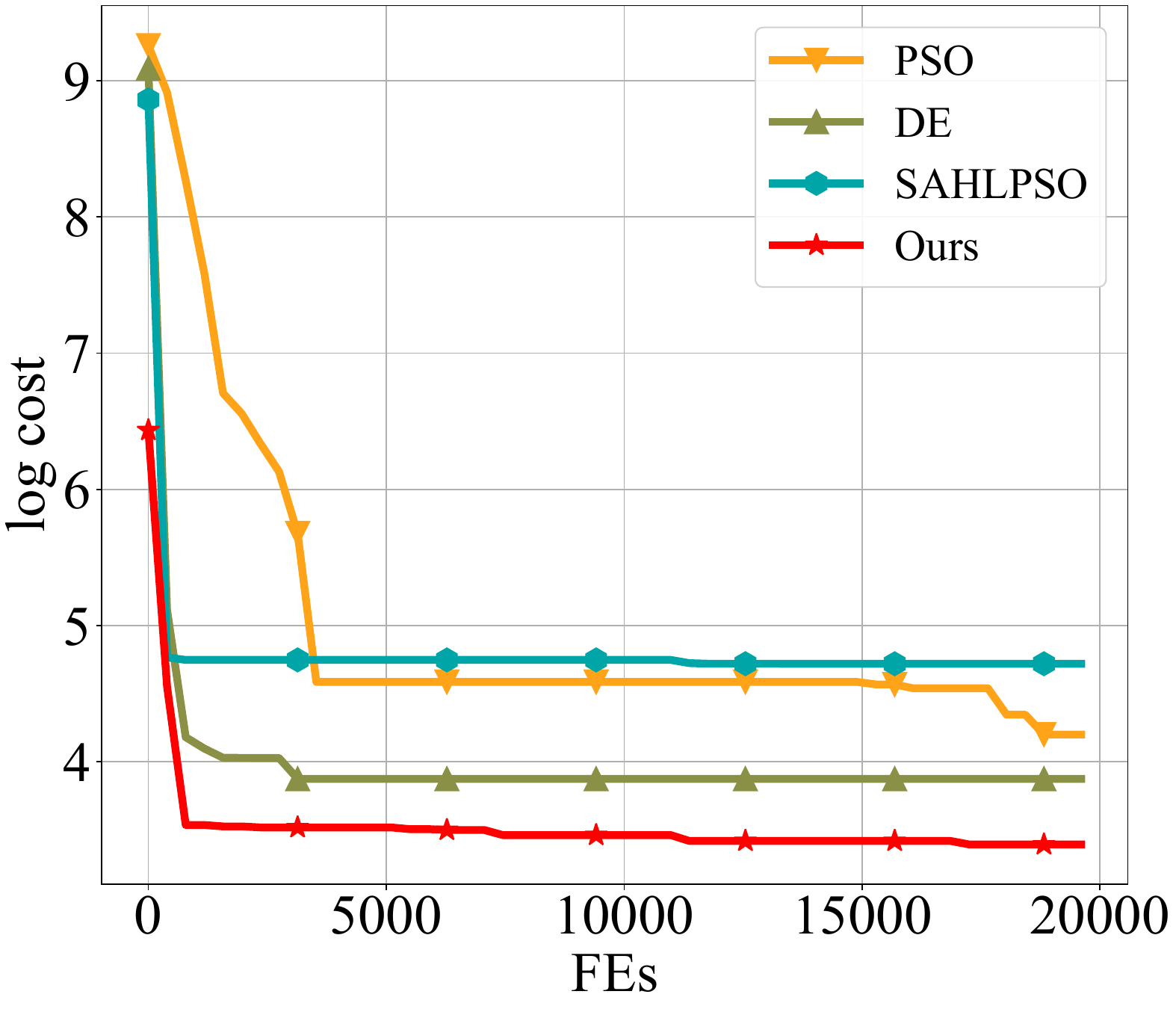}
		\caption*{Discus}
	\end{subfigure}
	\begin{subfigure}[b]{0.22\textwidth}
		\includegraphics[width=\linewidth]{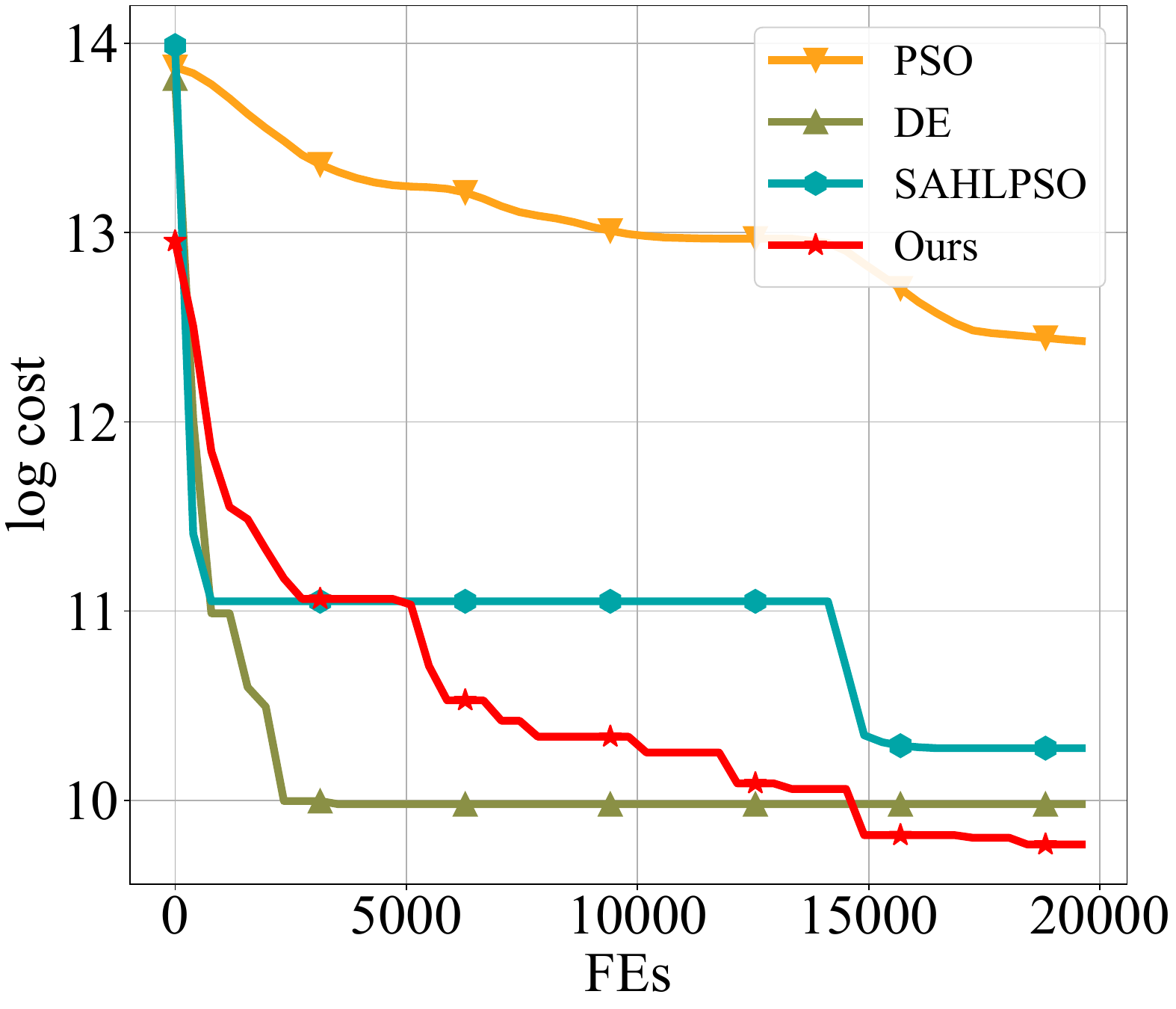}
		\caption*{Ellipsoidal}
	\end{subfigure}
	\begin{subfigure}[b]{0.22\textwidth}
		\includegraphics[width=\linewidth]{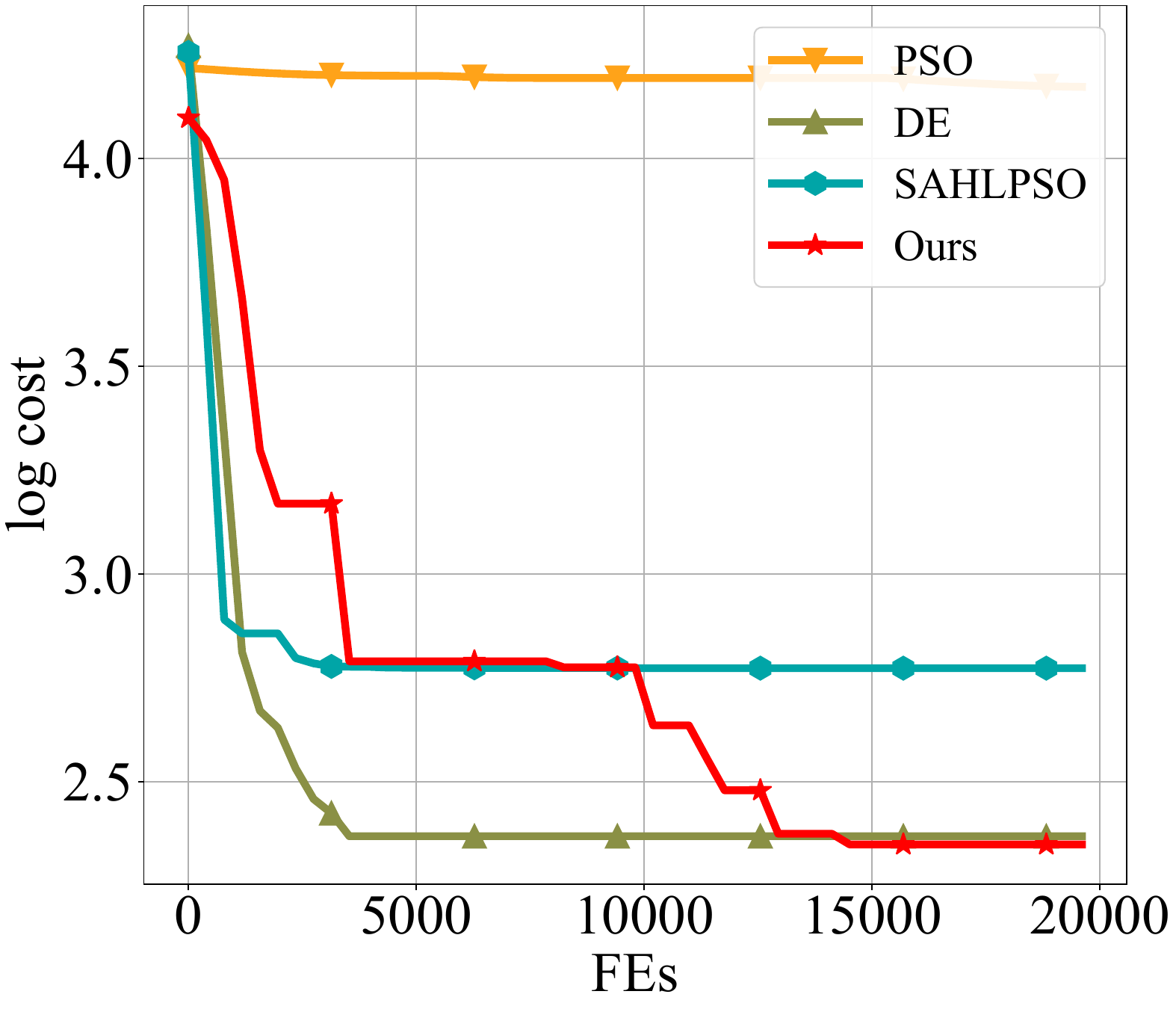}
		\caption*{Gallagher 21Peaks}
	\end{subfigure}
	
	\begin{subfigure}[b]{0.22\textwidth}
		\includegraphics[width=\linewidth]{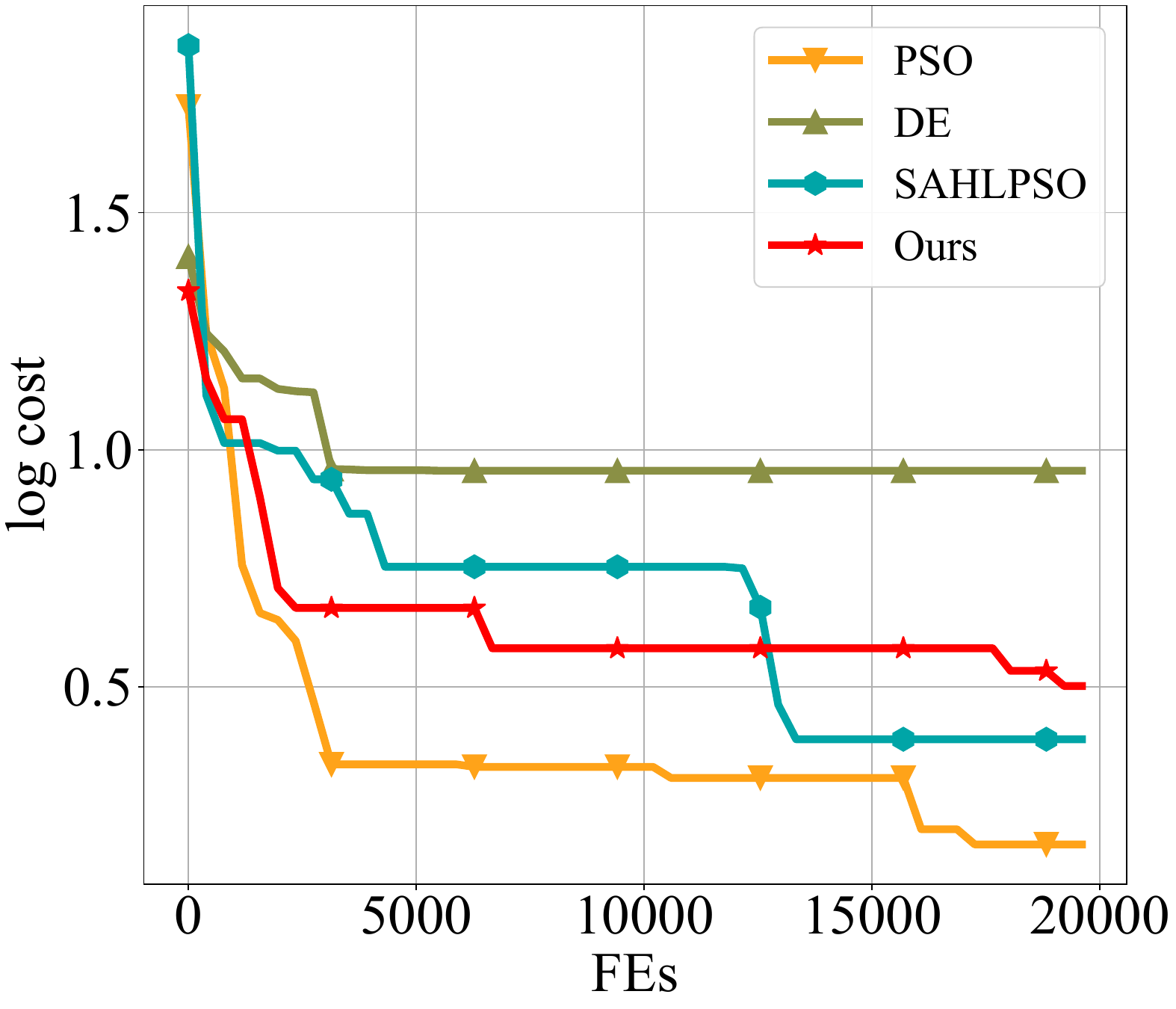}
		\caption*{Katsuura}
	\end{subfigure}
	\begin{subfigure}[b]{0.22\textwidth}
		\includegraphics[width=\linewidth]{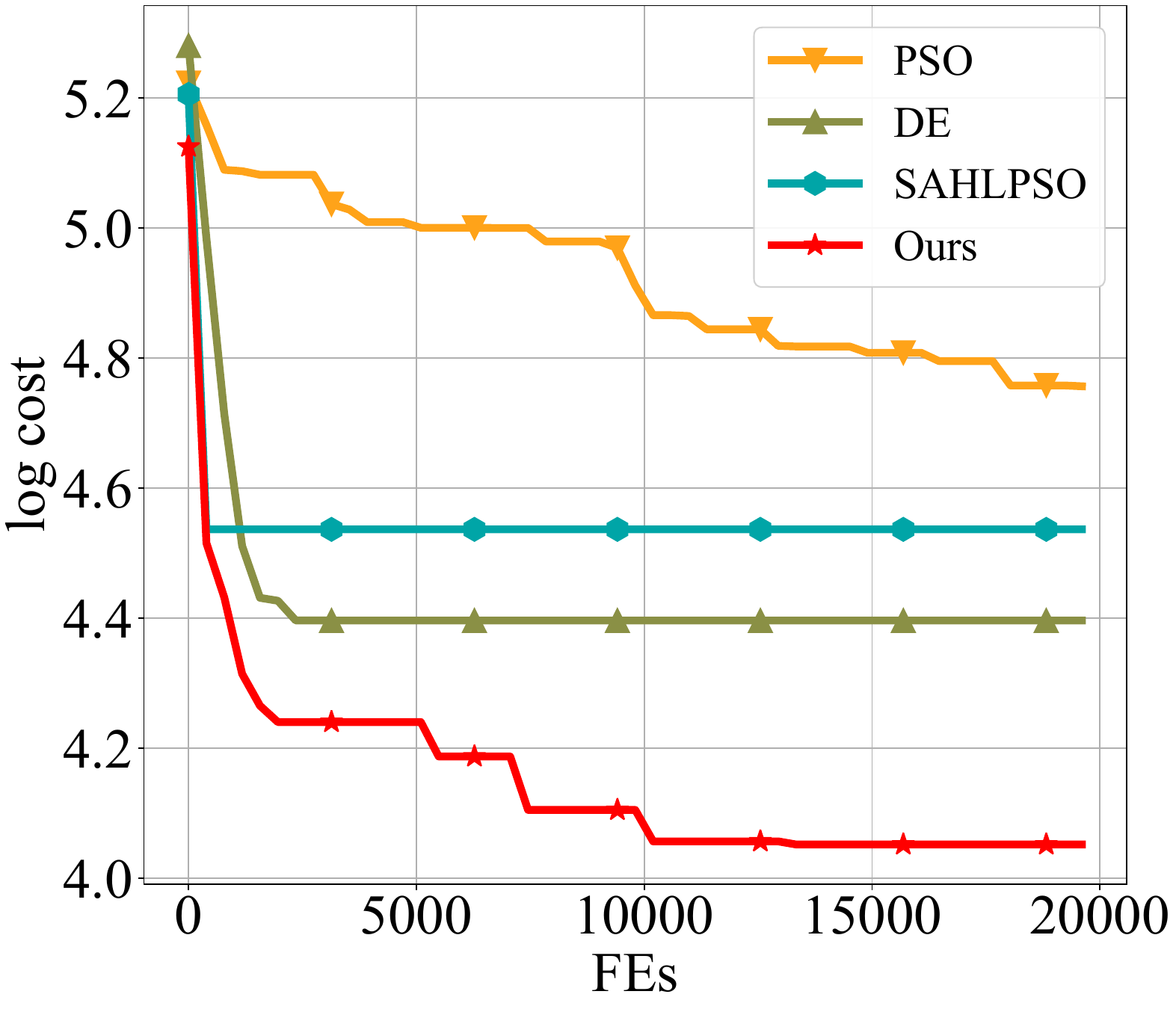}
		\caption*{Lunacek bi-Rastrigin}
	\end{subfigure}
	\begin{subfigure}[b]{0.22\textwidth}
		\includegraphics[width=\linewidth]{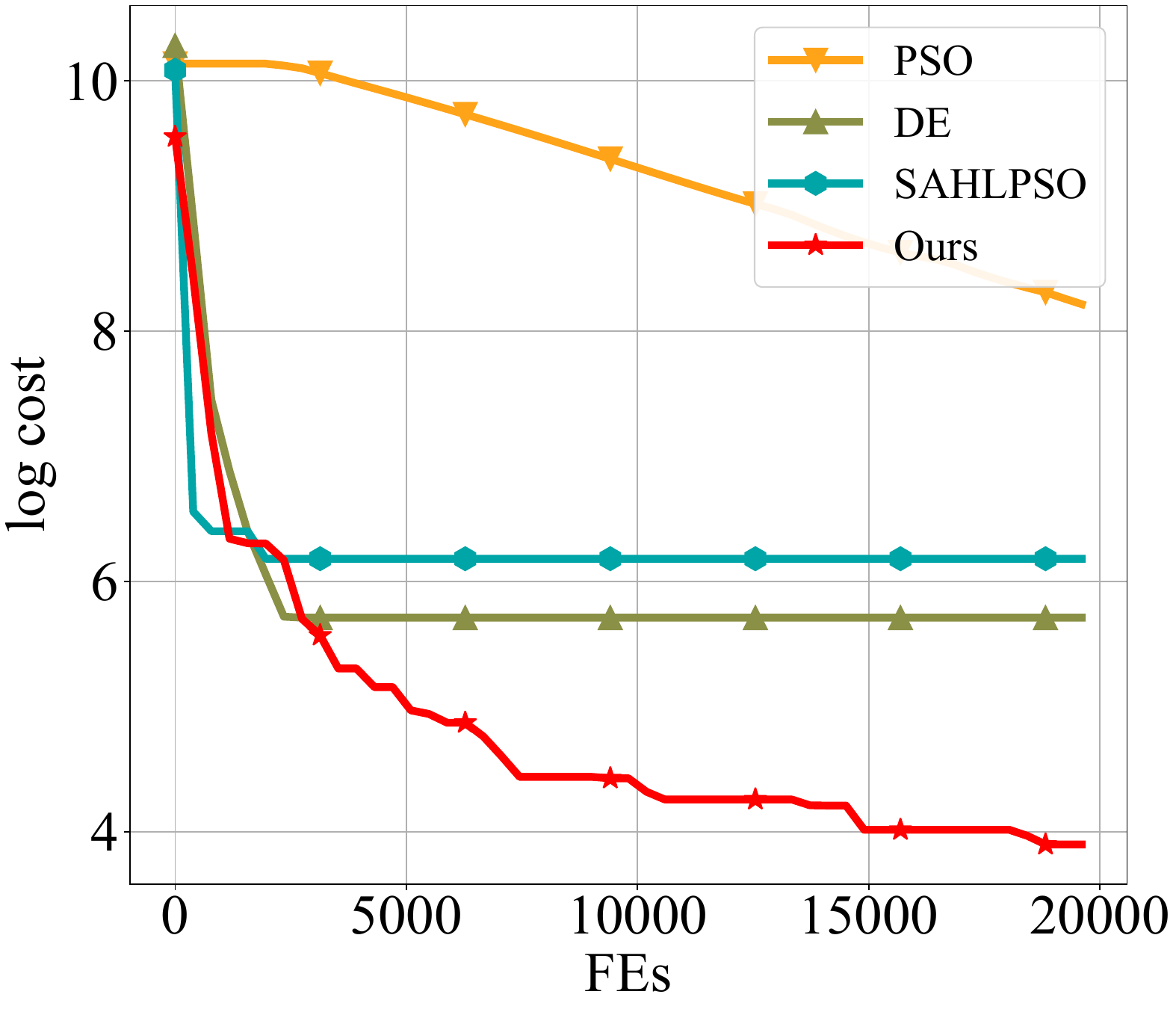}
		\caption*{Rosenbrock Original}
	\end{subfigure}
	\begin{subfigure}[b]{0.22\textwidth}
		\includegraphics[width=\linewidth]{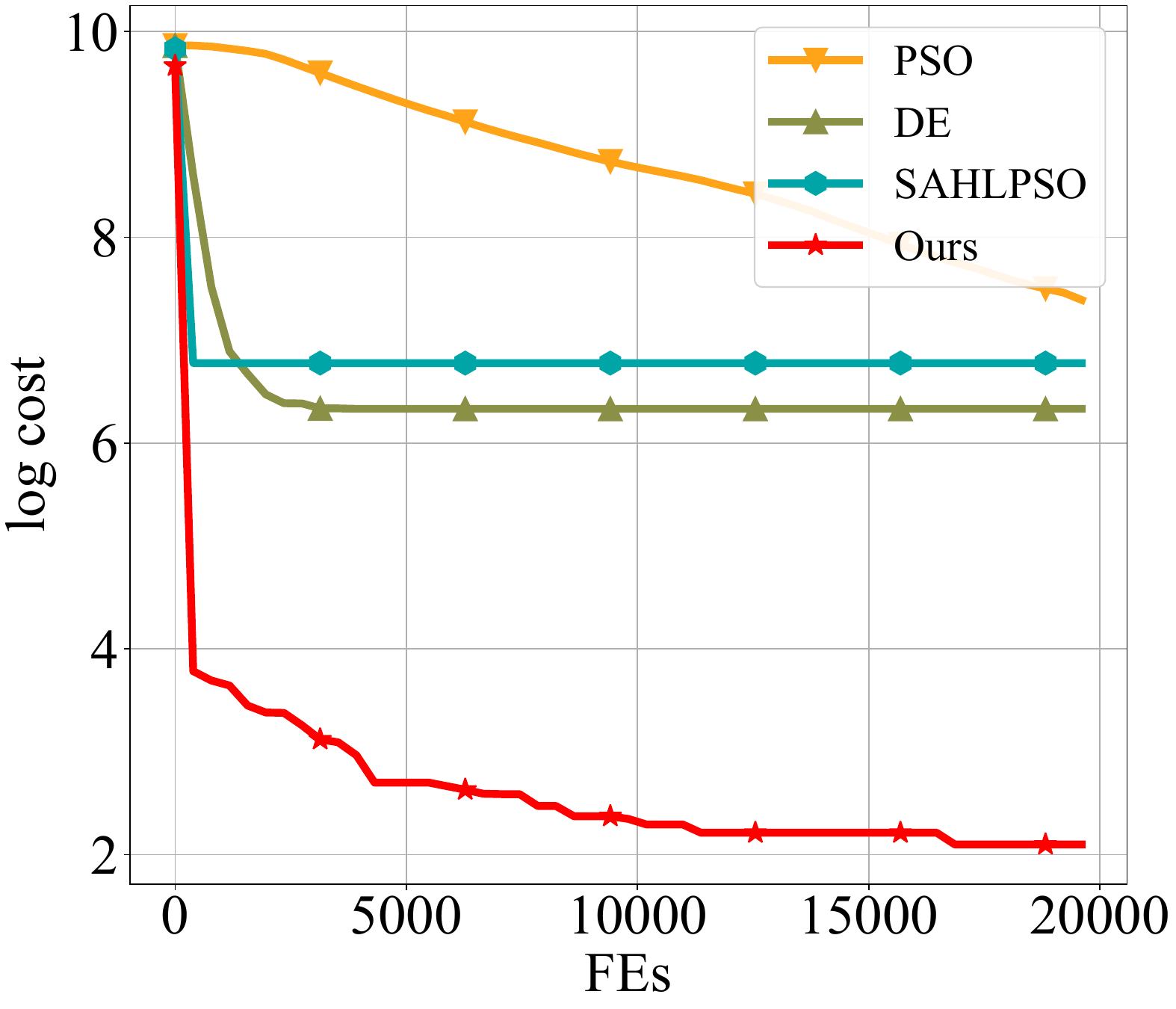}
		\caption*{Rosenbrock Rotated}
	\end{subfigure}
	
	\begin{subfigure}[b]{0.22\textwidth}
		\includegraphics[width=\linewidth]{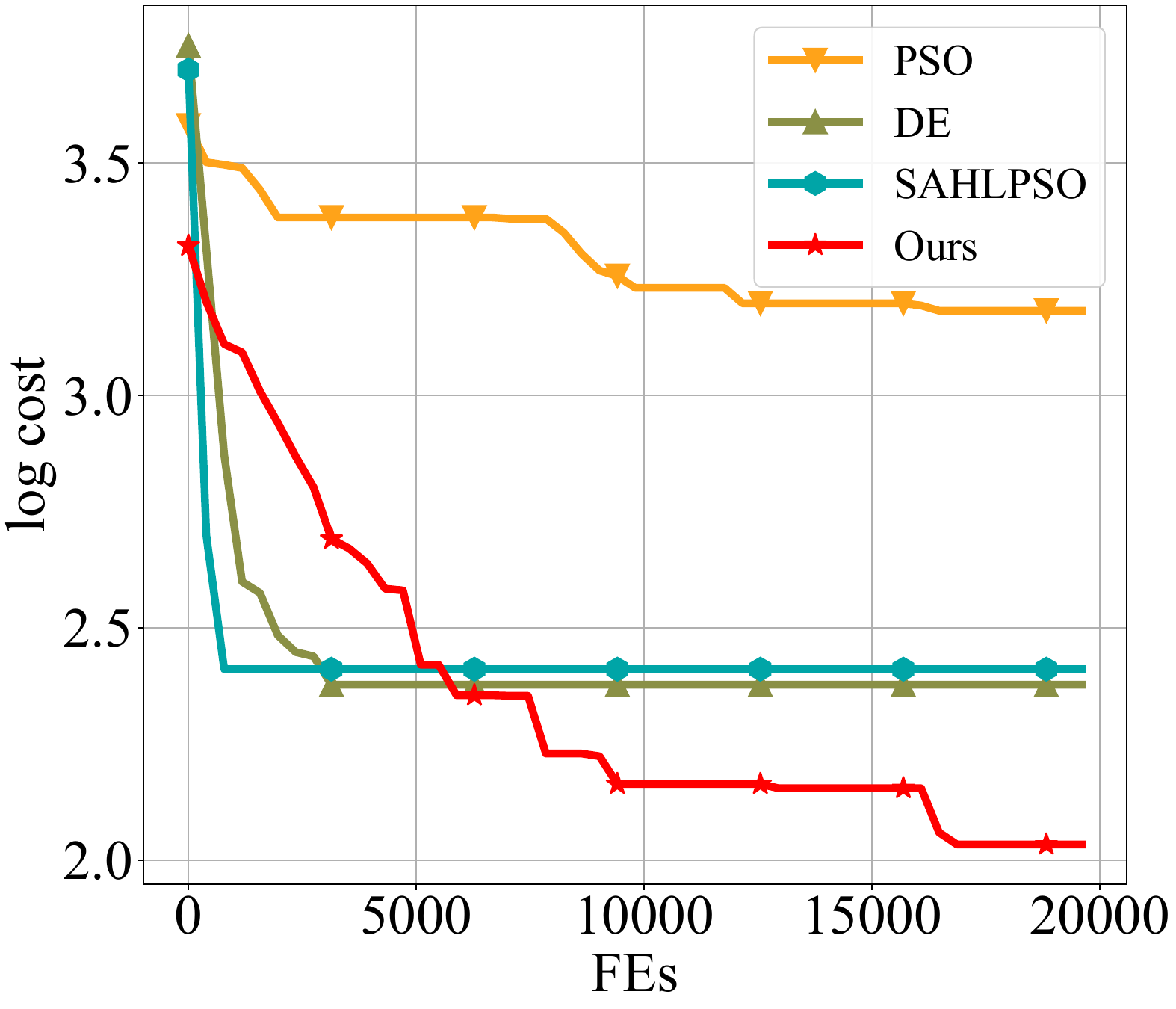}
		\caption*{Schaffers}
	\end{subfigure}
	\begin{subfigure}[b]{0.22\textwidth}
		\includegraphics[width=\linewidth]{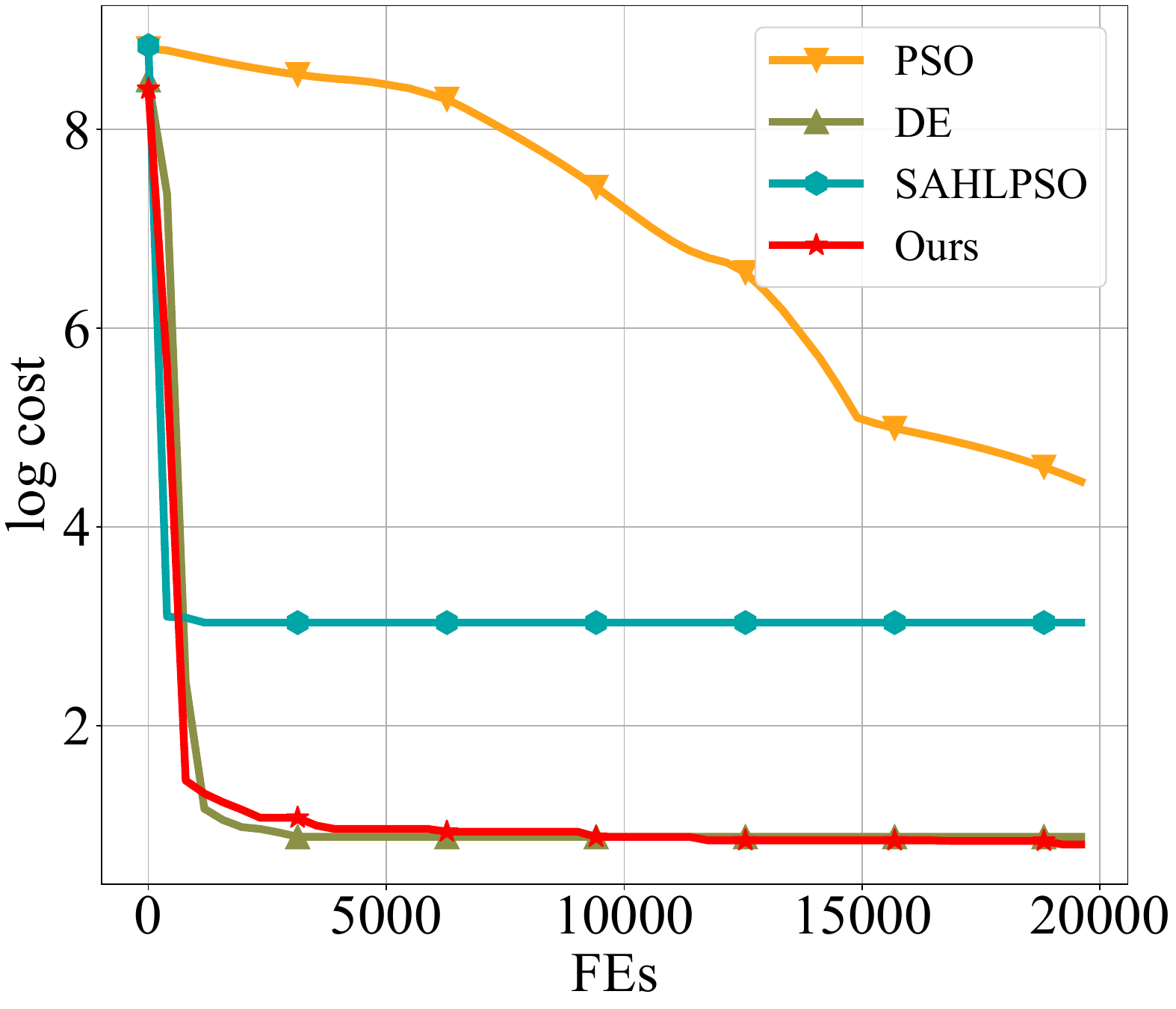}
		\caption*{Schwefel}
	\end{subfigure}
	\begin{subfigure}[b]{0.22\textwidth}
		\includegraphics[width=\linewidth]{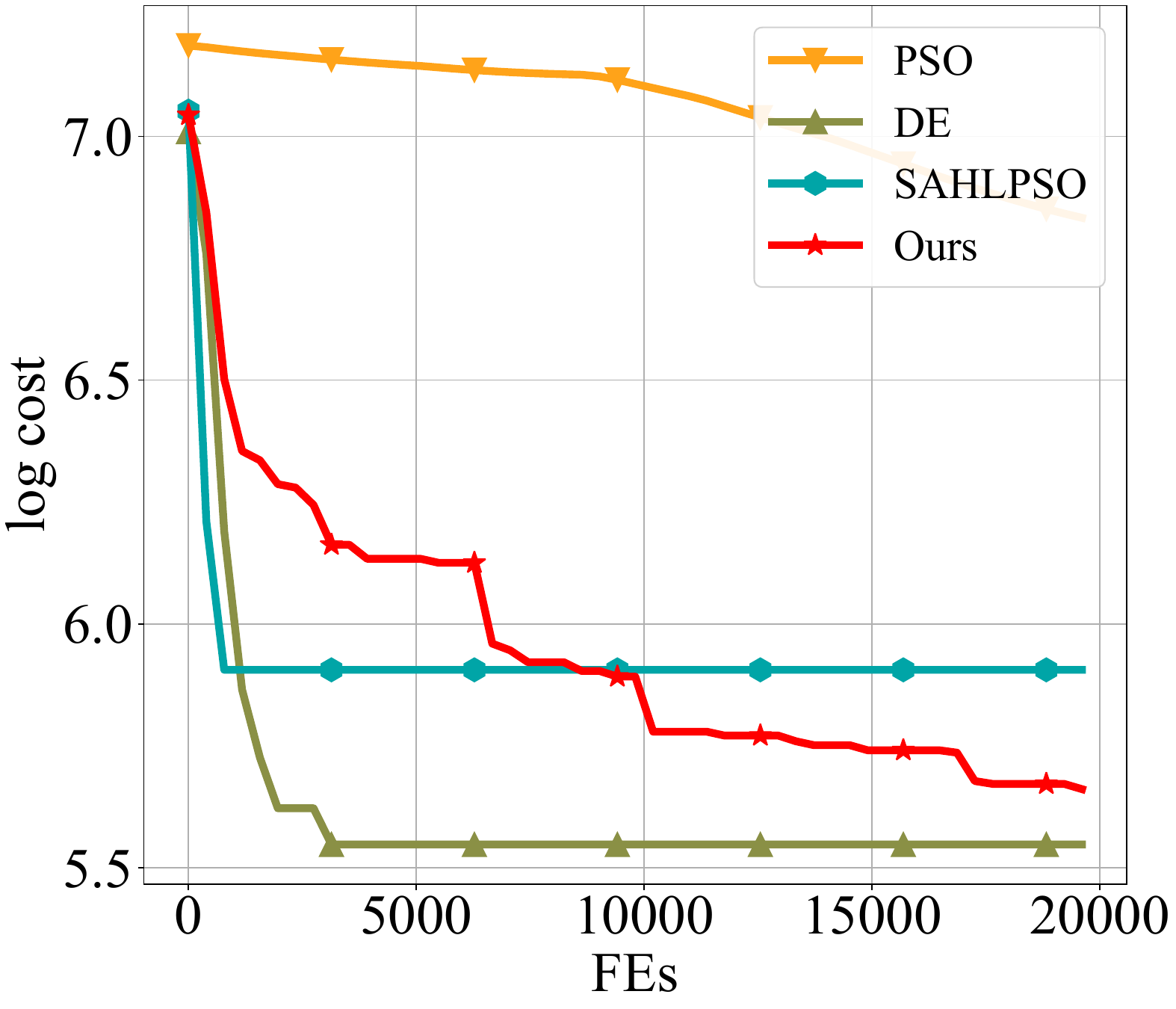}
		\caption*{Sharp Ridge}
	\end{subfigure}
	\begin{subfigure}[b]{0.22\textwidth}
		\includegraphics[width=\linewidth]{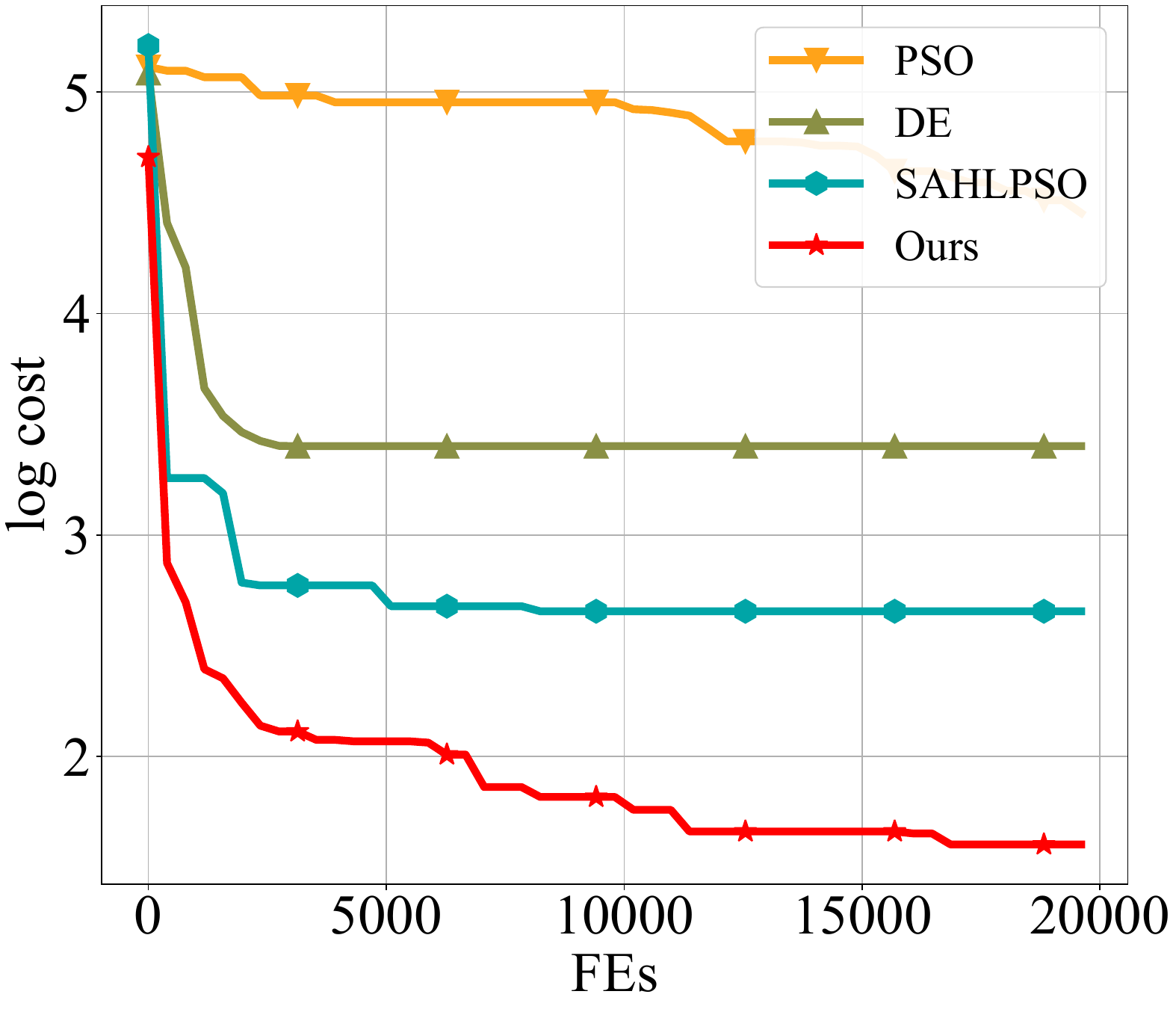}
		\caption*{Step Ellipsoidal}
	\end{subfigure}
	
	\caption{Log-scaled convergence curves of various traditional representative EA methods on \emph{BBOB-10D}~\cite{hansen2021coco}.}
	\label{fig:supp-bbob-10d-diff-16-traditional}
\end{figure*}
}

\newcommand{\FigSupbbobTenSurrogatecurve}{%
\begin{figure*}[htbp]
	\centering
	
	\begin{subfigure}[b]{0.22\textwidth}
		\includegraphics[width=\linewidth]{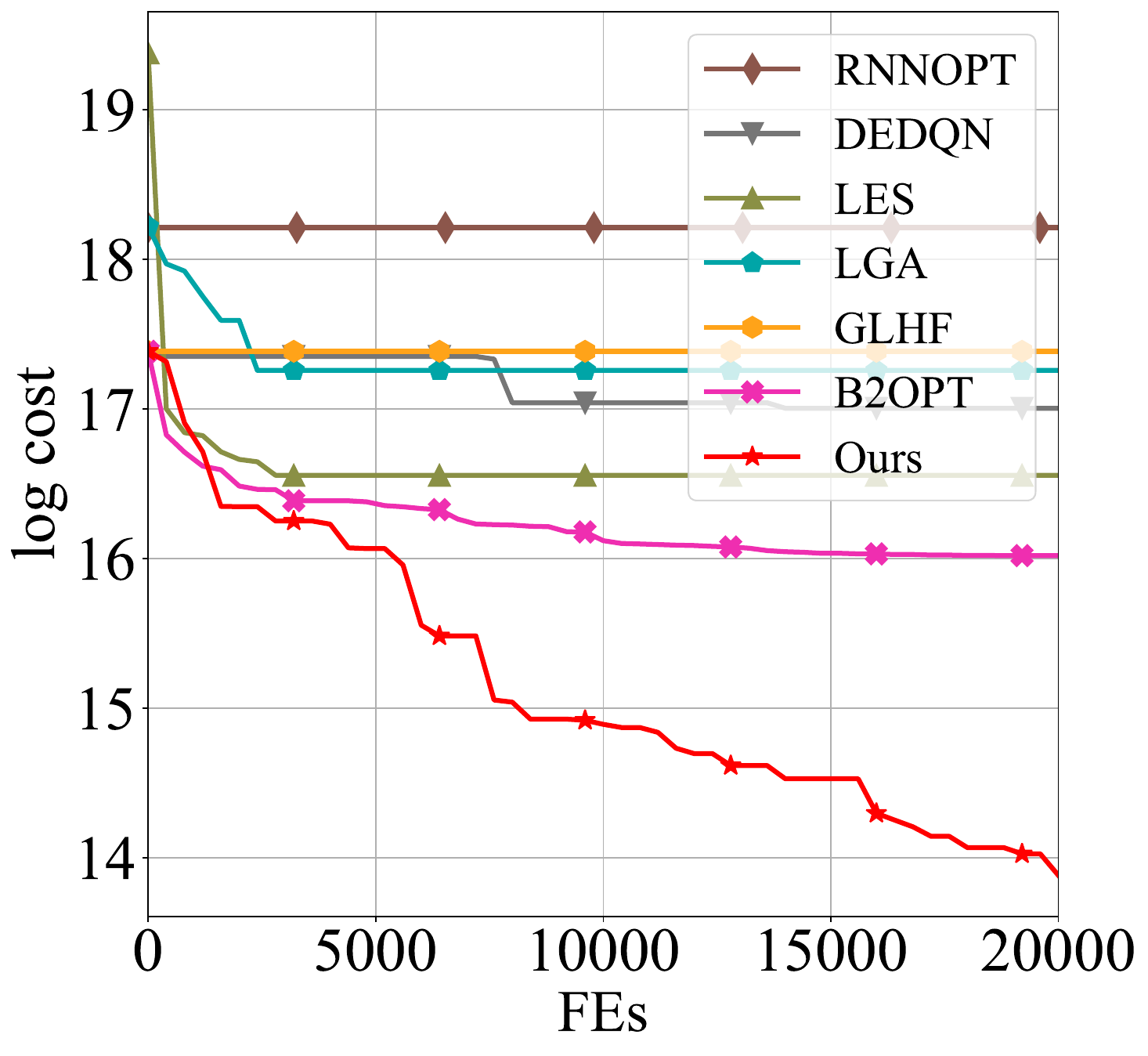}
		\caption*{Bent Cigar}
	\end{subfigure}
	\begin{subfigure}[b]{0.22\textwidth}
		\includegraphics[width=\linewidth]{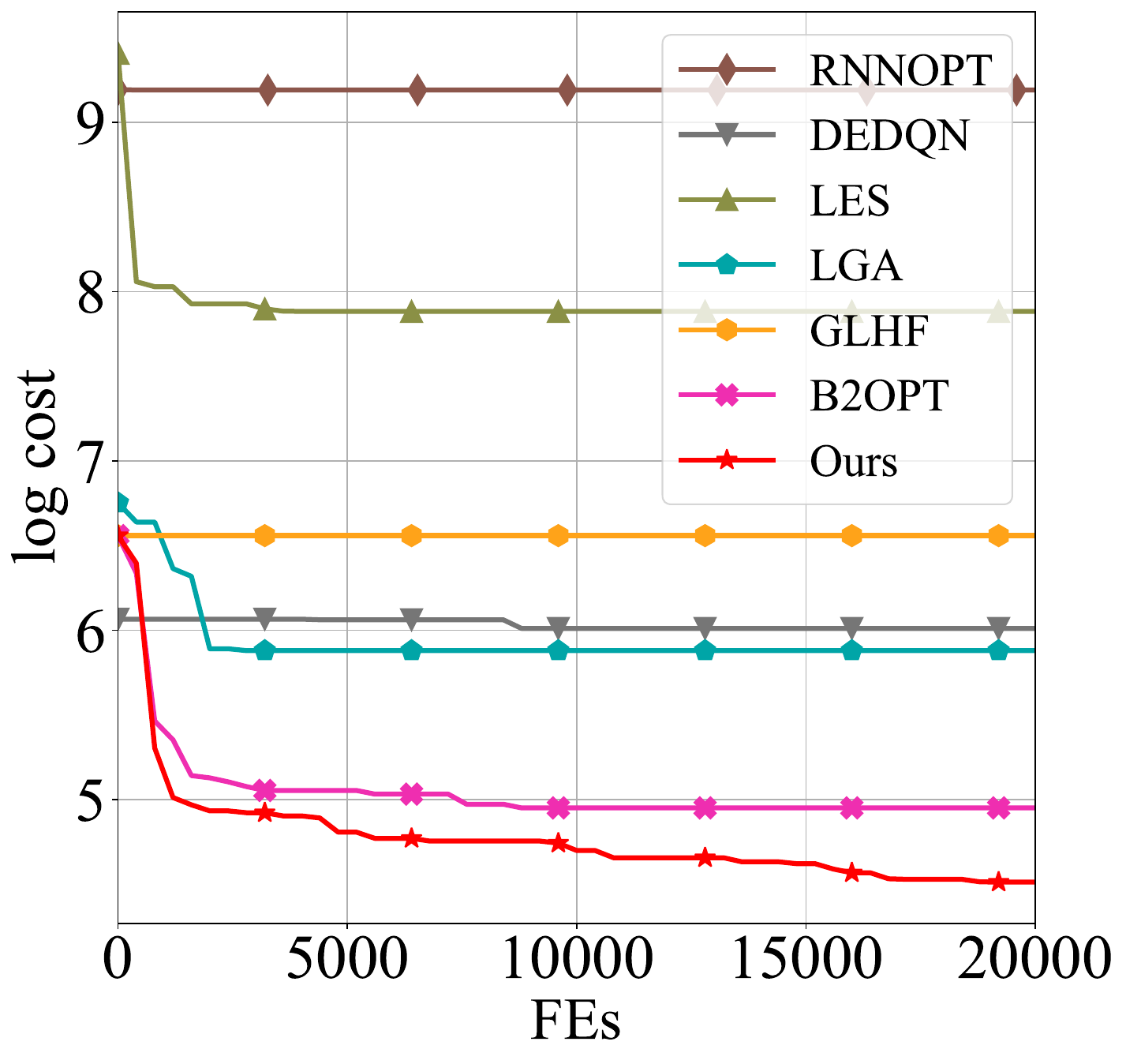}
		\caption*{Buche Rastrigin}
	\end{subfigure}
	\begin{subfigure}[b]{0.22\textwidth}
		\includegraphics[width=\linewidth]{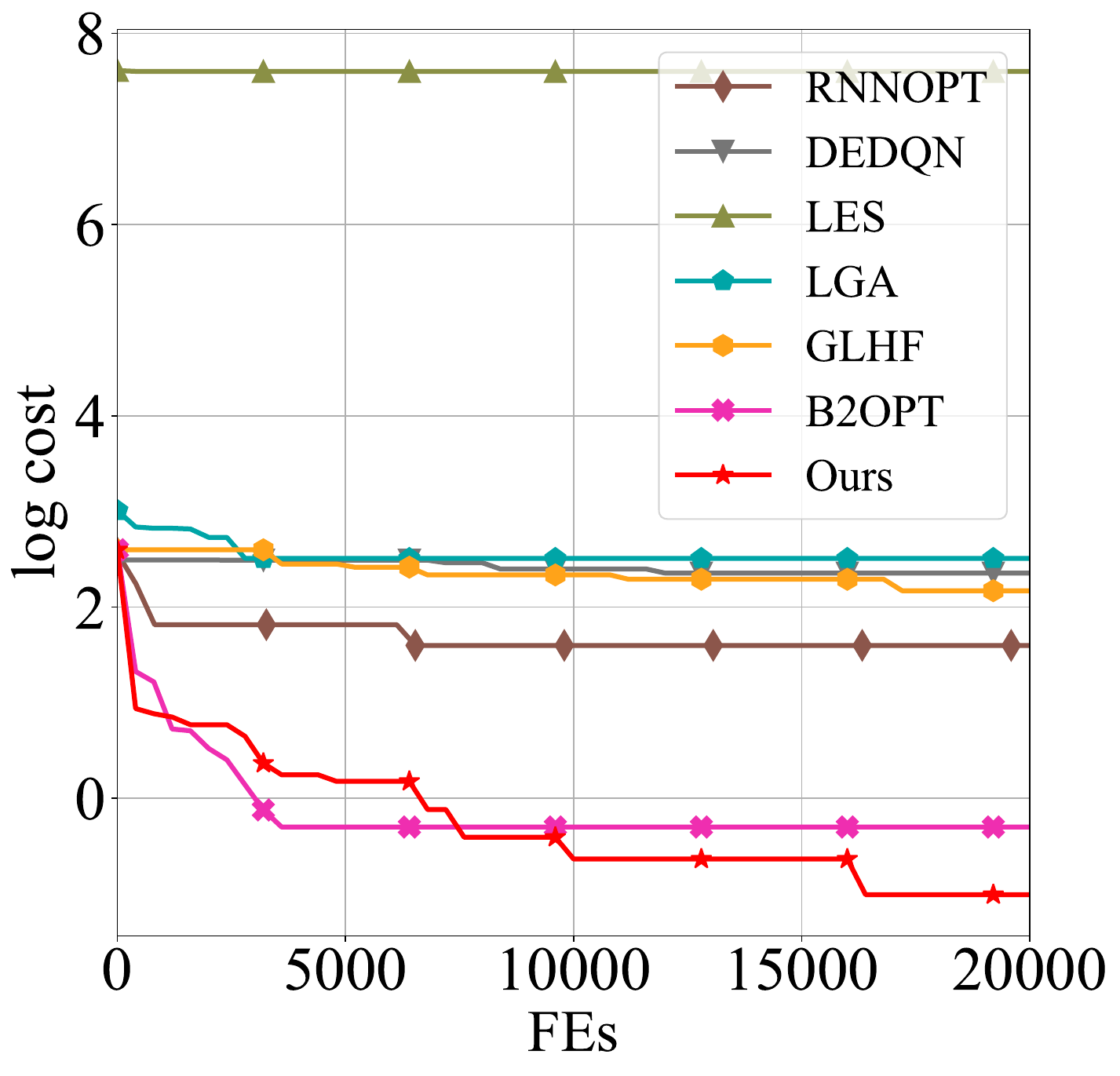}
		\caption*{Composite Grie-Rosen}
	\end{subfigure}
	\begin{subfigure}[b]{0.22\textwidth}
		\includegraphics[width=\linewidth]{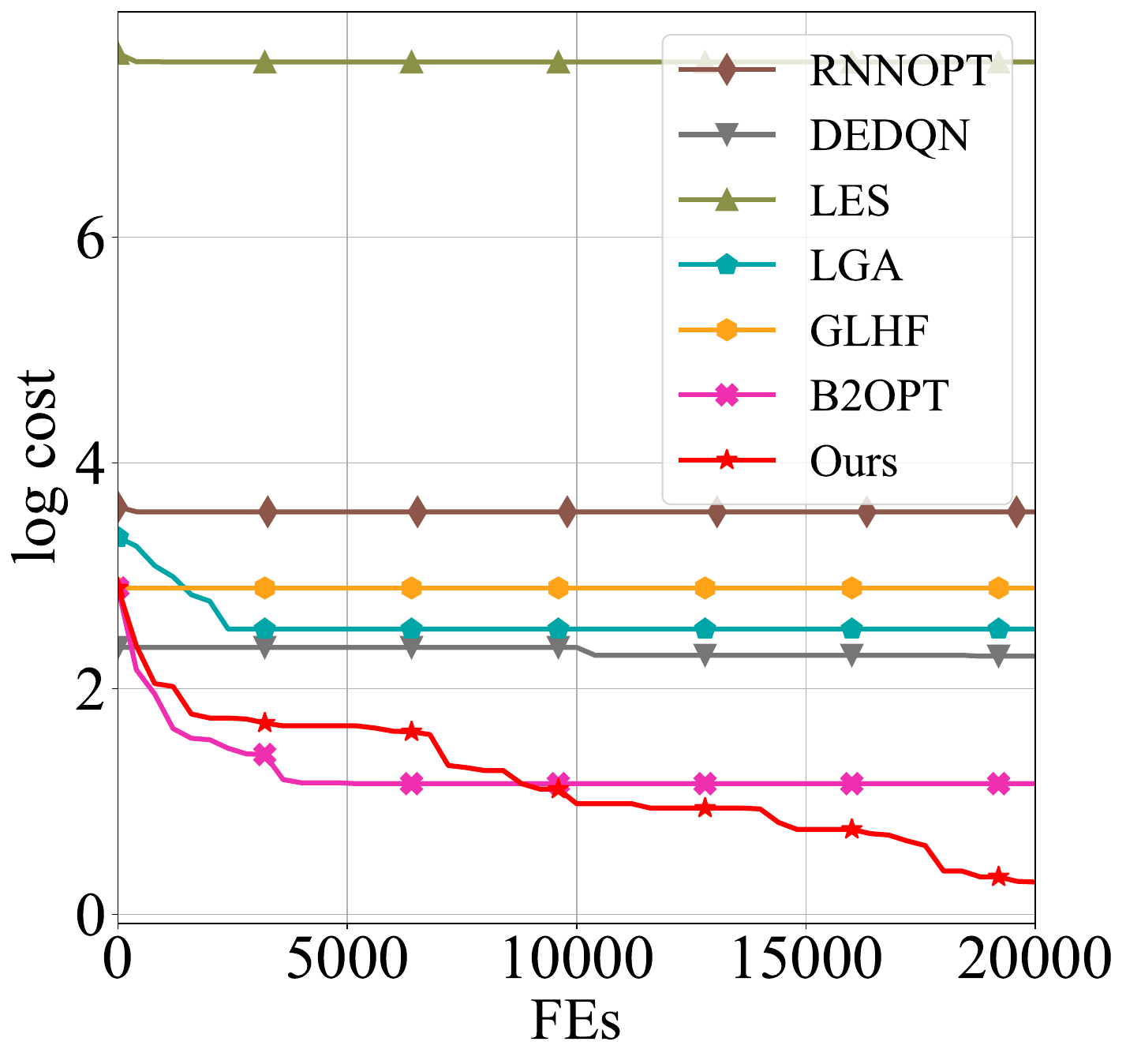}
		\caption*{Different Powers}
	\end{subfigure}
	
	\begin{subfigure}[b]{0.22\textwidth}
		\includegraphics[width=\linewidth]{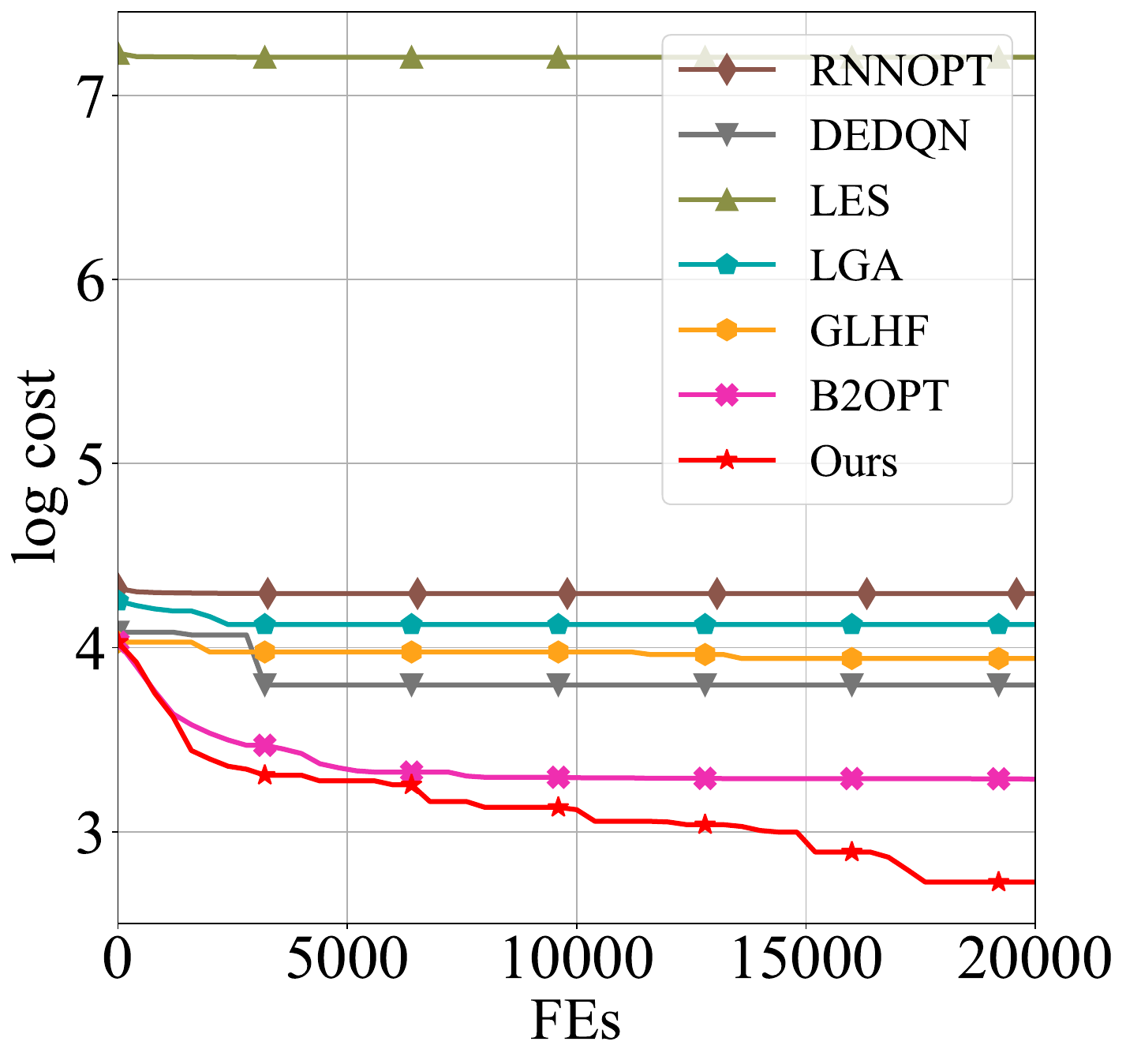}
		\caption*{Gallagher 21Peaks}
	\end{subfigure}
	\begin{subfigure}[b]{0.22\textwidth}
		\includegraphics[width=\linewidth]{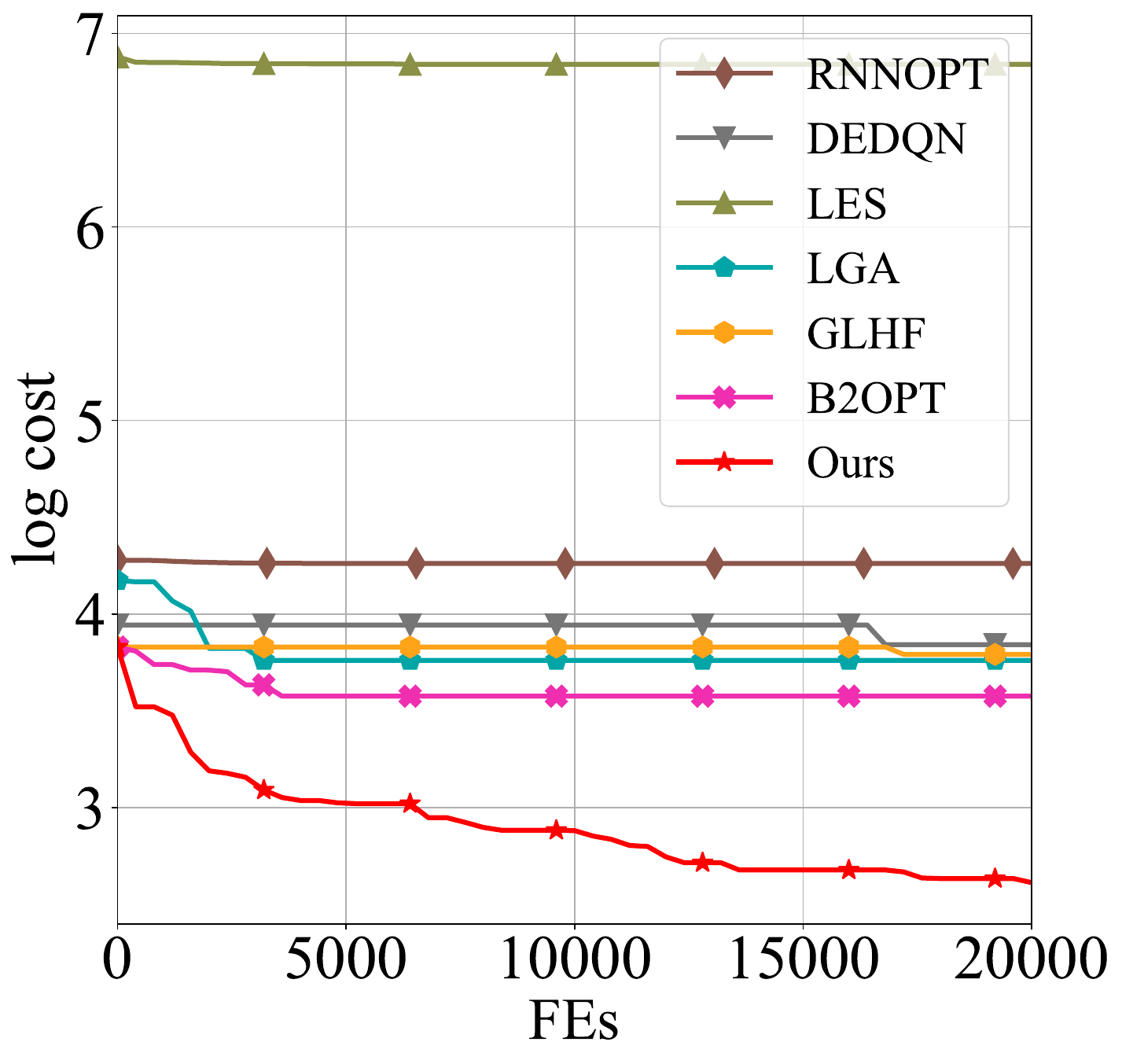}
		\caption*{Gallagher 101Peaks}
	\end{subfigure}
	\begin{subfigure}[b]{0.22\textwidth}
		\includegraphics[width=\linewidth]{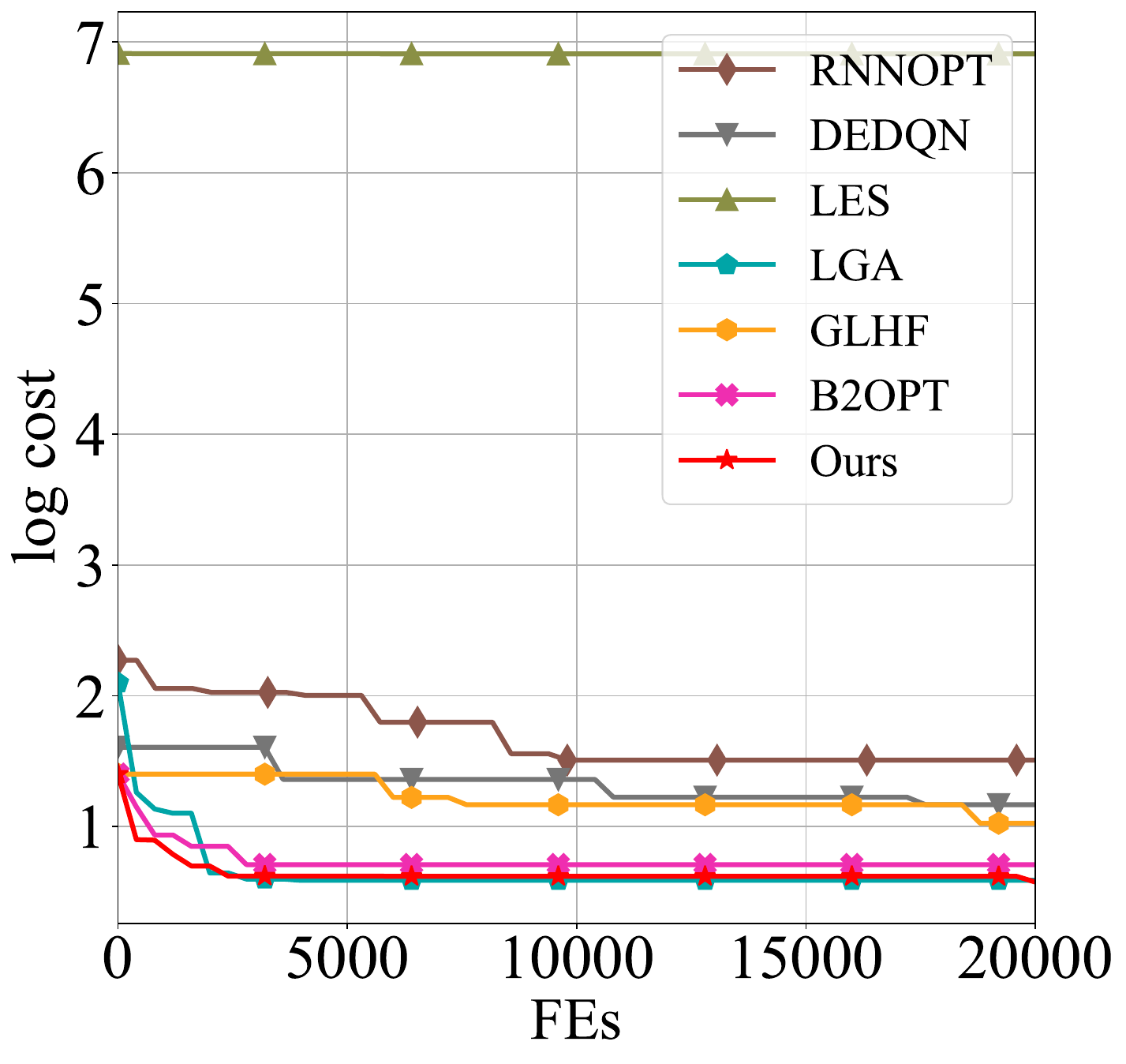}
		\caption*{Katsuura}
	\end{subfigure}
	\begin{subfigure}[b]{0.22\textwidth}
		\includegraphics[width=\linewidth]{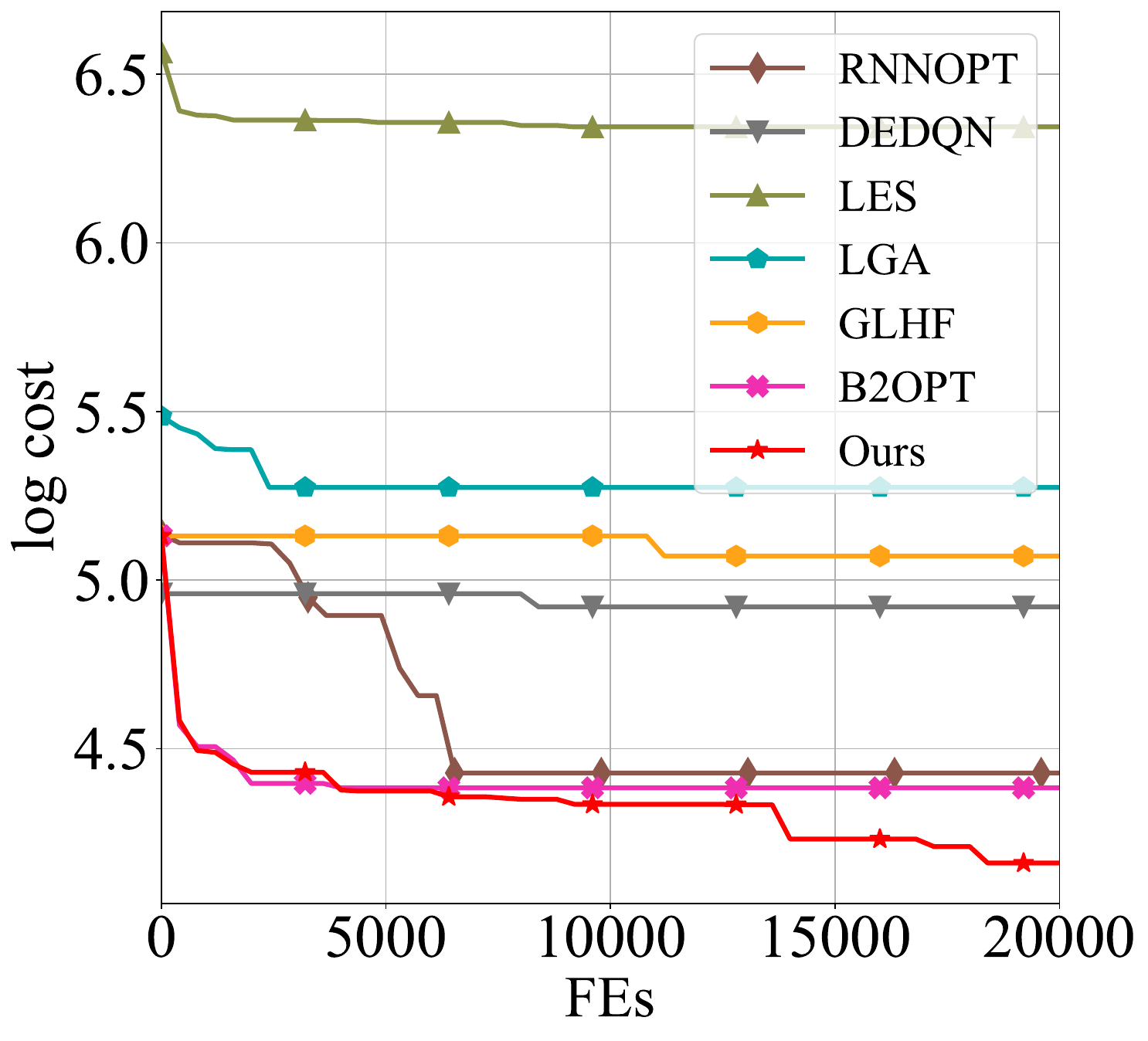}
		\caption*{Lunacek bi-Rastrigin}
	\end{subfigure}
	
	\begin{subfigure}[b]{0.22\textwidth}
		\includegraphics[width=\linewidth]{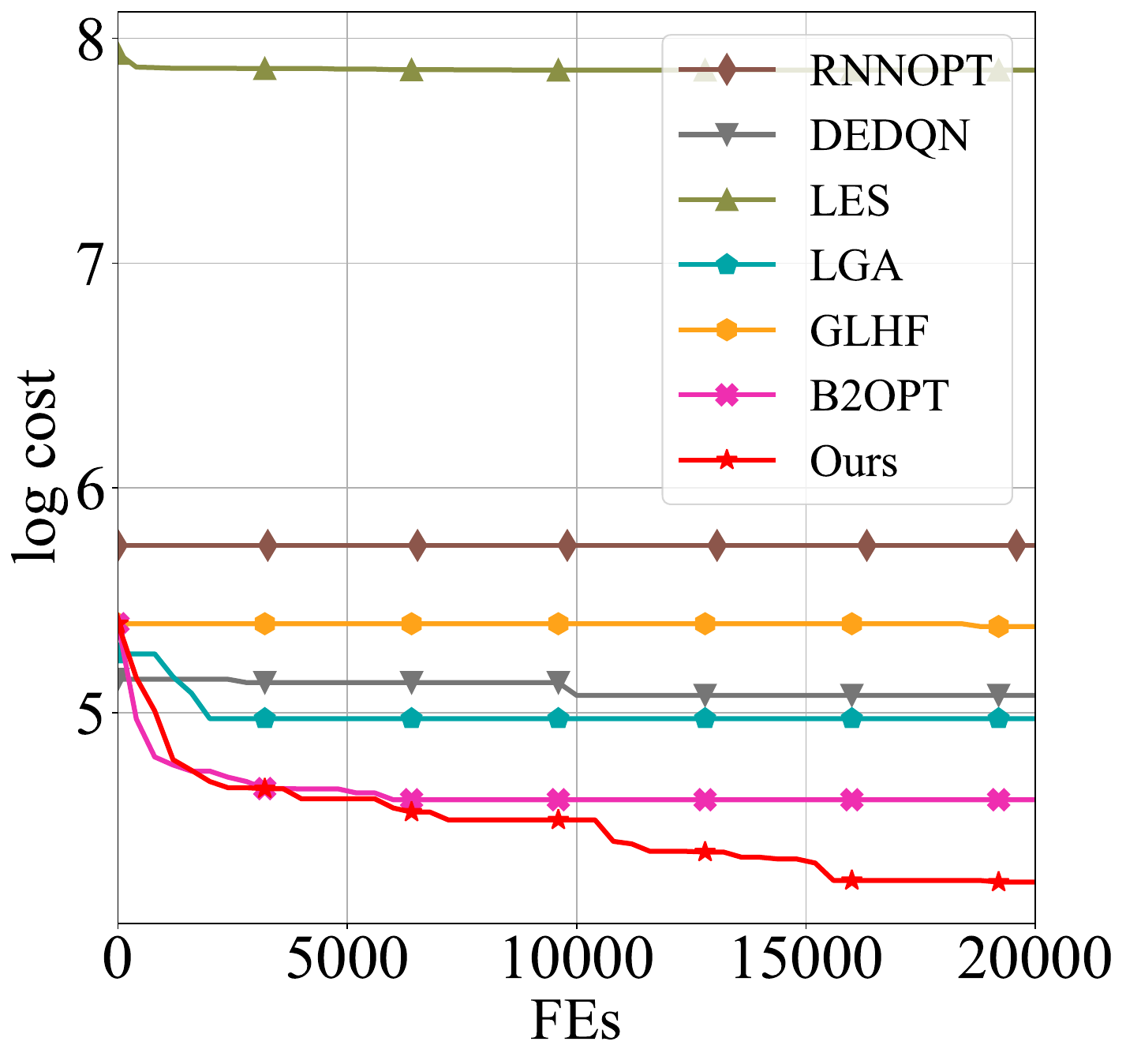}
		\caption*{Rastrigin F15}
	\end{subfigure}
	\begin{subfigure}[b]{0.22\textwidth}
		\includegraphics[width=\linewidth]{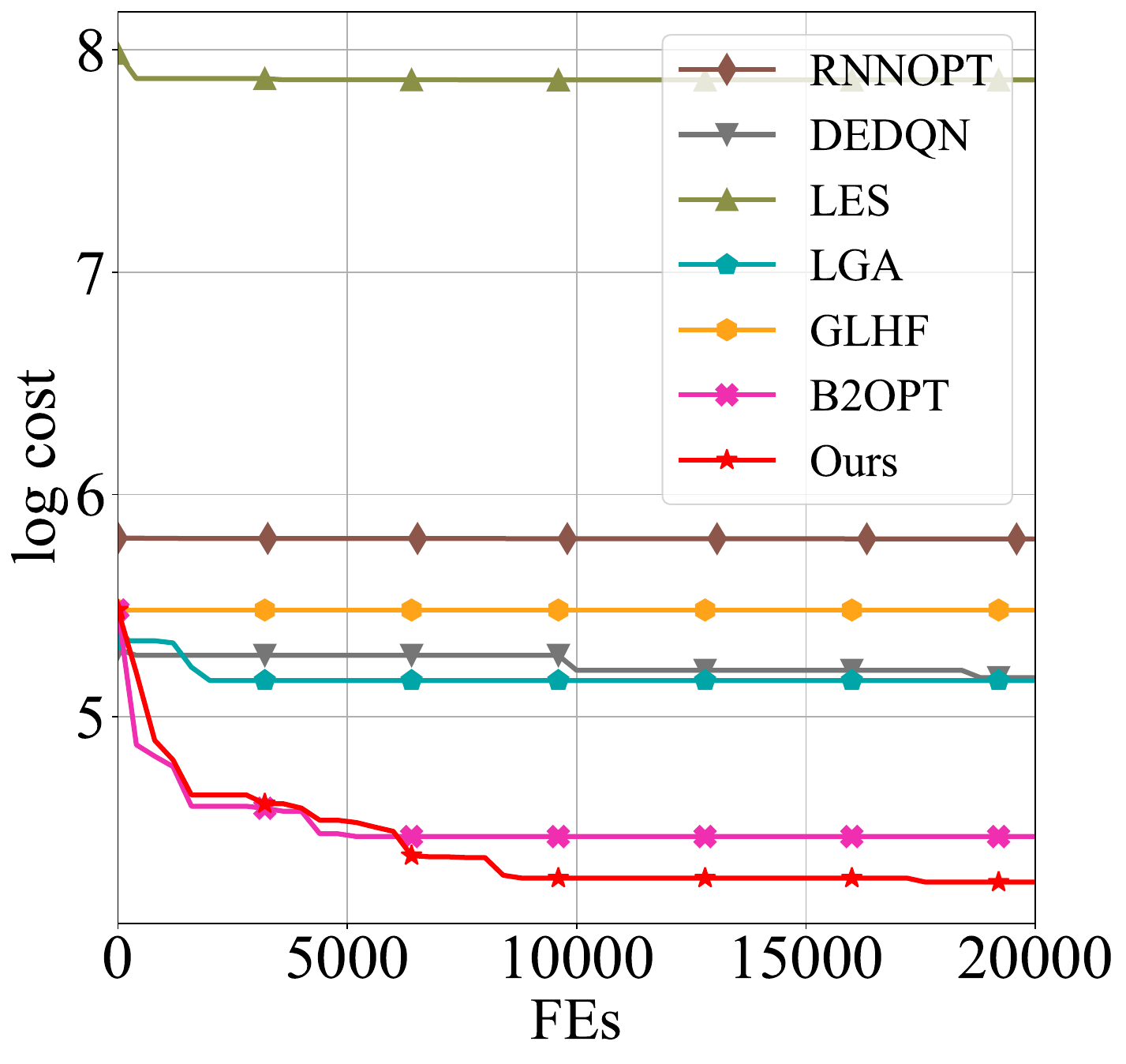}
		\caption*{Rastrigin}
	\end{subfigure}
	\begin{subfigure}[b]{0.22\textwidth}
		\includegraphics[width=\linewidth]{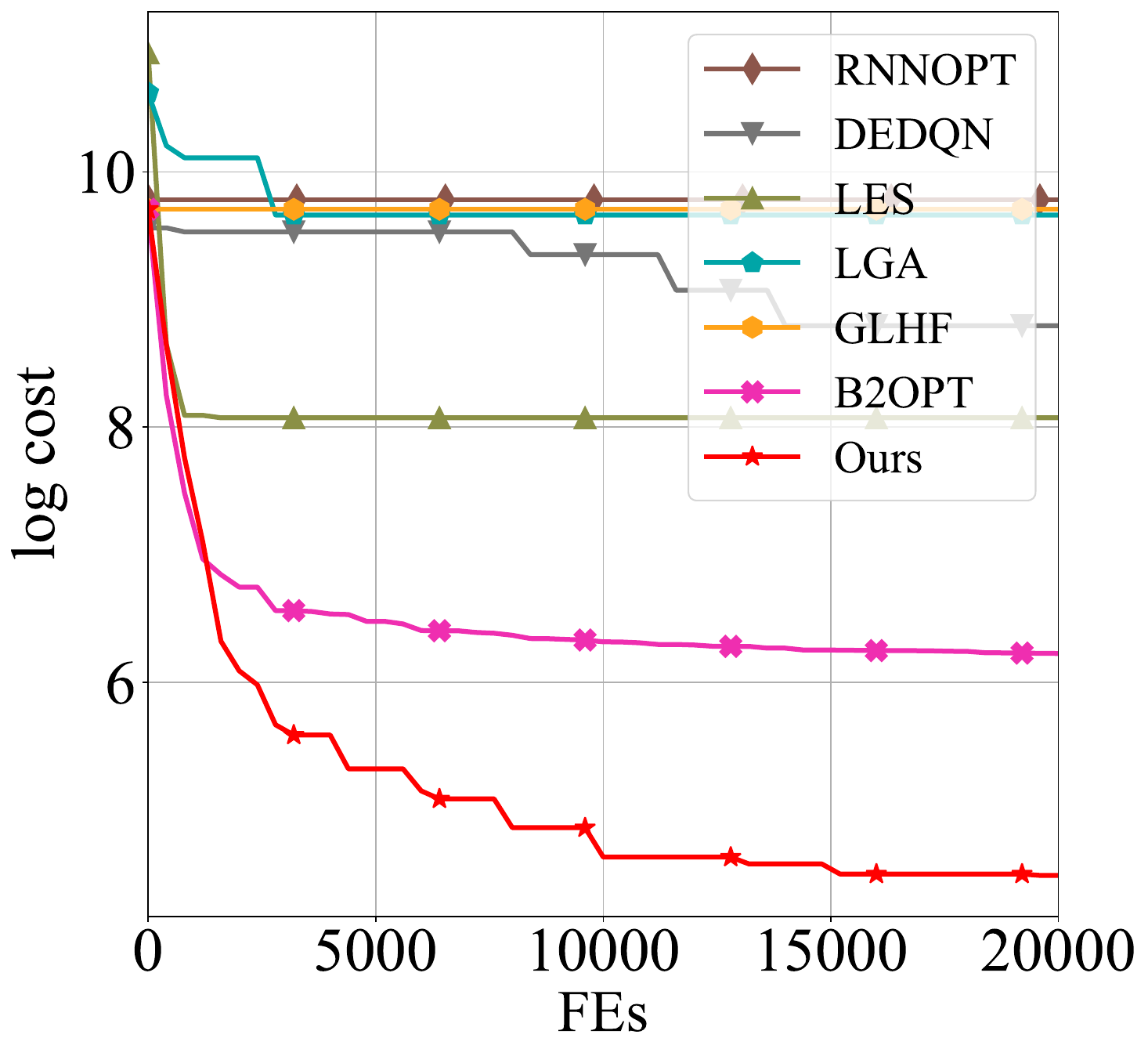}
		\caption*{Rosenbrock Original}
	\end{subfigure}
	\begin{subfigure}[b]{0.22\textwidth}
		\includegraphics[width=\linewidth]{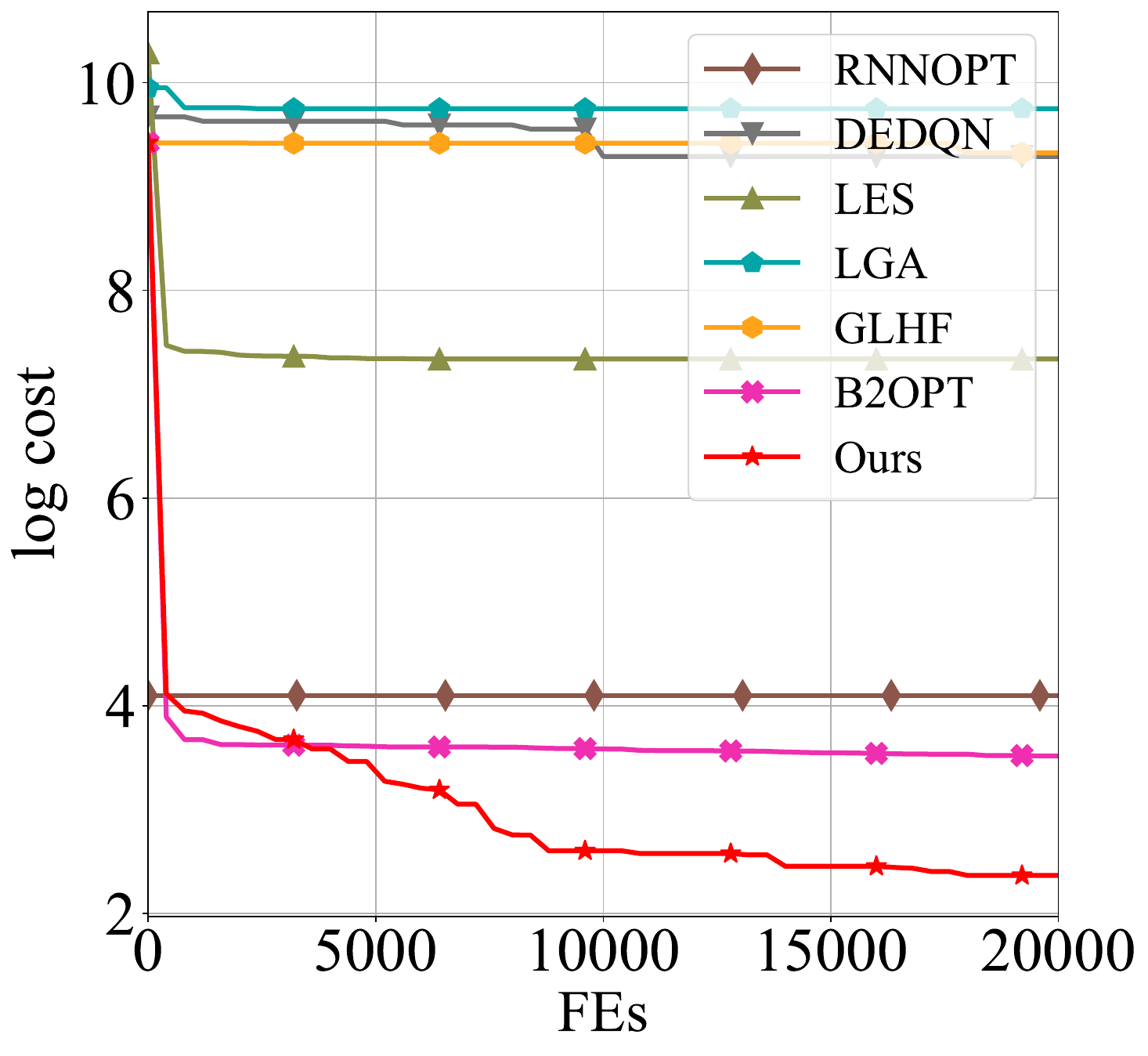}
		\caption*{Rosenbrock Rotated}
	\end{subfigure}
	
	\begin{subfigure}[b]{0.22\textwidth}
		\includegraphics[width=\linewidth]{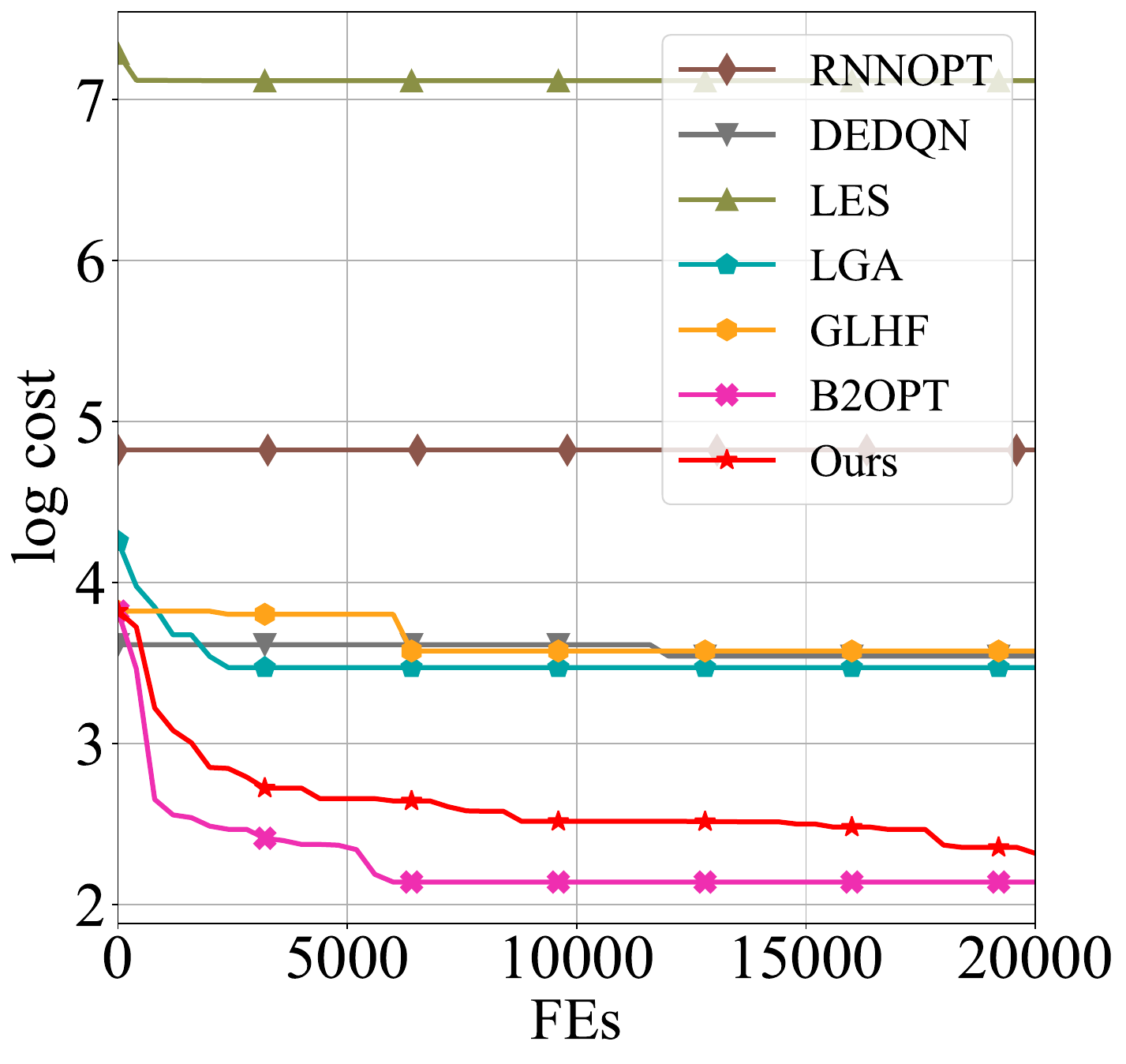}
		\caption*{Schaffers high cond}
	\end{subfigure}
	\begin{subfigure}[b]{0.22\textwidth}
		\includegraphics[width=\linewidth]{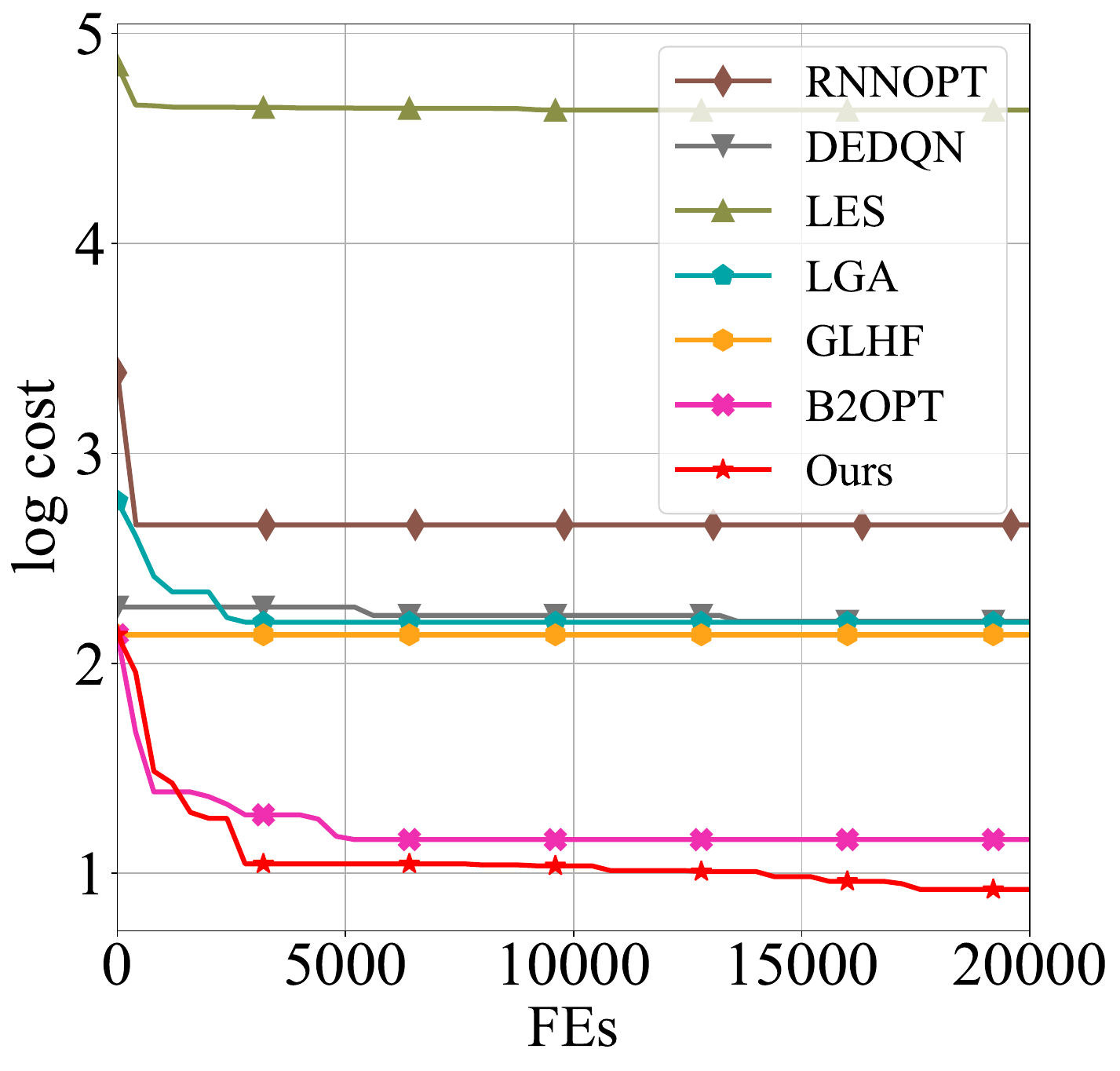}
		\caption*{Schaffers}
	\end{subfigure}
	\begin{subfigure}[b]{0.22\textwidth}
		\includegraphics[width=\linewidth]{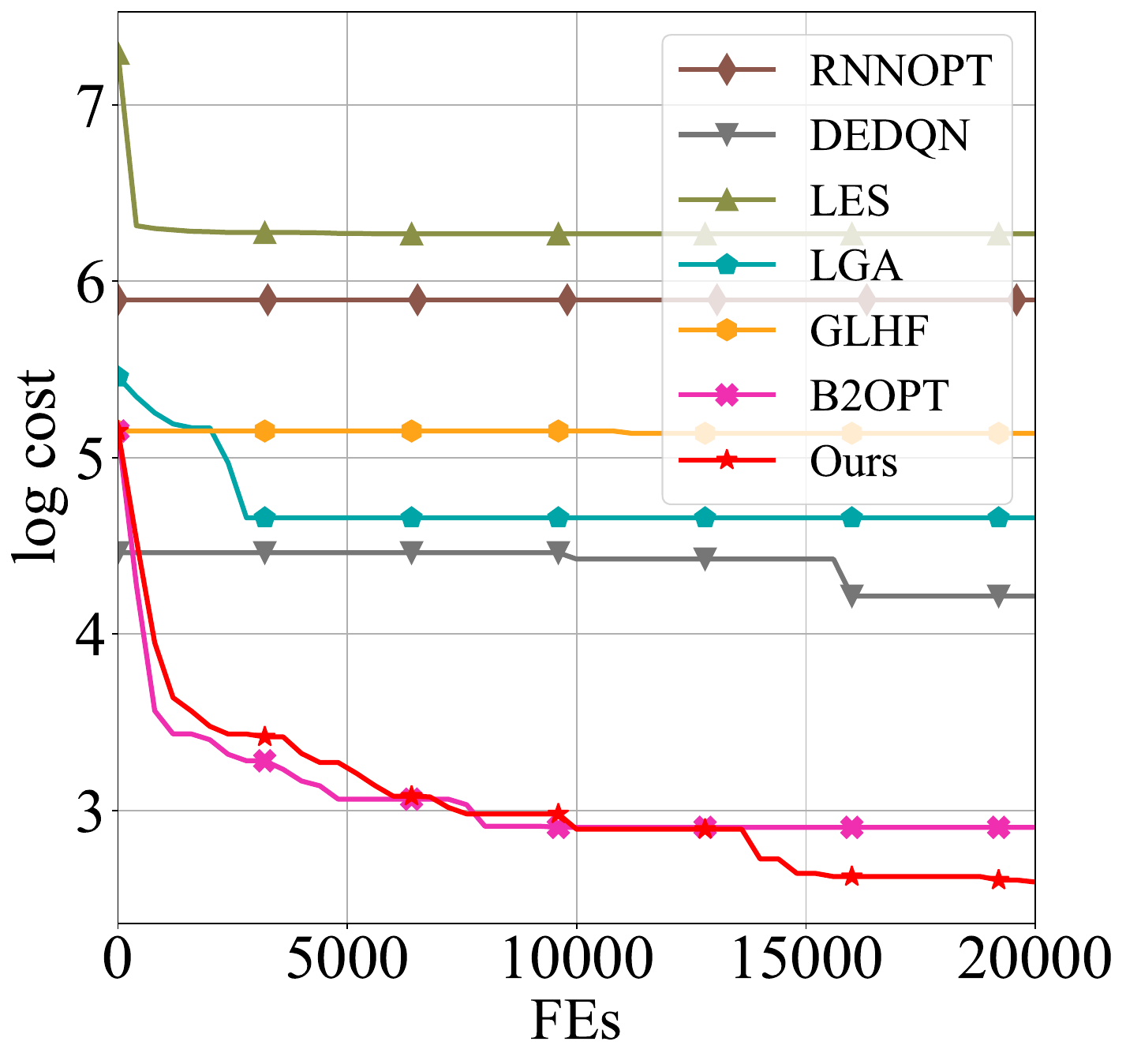}
		\caption*{Step Ellipsoidal}
	\end{subfigure}
	\begin{subfigure}[b]{0.22\textwidth}
		\includegraphics[width=\linewidth]{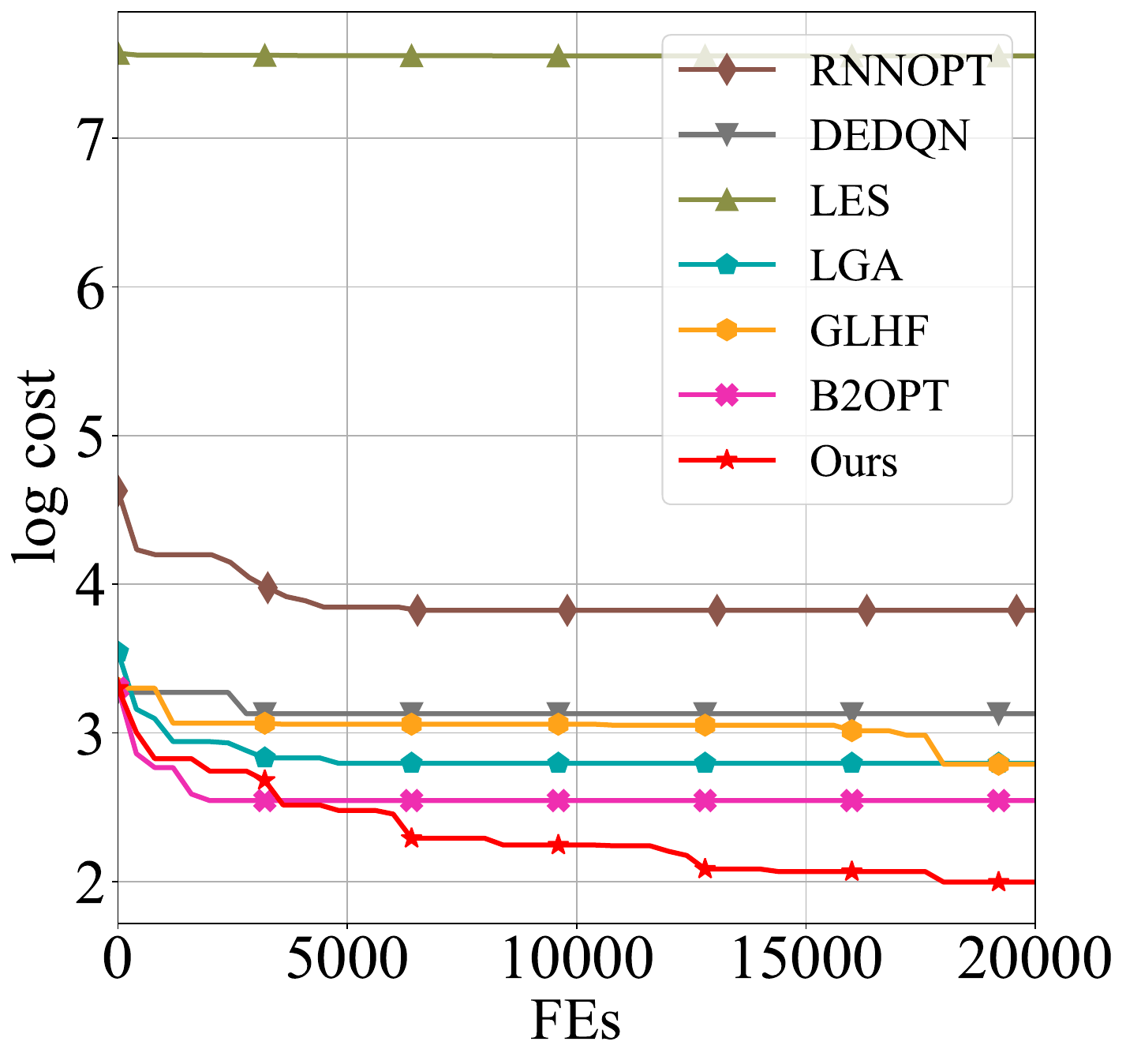}
		\caption*{Weierstrass}
	\end{subfigure}
	
	\caption{Out-of-distribution log-scaled convergence curves of various representative methods tested on \emph{BBOB-surrogate-10D}~\cite{hansen2021coco}.}
	\label{fig:supp-bbob-surrogate-10D}
\end{figure*}
}

\newcommand{\FigSupbbobThirtycurve}{%
\begin{figure*}[htbp]
	\centering
	
	\begin{subfigure}[b]{0.22\textwidth}
		\includegraphics[width=\linewidth]{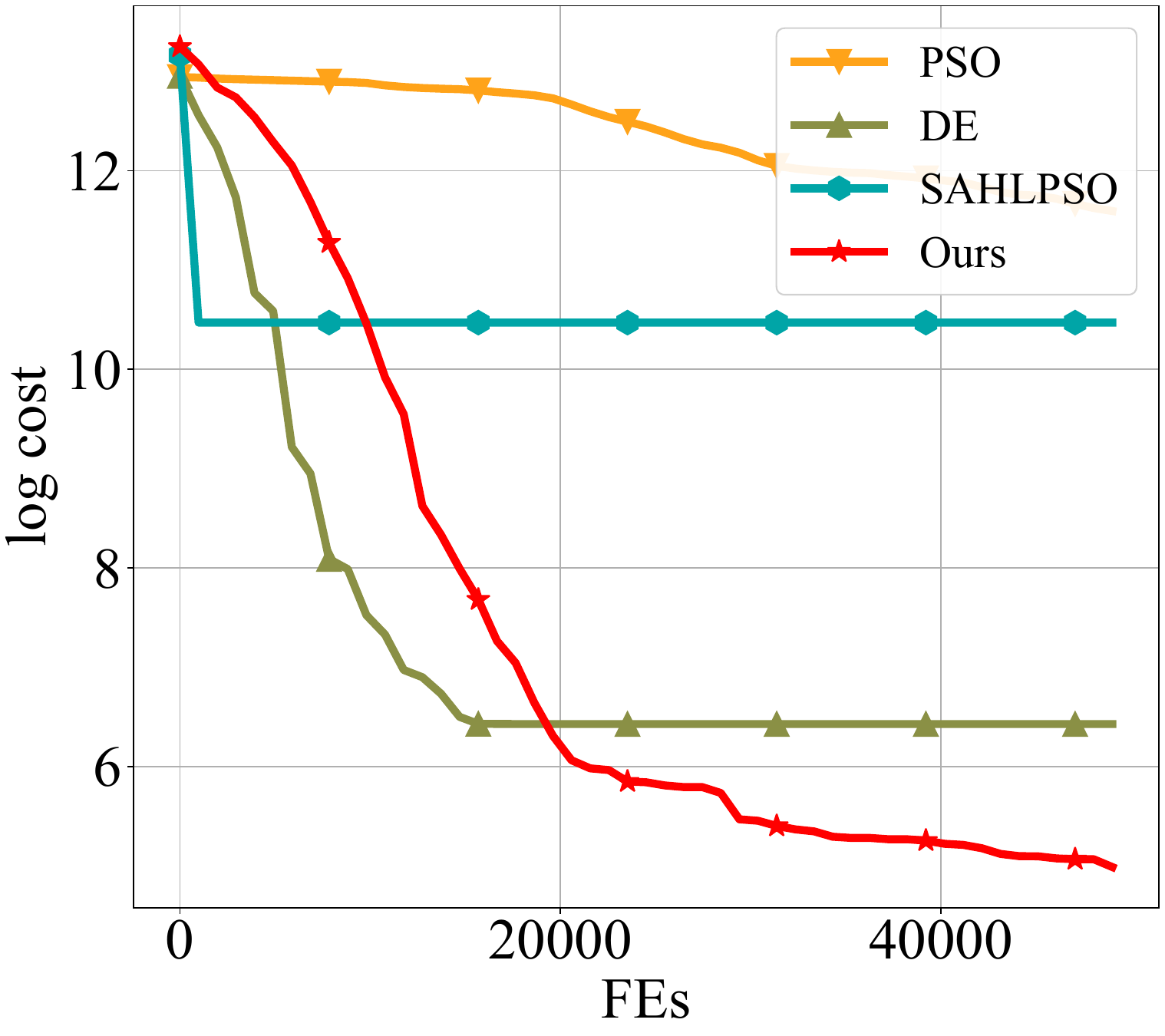}
		\caption*{Attractive Sector}
	\end{subfigure}
	\begin{subfigure}[b]{0.22\textwidth}
		\includegraphics[width=\linewidth]{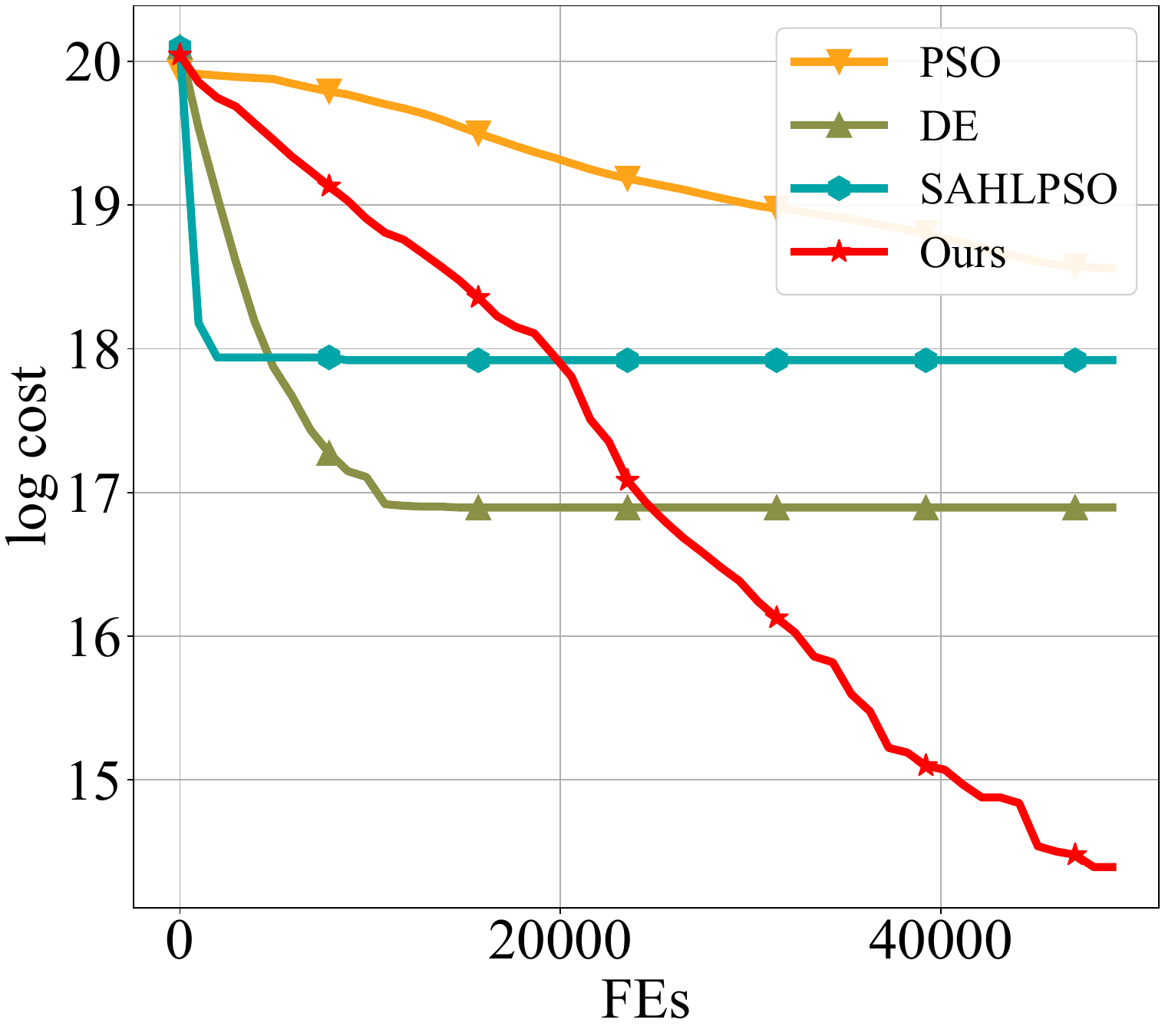}
		\caption*{Bent Cigar}
	\end{subfigure}
	\begin{subfigure}[b]{0.22\textwidth}
		\includegraphics[width=\linewidth]{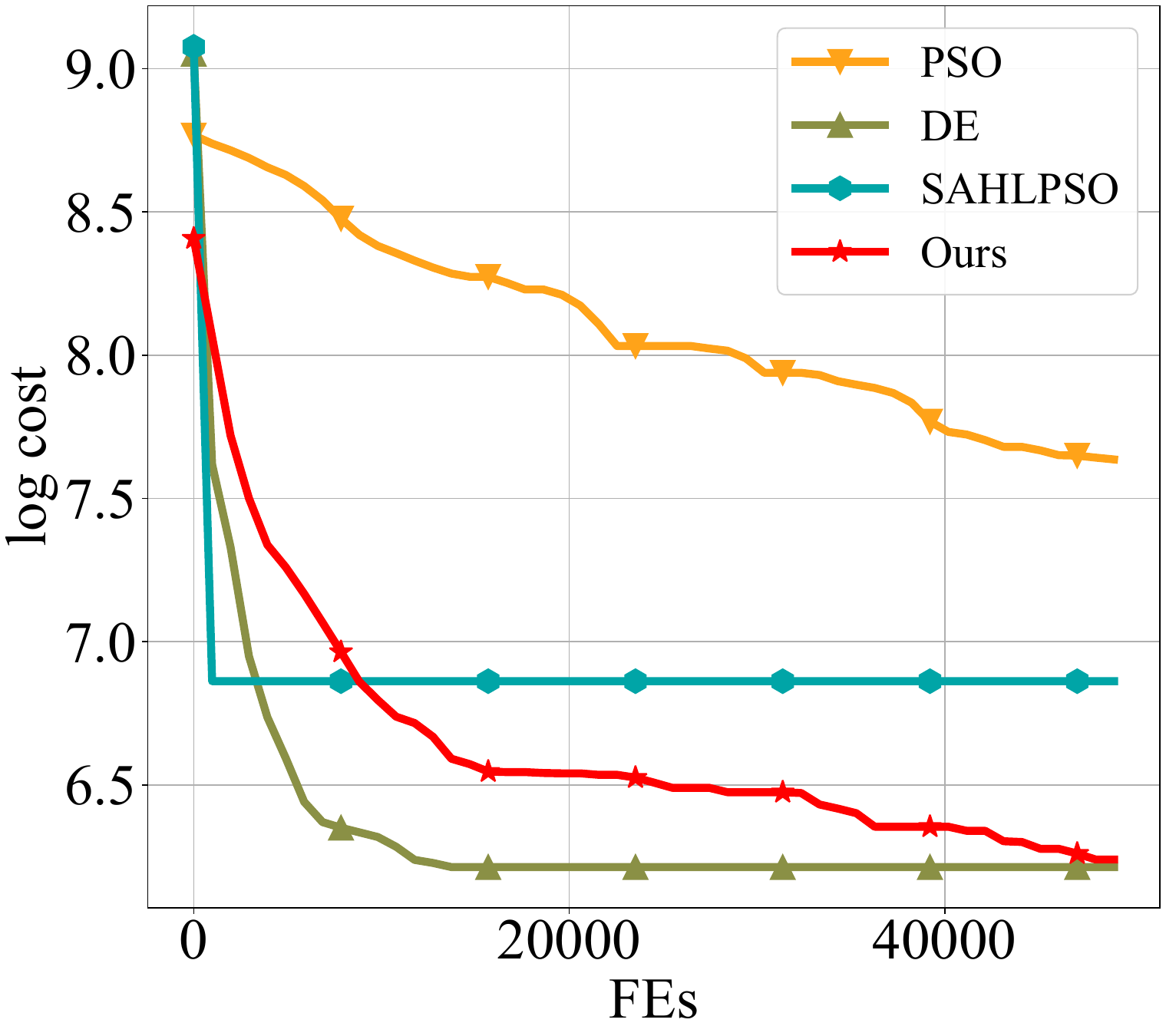}
		\caption*{Buche Rastrigin}
	\end{subfigure}
	\begin{subfigure}[b]{0.22\textwidth}
		\includegraphics[width=\linewidth]{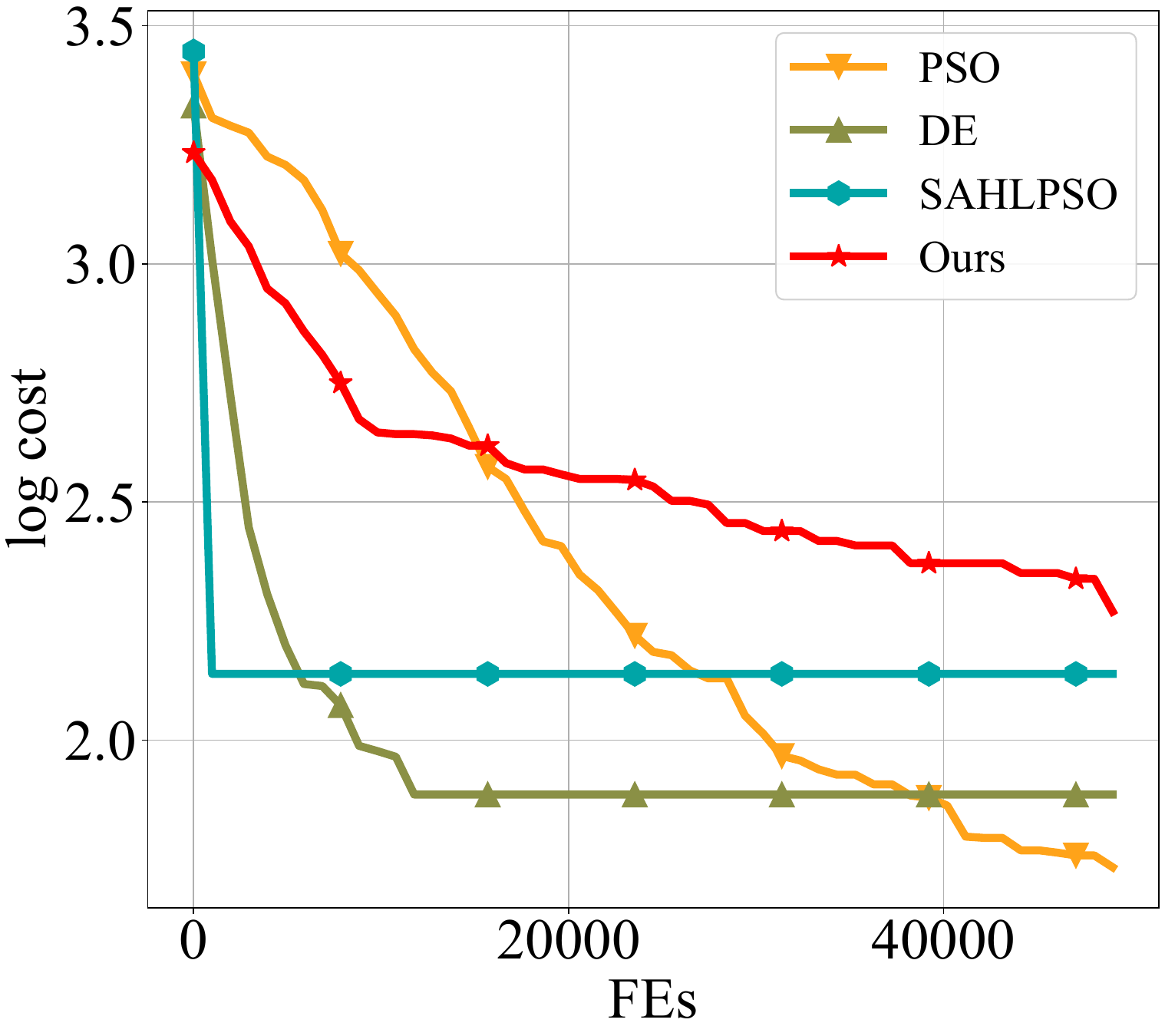}
		\caption*{Composite Grie-Rosen}
	\end{subfigure}
	
	\begin{subfigure}[b]{0.22\textwidth}
		\includegraphics[width=\linewidth]{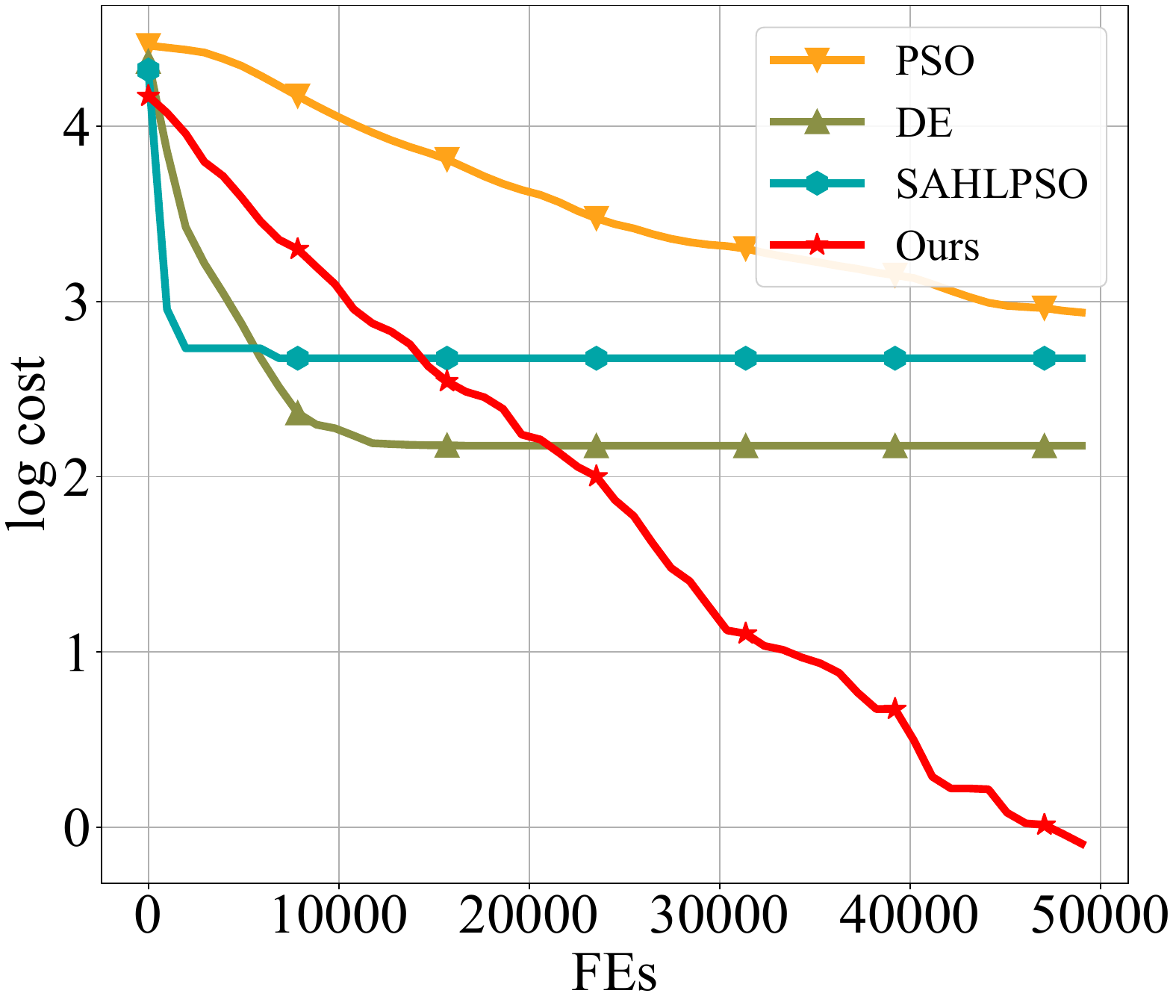}
		\caption*{Different Powers}
	\end{subfigure}
	\begin{subfigure}[b]{0.22\textwidth}
		\includegraphics[width=\linewidth]{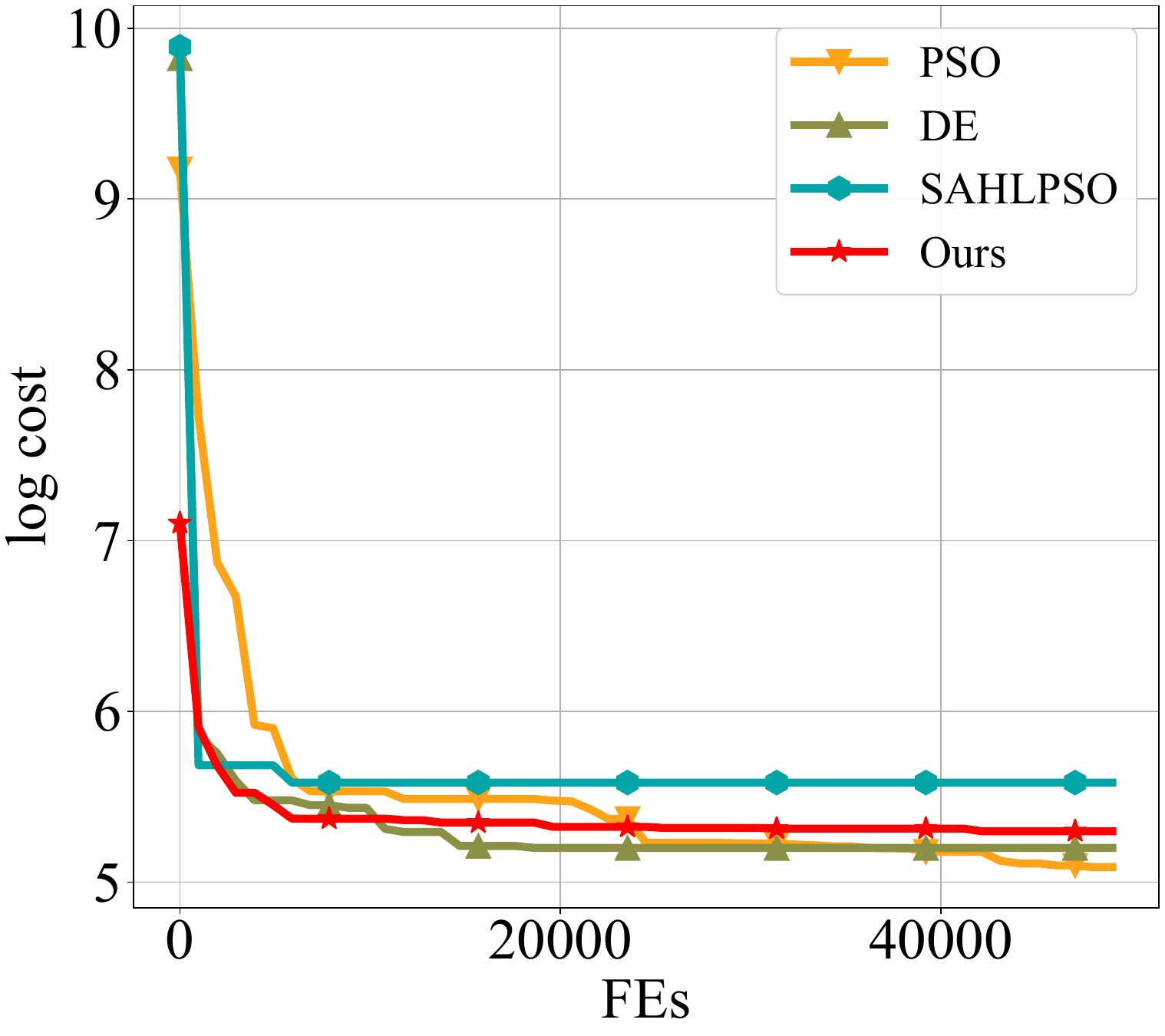}
		\caption*{Discus}
	\end{subfigure}
	\begin{subfigure}[b]{0.22\textwidth}
		\includegraphics[width=\linewidth]{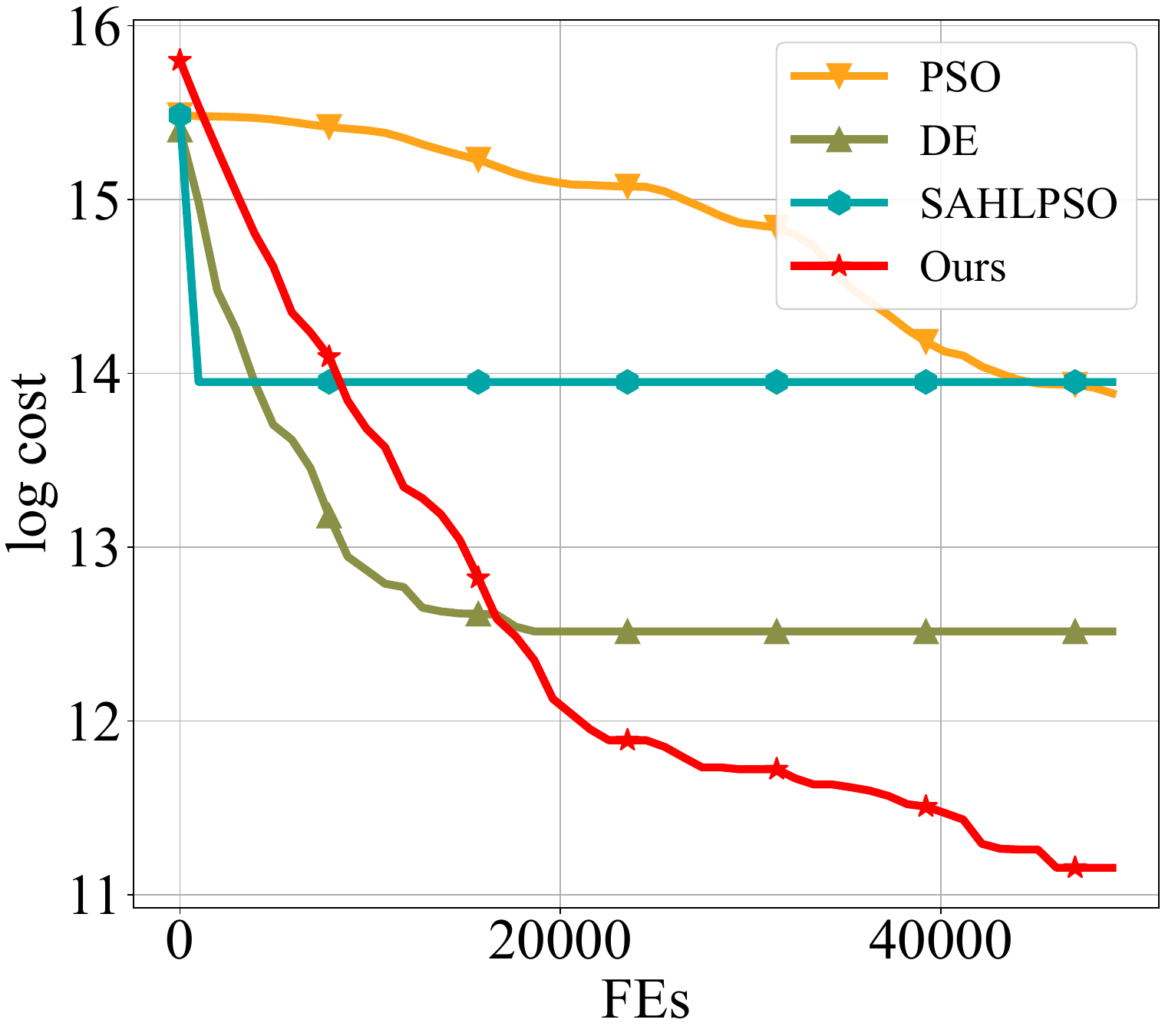}
		\caption*{Ellipsoidal}
	\end{subfigure}
	\begin{subfigure}[b]{0.22\textwidth}
		\includegraphics[width=\linewidth]{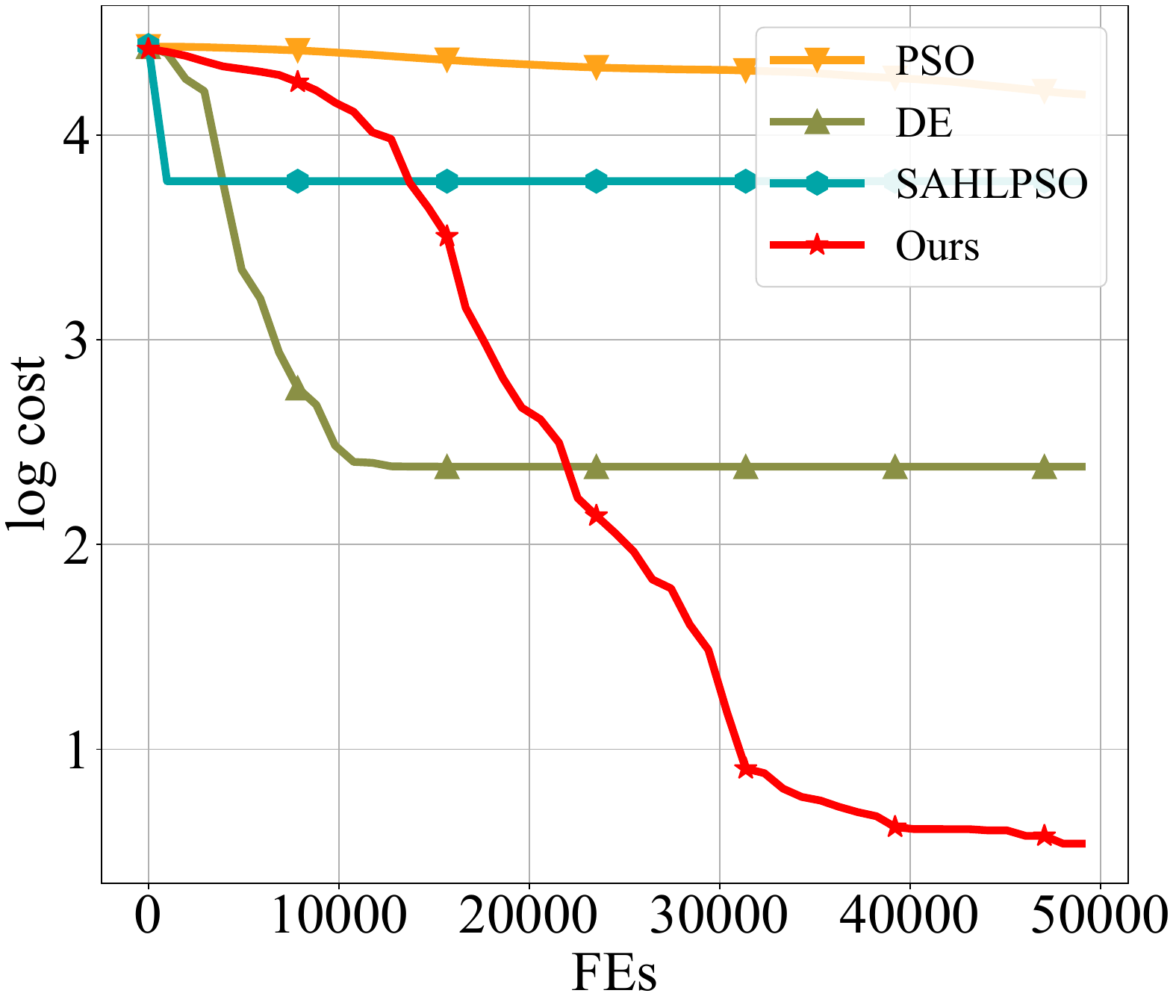}
		\caption*{Gallagher 21Peaks}
	\end{subfigure}
	
	\begin{subfigure}[b]{0.22\textwidth}
		\includegraphics[width=\linewidth]{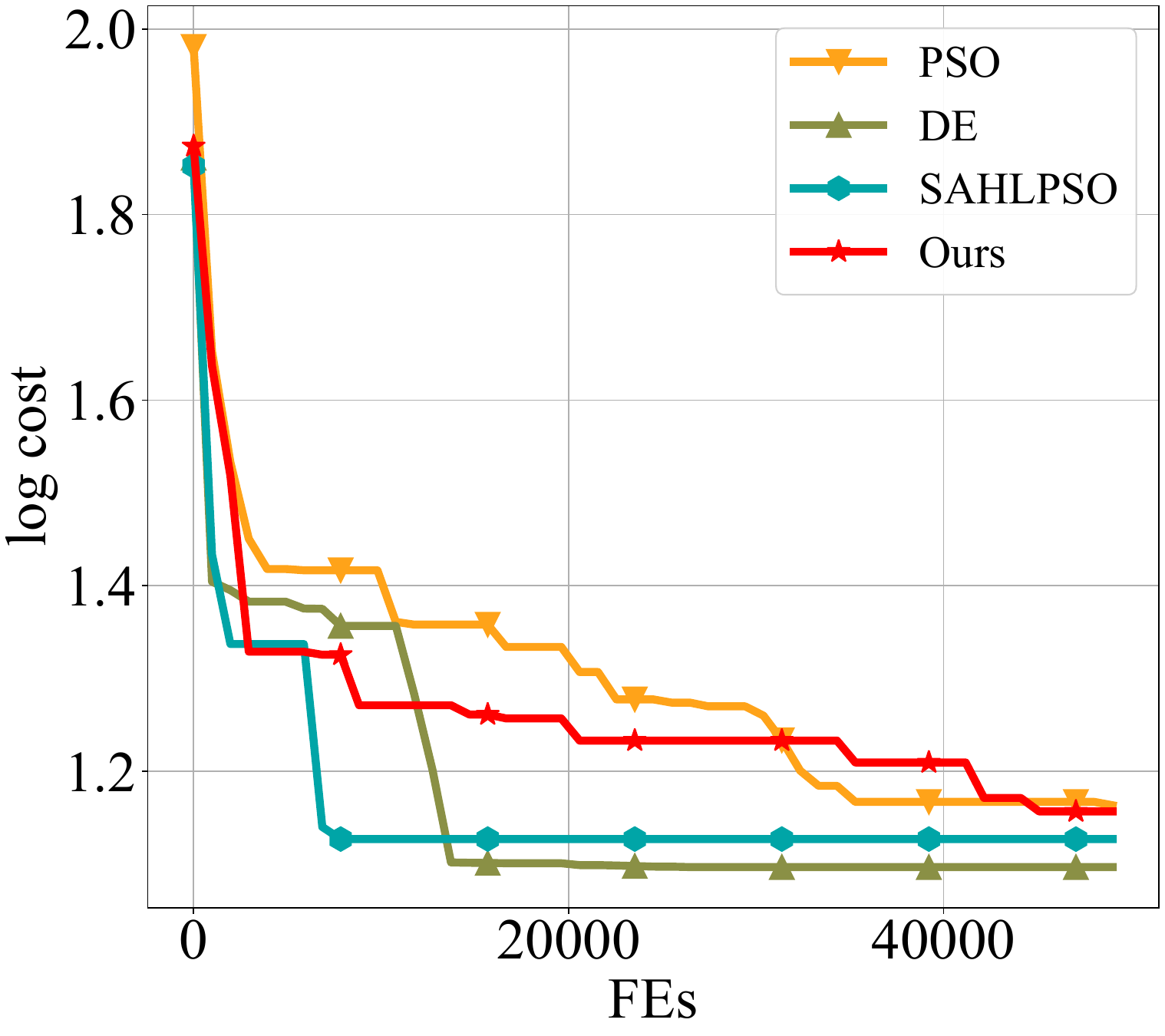}
		\caption*{Katsuura}
	\end{subfigure}
	\begin{subfigure}[b]{0.22\textwidth}
		\includegraphics[width=\linewidth]{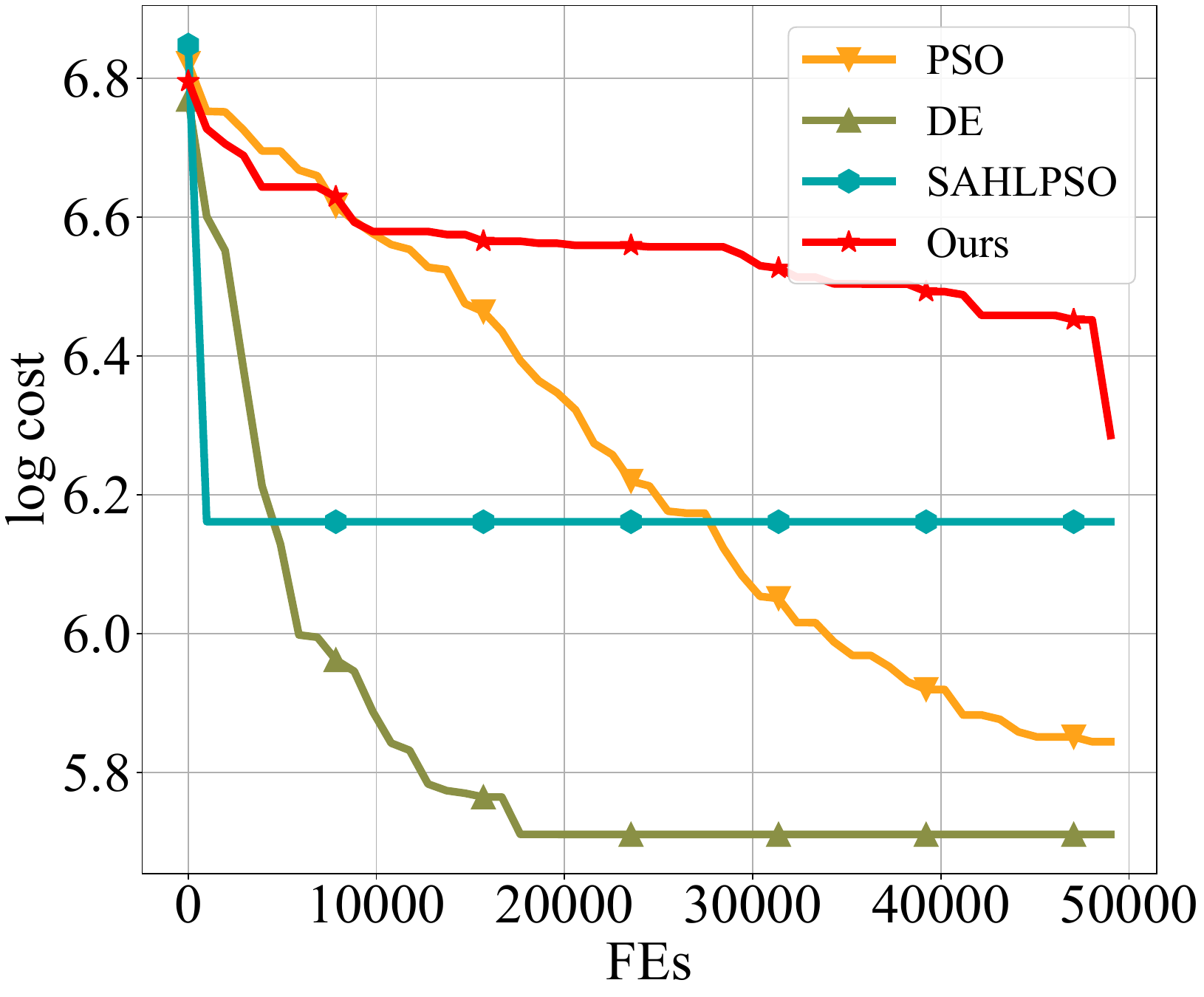}
		\caption*{Lunacek bi-Rastrigin}
	\end{subfigure}
	\begin{subfigure}[b]{0.22\textwidth}
		\includegraphics[width=\linewidth]{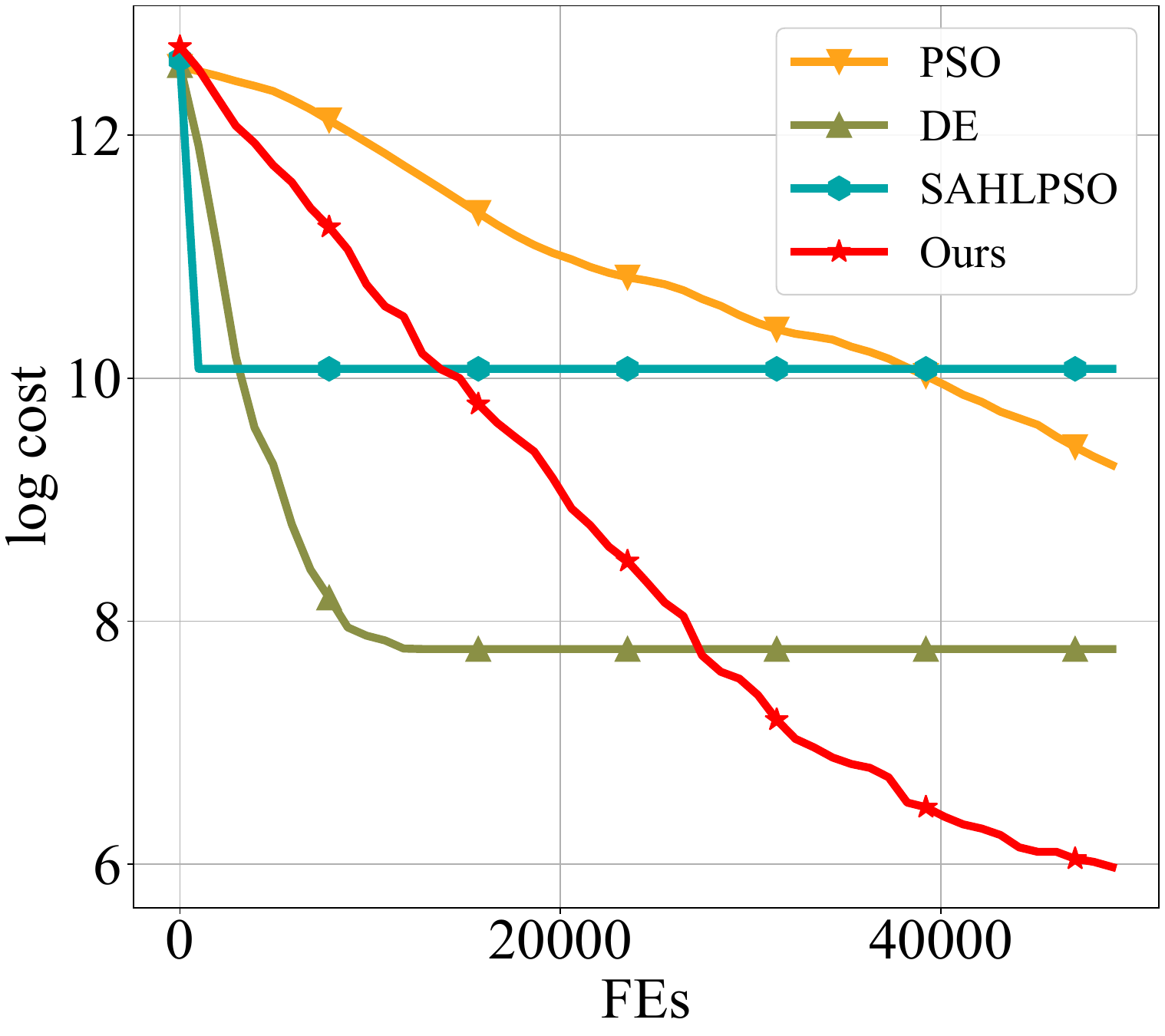}
		\caption*{Rosenbrock Original}
	\end{subfigure}
	\begin{subfigure}[b]{0.22\textwidth}
		\includegraphics[width=\linewidth]{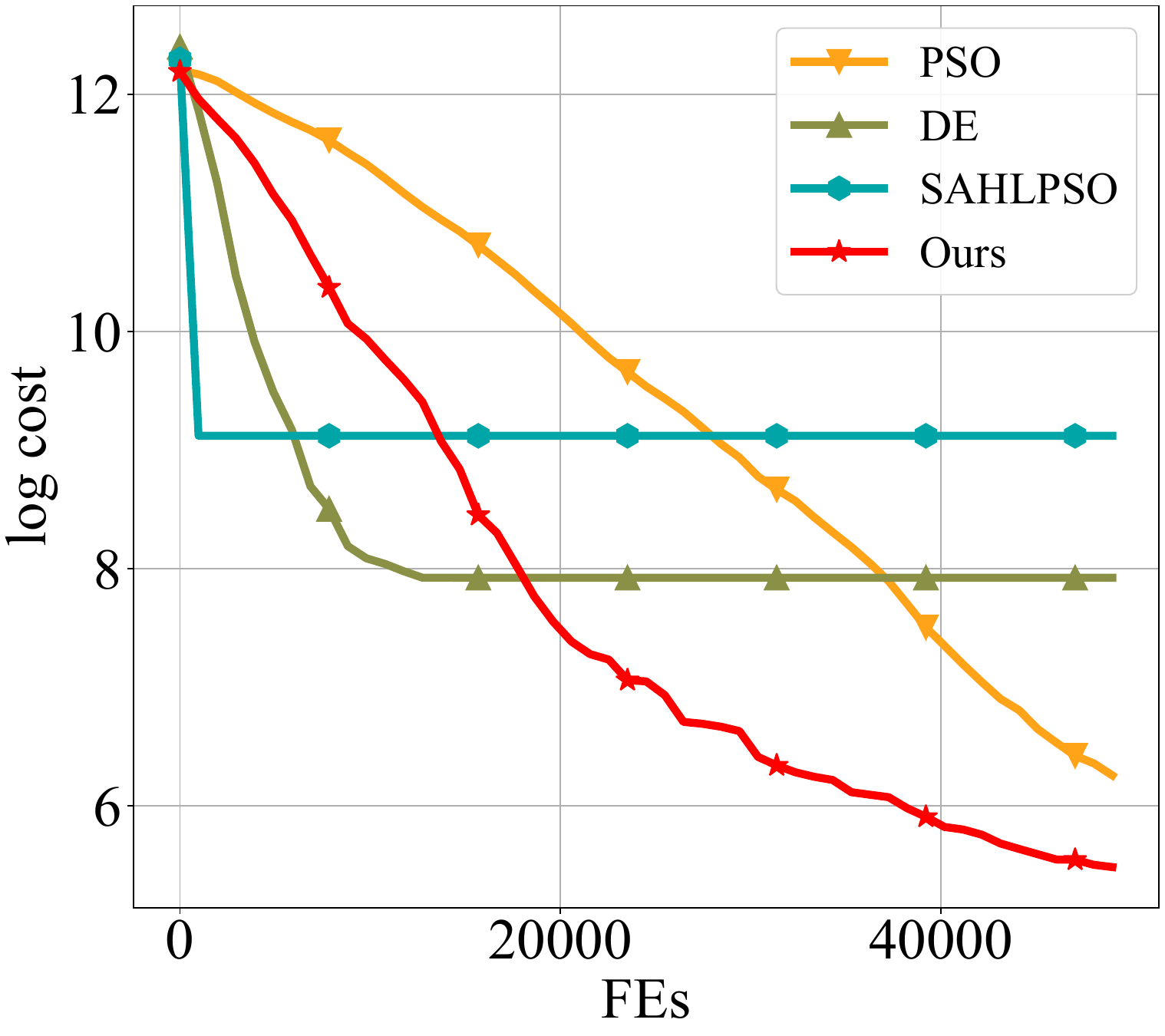}
		\caption*{Rosenbrock Rotated}
	\end{subfigure}
	
	\begin{subfigure}[b]{0.22\textwidth}
		\includegraphics[width=\linewidth]{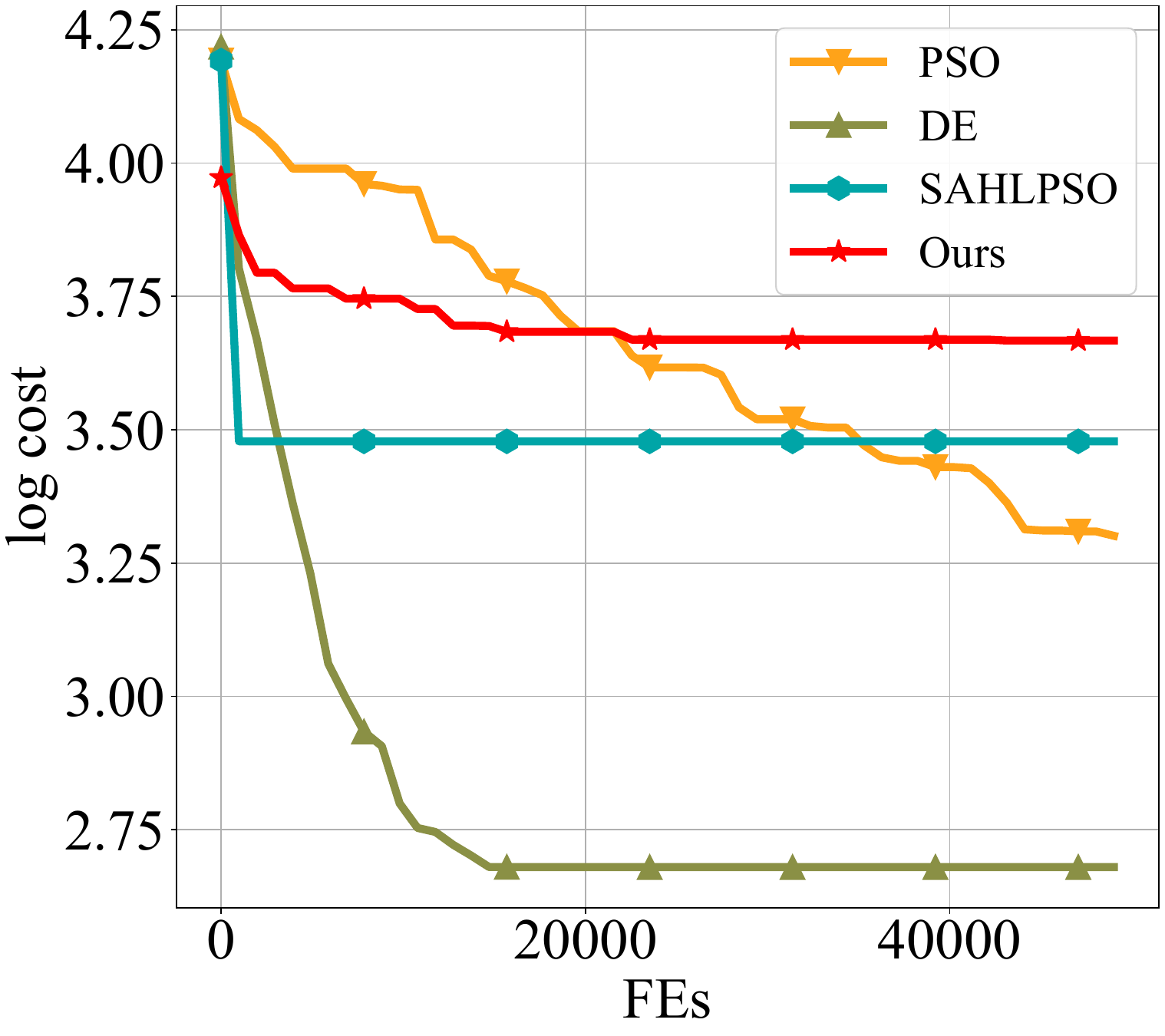}
		\caption*{Schaffers}
	\end{subfigure}
	\begin{subfigure}[b]{0.22\textwidth}
		\includegraphics[width=\linewidth]{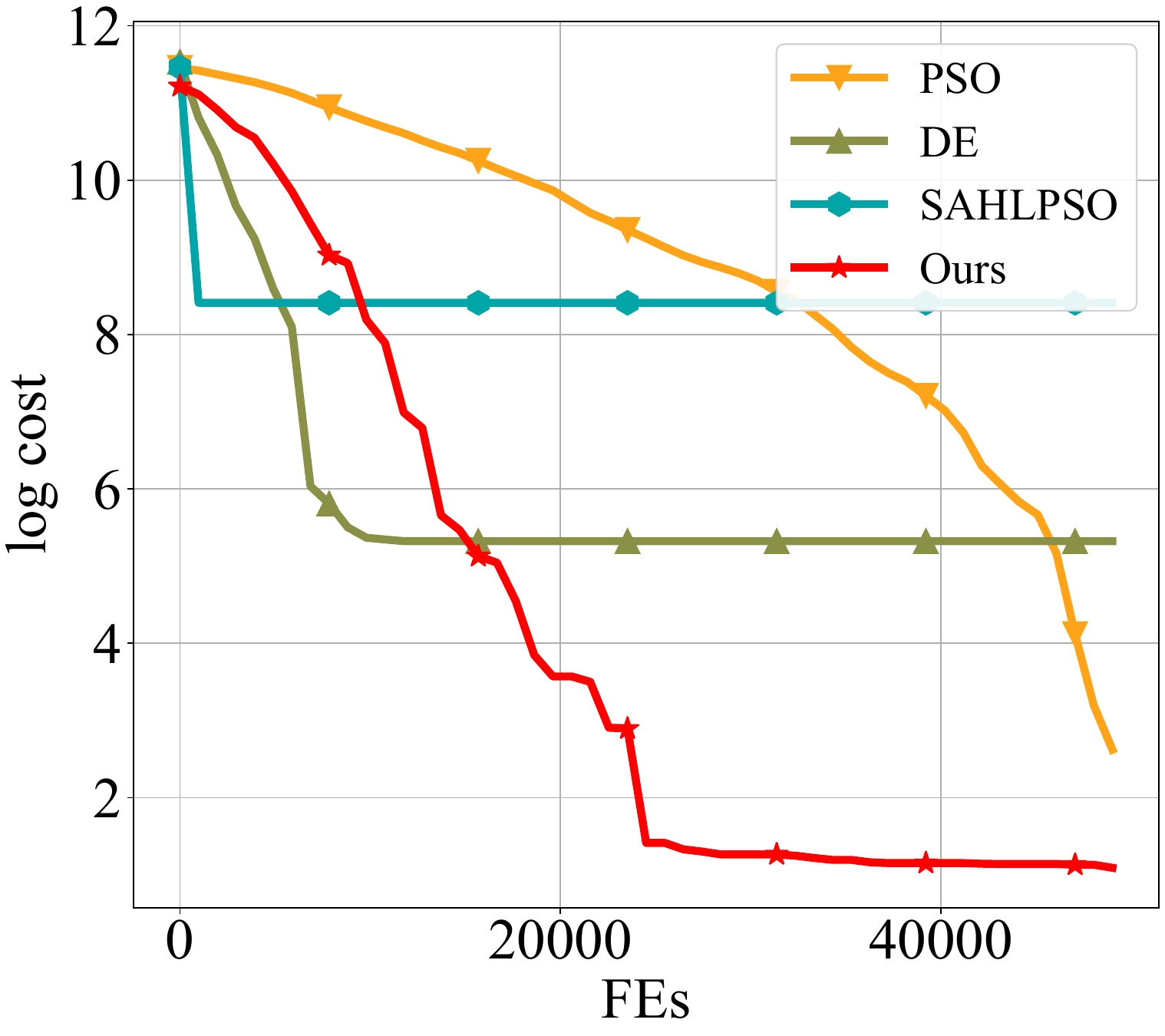}
		\caption*{Schwefel}
	\end{subfigure}
	\begin{subfigure}[b]{0.22\textwidth}
		\includegraphics[width=\linewidth]{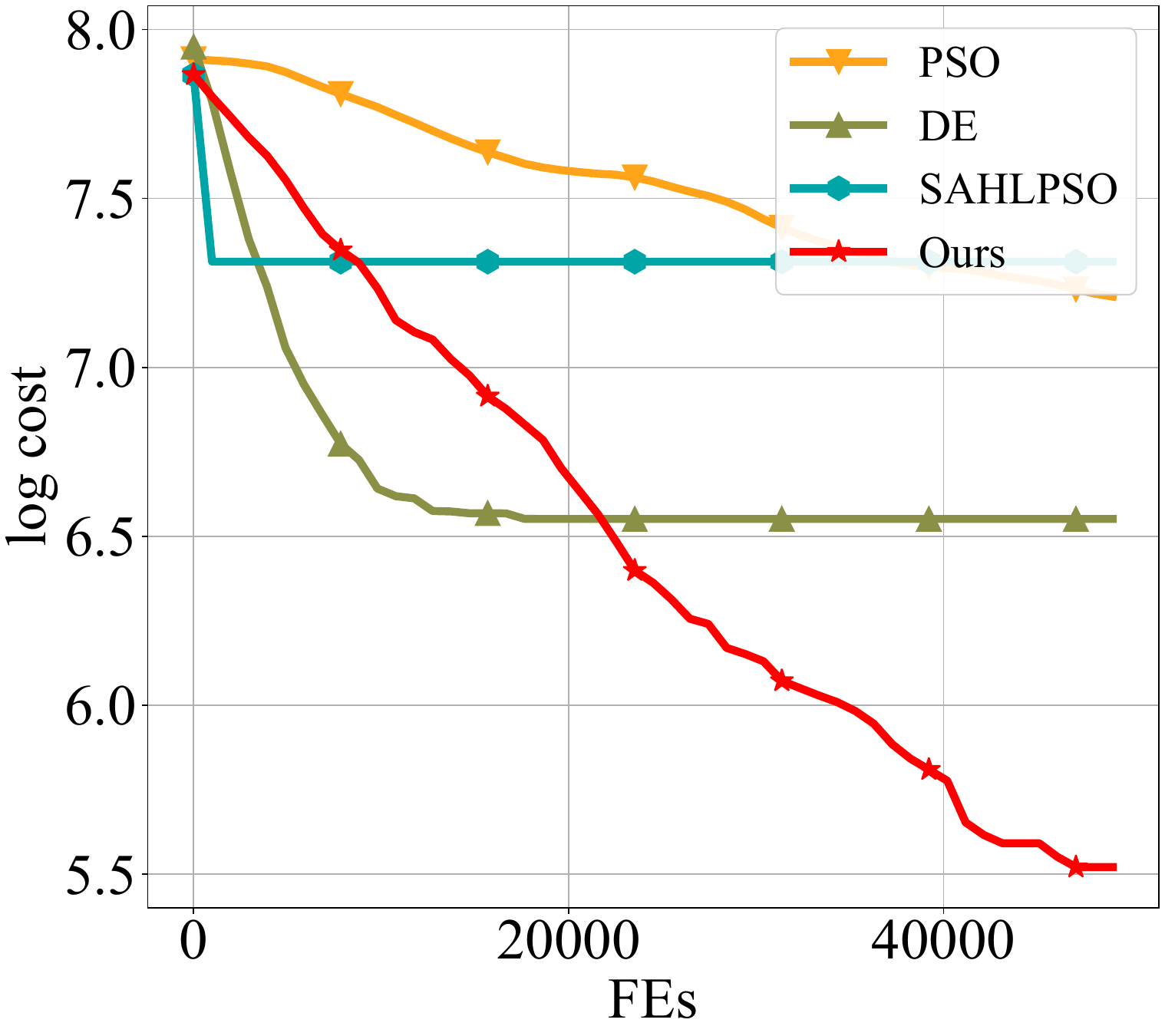}
		\caption*{Sharp Ridge}
	\end{subfigure}
	\begin{subfigure}[b]{0.22\textwidth}
		\includegraphics[width=\linewidth]{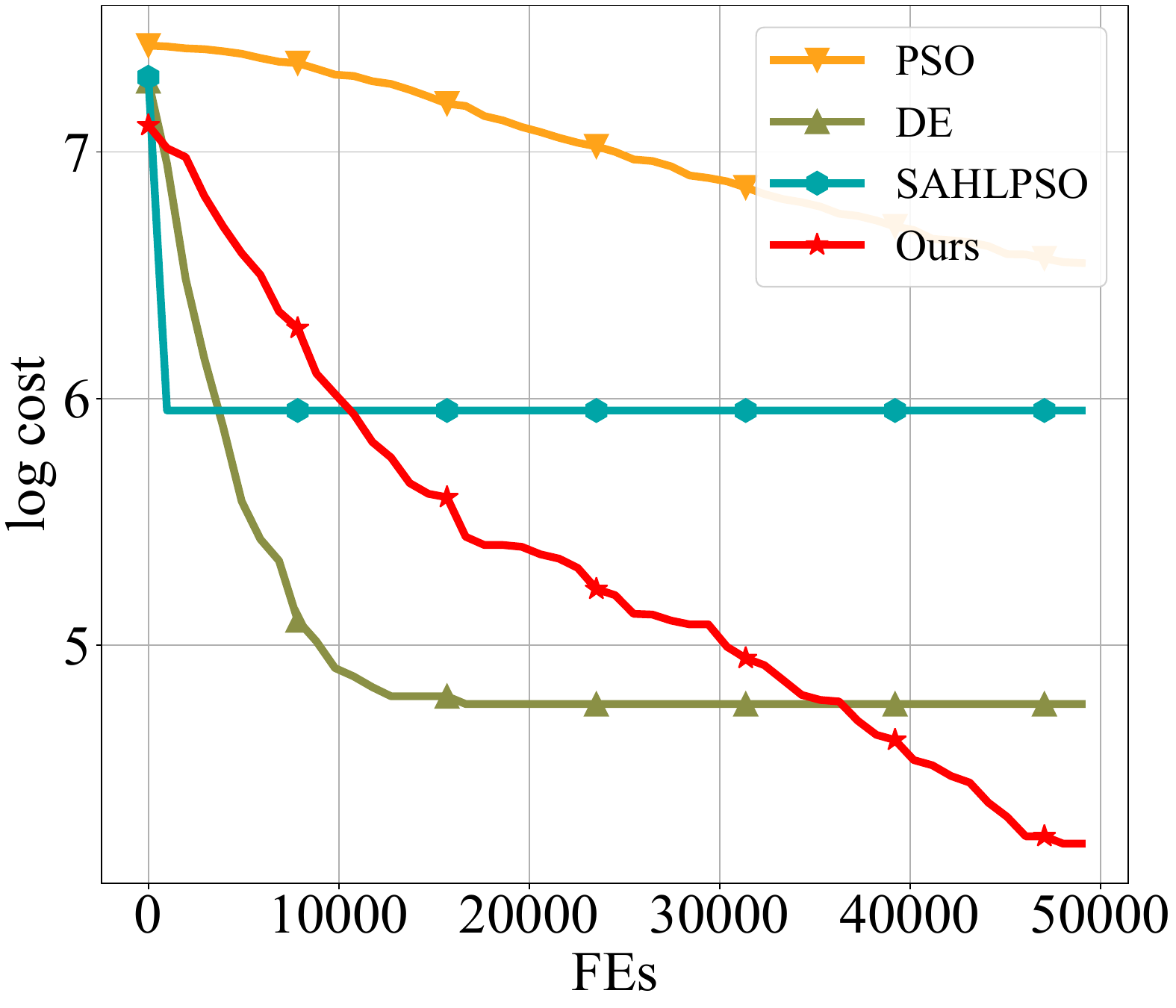}
		\caption*{Step Ellipsoidal}
	\end{subfigure}
	
	\caption{Log-scaled convergence curves of various traditional representative EA methods on \emph{BBOB-30D}~\cite{hansen2021coco}.}
	\label{fig:supp-bbob-30d-diff-16-traditional}
\end{figure*}
}

\newcommand{\FigSupbbobThirtycurvelog}{%
\begin{figure*}[htbp]
	\centering
	
	\begin{subfigure}[b]{0.22\textwidth}
		\includegraphics[width=\linewidth]{figures/bbob-30d-diff-5w/pics/Attractive_Sector_log_cost_curve.pdf}
		\caption*{Attractive Sector}
	\end{subfigure}
	\begin{subfigure}[b]{0.22\textwidth}
		\includegraphics[width=\linewidth]{figures/bbob-30d-diff-5w/pics/Bent_Cigar_log_cost_curve.pdf}
		\caption*{Bent Cigar}
	\end{subfigure}
	\begin{subfigure}[b]{0.22\textwidth}
		\includegraphics[width=\linewidth]{figures/bbob-30d-diff-5w/pics/Buche_Rastrigin_log_cost_curve.pdf}
		\caption*{Buche Rastrigin}
	\end{subfigure}
	\begin{subfigure}[b]{0.22\textwidth}
		\includegraphics[width=\linewidth]{figures/bbob-30d-diff-5w/pics/Composite_Grie_rosen_log_cost_curve.pdf}
		\caption*{Composite Grie-Rosen}
	\end{subfigure}
	
	\begin{subfigure}[b]{0.22\textwidth}
		\includegraphics[width=\linewidth]{figures/bbob-30d-diff-5w/pics/Different_Powers_log_cost_curve.pdf}
		\caption*{Different Powers}
	\end{subfigure}
	\begin{subfigure}[b]{0.22\textwidth}
		\includegraphics[width=\linewidth]{figures/bbob-30d-diff-5w/pics/Discus_log_cost_curve.pdf}
		\caption*{Discus}
	\end{subfigure}
	\begin{subfigure}[b]{0.22\textwidth}
		\includegraphics[width=\linewidth]{figures/bbob-30d-diff-5w/pics/Ellipsoidal_high_cond_log_cost_curve.pdf}
		\caption*{Ellipsoidal}
	\end{subfigure}
	\begin{subfigure}[b]{0.22\textwidth}
		\includegraphics[width=\linewidth]{figures/bbob-30d-diff-5w/pics/Gallagher_21Peaks_log_cost_curve.pdf}
		\caption*{Gallagher 21Peaks}
	\end{subfigure}
	
	\begin{subfigure}[b]{0.22\textwidth}
		\includegraphics[width=\linewidth]{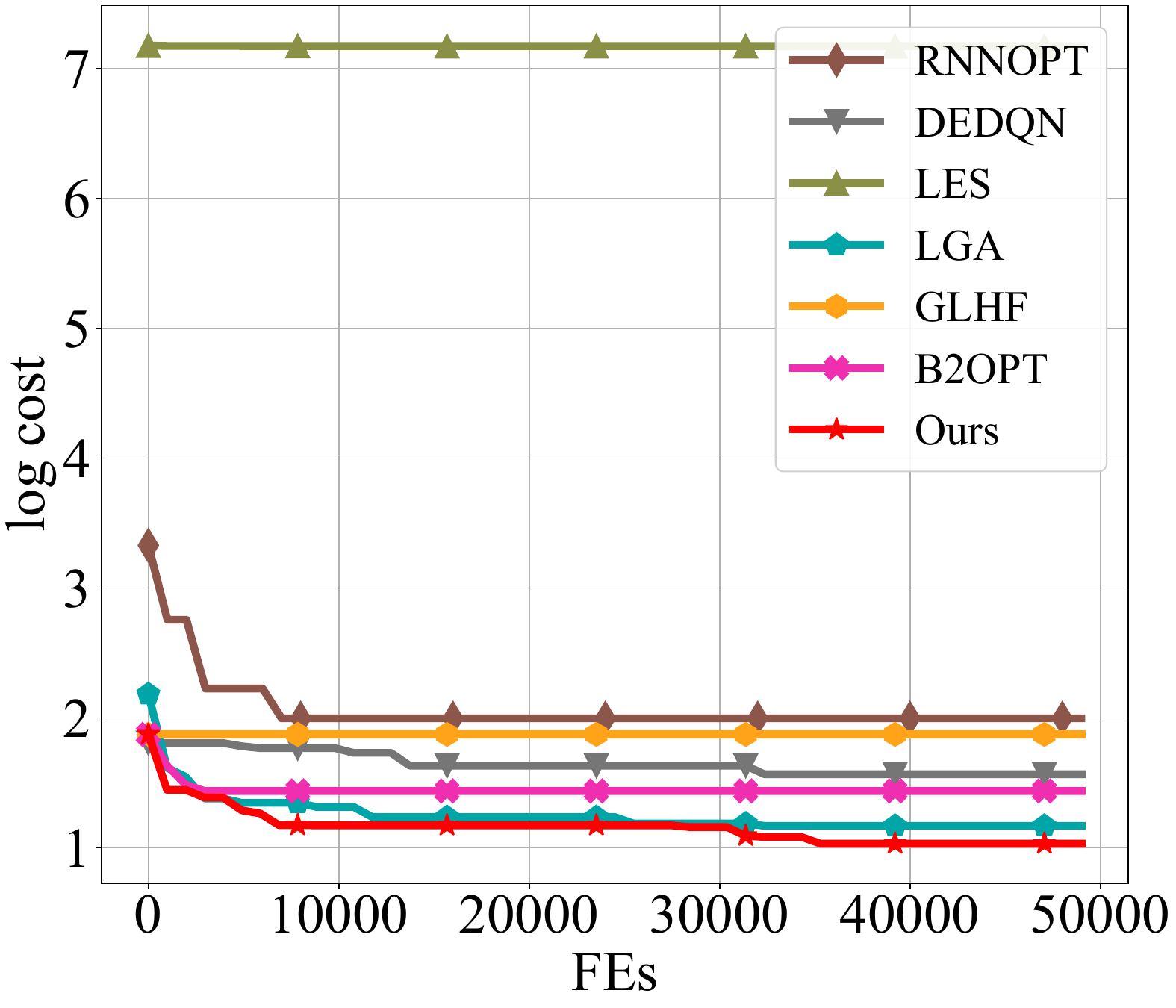}
		\caption*{Katsuura}
	\end{subfigure}
	\begin{subfigure}[b]{0.22\textwidth}
		\includegraphics[width=\linewidth]{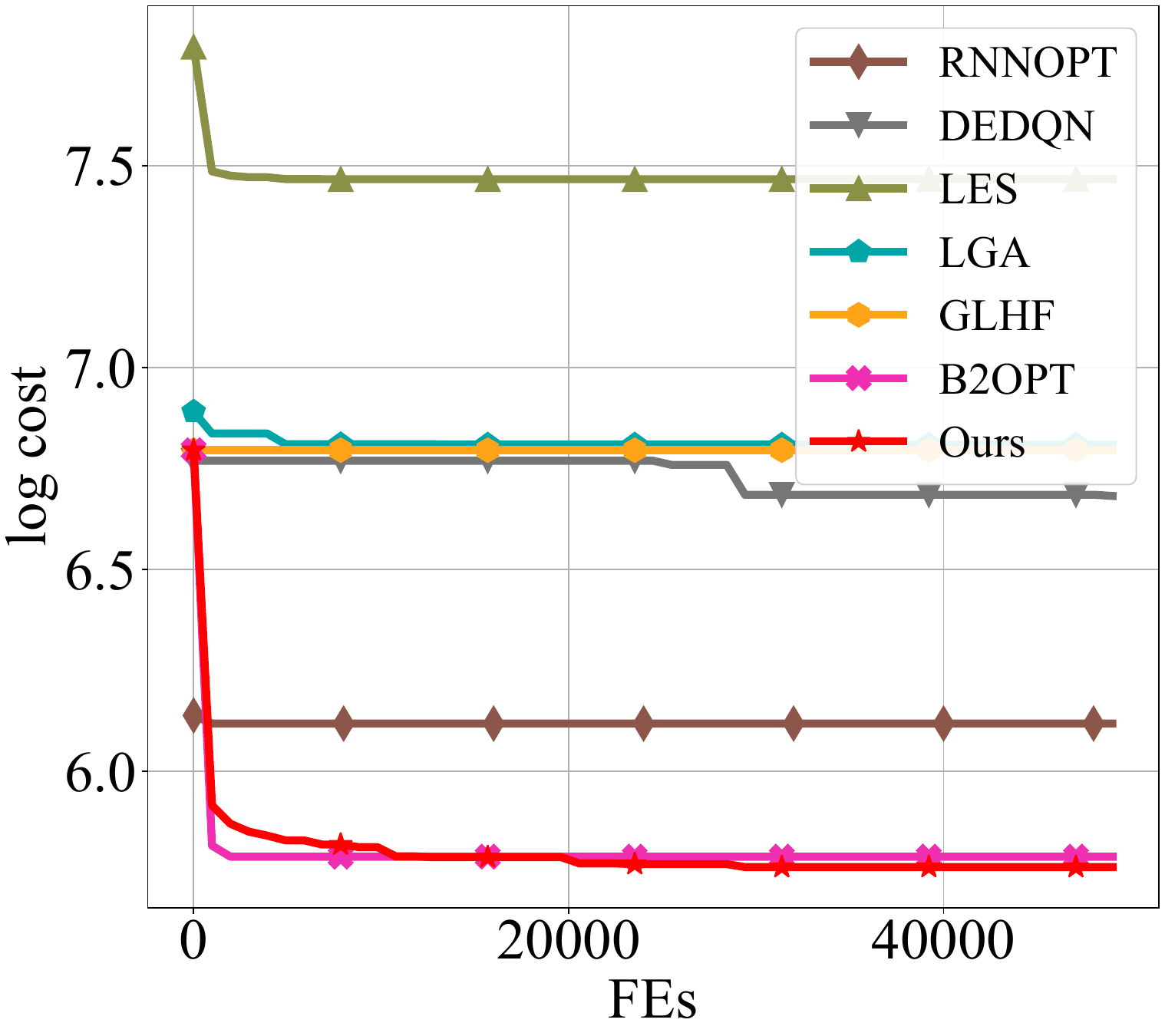}
		\caption*{Lunacek bi-Rastrigin}
	\end{subfigure}
	\begin{subfigure}[b]{0.22\textwidth}
		\includegraphics[width=\linewidth]{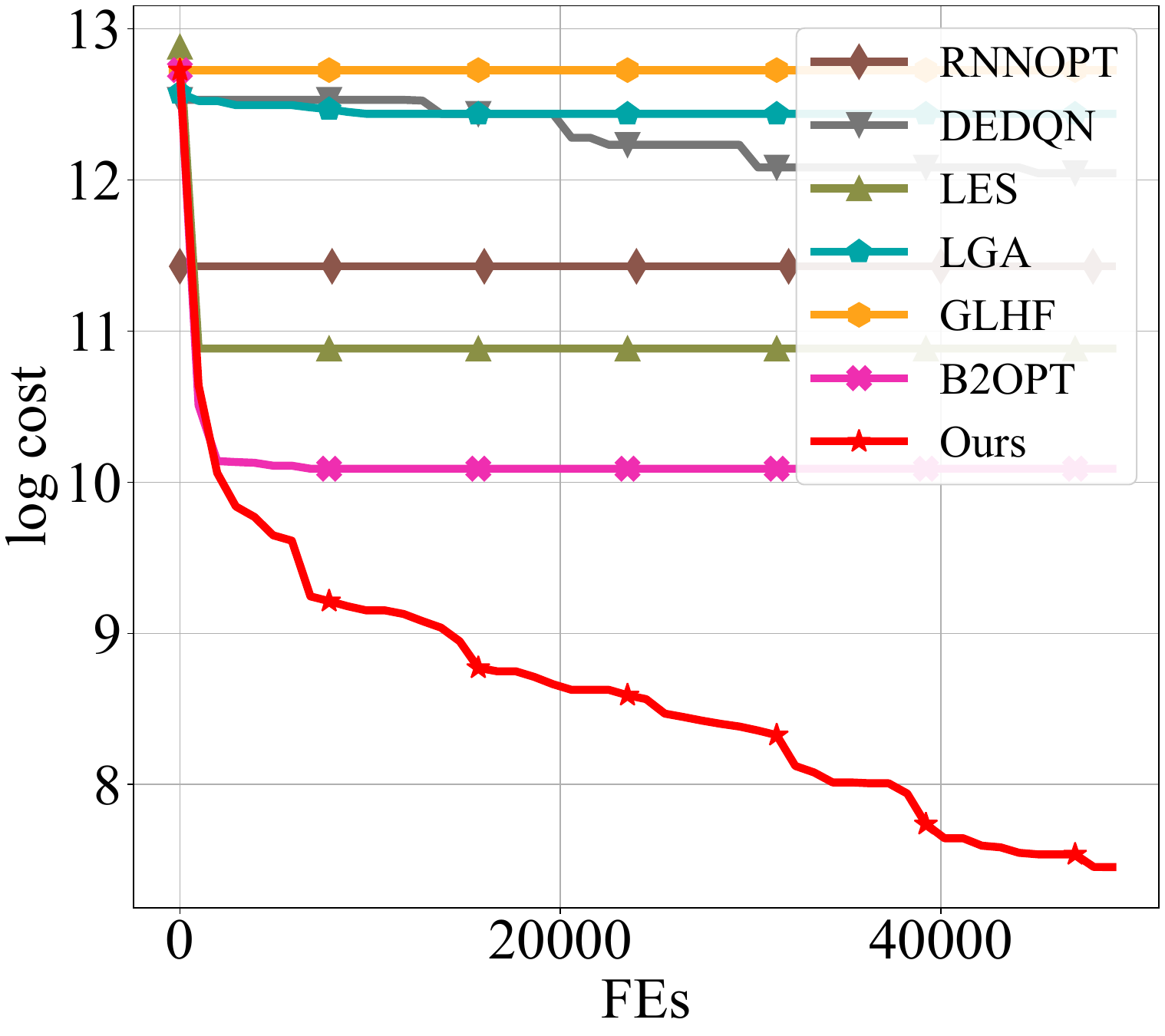}
		\caption*{Rosenbrock Original}
	\end{subfigure}
	\begin{subfigure}[b]{0.22\textwidth}
		\includegraphics[width=\linewidth]{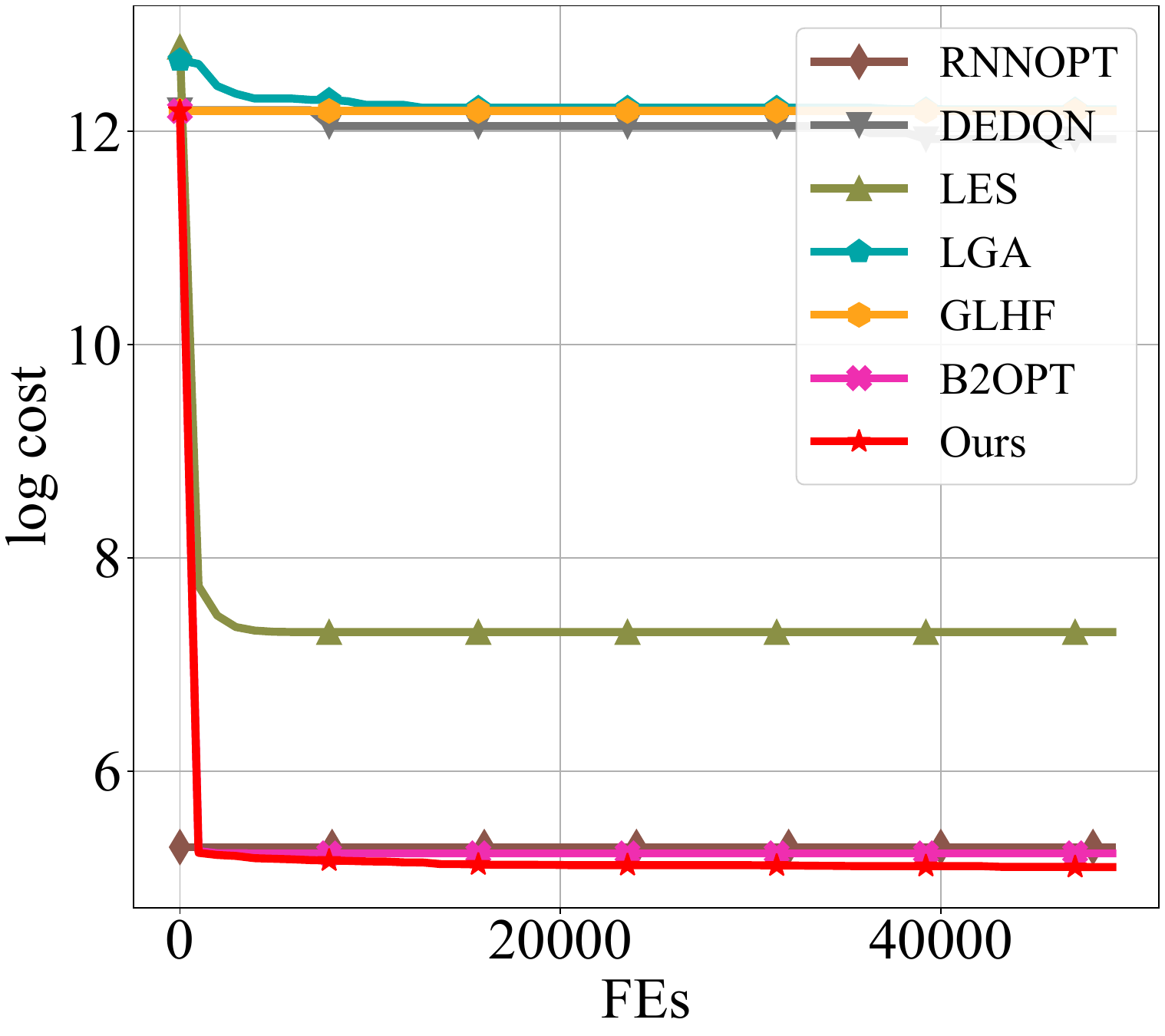}
		\caption*{Rosenbrock Rotated}
	\end{subfigure}
	
	\begin{subfigure}[b]{0.22\textwidth}
		\includegraphics[width=\linewidth]{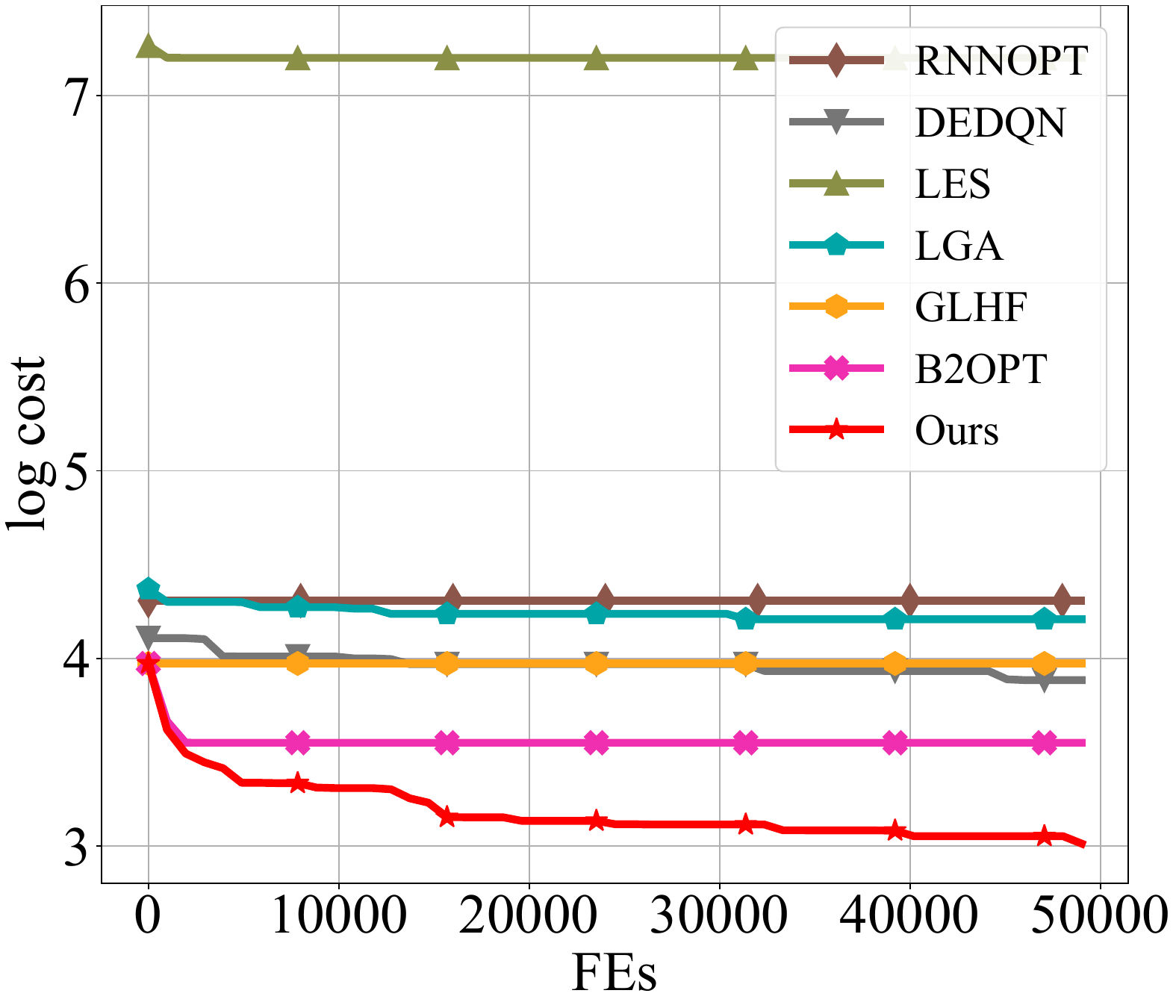}
		\caption*{Schaffers}
	\end{subfigure}
	\begin{subfigure}[b]{0.22\textwidth}
		\includegraphics[width=\linewidth]{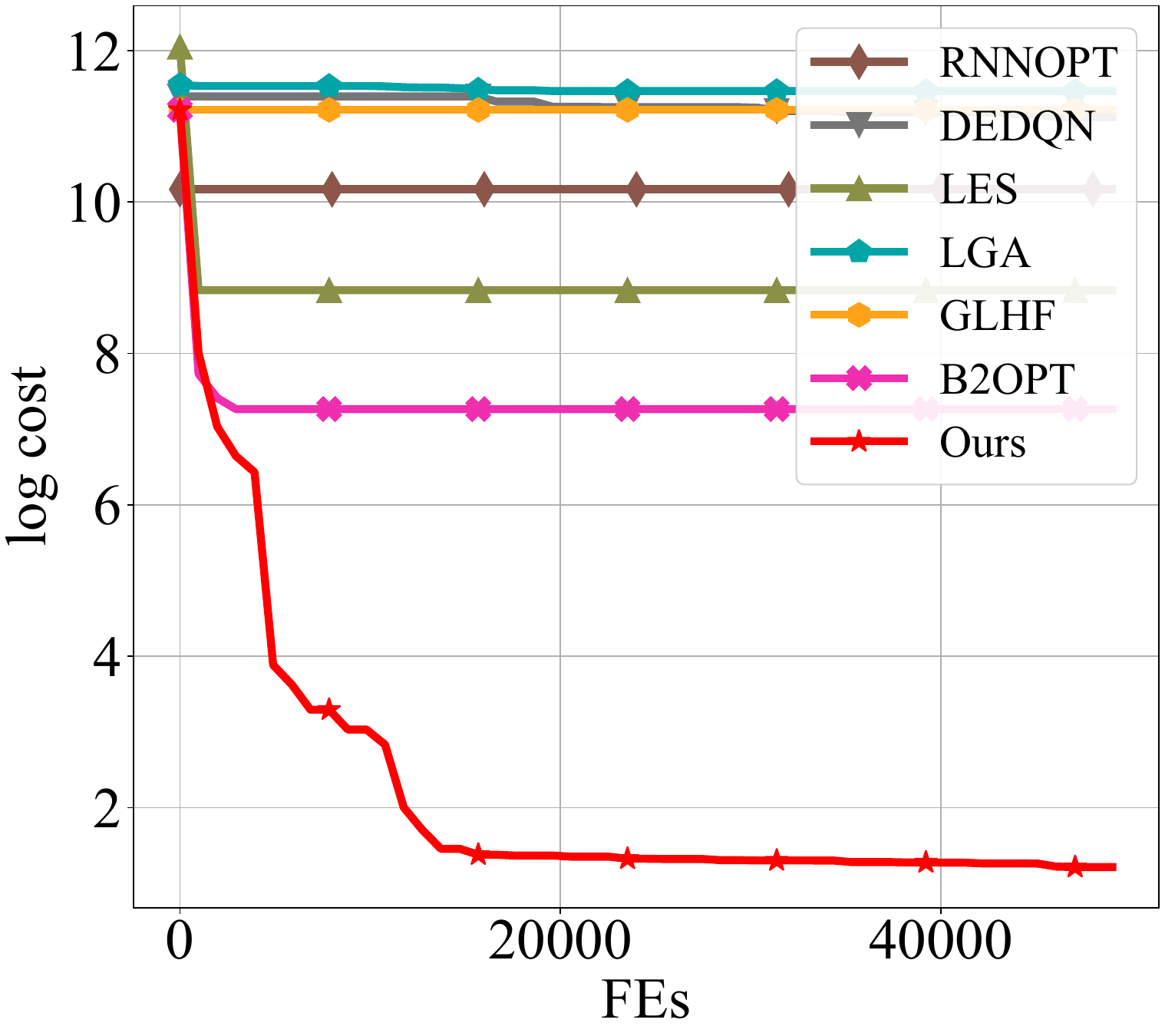}
		\caption*{Schwefel}
	\end{subfigure}
	\begin{subfigure}[b]{0.22\textwidth}
		\includegraphics[width=\linewidth]{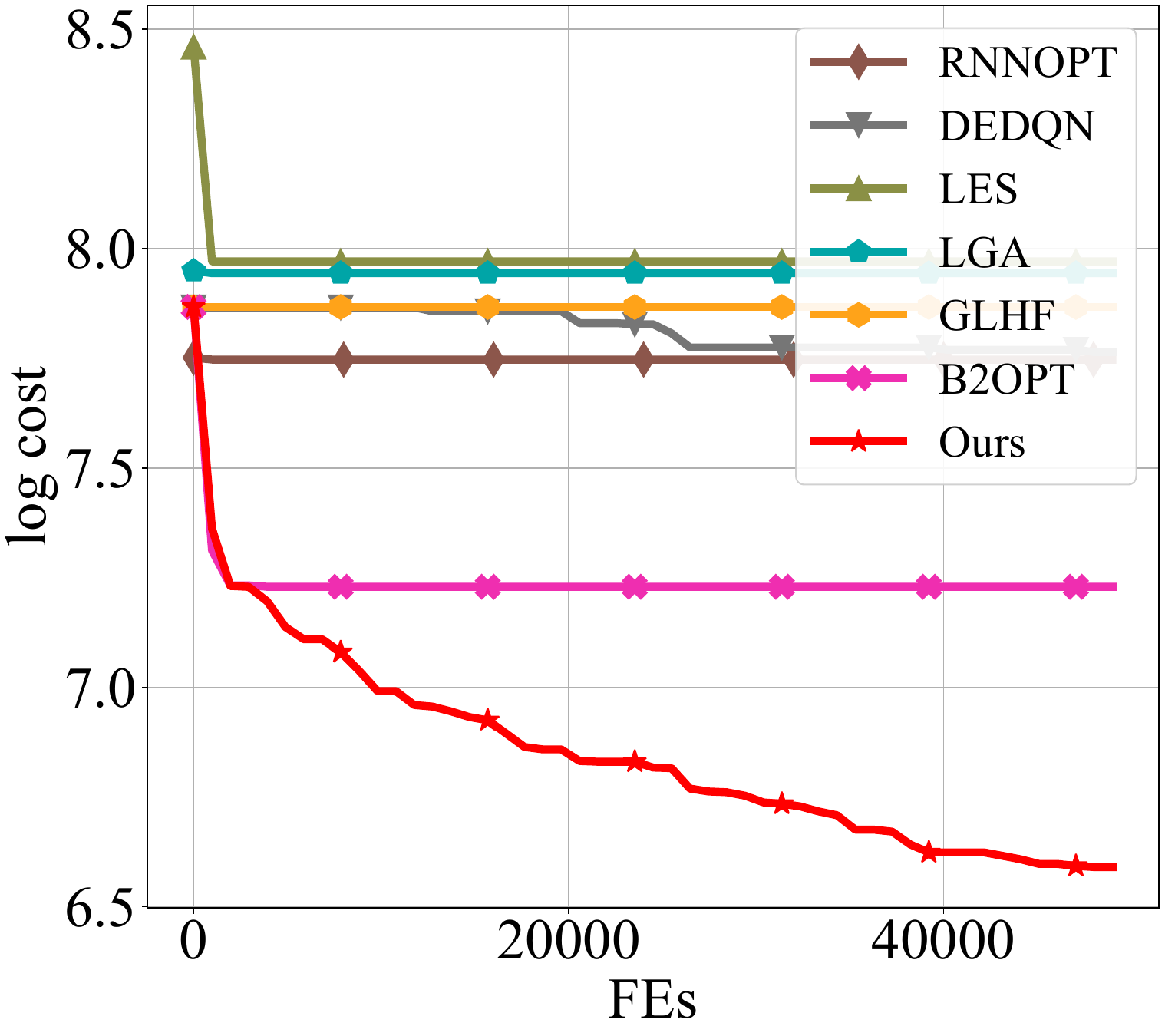}
		\caption*{Sharp Ridge}
	\end{subfigure}
	\begin{subfigure}[b]{0.22\textwidth}
		\includegraphics[width=\linewidth]{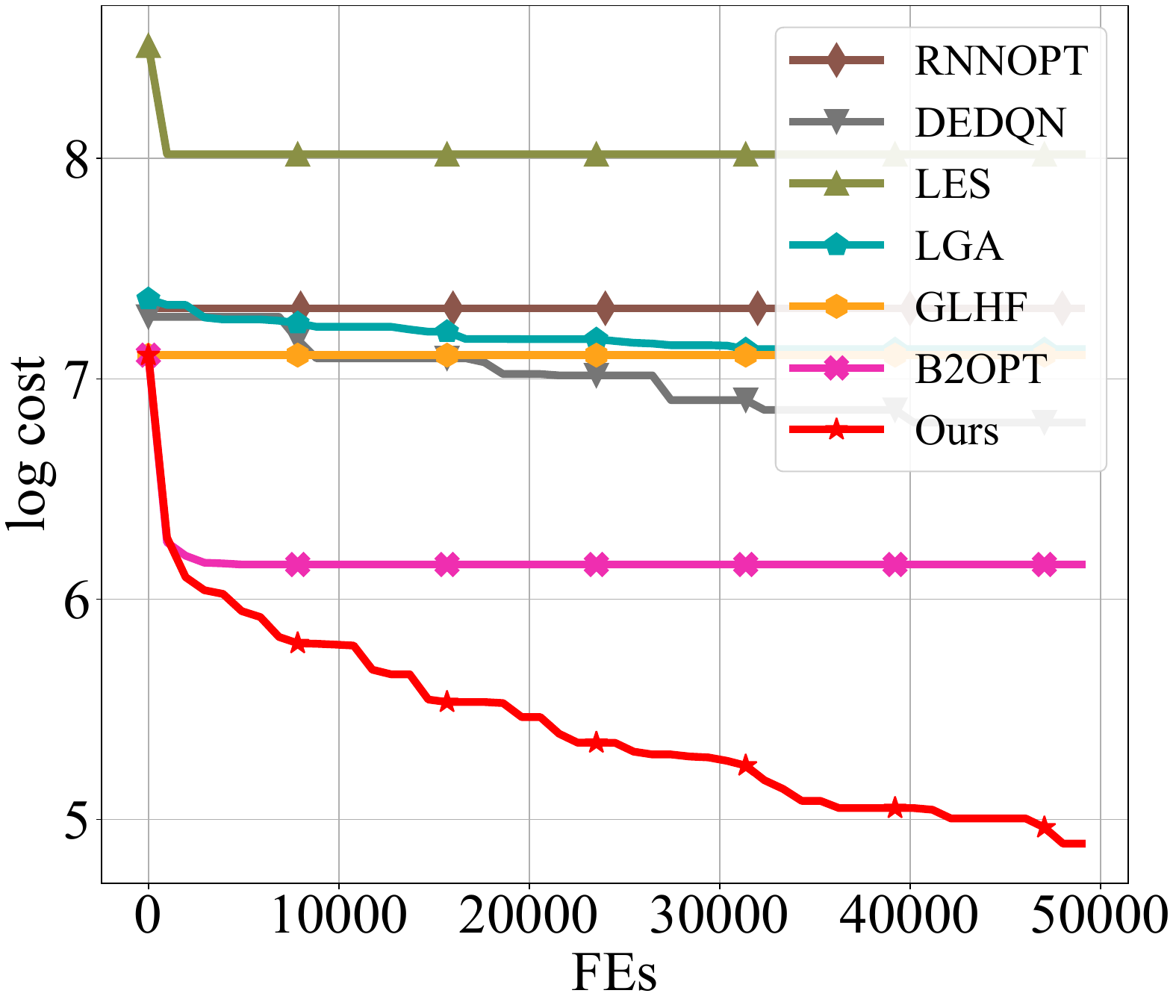}
		\caption*{Step Ellipsoidal}
	\end{subfigure}
	
	\caption{Log-scaled convergence curves of various representative methods on \emph{BBOB-30D}~\cite{hansen2021coco}.}
	\label{fig:supp-bbob-30d-diff-16}
\end{figure*}
}

\newcommand{\FigSupLsgocurve}{%
	\begin{figure*}[htbp]
		\centering
		
		\begin{subfigure}[b]{0.3\textwidth}
			\includegraphics[width=\linewidth]{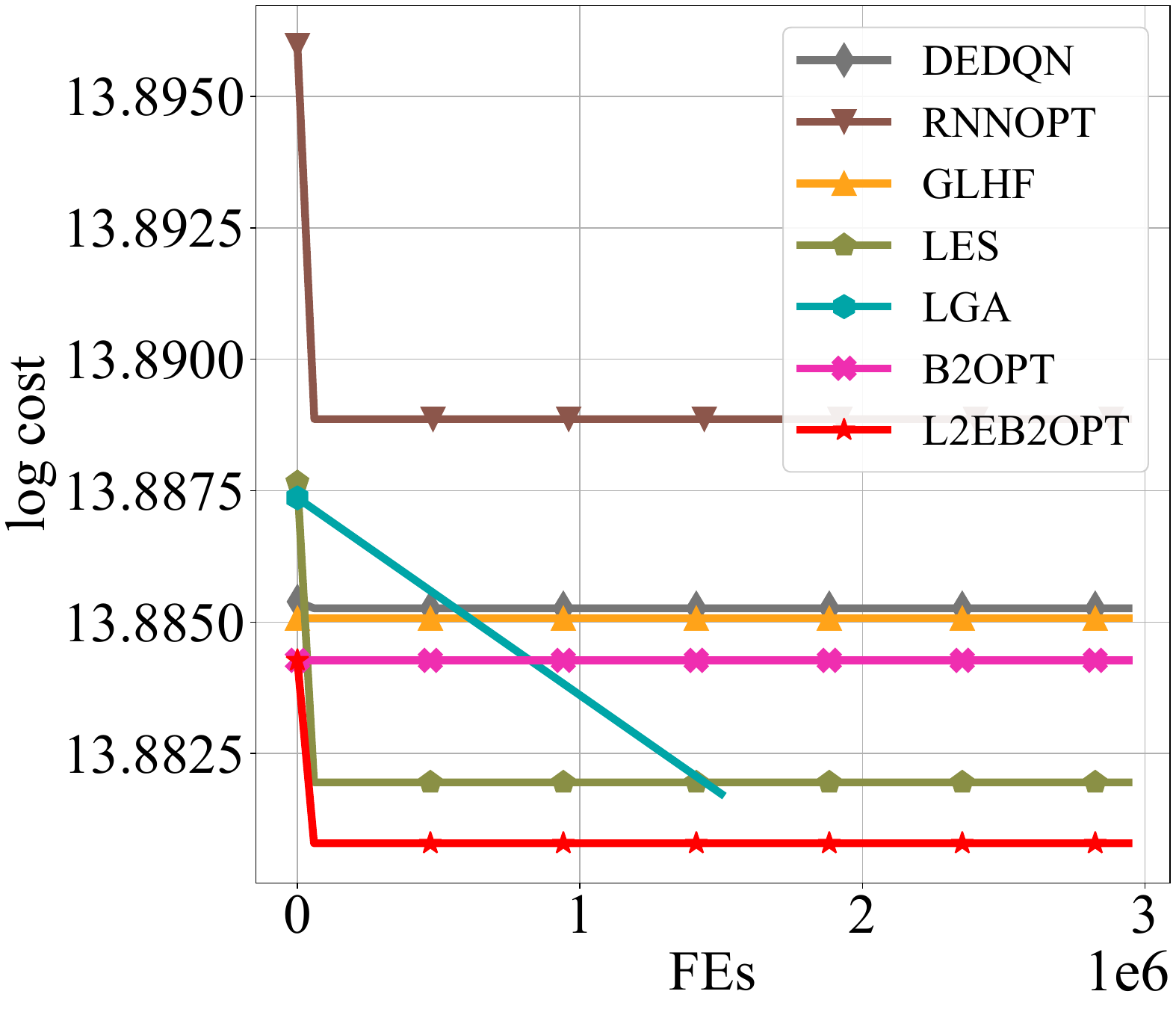}
			\caption*{Rot. Ackley}
		\end{subfigure}
		\begin{subfigure}[b]{0.3\textwidth}
			\includegraphics[width=\linewidth]{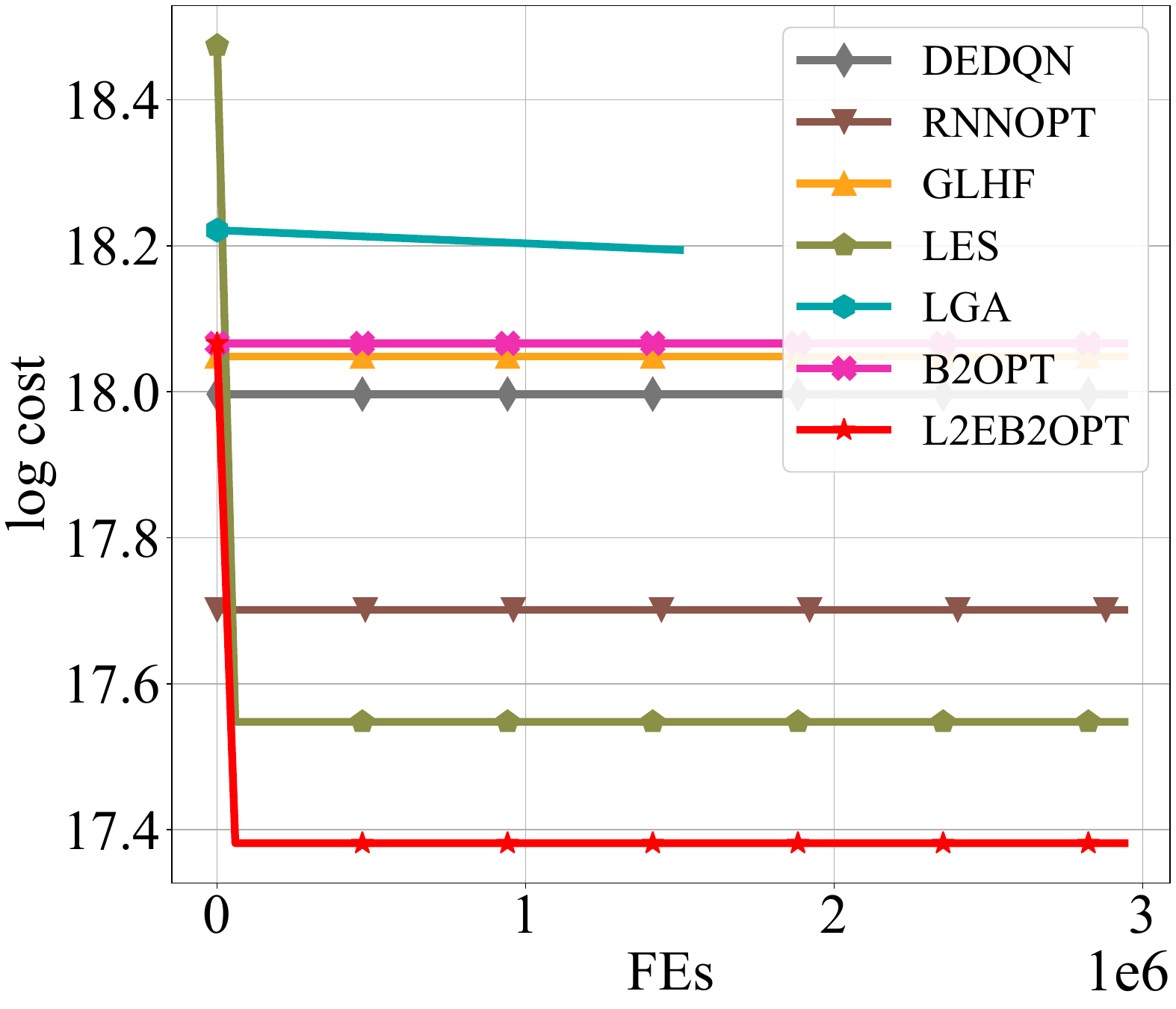}
			\caption*{Rot. Rastrigin}
		\end{subfigure}
		\begin{subfigure}[b]{0.3\textwidth}
			\includegraphics[width=\linewidth]{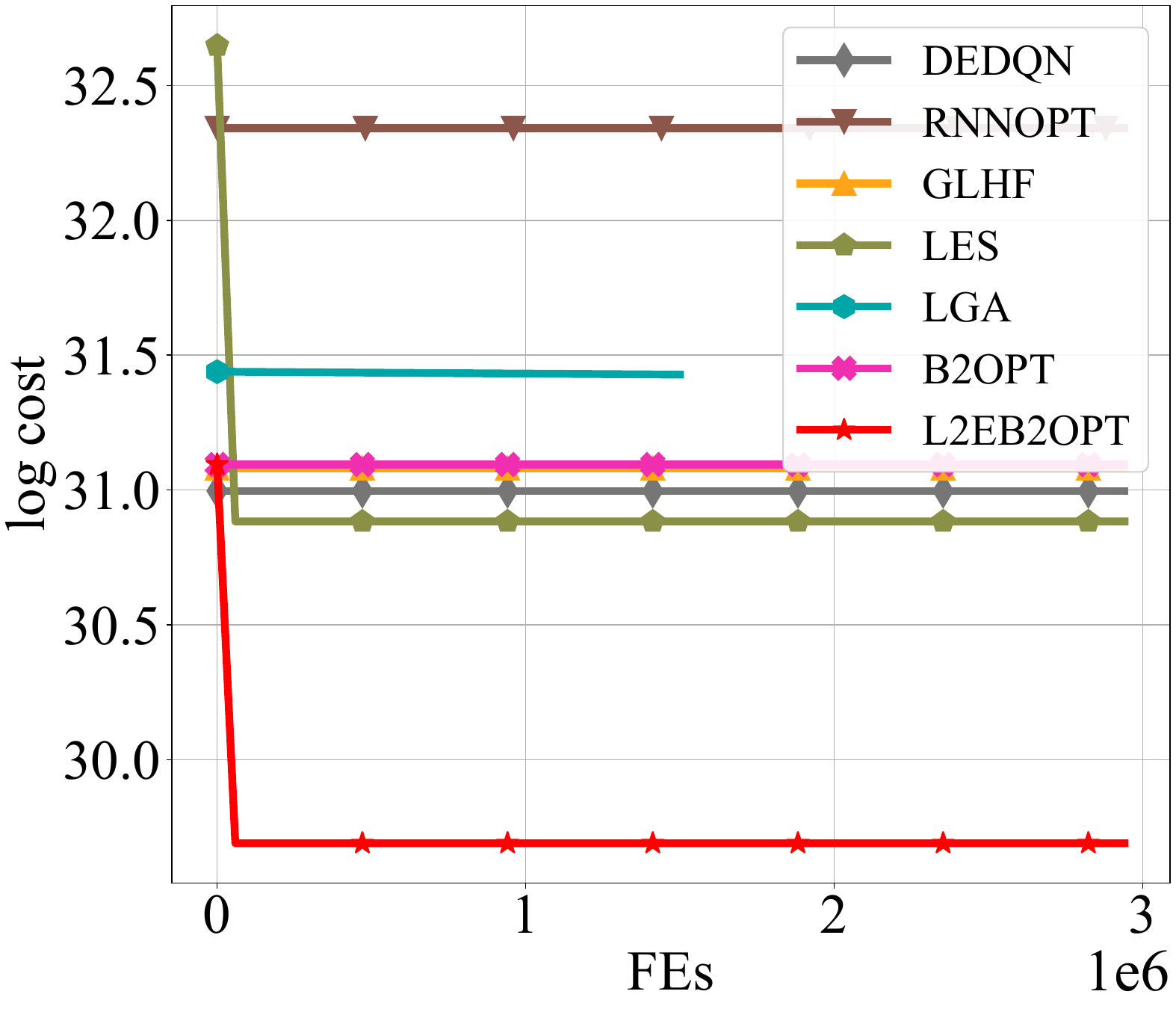}
			\caption*{Rot. Elliptic}
		\end{subfigure}
		
		\vspace{0.5em}  
		
		\begin{subfigure}[b]{0.3\textwidth}
			\includegraphics[width=\linewidth]{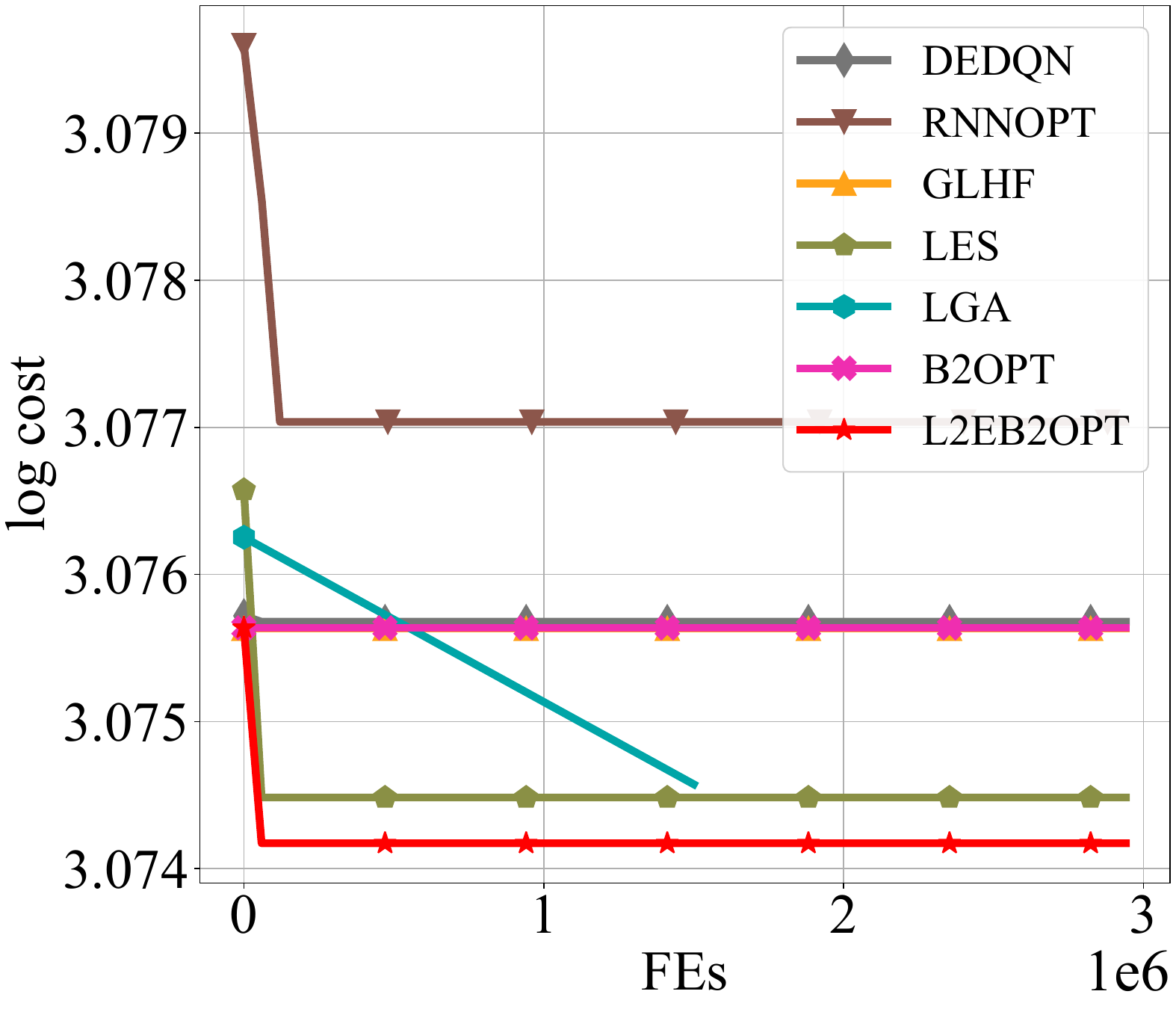}
			\caption*{Shifted Ackley}
		\end{subfigure}
		\begin{subfigure}[b]{0.3\textwidth}
			\includegraphics[width=\linewidth]{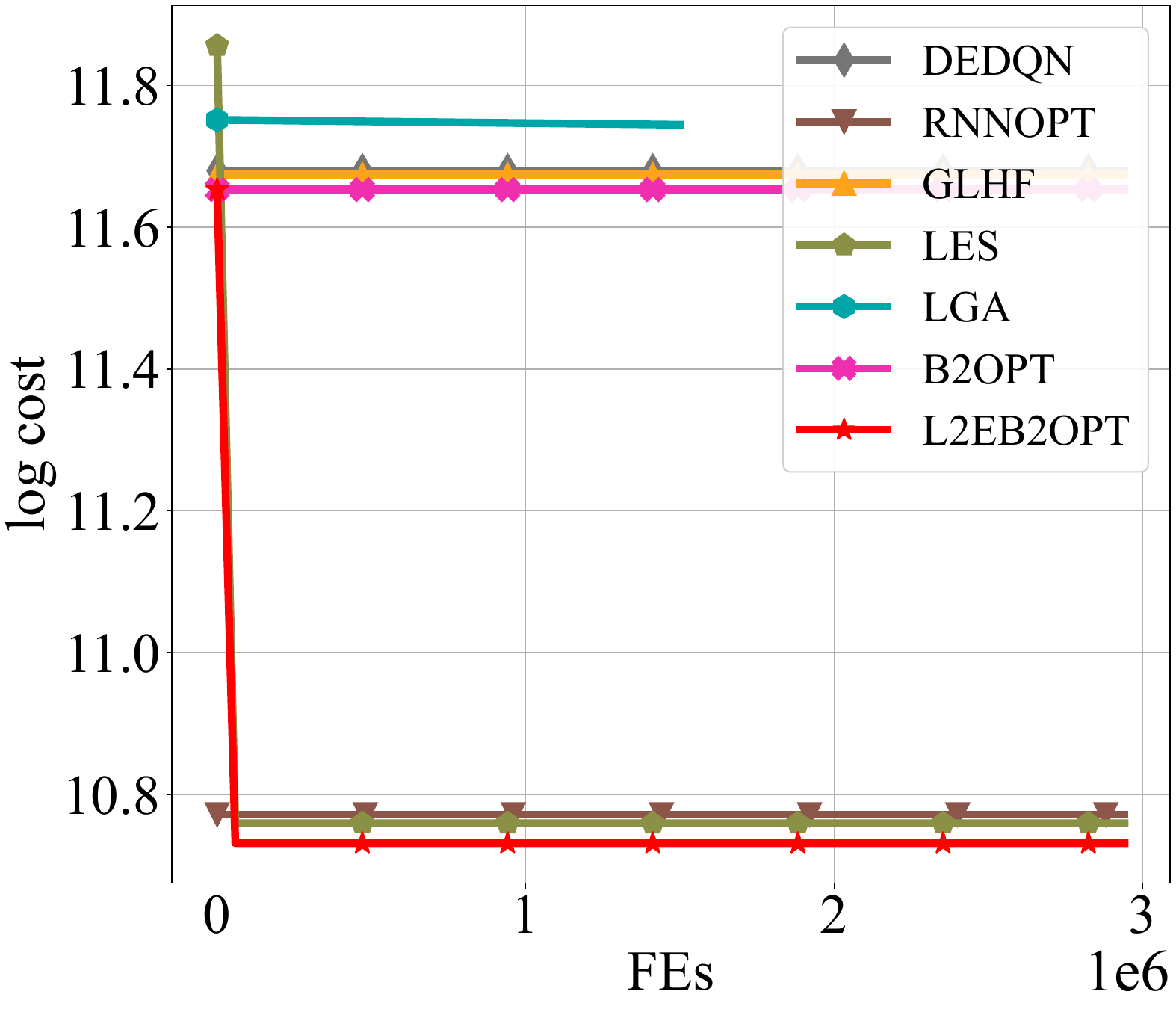}
			\caption*{Shifted Rastrigin}
		\end{subfigure}
		\begin{subfigure}[b]{0.3\textwidth}
			\includegraphics[width=\linewidth]{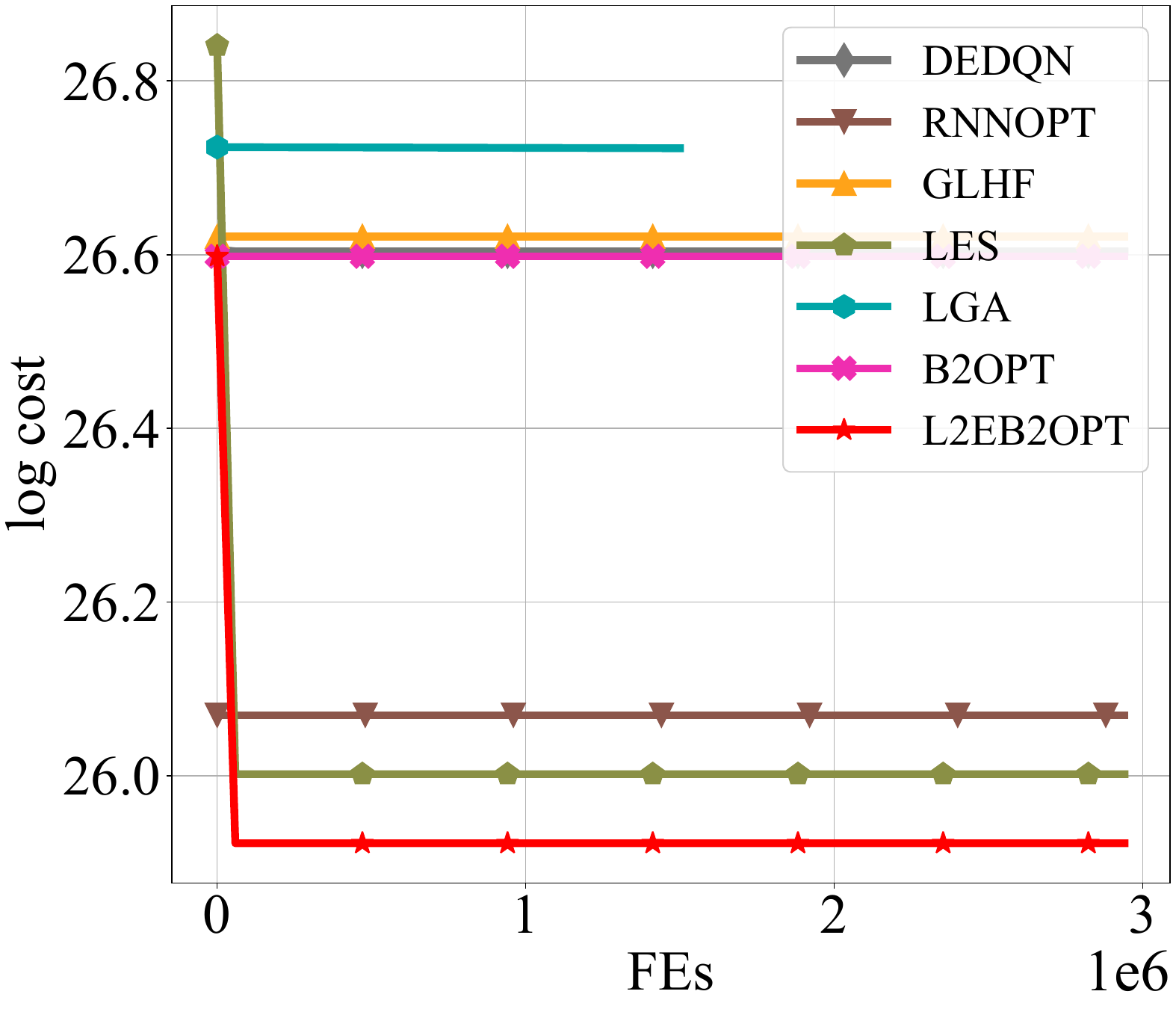}
			\caption*{Shifted Elliptic}
		\end{subfigure}
		
		\caption{Log-scaled convergence curves of various representative methods on \emph{LSGO-1000D}~\cite{li2013benchmark}.}
		\label{fig:supp-lsgo-1000d}
	\end{figure*}
}

\newcommand{\FigSupLsgoECDFcurve}{%
	\begin{figure*}[htbp]
		\centering
		
		\begin{subfigure}[b]{0.3\textwidth}
			\includegraphics[width=\linewidth]{figures/lsgo-run-10/pics/ECDF_7-nonseparable_1-separableSRA.pdf}
			\caption*{Rot. Ackley}
		\end{subfigure}
		\begin{subfigure}[b]{0.3\textwidth}
			\includegraphics[width=\linewidth]{figures/lsgo-run-10/pics/ECDF_7-nonseparable_1-separableSRR.pdf}
			\caption*{Rot. Rastrigin}
		\end{subfigure}
		\begin{subfigure}[b]{0.3\textwidth}
			\includegraphics[width=\linewidth]{figures/lsgo-run-10/pics/ECDF_7-nonseparable_1-separableSRE.pdf}
			\caption*{Rot. Elliptic}
		\end{subfigure}
		
		\vspace{0.8em} 
		
		\begin{subfigure}[b]{0.3\textwidth}
			\includegraphics[width=\linewidth]{figures/lsgo-run-10/pics/ECDF_Shifted_Ackley.pdf}
			\caption*{Shifted Ackley}
		\end{subfigure}
		\begin{subfigure}[b]{0.3\textwidth}
			\includegraphics[width=\linewidth]{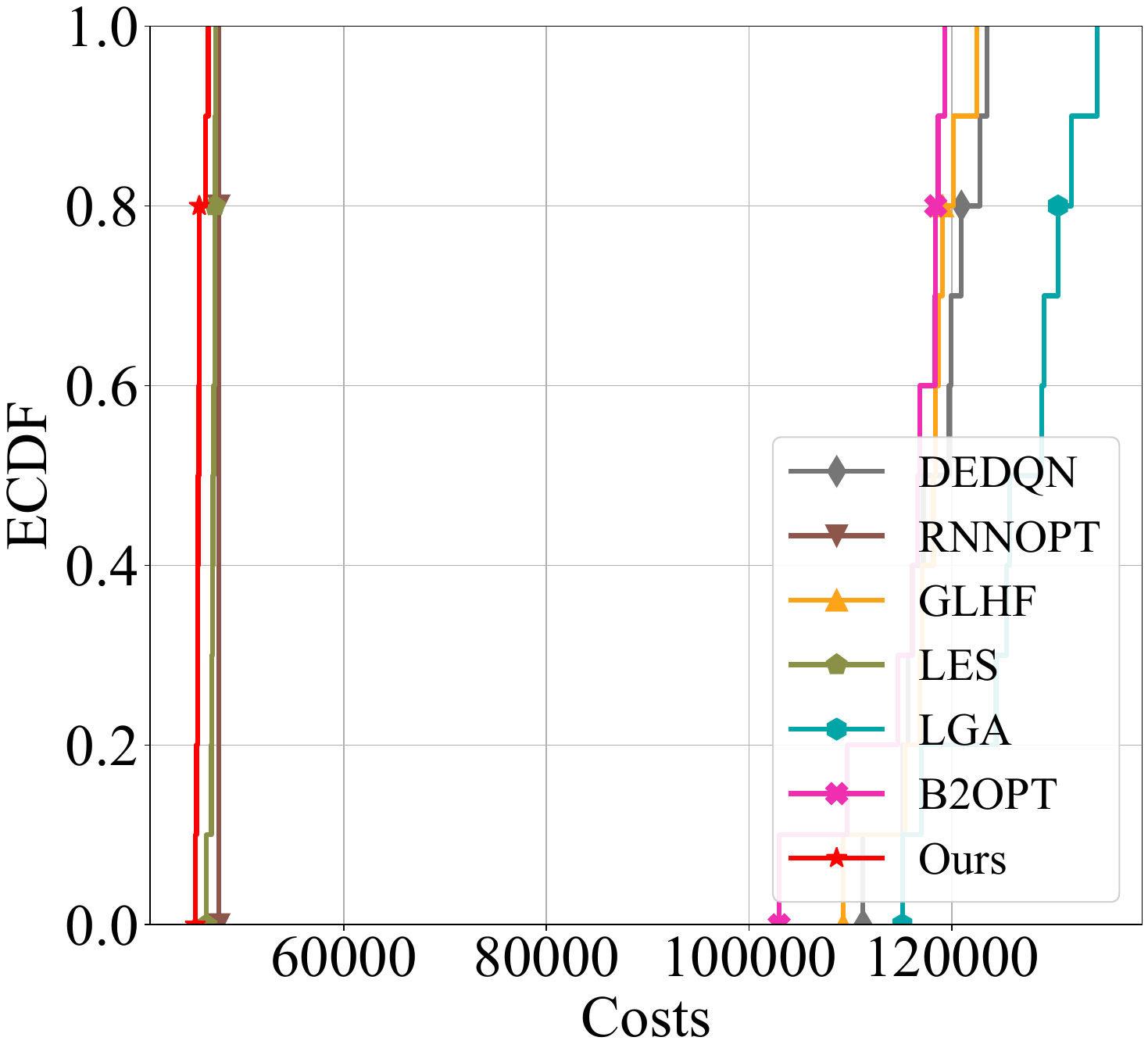}
			\caption*{Shifted Rastrigin}
		\end{subfigure}
		\begin{subfigure}[b]{0.3\textwidth}
			\includegraphics[width=\linewidth]{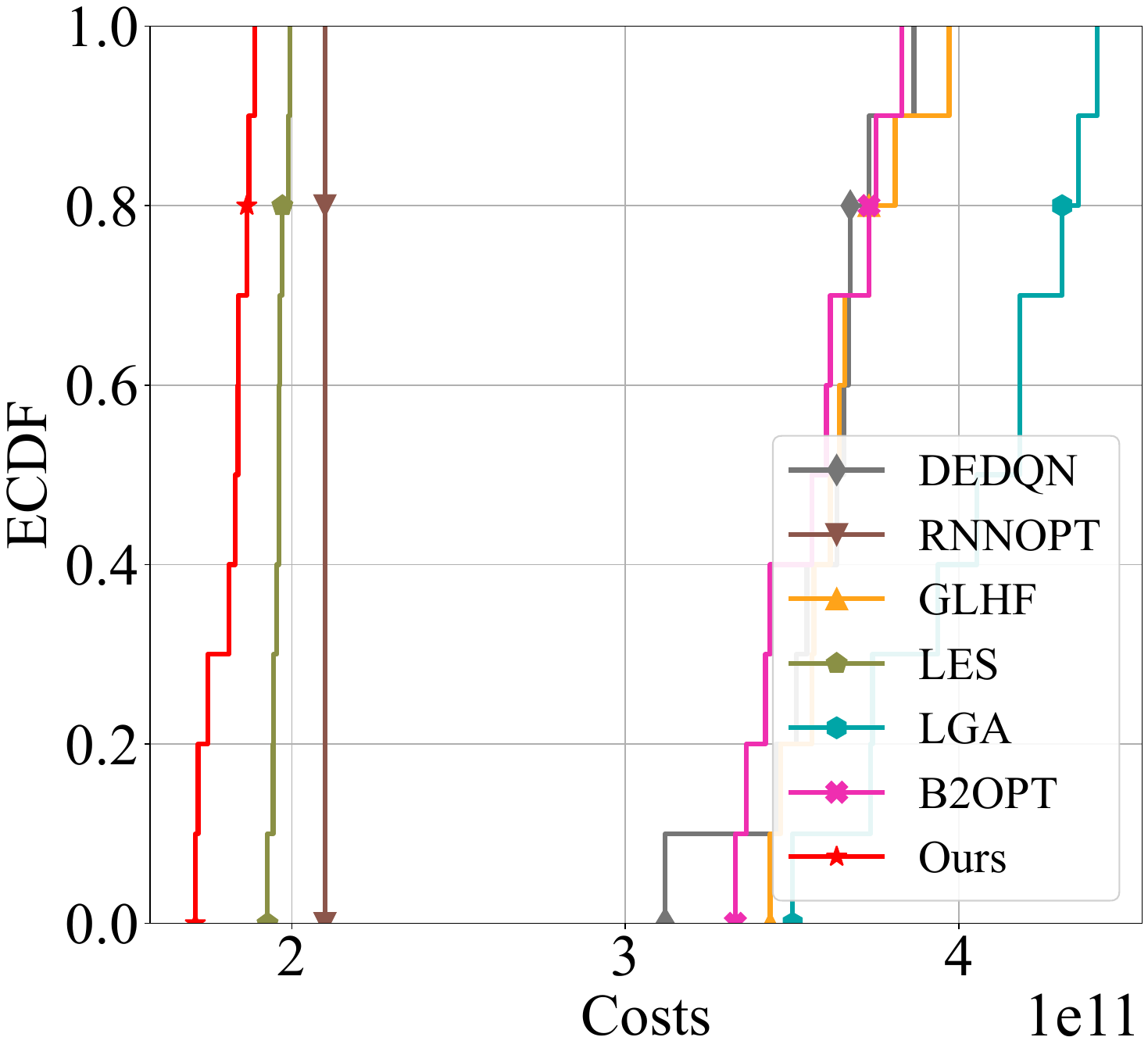}
			\caption*{Shifted Elliptic}
		\end{subfigure}
		
		\caption{Empirical Cumulative Distribution Functions (ECDF) of various representative methods  \emph{LSGO-1000D}~\cite{li2013benchmark}.}
		\label{fig:supp-ecdf-lsgo-1000D}
	\end{figure*}
}

\newcommand{\FigSupLSGOBox}{%
	\begin{figure*}[htbp]
		\centering
		
		\begin{subfigure}[b]{0.3\textwidth}
			\includegraphics[width=\linewidth]{figures/lsgo-run-10/pics/7-nonseparable_1-separableSRA_boxplot.pdf}
			\caption*{Rot. Ackley}
		\end{subfigure}
		\begin{subfigure}[b]{0.3\textwidth}
			\includegraphics[width=\linewidth]{figures/lsgo-run-10/pics/7-nonseparable_1-separableSRR_boxplot.pdf}
			\caption*{Rot. Rastrigin}
		\end{subfigure}
		\begin{subfigure}[b]{0.3\textwidth}
			\includegraphics[width=\linewidth]{figures/lsgo-run-10/pics/7-nonseparable_1-separableSRE_boxplot.pdf}
			\caption*{Rot. Elliptic}
		\end{subfigure}
		
		\vspace{0.8em} 
		
		\begin{subfigure}[b]{0.3\textwidth}
			\includegraphics[width=\linewidth]{figures/lsgo-run-10/pics/Shifted_Ackley_boxplot.pdf}
			\caption*{Shifted Ackley}
		\end{subfigure}
		\begin{subfigure}[b]{0.3\textwidth}
			\includegraphics[width=\linewidth]{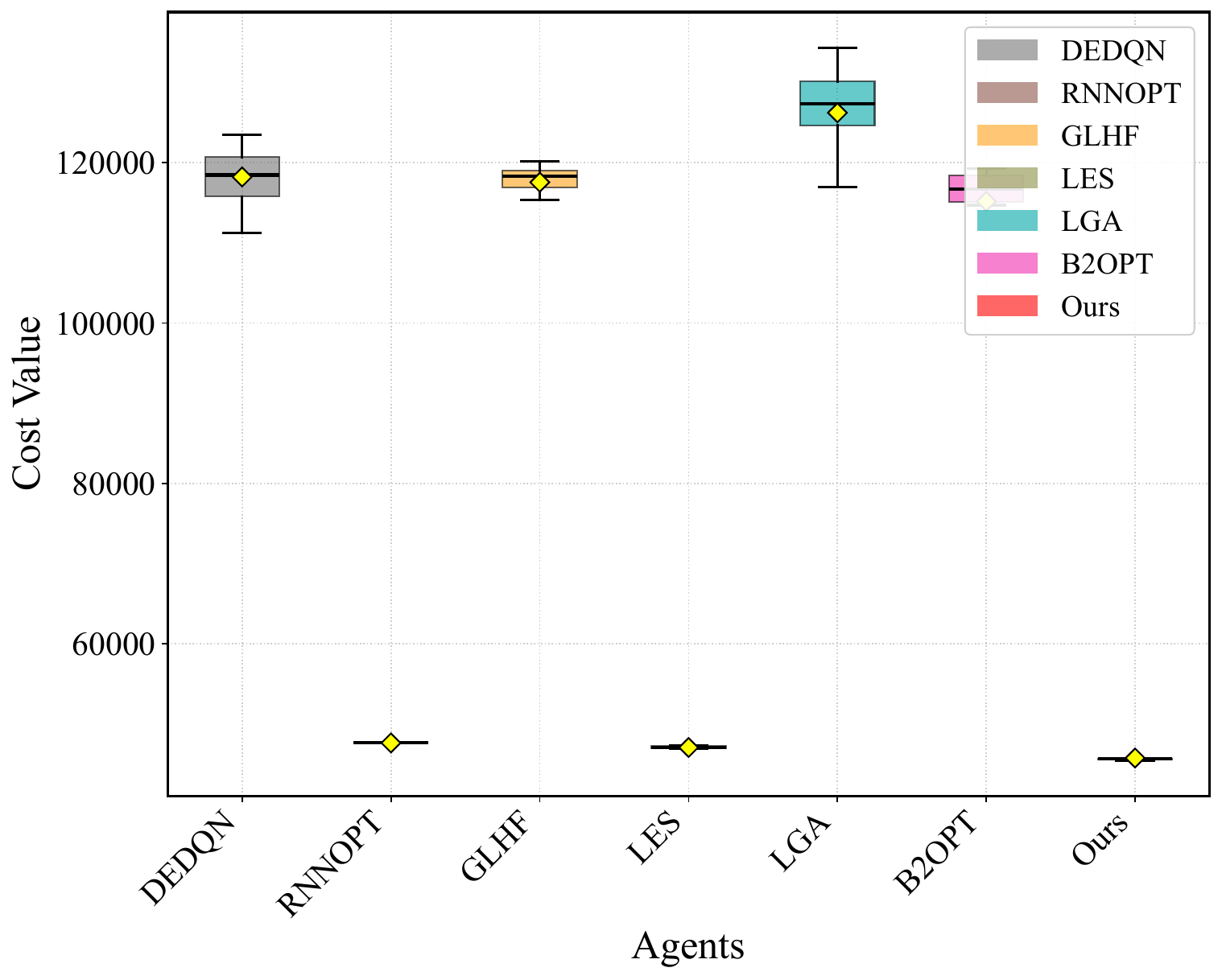}
			\caption*{Shifted Rastrigin}
		\end{subfigure}
		\begin{subfigure}[b]{0.3\textwidth}
			\includegraphics[width=\linewidth]{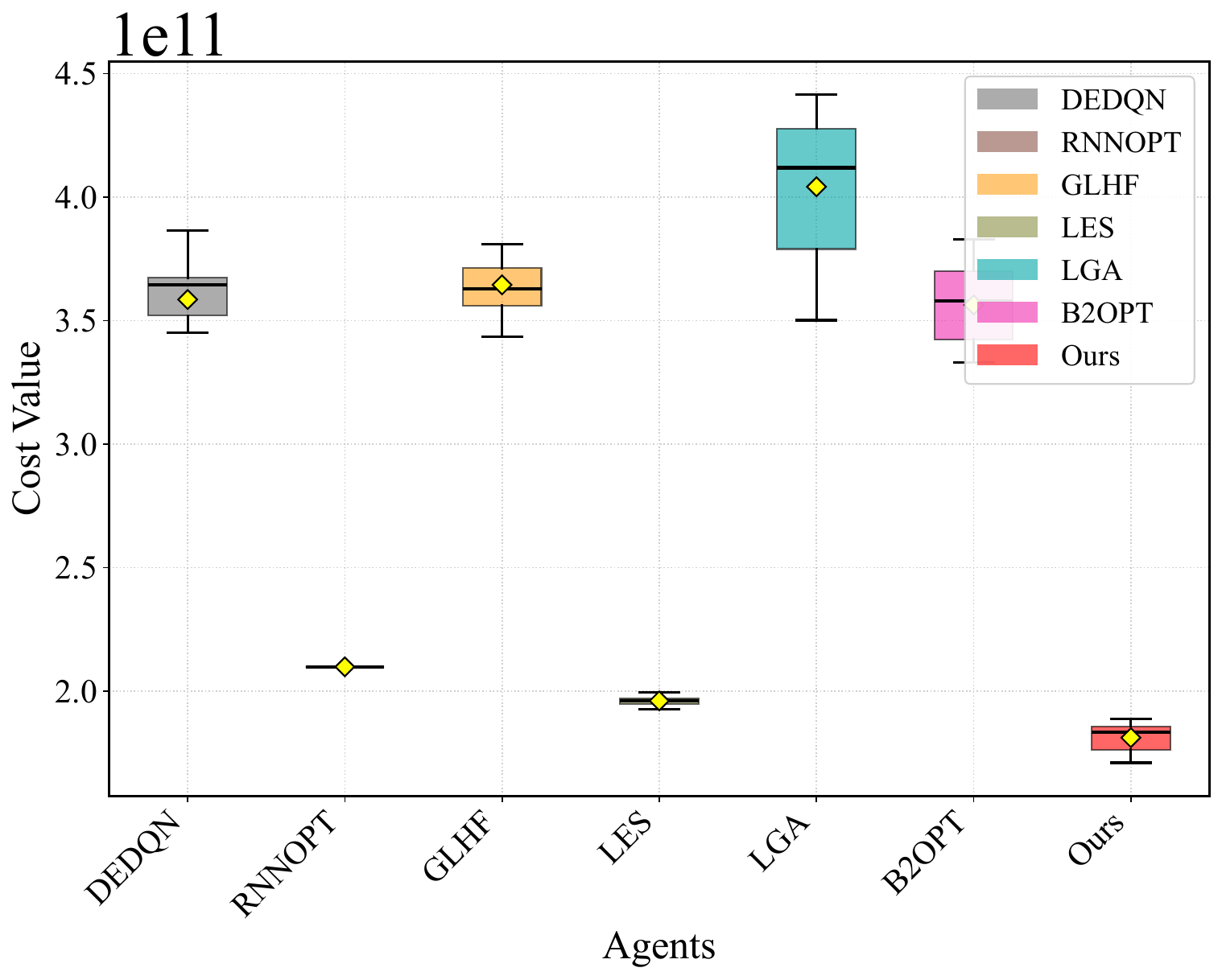}
			\caption*{Shifted Elliptic}
		\end{subfigure}
		
		\caption{Boxplots of various representative methods trained and tested on \emph{LSGO-1000D}~\cite{li2013benchmark}.}
		\label{fig:supp-boxplot-lsgo-1000D}
	\end{figure*}
}

\newcommand{\TabSupAblShared}{%
\begin{table*}[tbp]
	\centering
	\scriptsize
	\setlength{\tabcolsep}{8pt} 
	\caption{
		Evaluation results (mean ± std) of various representative methods on the \emph{UAV} benchmark, consisting of 56 distinct terrain scenarios. 
		Each value is averaged over 10 independent test runs. 
		Only the first 40 terrains are shown in this table.
	}
	\label{tab:UAV}
	\begin{tabular}{lcccccccc}
		\toprule
		\textbf{Method} & \texttt{Terrain 1} & \texttt{Terrain 2} & \texttt{Terrain 3} & \texttt{Terrain 4} & \texttt{Terrain 5} & \texttt{Terrain 6} & \texttt{Terrain 7} & \texttt{Terrain 8} \\
		\midrule
		\midrule
		\multirow{1}{*}{RNNOPT}  & 5.690E+04 & 1.558E+05 & 1.067E+05 & 1.657E+05 & 2.565E+04 & 7.611E+04 & 2.565E+04 & 7.606E+04 \\
		& ($\pm$0.000E+00) & ($\pm$0.000E+00) & ($\pm$0.000E+00) & ($\pm$0.000E+00) & ($\pm$0.000E+00) & ($\pm$0.000E+00) & ($\pm$0.000E+00) & ($\pm$0.000E+00) \\
		\multirow{1}{*}{DEDQN}  & 2.066E+04 & 5.283E+04 & 2.641E+04 & 8.209E+04 & 1.263E+04 & 5.557E+04 & 1.298E+04 & 5.568E+04 \\
		& ($\pm$8.082E+03) & ($\pm$1.241E+04) & ($\pm$4.780E+03) & ($\pm$3.946E+03) & ($\pm$3.144E+03) & ($\pm$4.936E+03) & ($\pm$3.057E+03) & ($\pm$4.900E+03) \\
		\multirow{1}{*}{GLHF}  & 2.116E+04 & 2.712E+04 & 2.588E+04 & 2.567E+04 & 1.827E+04 & 2.594E+04 & 1.932E+04 & 2.663E+04 \\
		& ($\pm$6.670E+03) & ($\pm$4.553E+02) & ($\pm$1.734E+03) & ($\pm$4.919E+02) & ($\pm$1.610E+03) & ($\pm$3.167E+02) & ($\pm$2.756E+02) & ($\pm$2.866E+02) \\
		\multirow{1}{*}{LES}  & 4.619E+04 & 5.573E+04 & 3.556E+04 & 5.550E+04 & 1.983E+04 & 3.831E+04 & 1.985E+04 & 3.826E+04 \\
		& ($\pm$1.855E+01) & ($\pm$3.643E+01) & ($\pm$2.998E-02) & ($\pm$4.134E+01) & ($\pm$7.404E+02) & ($\pm$9.534E+02) & ($\pm$7.894E+02) & ($\pm$9.289E+02) \\
		\multirow{1}{*}{LGA}  & 1.416E+04 & 4.659E+04 & 1.626E+04 & 5.276E+04 & 1.191E+04 & 4.079E+04 & 1.192E+04 & 4.047E+04 \\
		& ($\pm$5.311E+03) & ($\pm$9.300E+03) & ($\pm$4.912E+03) & ($\pm$4.081E+03) & ($\pm$3.661E+03) & ($\pm$1.008E+03) & ($\pm$3.784E+03) & ($\pm$9.781E+02) \\
		\multirow{1}{*}{B2OPT}  & 1.434E+04 & 2.502E+04 & 1.232E+04 & \textbf{2.002E+04} & 1.428E+04 & \textbf{1.656E+04} & 1.283E+04 & 2.347E+04 \\
		& ($\pm$1.984E+03) & ($\pm$5.066E+03) & ($\pm$8.766E+02) & ($\pm$6.043E+03) & ($\pm$2.112E+03) & ($\pm$4.990E+03) & ($\pm$1.564E+03) & ($\pm$9.482E+03) \\
		\midrule
		\rowcolor{blue!5} \multirow{1}{*}{Ours}   & \textbf{1.021E+04} & \textbf{1.576E+04} & \textbf{1.145E+04} & 2.216E+04 & \textbf{8.837E+03} & 2.034E+04 & \textbf{8.883E+03} & \textbf{1.868E+04} \\
		\rowcolor{blue!5} & ($\pm$5.697E+02) & ($\pm$4.993E+03) & ($\pm$4.824E+02) & ($\pm$8.081E+03) & ($\pm$6.256E+02) & ($\pm$6.604E+03) & ($\pm$5.821E+02) & ($\pm$4.746E+03) \\ 
		
		\toprule
		\textbf{Method} & \texttt{Terrain 9} & \texttt{Terrain 10} & \texttt{Terrain 11} & \texttt{Terrain 12} & \texttt{Terrain 13} & \texttt{Terrain 14} & \texttt{Terrain 15} & \texttt{Terrain 16} \\
		\midrule
		\midrule
		\multirow{1}{*}{RNNOPT}  & 8.692E+04 & 8.588E+04 & 8.693E+04 & 8.592E+04 & 5.798E+04 & 1.564E+05 & 8.594E+04 & 9.496E+04 \\
		& ($\pm$0.000E+00) & ($\pm$0.000E+00) & ($\pm$0.000E+00) & ($\pm$0.000E+00) & ($\pm$0.000E+00) & ($\pm$0.000E+00) & ($\pm$0.000E+00) & ($\pm$0.000E+00) \\
		\multirow{1}{*}{DEDQN}  & 4.384E+04 & 3.604E+04 & 4.416E+04 & 3.553E+04 & 2.061E+04 & 5.317E+04 & 9.008E+03 & 3.909E+04 \\
		& ($\pm$4.744E+03) & ($\pm$1.298E+04) & ($\pm$4.250E+03) & ($\pm$1.220E+04) & ($\pm$8.004E+03) & ($\pm$1.273E+04) & ($\pm$5.923E+02) & ($\pm$6.614E+03) \\
		\multirow{1}{*}{GLHF}  & 2.679E+04 & 2.355E+04 & 2.646E+04 & 2.354E+04 & 2.082E+04 & 2.659E+04 & 9.435E+03 & 2.469E+04 \\
		& ($\pm$5.741E+02) & ($\pm$3.839E+03) & ($\pm$6.509E+02) & ($\pm$4.007E+03) & ($\pm$6.970E+03) & ($\pm$4.403E+02) & ($\pm$3.265E+02) & ($\pm$9.598E+02) \\
		\multirow{1}{*}{LES}  & 5.236E+04 & 8.793E+03 & 5.733E+04 & 8.795E+03 & 4.682E+04 & 5.606E+04 & 7.885E+03 & 4.464E+04 \\
		& ($\pm$3.927E+03) & ($\pm$4.534E+03) & ($\pm$7.693E+03) & ($\pm$4.520E+03) & ($\pm$7.855E+01) & ($\pm$1.056E+02) & ($\pm$2.031E+03) & ($\pm$7.187E-08) \\
		\multirow{1}{*}{LGA}  & 3.696E+04 & 9.106E+03 & 3.661E+04 & 8.827E+03 & 1.439E+04 & 4.721E+04 & 9.093E+03 & 2.679E+04 \\
		& ($\pm$9.127E+03) & ($\pm$4.517E+02) & ($\pm$9.261E+03) & ($\pm$1.275E+02) & ($\pm$5.417E+03) & ($\pm$9.391E+03) & ($\pm$4.048E+02) & ($\pm$5.392E+03) \\
		\multirow{1}{*}{B2OPT}  & 1.757E+04 & \textbf{7.690E+03} & 1.363E+04 & 8.942E+03 & 1.261E+04 & 3.359E+04 & 8.989E+03 & 2.299E+04 \\
		& ($\pm$2.087E+03) & ($\pm$2.264E+03) & ($\pm$2.491E+03) & ($\pm$1.675E+03) & ($\pm$1.557E+03) & ($\pm$8.785E+03) & ($\pm$7.819E+02) & ($\pm$9.526E+03) \\
		\midrule
		\rowcolor{blue!5} \multirow{1}{*}{Ours}  & \textbf{1.448E+04} & 7.972E+03 & \textbf{1.191E+04} & \textbf{7.872E+03} & \textbf{1.094E+04} & \textbf{1.947E+04} & \textbf{8.573E+03} & \textbf{1.694E+04} \\
		\rowcolor{blue!5} & ($\pm$3.823E+03) & ($\pm$5.873E+02) & ($\pm$2.785E+02) & ($\pm$3.508E+02) & ($\pm$1.510E+02) & ($\pm$4.620E+03) & ($\pm$2.306E+02) & ($\pm$8.933E+03) \\
		
		\toprule
		\textbf{Method} & \texttt{Terrain 17} & \texttt{Terrain 18} & \texttt{Terrain 19} & \texttt{Terrain 20} & \texttt{Terrain 21} & \texttt{Terrain 22} & \texttt{Terrain 23} & \texttt{Terrain 24} \\
		\midrule
		\midrule
		\multirow{1}{*}{RNNOPT}  & 2.572E+04 & 7.614E+04 & 8.716E+04 & 8.605E+04 & 5.848E+04 & 1.566E+05 & 7.811E+04 & 5.656E+04 \\
		& ($\pm$0.000E+00) & ($\pm$0.000E+00) & ($\pm$0.000E+00) & ($\pm$0.000E+00) & ($\pm$0.000E+00) & ($\pm$0.000E+00) & ($\pm$0.000E+00) & ($\pm$7.276E-12) \\
		\multirow{1}{*}{DEDQN}  & 1.318E+04 & 5.571E+04 & 4.424E+04 & 3.625E+04 & 2.080E+04 & 5.335E+04 & 9.639E+03 & 5.663E+04 \\
		& ($\pm$2.964E+03) & ($\pm$4.854E+03) & ($\pm$4.318E+03) & ($\pm$1.285E+04) & ($\pm$8.076E+03) & ($\pm$1.279E+04) & ($\pm$5.344E+02) & ($\pm$8.863E+03) \\
		\multirow{1}{*}{GLHF}  & 1.938E+04 & 2.630E+04 & 2.602E+04 & 2.427E+04 & 2.105E+04 & 2.758E+04 & 1.144E+04 & 2.637E+04 \\
		& ($\pm$2.861E+02) & ($\pm$4.491E+02) & ($\pm$1.420E+03) & ($\pm$4.544E+03) & ($\pm$7.118E+03) & ($\pm$2.324E+02) & ($\pm$1.558E+03) & ($\pm$1.186E+03) \\
		\multirow{1}{*}{LES}  & 2.006E+04 & 3.843E+04 & 5.741E+04 & 8.844E+03 & 4.891E+04 & 5.621E+04 & 5.386E+03 & 5.495E+04 \\
		& ($\pm$6.790E+02) & ($\pm$8.750E+02) & ($\pm$7.604E+03) & ($\pm$4.502E+03) & ($\pm$2.615E+03) & ($\pm$1.118E+02) & ($\pm$4.109E+01) & ($\pm$8.139E-08) \\
		\multirow{1}{*}{LGA}  & 1.199E+04 & 4.045E+04 & 3.700E+04 & 8.823E+03 & 1.438E+04 & 4.743E+04 & 9.984E+03 & 5.397E+04 \\
		& ($\pm$3.765E+03) & ($\pm$9.730E+02) & ($\pm$9.602E+03) & ($\pm$1.375E+02) & ($\pm$5.445E+03) & ($\pm$9.373E+03) & ($\pm$8.206E+02) & ($\pm$5.122E+03) \\
		\multirow{1}{*}{B2OPT}  & 1.478E+04 & 2.230E+04 & 1.455E+04 & 9.622E+03 & 1.281E+04 & \textbf{1.884E+04} & \textbf{5.600E+03} & 1.598E+04 \\
		& ($\pm$1.986E+03) & ($\pm$4.793E+03) & ($\pm$2.085E+03) & ($\pm$2.744E+03) & ($\pm$1.305E+03) & ($\pm$4.703E+03) & ($\pm$7.442E+01) & ($\pm$5.684E+03)\\
		\midrule
		\rowcolor{blue!5} \multirow{1}{*}{Ours}  & \textbf{9.485E+03} & \textbf{1.869E+04} & \textbf{1.197E+04} & \textbf{7.906E+03} & \textbf{1.105E+04} & 1.959E+04 & 9.635E+03 & \textbf{1.275E+04} \\
		\rowcolor{blue!5} & ($\pm$4.748E+02) & ($\pm$4.687E+03) & ($\pm$3.189E+02) & ($\pm$3.190E+02) & ($\pm$7.180E+01) & ($\pm$4.574E+03) & ($\pm$3.123E+02) & ($\pm$2.429E+02) \\

		\toprule
		\textbf{Method} & \texttt{Terrain 25} & \texttt{Terrain 26} & \texttt{Terrain 27} & \texttt{Terrain 28} & \texttt{Terrain 29} & \texttt{Terrain 30} & \texttt{Terrain 31} & \texttt{Terrain 32} \\
		\midrule
		\midrule
		\multirow{1}{*}{RNNOPT}  & 1.075E+05 & 1.663E+05 & 1.557E+05 & 1.268E+05 & 1.283E+05 & 1.073E+05 & 2.565E+04 & 7.610E+04 \\
		& ($\pm$0.000E+00) & ($\pm$0.000E+00) & ($\pm$0.000E+00) & ($\pm$1.455E-11) & ($\pm$0.000E+00) & ($\pm$0.000E+00) & ($\pm$0.000E+00) & ($\pm$0.000E+00) \\
		\multirow{1}{*}{DEDQN}  & 2.703E+04 & 8.210E+04 & 3.657E+04 & 8.004E+04 & 3.076E+04 & 5.740E+04 & 1.309E+04 & 5.576E+04 \\
		& ($\pm$4.661E+03) & ($\pm$3.986E+03) & ($\pm$5.295E+03) & ($\pm$8.886E+03) & ($\pm$8.143E+03) & ($\pm$5.678E+03) & ($\pm$3.133E+03) & ($\pm$4.886E+03) \\
		\multirow{1}{*}{GLHF}  & 2.640E+04 & 2.568E+04 & 2.579E+04 & 2.727E+04 & 2.578E+04 & 2.719E+04 & 1.934E+04 & 2.671E+04 \\
		& ($\pm$1.018E+03) & ($\pm$7.653E+02) & ($\pm$1.252E+03) & ($\pm$6.339E+02) & ($\pm$1.997E+02) & ($\pm$1.974E+02) & ($\pm$2.962E+02) & ($\pm$6.828E+02) \\
		\multirow{1}{*}{LES}  & 3.557E+04 & 5.693E+04 & 1.539E+04 & 6.578E+04 & 3.154E+04 & 8.512E+04 & 1.986E+04 & 3.831E+04 \\
		& ($\pm$2.997E-02) & ($\pm$9.930E+02) & ($\pm$8.869E+01) & ($\pm$2.669E+01) & ($\pm$3.163E+03) & ($\pm$5.447E-08) & ($\pm$8.531E+02) & ($\pm$9.005E+02) \\
		\multirow{1}{*}{LGA}  & 1.682E+04 & 5.325E+04 & 2.345E+04 & 5.933E+04 & 1.742E+04 & 4.371E+04 & 1.206E+04 & 4.053E+04 \\
		& ($\pm$4.730E+03) & ($\pm$4.207E+03) & ($\pm$4.544E+03) & ($\pm$4.942E+03) & ($\pm$4.493E+03) & ($\pm$9.596E+03) & ($\pm$3.835E+03) & ($\pm$9.651E+02) \\
		\multirow{1}{*}{B2OPT}  & 1.518E+04 & 2.220E+04 & 1.748E+04 & 2.399E+04 & 1.267E+04 & 2.490E+04 & 1.387E+04 & 1.843E+04 \\
		& ($\pm$2.285E+03) & ($\pm$1.127E+04) & ($\pm$9.842E+02) & ($\pm$1.483E+03) & ($\pm$5.207E+02) & ($\pm$8.359E+03) & ($\pm$1.222E+03) & ($\pm$5.783E+03)  \\
		\midrule
		\rowcolor{blue!5} \multirow{1}{*}{Ours} & \textbf{1.169E+04} & \textbf{2.127E+04} & \textbf{1.641E+04} & \textbf{2.166E+04} & \textbf{9.834E+03} & \textbf{1.611E+04} & \textbf{8.976E+03} & \textbf{1.320E+04} \\
		\rowcolor{blue!5} & ($\pm$6.543E+02) & ($\pm$1.324E+04) & ($\pm$3.218E+03) & ($\pm$7.680E+03) & ($\pm$3.420E+02) & ($\pm$5.398E+03) & ($\pm$6.216E+02) & ($\pm$9.004E+02) \\

		\toprule
		\textbf{Method} & \texttt{Terrain 33} & \texttt{Terrain 34} & \texttt{Terrain 35} & \texttt{Terrain 36} & \texttt{Terrain 37} & \texttt{Terrain 38} & \texttt{Terrain 39} & \texttt{Terrain 40} \\
		\midrule
		\midrule
		\multirow{1}{*}{RNNOPT}  & 5.770E+04 & 1.563E+05 & 5.535E+04 & 2.067E+05 & 1.074E+05 & 1.663E+05 & 1.077E+05 & 1.665E+05 \\
		& ($\pm$7.276E-12) & ($\pm$0.000E+00) & ($\pm$0.000E+00) & ($\pm$2.910E-11) & ($\pm$0.000E+00) & ($\pm$0.000E+00) & ($\pm$0.000E+00) & ($\pm$0.000E+00) \\
		\multirow{1}{*}{DEDQN}  & 2.114E+04 & 5.311E+04 & 8.795E+03 & 7.363E+04 & 2.691E+04 & 8.244E+04 & 2.673E+04 & 8.224E+04 \\
		& ($\pm$8.180E+03) & ($\pm$1.245E+04) & ($\pm$9.931E+01) & ($\pm$4.892E+03) & ($\pm$4.560E+03) & ($\pm$3.816E+03) & ($\pm$4.813E+03) & ($\pm$3.905E+03) \\
		\multirow{1}{*}{GLHF}  & 2.138E+04 & 2.753E+04 & 9.151E+03 & 2.594E+04 & 2.553E+04 & 2.741E+04 & 2.617E+04 & 2.670E+04 \\
		& ($\pm$6.367E+03) & ($\pm$9.987E+02) & ($\pm$3.592E+02) & ($\pm$1.448E+03) & ($\pm$1.363E+03) & ($\pm$9.253E+02) & ($\pm$7.117E+02) & ($\pm$4.503E+02) \\
		\multirow{1}{*}{LES}  & 4.643E+04 & 5.591E+04 & 1.505E+04 & 8.099E+04 & 3.557E+04 & 5.553E+04 & 4.123E+04 & 6.552E+04 \\
		& ($\pm$2.953E+01) & ($\pm$4.592E+01) & ($\pm$4.304E+03) & ($\pm$1.280E+04) & ($\pm$3.013E-02) & ($\pm$5.565E+01) & ($\pm$3.239E+03) & ($\pm$4.133E+01) \\
		\multirow{1}{*}{LGA}  & 1.451E+04 & 4.678E+04 & 8.497E+03 & 5.835E+04 & 2.025E+04 & 5.300E+04 & 1.665E+04 & 5.307E+04 \\
		& ($\pm$5.359E+03) & ($\pm$9.310E+03) & ($\pm$4.192E+02) & ($\pm$8.511E+03) & ($\pm$4.677E+02) & ($\pm$4.224E+03) & ($\pm$4.764E+03) & ($\pm$4.185E+03) \\
		\multirow{1}{*}{B2OPT}  & 1.566E+04 & 2.298E+04 & 9.151E+03 & 1.814E+04 & 1.407E+04 & \textbf{1.524E+04} & 1.330E+04 & \textbf{1.964E+04} \\
		& ($\pm$3.716E+03) & ($\pm$6.170E+03) & ($\pm$3.592E+02) & ($\pm$4.785E+03) & ($\pm$2.307E+03) & ($\pm$1.760E+03) & ($\pm$1.433E+03) & ($\pm$8.551E+03) \\
		\midrule
		\rowcolor{blue!5} \multirow{1}{*}{Ours}   & \textbf{1.043E+04} & \textbf{1.592E+04} & \textbf{8.330E+03} & \textbf{1.152E+04} & \textbf{1.179E+04} & 2.477E+04 & \textbf{1.156E+04} & 2.137E+04 \\
		\rowcolor{blue!5} & ($\pm$4.851E+02) & ($\pm$5.090E+03) & ($\pm$1.275E+02) & ($\pm$5.383E+02) & ($\pm$4.478E+02) & ($\pm$1.137E+04) & ($\pm$3.861E+02) & ($\pm$1.302E+04) \\

		\bottomrule
	\end{tabular} 
\end{table*}
}

\newcommand{\TabSupAblUnShared}{%
\begin{table*}[tbp]
	\centering
	\scriptsize
	\setlength{\tabcolsep}{8pt}  
	\caption{
		Evaluation results (mean ± std) of various representative methods on the \emph{UAV} benchmark, consisting of 56 distinct terrain scenarios. 
		Each value is averaged over 10 independent test runs. 
		The last 16 terrains are shown in this table.
	}
	\label{tab:UAV-2}
	\begin{tabular}{lcccccccc} 
		\toprule
		\textbf{Method} & \texttt{Terrain 41} & \texttt{Terrain 42} & \texttt{Terrain 43} & \texttt{Terrain 44} & \texttt{Terrain 45} & \texttt{Terrain 46} & \texttt{Terrain 47} & \texttt{Terrain 48} \\
		\midrule
		\midrule
		\multirow{1}{*}{RNNOPT}  & 6.601E+04 & 1.868E+05 & 3.685E+04 & 1.067E+05 & 7.903E+04 & 5.708E+04 & 6.631E+04 & 1.869E+05 \\
		& ($\pm$0.000E+00) & ($\pm$0.000E+00) & ($\pm$0.000E+00) & ($\pm$0.000E+00) & ($\pm$0.000E+00) & ($\pm$0.000E+00) & ($\pm$0.000E+00) & ($\pm$0.000E+00) \\
		\multirow{1}{*}{DEDQN}  & 1.924E+04 & 7.102E+04 & 1.265E+04 & 3.652E+04 & 1.056E+04 & 5.700E+04 & 1.941E+04 & 7.140E+04 \\
		& ($\pm$4.827E+02) & ($\pm$7.717E+03) & ($\pm$5.148E+03) & ($\pm$1.244E+04) & ($\pm$1.275E+03) & ($\pm$8.795E+03) & ($\pm$8.967E+01) & ($\pm$7.812E+03) \\
		\multirow{1}{*}{GLHF}  & 2.155E+04 & 2.442E+04 & 1.708E+04 & 2.904E+04 & 1.222E+04 & 2.669E+04 & 2.156E+04 & 2.607E+04 \\
		& ($\pm$3.149E+03) & ($\pm$1.618E+03) & ($\pm$4.105E+03) & ($\pm$7.656E+03) & ($\pm$1.523E+03) & ($\pm$5.884E+02) & ($\pm$2.720E+03) & ($\pm$4.485E+02) \\
		\multirow{1}{*}{LES}  & 1.506E+04 & 7.594E+04 & 1.614E+04 & 2.642E+04 & 5.485E+03 & 6.494E+04 & 1.509E+04 & 7.605E+04 \\
		& ($\pm$3.808E+01) & ($\pm$7.400E+03) & ($\pm$7.082E+01) & ($\pm$4.099E+03) & ($\pm$4.525E+01) & ($\pm$8.137E-08) & ($\pm$3.570E+01) & ($\pm$7.294E+03) \\
		\multirow{1}{*}{LGA}  & 1.949E+04 & 1.798E+04 & 9.028E+03 & 2.732E+04 & 1.045E+04 & 5.427E+04 & 1.973E+04 & 1.818E+04 \\
		& ($\pm$6.219E+02) & ($\pm$9.140E+03) & ($\pm$1.456E+03) & ($\pm$9.094E+03) & ($\pm$1.016E+03) & ($\pm$5.029E+03) & ($\pm$5.082E+02) & ($\pm$9.160E+03) \\
		\multirow{1}{*}{B2OPT}  & 1.474E+04 & 1.368E+04 & 1.226E+04 & 2.426E+04 & \textbf{5.524E+03} & 1.970E+04 & 1.396E+04 & 1.588E+04 \\
		& ($\pm$9.304E+02) & ($\pm$1.287E+03) & ($\pm$9.304E+02) & ($\pm$5.410E+03) & ($\pm$4.132E+01) & ($\pm$4.988E+03) & ($\pm$1.215E+03) & ($\pm$4.696E+03) \\
		\midrule
		\rowcolor{blue!5} \multirow{1}{*}{Ours}   & \textbf{1.129E+04} & \textbf{1.070E+04} & \textbf{9.214E+03} & \textbf{1.881E+04} & 9.921E+03 & \textbf{1.281E+04} & \textbf{1.172E+04} & \textbf{1.079E+04} \\
		\rowcolor{blue!5} & ($\pm$1.076E+03) & ($\pm$7.209E+02) & ($\pm$1.849E+02) & ($\pm$1.762E+02) & ($\pm$5.614E+02) & ($\pm$2.031E+02) & ($\pm$1.104E+03) & ($\pm$6.357E+02)  \\
		
		\toprule
		\textbf{Method} & \texttt{Terrain 49} & \texttt{Terrain 50} & \texttt{Terrain 51} & \texttt{Terrain 52} & \texttt{Terrain 53} & \texttt{Terrain 54} & \texttt{Terrain 55} & \texttt{Terrain 56} \\
		\midrule
		\midrule
		\multirow{1}{*}{RNNOPT}  & 2.588E+04 & 1.068E+05 & 6.876E+04 & 1.881E+05 & 6.683E+04 & 1.871E+05 & 3.795E+04 & 1.077E+05 \\
		& ($\pm$0.000E+00) & ($\pm$1.455E-11) & ($\pm$0.000E+00) & ($\pm$0.000E+00) & ($\pm$0.000E+00) & ($\pm$0.000E+00) & ($\pm$0.000E+00) & ($\pm$0.000E+00) \\
		\multirow{1}{*}{DEDQN}  & 1.940E+04 & 4.706E+04 & 2.115E+04 & 7.279E+04 & 1.977E+04 & 7.124E+04 & 1.351E+04 & 3.765E+04 \\
		& ($\pm$3.528E+02) & ($\pm$4.332E+03) & ($\pm$1.526E+02) & ($\pm$8.366E+03) & ($\pm$5.183E+02) & ($\pm$7.790E+03) & ($\pm$5.257E+03) & ($\pm$1.220E+04) \\
		\multirow{1}{*}{GLHF}  & 1.296E+04 & 2.596E+04 & 2.340E+04 & 2.691E+04 & 2.211E+04 & 2.495E+04 & 1.765E+04 & 3.027E+04 \\
		& ($\pm$4.324E+03) & ($\pm$1.405E+02) & ($\pm$2.258E+03) & ($\pm$5.916E+02) & ($\pm$3.318E+03) & ($\pm$3.760E+02) & ($\pm$4.027E+03) & ($\pm$7.403E+03) \\
		\multirow{1}{*}{LES}  & 9.874E+03 & 6.591E+04 & 1.537E+04 & 7.715E+04 & 2.143E+04 & 7.955E+04 & 1.802E+04 & 3.088E+04 \\
		& ($\pm$3.625E+03) & ($\pm$6.537E+03) & ($\pm$3.658E+01) & ($\pm$6.604E+03) & ($\pm$2.665E+03) & ($\pm$1.155E+04) & ($\pm$6.166E+02) & ($\pm$6.181E+03) \\
		\multirow{1}{*}{LGA}  & 8.652E+03 & 2.203E+04 & 2.151E+04 & 1.951E+04 & 1.989E+04 & 1.830E+04 & 9.623E+03 & 2.847E+04 \\
		& ($\pm$4.710E+02) & ($\pm$7.832E+03) & ($\pm$7.476E+02) & ($\pm$8.936E+03) & ($\pm$6.015E+02) & ($\pm$9.152E+03) & ($\pm$1.176E+03) & ($\pm$9.489E+03) \\
		\multirow{1}{*}{B2OPT}  & 1.108E+04 & 1.849E+04 & \textbf{1.586E+04} & 2.384E+04 & \textbf{1.365E+04} & 1.640E+04 & 1.363E+04 & 2.618E+04 \\
		& ($\pm$1.683E+03) & ($\pm$4.349E+03) & ($\pm$3.884E+02) & ($\pm$7.014E+02) & ($\pm$1.408E+03) & ($\pm$5.132E+03) & ($\pm$2.530E+03) & ($\pm$9.439E+03) \\
		\midrule
		\rowcolor{blue!5} \multirow{1}{*}{Ours} & \textbf{8.500E+03} & \textbf{1.151E+04} & 1.778E+04 & \textbf{1.133E+04} & 1.420E+04 & \textbf{1.088E+04} & \textbf{9.949E+03} & \textbf{1.976E+04} \\
		\rowcolor{blue!5} & ($\pm$5.858E+02) & ($\pm$5.849E+02) & ($\pm$3.401E+03) & ($\pm$6.799E+02) & ($\pm$3.180E+03) & ($\pm$6.448E+02) & ($\pm$5.330E+01) & ($\pm$1.875E+02) \\ 
		
		\bottomrule
	\end{tabular} 
\end{table*}
}

\newcommand{\TabSupbbobTensurogate}{%
\begin{table*}[htbp]
	\centering
	\scriptsize
	\setlength{\tabcolsep}{2pt}  
	\caption{Out-of-distribution optimization performance (mean ± std) of various representative methods tested on \emph{BBOB-10D-surogate}. Each entry is calculated from 10 independent test runs. The best-performing entries are highlighted in bold.}
	\label{tab:bbob-10d-surogate}
	\begin{tabular}{lcccccccc}
		\toprule
		\textbf{Method} & \texttt{Rastrigin} & \texttt{Buche\_Ras} & \texttt{Step\_Ell} & \texttt{Rosenbrock\_ori} & \texttt{Rosenbrock\_rot} & \texttt{Bent\_Cigar} & \texttt{Different\_Pow} & \texttt{Rastrigin\_F15} \\
		\midrule
		\multirow{1}{*}{RNNOPT}  & 3.304E+02 & 9.794E+03 & 3.630E+02 & 1.770E+04 & 6.023E+01 & 8.109E+07 & 3.539E+01 & 3.122E+02 \\
		& ($\pm$0.000E+00) & ($\pm$0.000E+00) & ($\pm$0.000E+00) & ($\pm$0.000E+00) & ($\pm$0.000E+00) & ($\pm$0.000E+00) & ($\pm$0.000E+00) & ($\pm$0.000E+00) \\
		\multirow{1}{*}{DEDQN}  & 1.787E+02 & 4.325E+02 & 6.788E+01 & 6.970E+03 & 1.115E+04 & 2.643E+07 & 1.009E+01 & 1.762E+02 \\
		& ($\pm$2.505E+01) & ($\pm$1.378E+02) & ($\pm$5.073E+00) & ($\pm$2.093E+03) & ($\pm$2.527E+03) & ($\pm$1.157E+07) & ($\pm$2.227E+00) & ($\pm$6.962E+01) \\
		\multirow{1}{*}{LES}  & 2.605E+03 & 2.659E+03 & 5.276E+02 & 3.211E+03 & 1.538E+03 & 1.554E+07 & 1.906E+03 & 2.590E+03 \\
		& ($\pm$1.254E+01) & ($\pm$2.124E+02) & ($\pm$4.530E+00) & ($\pm$1.459E+02) & ($\pm$2.759E+00) & ($\pm$1.270E+06) & ($\pm$3.548E-01) & ($\pm$9.267E+00) \\
		\multirow{1}{*}{LGA}  & 1.794E+02 & 3.983E+02 & 1.074E+02 & 2.367E+04 & 1.847E+04 & 3.295E+07 & 1.254E+01 & 1.614E+02 \\
		& ($\pm$4.037E+01) & ($\pm$1.896E+02) & ($\pm$1.961E+01) & ($\pm$2.145E+04) & ($\pm$7.430E+03) & ($\pm$1.053E+07) & ($\pm$4.550E-01) & ($\pm$7.695E+01) \\
		\multirow{1}{*}{GLHF}  & 2.446E+02 & 8.649E+02 & 1.711E+02 & 1.912E+04 & 1.127E+04 & 3.711E+07 & 1.806E+01 & 2.215E+02 \\
		& ($\pm$4.885E+01) & ($\pm$5.709E+02) & ($\pm$1.664E+01) & ($\pm$9.174E+03) & ($\pm$1.288E+03) & ($\pm$1.151E+07) & ($\pm$1.153E+00) & ($\pm$4.170E+01) \\
		\multirow{1}{*}{B2OPT}  & 8.877E+01 & 1.416E+02 & 1.858E+01 & 5.183E+02 & 3.359E+01 & 9.190E+06 & 3.331E+00 & 1.008E+02 \\
		& ($\pm$1.903E+01) & ($\pm$7.587E+00) & ($\pm$3.339E+00) & ($\pm$1.167E+02) & ($\pm$4.976E-01) & ($\pm$1.561E+06) & ($\pm$9.869E-01) & ($\pm$3.868E+00) \\
		\midrule
		\rowcolor{blue!5} \multirow{1}{*}{Ours}  & 7.165E+01 & 9.191E+01 & 1.348E+01 & 9.369E+01 & 1.087E+01 & 1.095E+06 & 1.337E+00 & 6.993E+01 \\
		\rowcolor{blue!5} & ($\pm$1.205E+01) & ($\pm$1.117E+01) & ($\pm$1.656E+00) & ($\pm$2.903E+01) & ($\pm$2.326E+00) & ($\pm$2.296E+05) & ($\pm$1.273E-01) & ($\pm$3.951E+00) \\
		\toprule
		\textbf{Method} & \texttt{Weierstrass} & \texttt{Schaffers} & \texttt{Schaffers\_hig} & \texttt{Composite\_Gri} & \texttt{Gallagher\_101Pea} & \texttt{Gallagher\_21Pea} & \texttt{Katsuura} & \texttt{Lunacek\_bi} \\
		\midrule
		\multirow{1}{*}{RNNOPT}  & 4.581E+01 & 1.429E+01 & 1.242E+02 & 4.951E+00 & 7.088E+01 & 7.325E+01 & 4.521E+00 & 8.376E+01 \\
		& ($\pm$0.000E+00) & ($\pm$0.000E+00) & ($\pm$0.000E+00) & ($\pm$0.000E+00) & ($\pm$0.000E+00) & ($\pm$0.000E+00) & ($\pm$0.000E+00) & ($\pm$0.000E+00) \\
		\multirow{1}{*}{DEDQN}  & 2.300E+01 & 9.034E+00 & 3.502E+01 & 1.058E+01 & 4.718E+01 & 4.790E+01 & 3.236E+00 & 1.403E+02 \\
		& ($\pm$2.691E+00) & ($\pm$3.442E-01) & ($\pm$5.593E+00) & ($\pm$7.033E-01) & ($\pm$7.828E+00) & ($\pm$1.606E+01) & ($\pm$3.434E-01) & ($\pm$2.937E+01) \\
		\multirow{1}{*}{LES}  & 1.908E+03 & 1.030E+02 & 1.230E+03 & 2.000E+03 & 9.346E+02 & 1.349E+03 & 1.002E+03 & 5.688E+02 \\
		& ($\pm$1.667E+00) & ($\pm$5.131E-01) & ($\pm$2.826E-01) & ($\pm$6.244E-03) & ($\pm$2.815E+00) & ($\pm$2.169E+00) & ($\pm$2.413E-01) & ($\pm$6.721E+00) \\
		\multirow{1}{*}{LGA}  & 1.653E+01 & 9.011E+00 & 3.239E+01 & 1.307E+01 & 4.406E+01 & 6.217E+01 & 1.991E+00 & 1.967E+02 \\
		& ($\pm$2.078E+00) & ($\pm$6.250E-01) & ($\pm$3.437E+00) & ($\pm$4.404E+00) & ($\pm$9.355E+00) & ($\pm$5.913E+00) & ($\pm$7.629E-01) & ($\pm$2.346E+01) \\
		\multirow{1}{*}{GLHF}  & 1.683E+01 & 8.558E+00 & 3.609E+01 & 8.791E+00 & 4.441E+01 & 5.212E+01 & 2.869E+00 & 1.594E+02 \\
		& ($\pm$4.295E+00) & ($\pm$1.248E+00) & ($\pm$5.200E+00) & ($\pm$5.094E-01) & ($\pm$3.220E+00) & ($\pm$8.082E+00) & ($\pm$6.562E-01) & ($\pm$5.139E+00) \\
		\multirow{1}{*}{B2OPT}  & 1.367E+01 & 3.199E+00 & 8.515E+00 & 7.589E-01 & 3.727E+01 & 2.759E+01 & 2.072E+00 & 8.035E+01 \\
		& ($\pm$4.833E+00) & ($\pm$2.136E-01) & ($\pm$4.683E-01) & ($\pm$1.661E-01) & ($\pm$1.005E+01) & ($\pm$6.623E+00) & ($\pm$3.789E-01) & ($\pm$5.748E+00) \\
		\midrule
		\rowcolor{blue!5} \multirow{1}{*}{Ours}  & 7.496E+00 & 2.595E+00 & 1.044E+01 & 3.979E-01 & 1.458E+01 & 1.536E+01 & 1.782E+00 & 6.457E+01 \\
		\rowcolor{blue!5} & ($\pm$1.332E+00) & ($\pm$6.233E-01) & ($\pm$2.331E+00) & ($\pm$1.711E-01) & ($\pm$4.802E+00) & ($\pm$1.555E+00) & ($\pm$8.599E-02) & ($\pm$7.252E+00) \\
		\bottomrule
	\end{tabular}
\end{table*}
}

\newcommand{\FigSupbbobConvershared}{%
\begin{figure*}[!tbp]
	\centering
	\begin{subfigure}[b]{0.24\textwidth}
		\includegraphics[width=\linewidth,
		trim=0 0 0 0, clip]{figures/abl/pics_9_SharedTrue_EMS/Ellipsoidal_high_cond_log_cost_curve.pdf}
	\end{subfigure}
	\begin{subfigure}[b]{0.24\textwidth}
		\includegraphics[width=\linewidth,
		trim=0 0 0 0, clip]{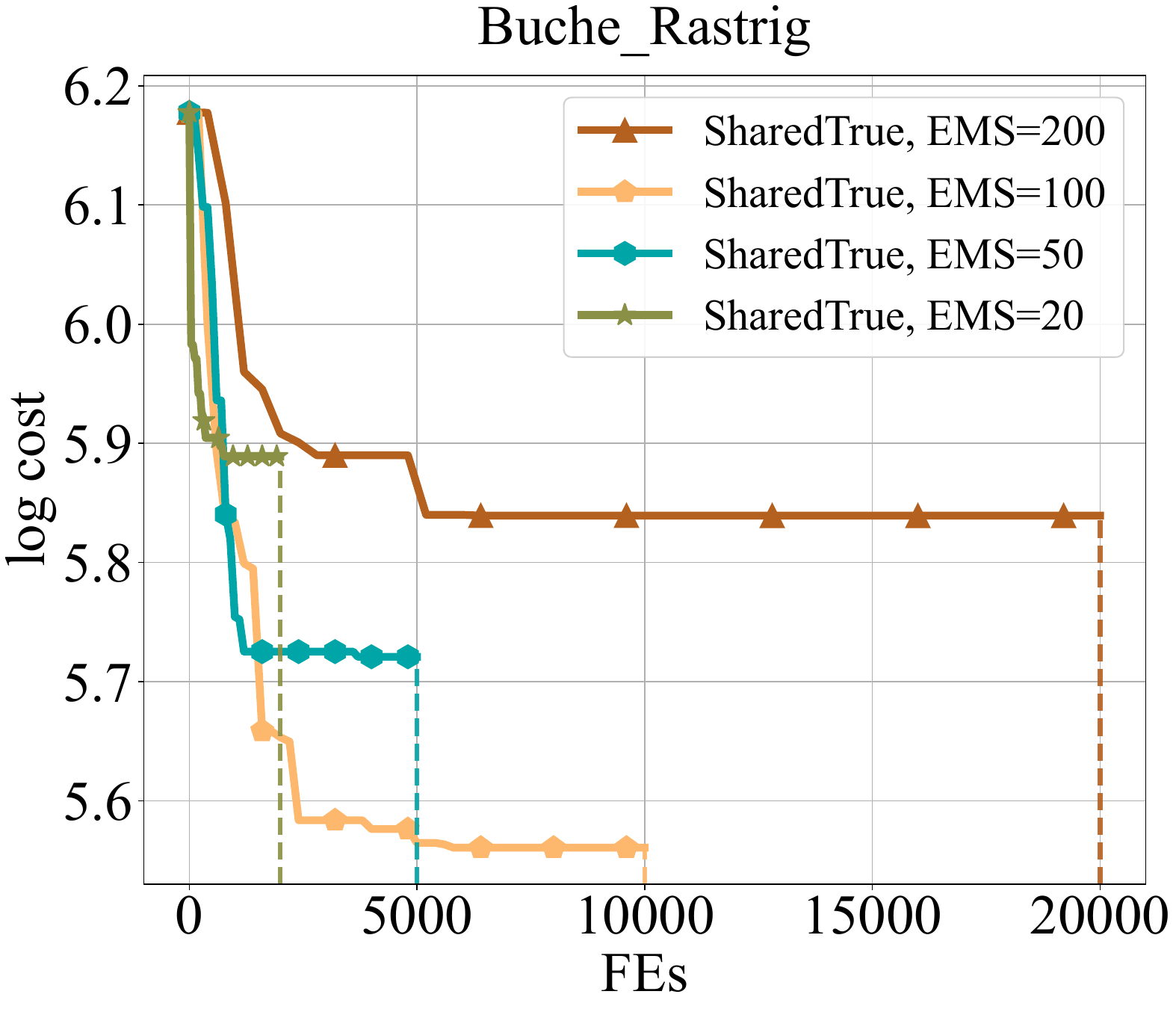}
	\end{subfigure} 
	\begin{subfigure}[b]{0.24\textwidth}
		\includegraphics[width=\linewidth,
		trim=0 0 0 0, clip]{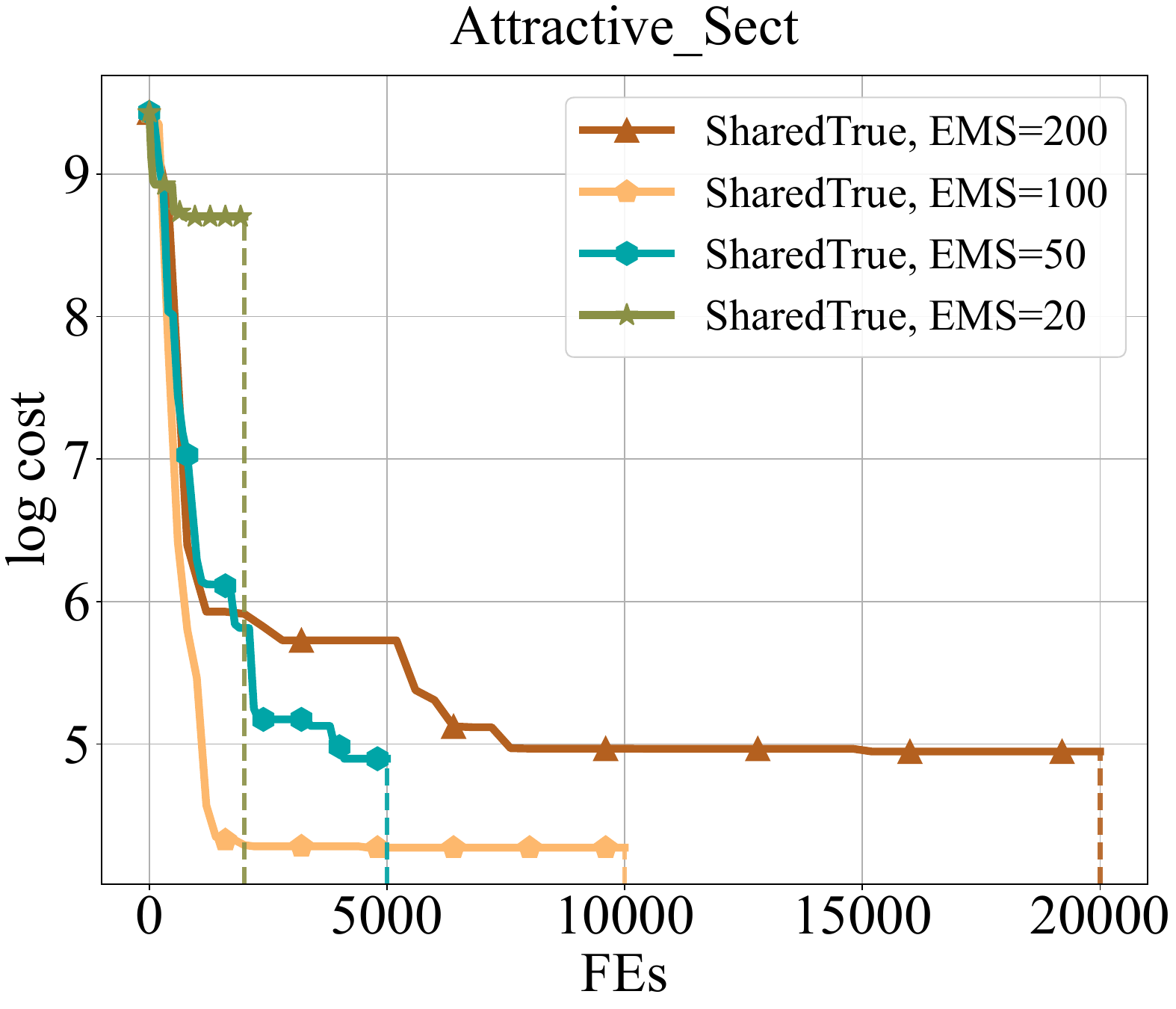}
	\end{subfigure}
	\begin{subfigure}[b]{0.24\textwidth}
		\includegraphics[width=\linewidth,
		trim=0 0 0 0, clip]{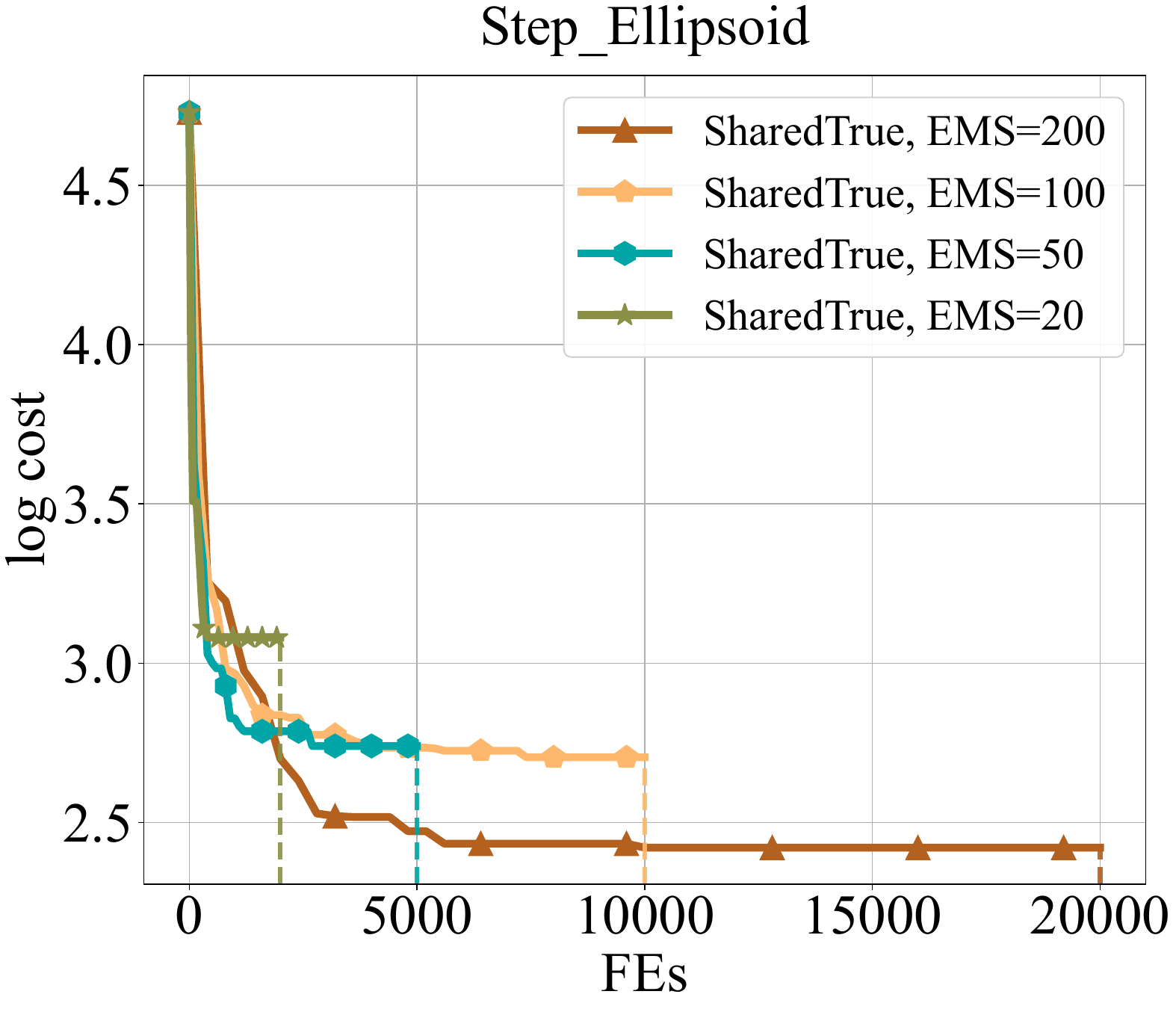}
	\end{subfigure}
	
	\begin{subfigure}[b]{0.24\textwidth}
		\includegraphics[width=\linewidth,
		trim=0 0 0 0, clip]{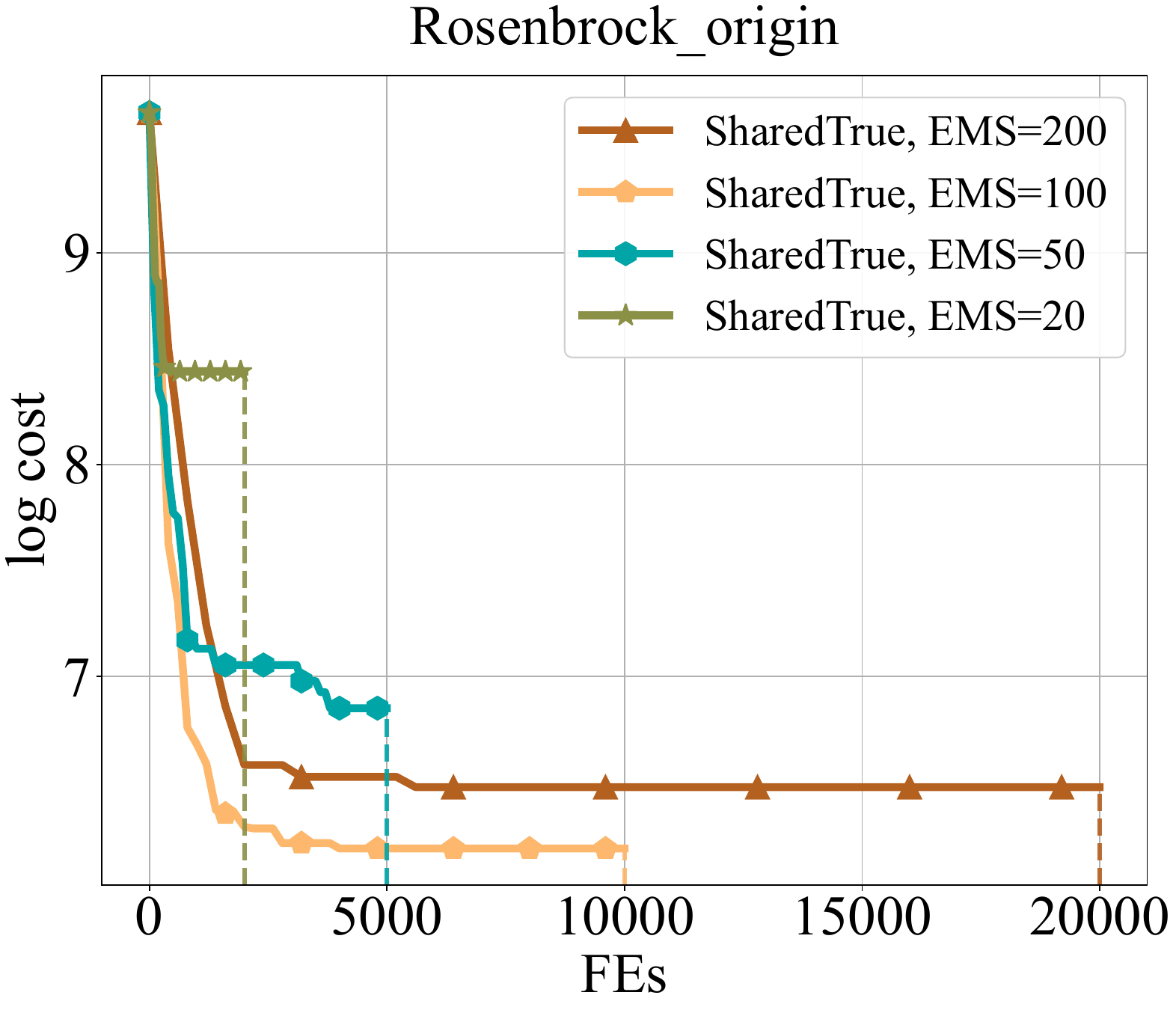}
	\end{subfigure}
	\begin{subfigure}[b]{0.24\textwidth}
		\includegraphics[width=\linewidth,
		trim=0 0 0 0, clip]{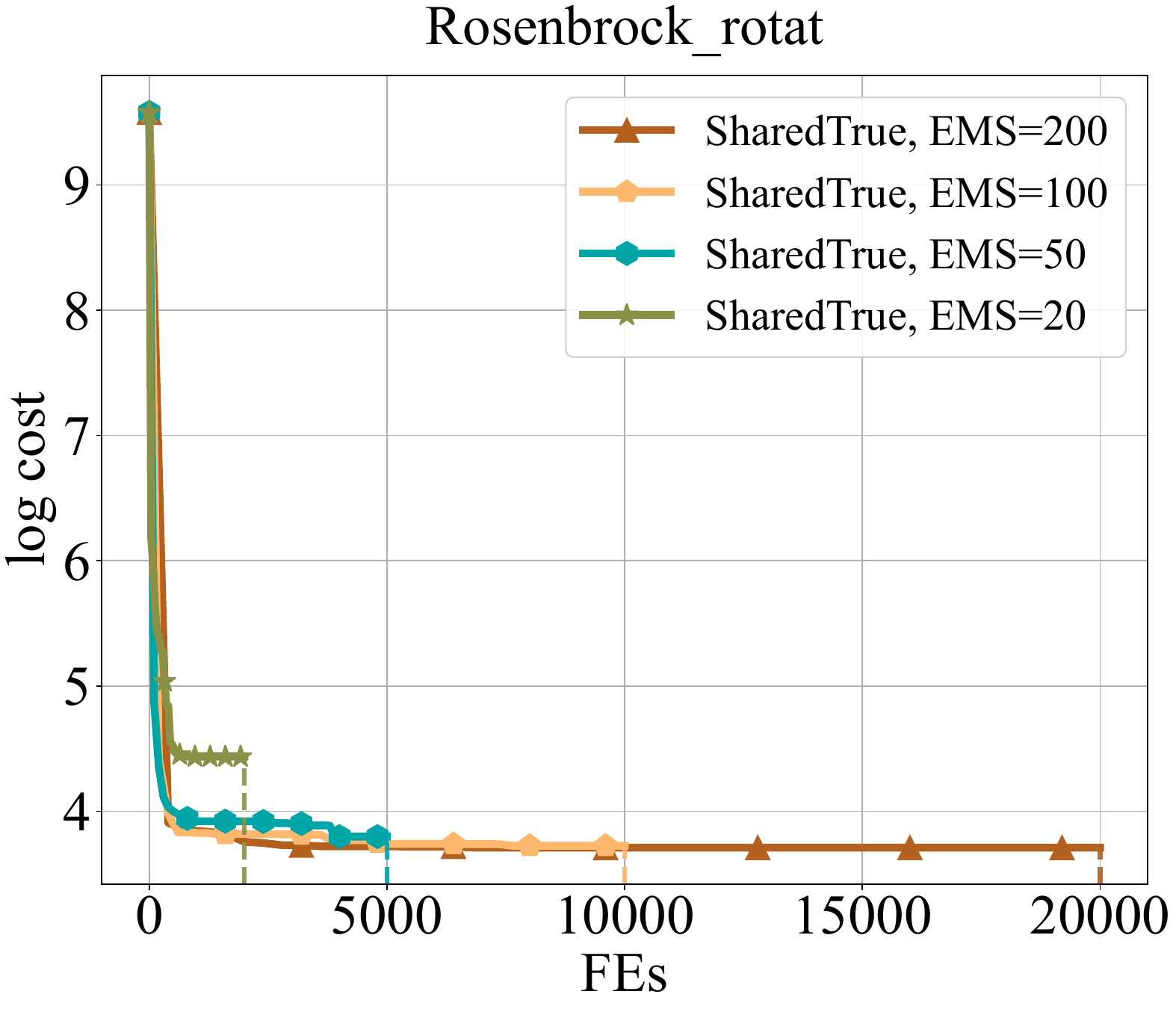}
	\end{subfigure} 
	\begin{subfigure}[b]{0.24\textwidth}
		\includegraphics[width=\linewidth,
		trim=0 0 0 0, clip]{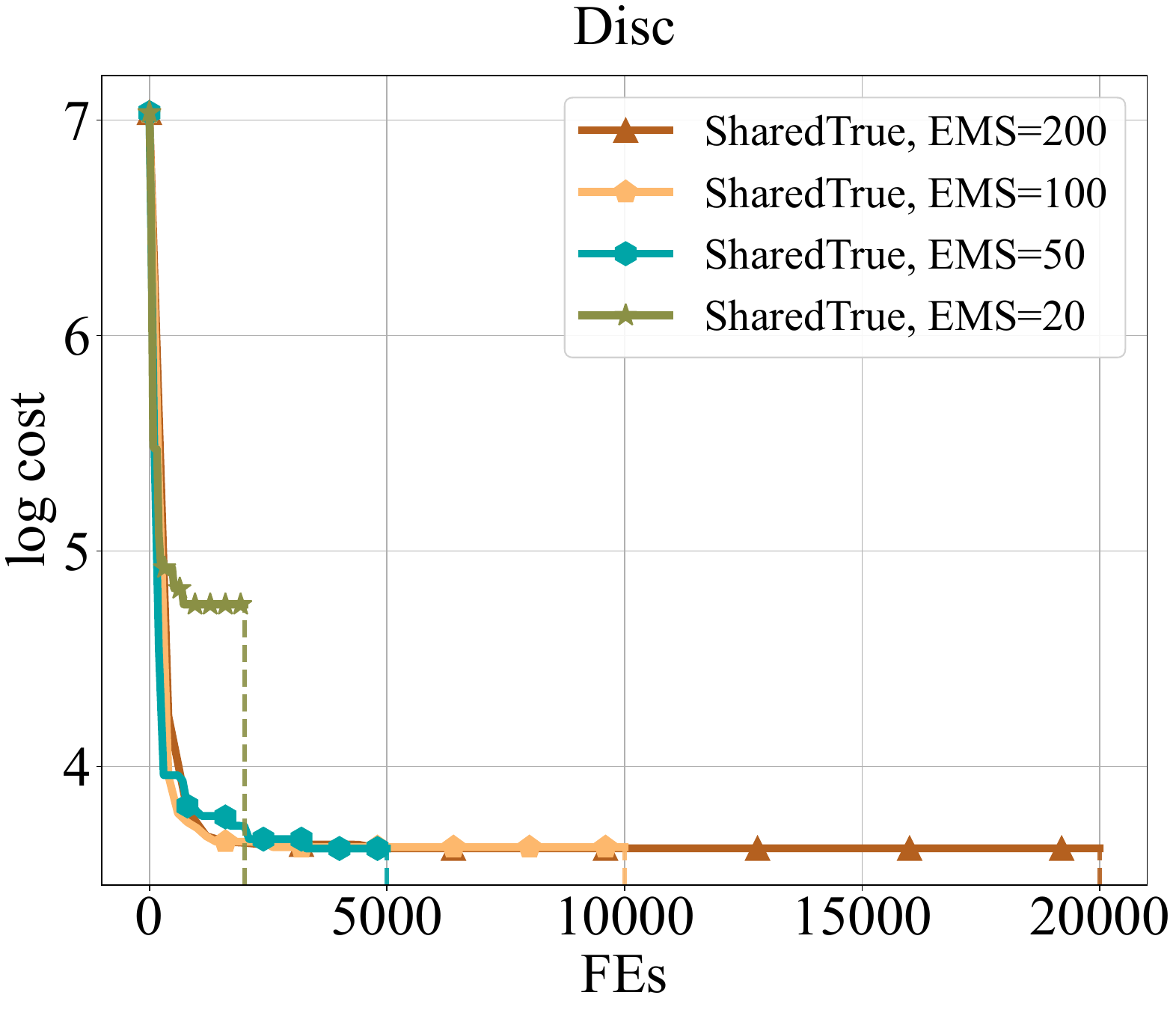}
	\end{subfigure}
	\begin{subfigure}[b]{0.24\textwidth}
		\includegraphics[width=\linewidth,
		trim=0 0 0 0, clip]{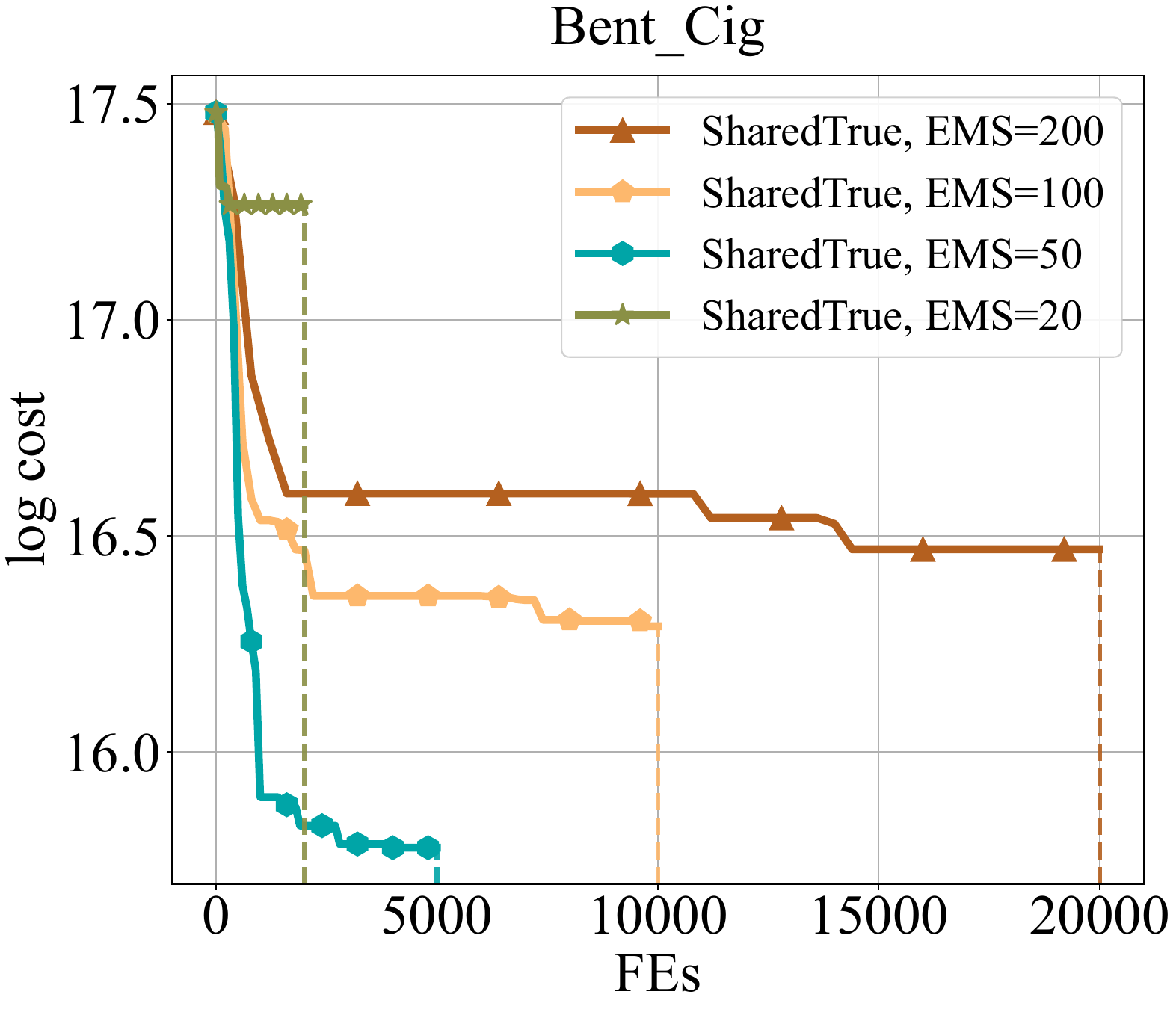}
	\end{subfigure}
	
	\begin{subfigure}[b]{0.24\textwidth}
		\includegraphics[width=\linewidth,
		trim=0 0 0 0, clip]{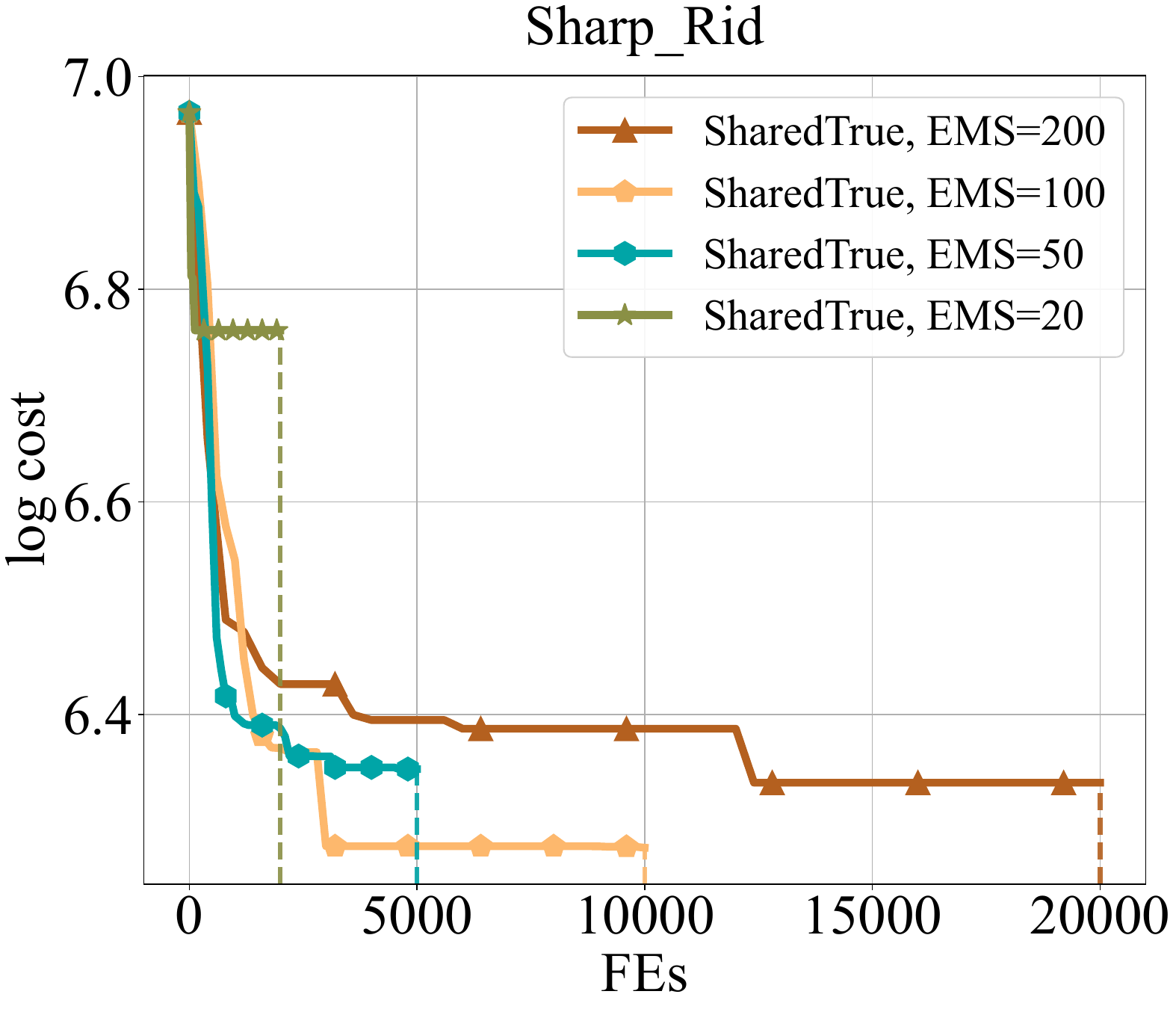}
	\end{subfigure}
	\begin{subfigure}[b]{0.24\textwidth}
		\includegraphics[width=\linewidth,
		trim=0 0 0 0, clip]{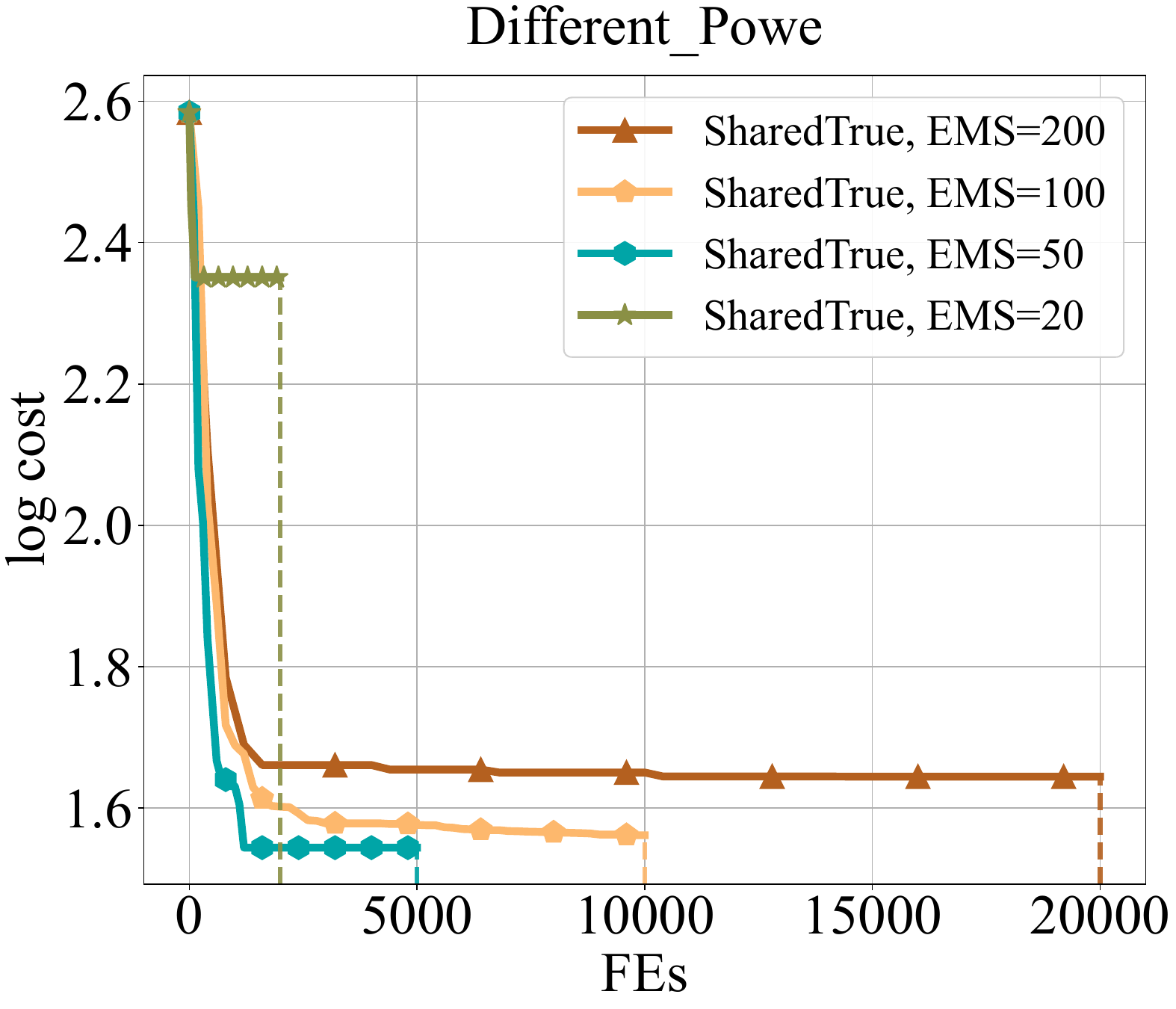}
	\end{subfigure} 
	\begin{subfigure}[b]{0.24\textwidth}
		\includegraphics[width=\linewidth,
		trim=0 0 0 0, clip]{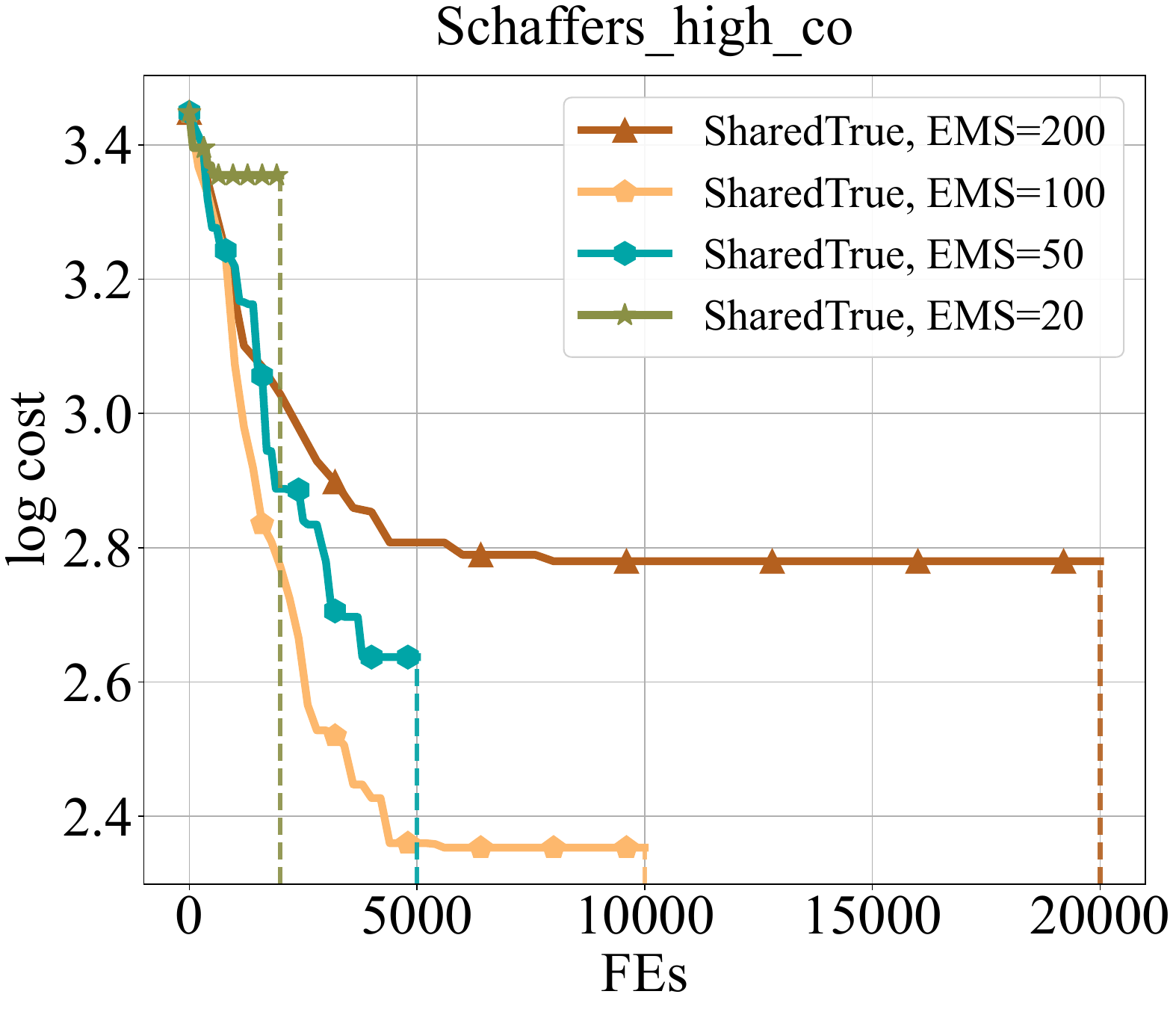}
	\end{subfigure}
	\begin{subfigure}[b]{0.24\textwidth}
		\includegraphics[width=\linewidth,
		trim=0 0 0 0, clip]{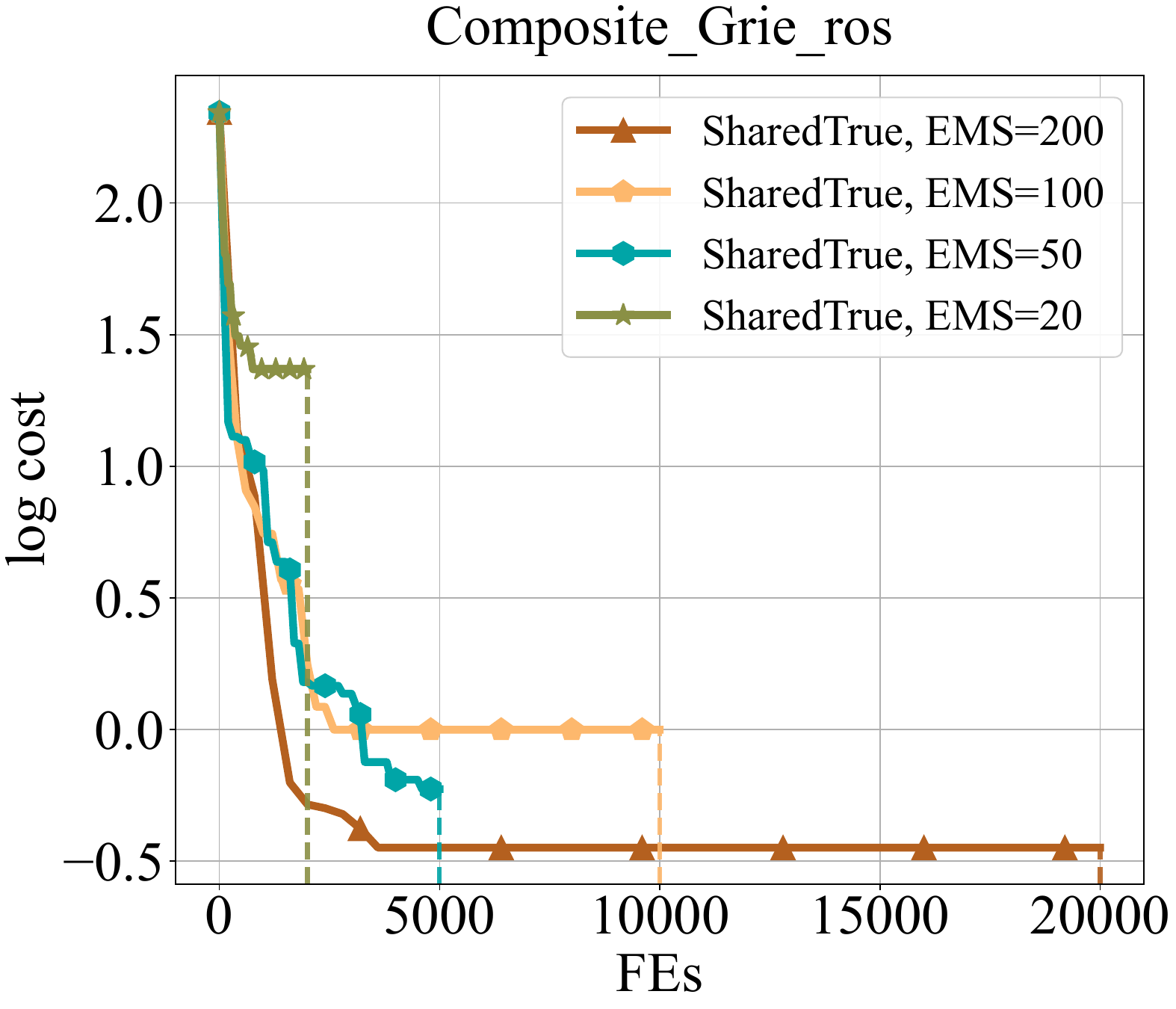}
	\end{subfigure}
	\begin{subfigure}[b]{0.24\textwidth}
		\includegraphics[width=\linewidth,
		trim=0 0 0 0, clip]{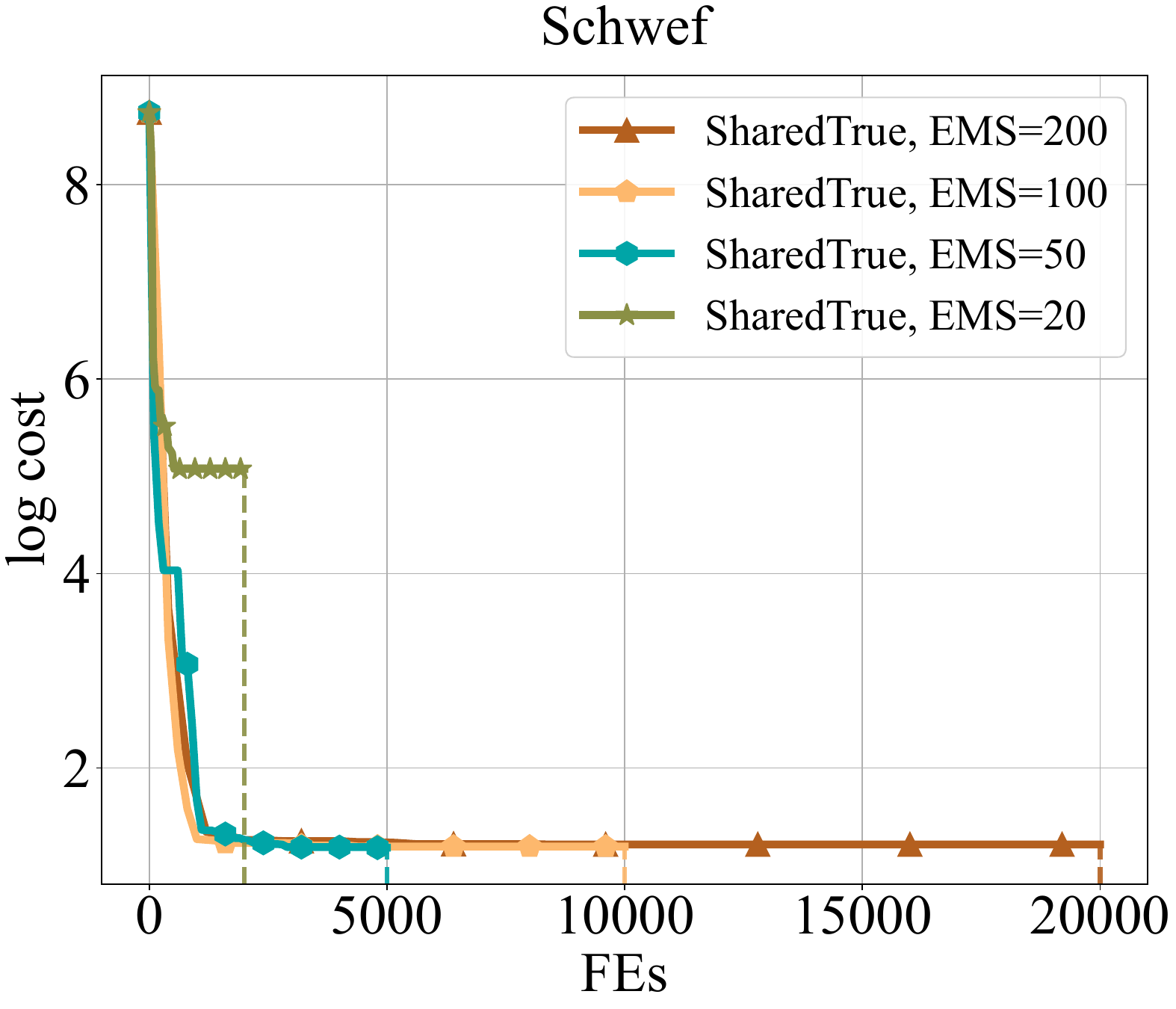}
	\end{subfigure}
	\begin{subfigure}[b]{0.24\textwidth}
		\includegraphics[width=\linewidth,
		trim=0 0 0 0, clip]{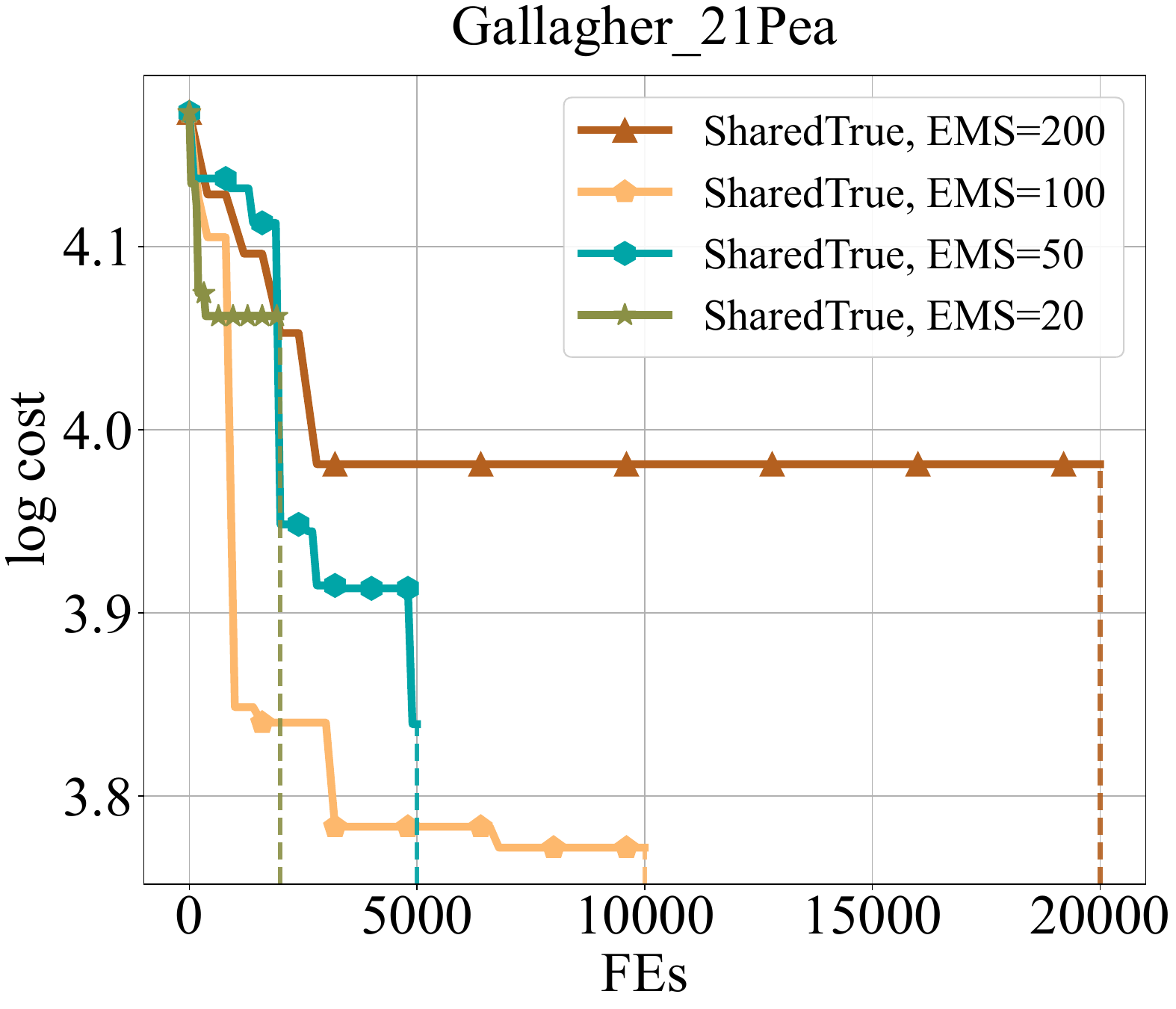}
	\end{subfigure} 
	\begin{subfigure}[b]{0.24\textwidth}
		\includegraphics[width=\linewidth,
		trim=0 0 0 0, clip]{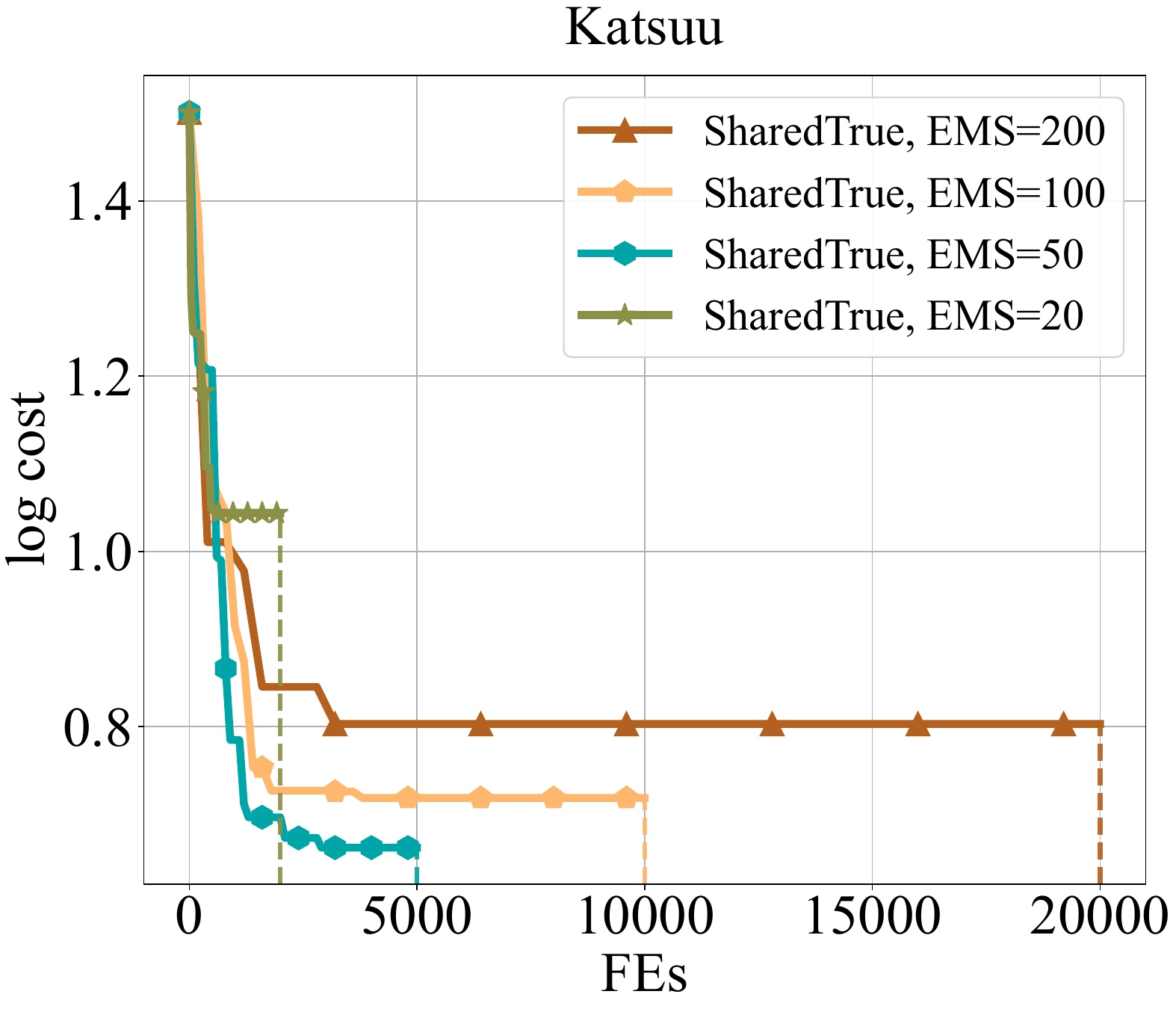}
	\end{subfigure}
	\begin{subfigure}[b]{0.24\textwidth}
		\includegraphics[width=\linewidth,
		trim=0 0 0 0, clip]{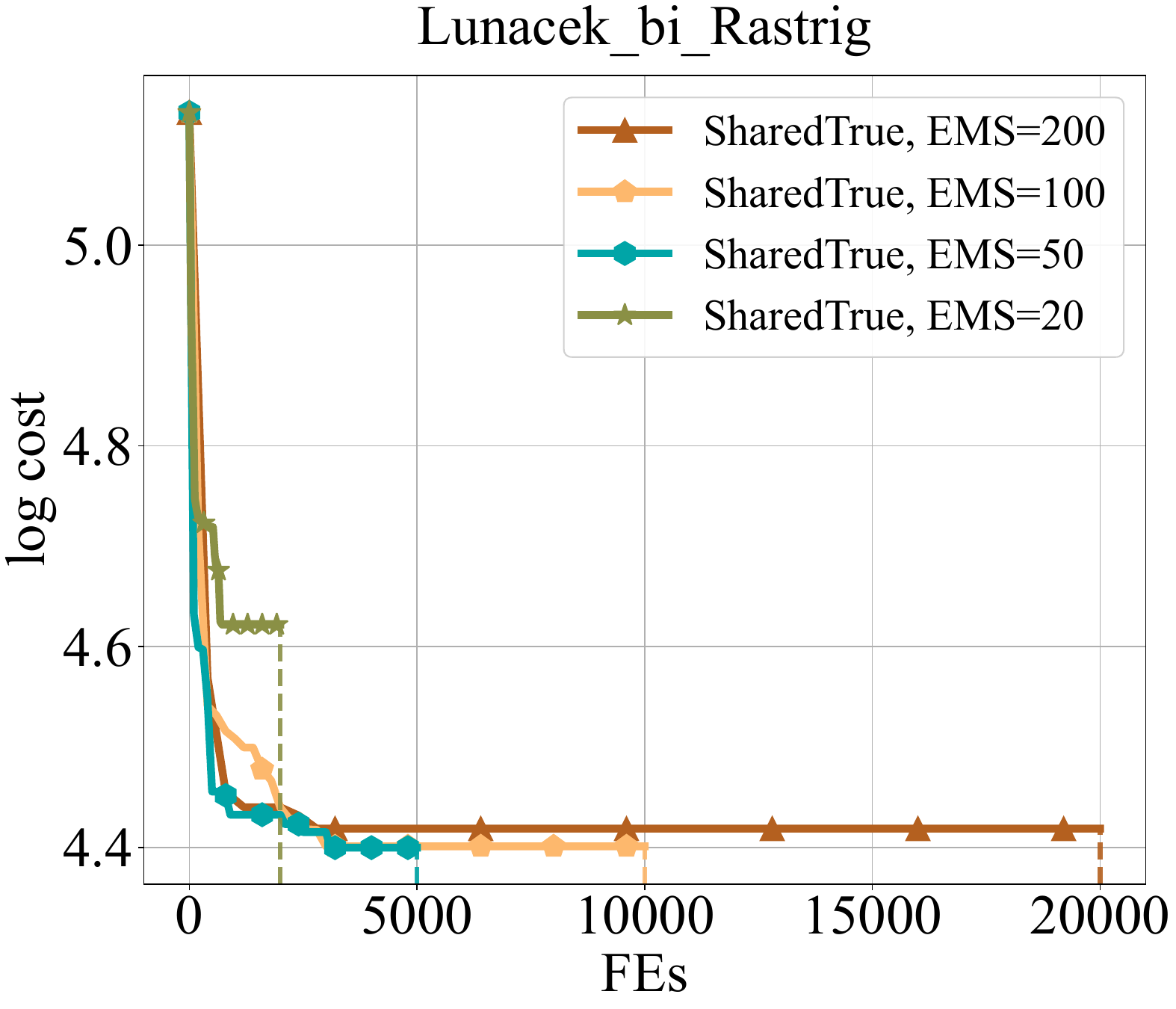}
	\end{subfigure} 
	\caption{
		Convergence behaviors under the \emph{shared}  parameterization of the 
		neural evolution operator~$\mathcal{O}_{\mathrm{evo}}$. 
		Log-cost convergence over all 16 BBOB-like test problems.  
		Vertical dashed lines denote the end of each agent’s evaluation budget.  
	}
	\label{fig:abl_shared}
\end{figure*}
}

\newcommand{\FigSupbbobConverunshared}{%
\begin{figure*}[!tbp]
	\centering
	\begin{subfigure}[b]{0.24\textwidth}
		\includegraphics[width=\linewidth,
		trim=0 0 0 0, clip]{figures/abl/pics_8_SharedFalse_EMS/Ellipsoidal_high_cond_log_cost_curve.pdf}
	\end{subfigure}
	\begin{subfigure}[b]{0.24\textwidth}
		\includegraphics[width=\linewidth,
		trim=0 0 0 0, clip]{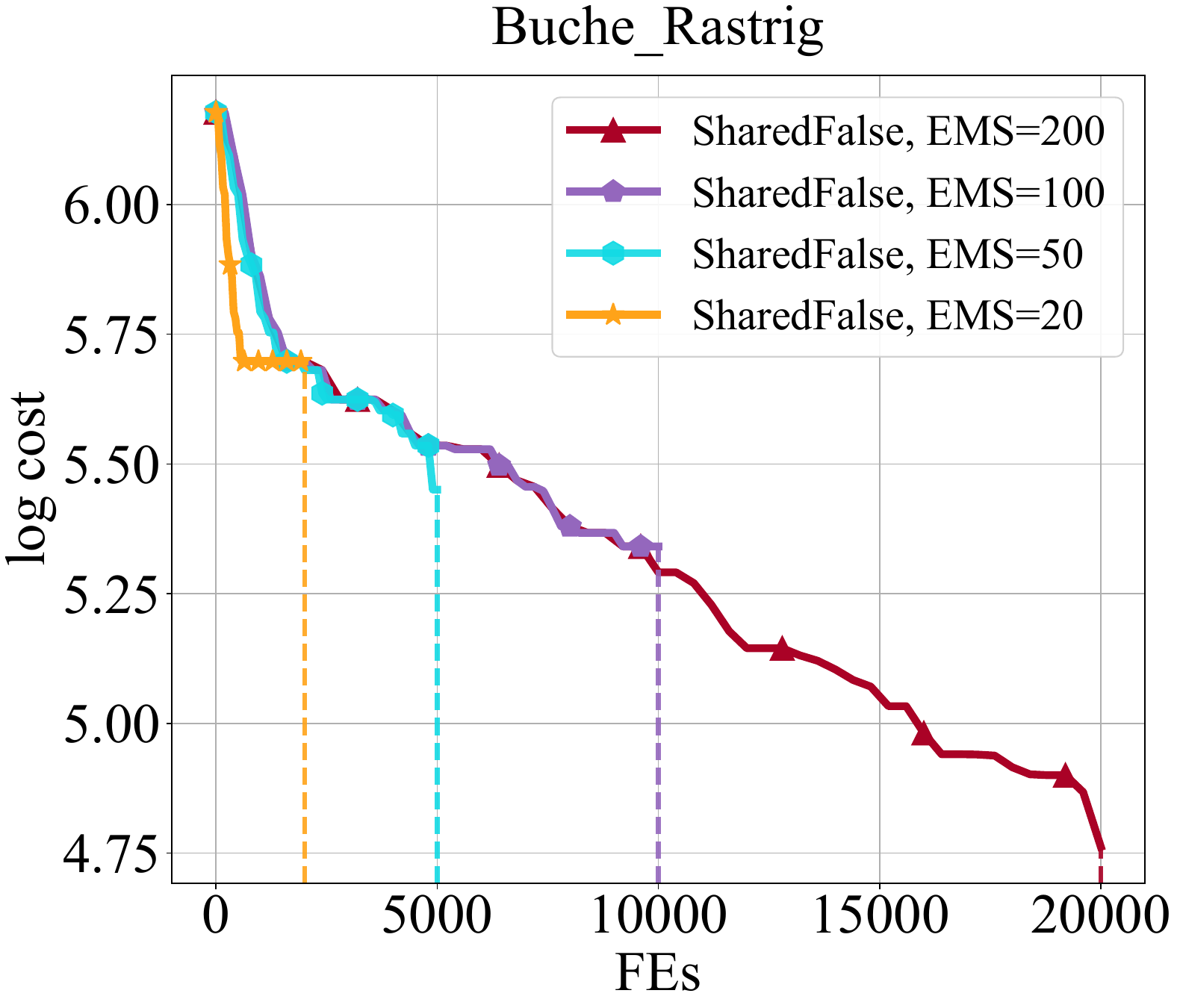}
	\end{subfigure} 
	\begin{subfigure}[b]{0.24\textwidth}
		\includegraphics[width=\linewidth,
		trim=0 0 0 0, clip]{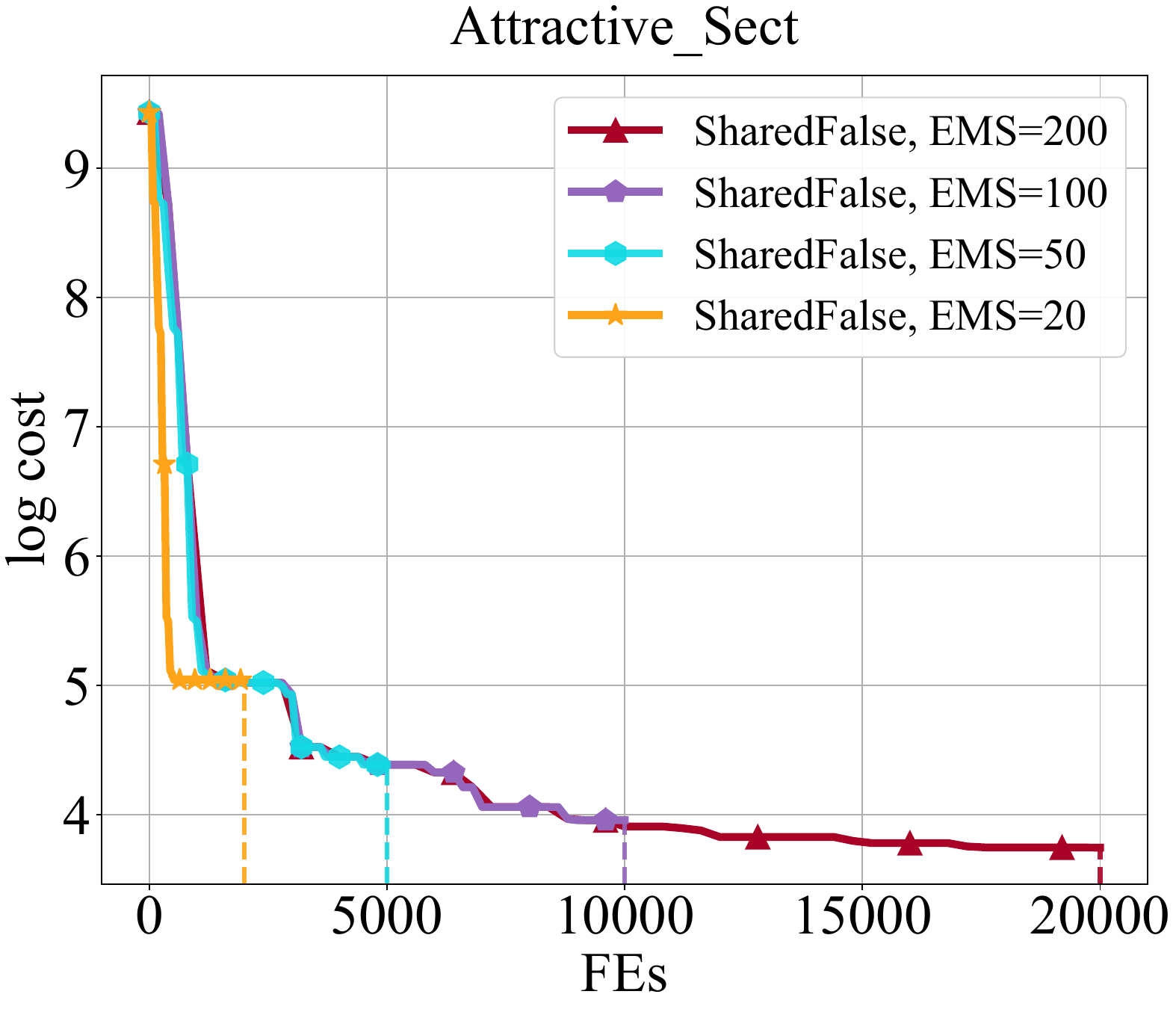}
	\end{subfigure}
	\begin{subfigure}[b]{0.24\textwidth}
		\includegraphics[width=\linewidth,
		trim=0 0 0 0, clip]{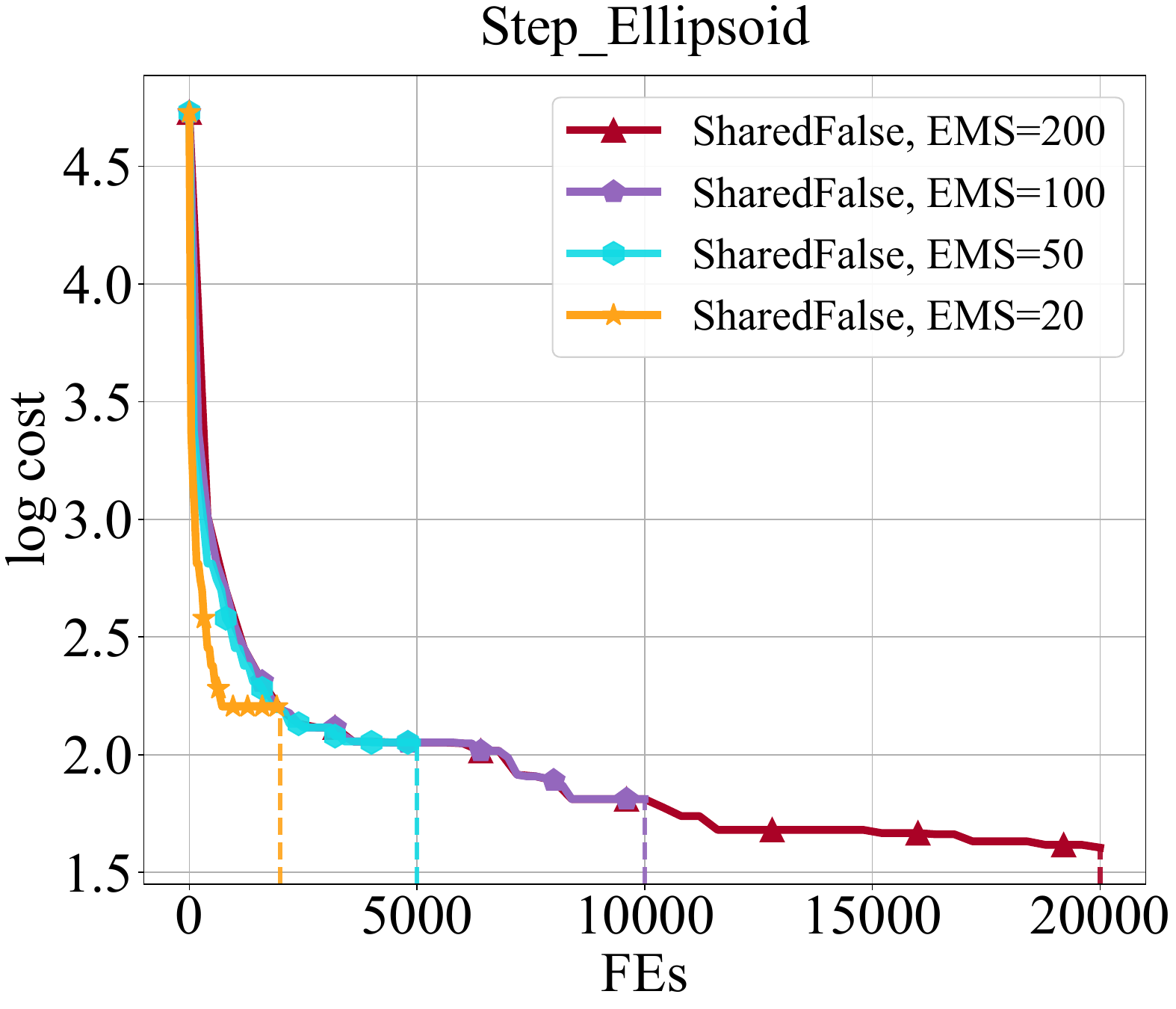}
	\end{subfigure}
	
	\begin{subfigure}[b]{0.24\textwidth}
		\includegraphics[width=\linewidth,
		trim=0 0 0 0, clip]{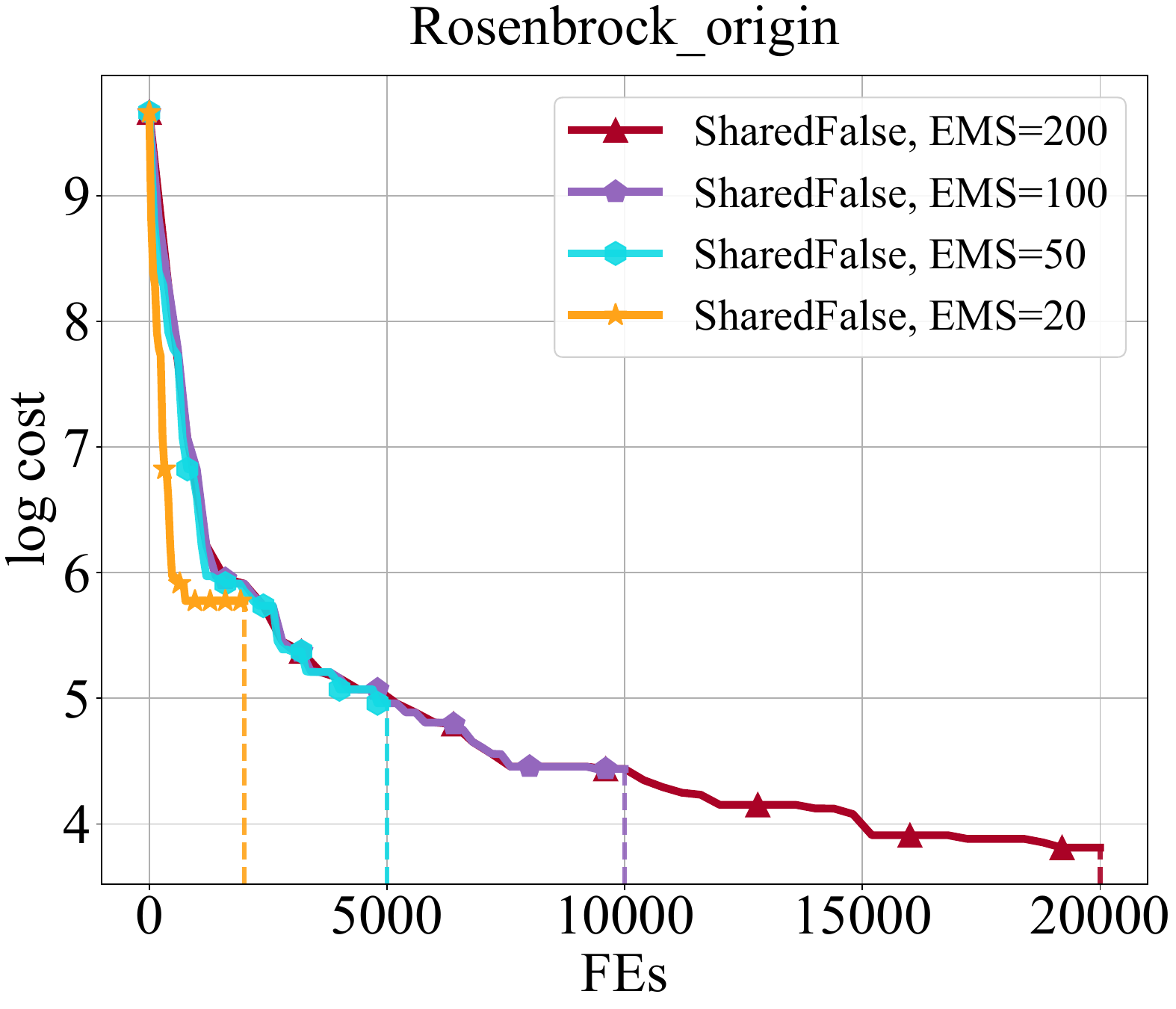}
	\end{subfigure}
	\begin{subfigure}[b]{0.24\textwidth}
		\includegraphics[width=\linewidth,
		trim=0 0 0 0, clip]{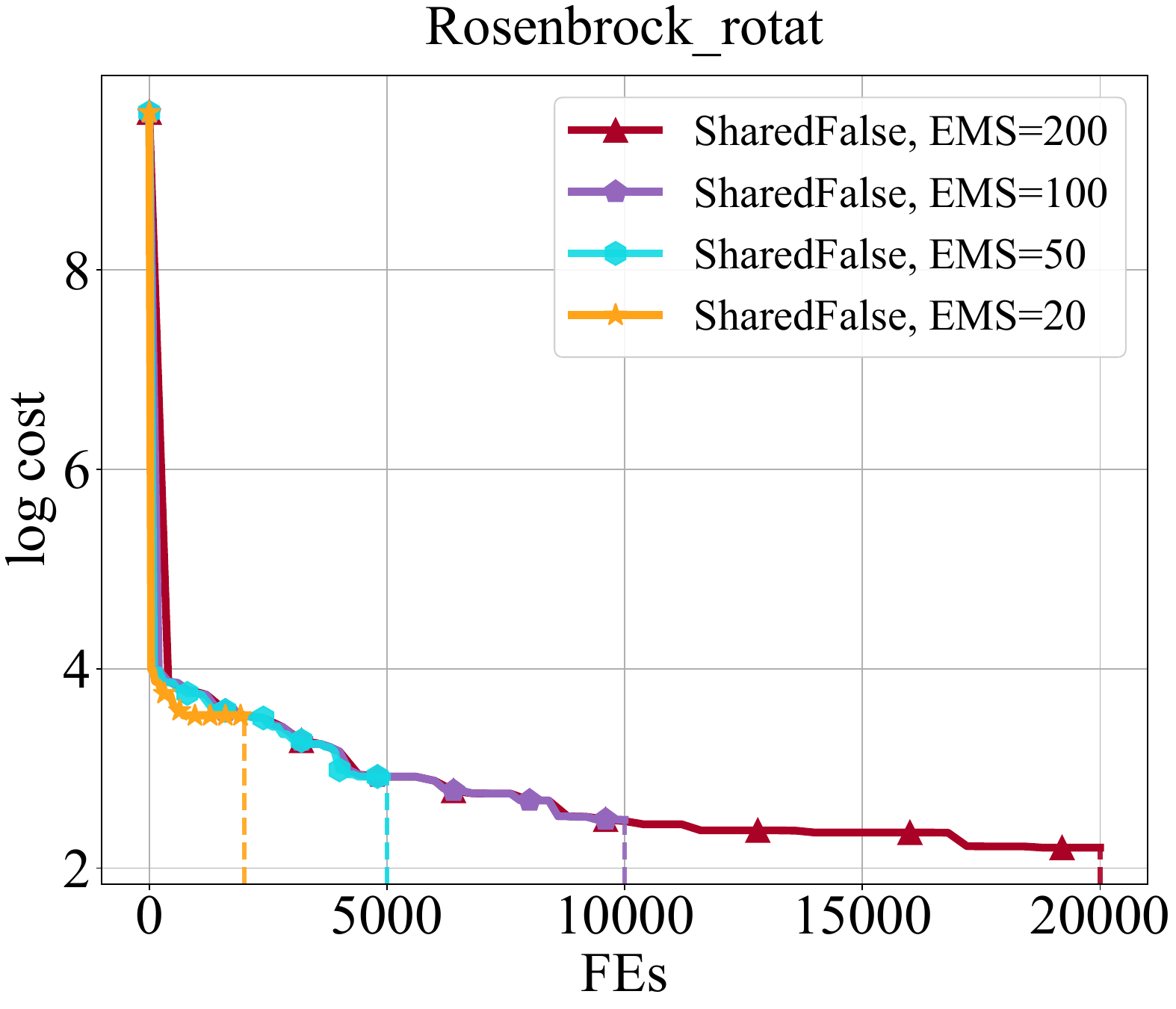}
	\end{subfigure} 
	\begin{subfigure}[b]{0.24\textwidth}
		\includegraphics[width=\linewidth,
		trim=0 0 0 0, clip]{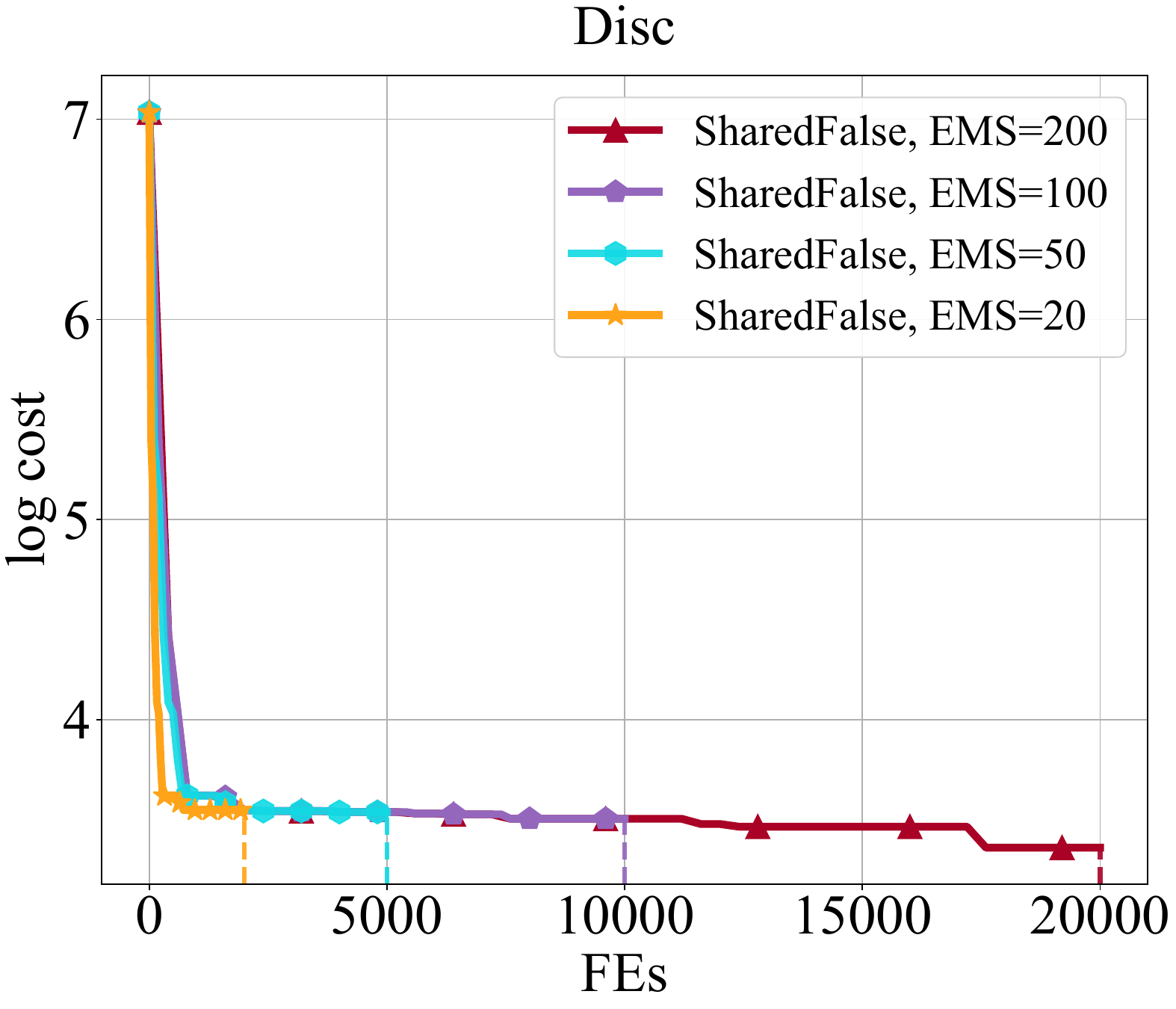}
	\end{subfigure}
	\begin{subfigure}[b]{0.24\textwidth}
		\includegraphics[width=\linewidth,
		trim=0 0 0 0, clip]{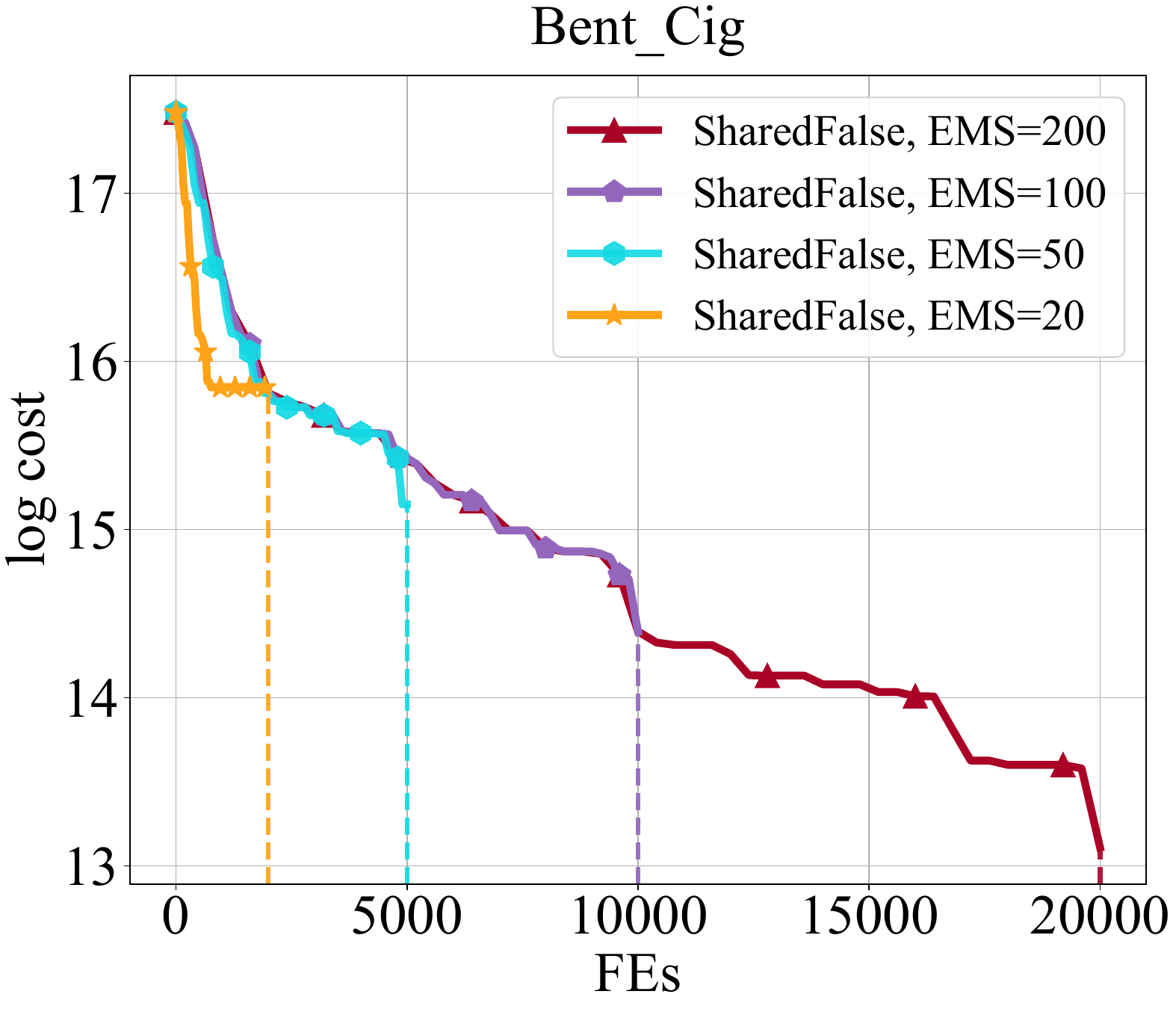}
	\end{subfigure}
	
	\begin{subfigure}[b]{0.24\textwidth}
		\includegraphics[width=\linewidth,
		trim=0 0 0 0, clip]{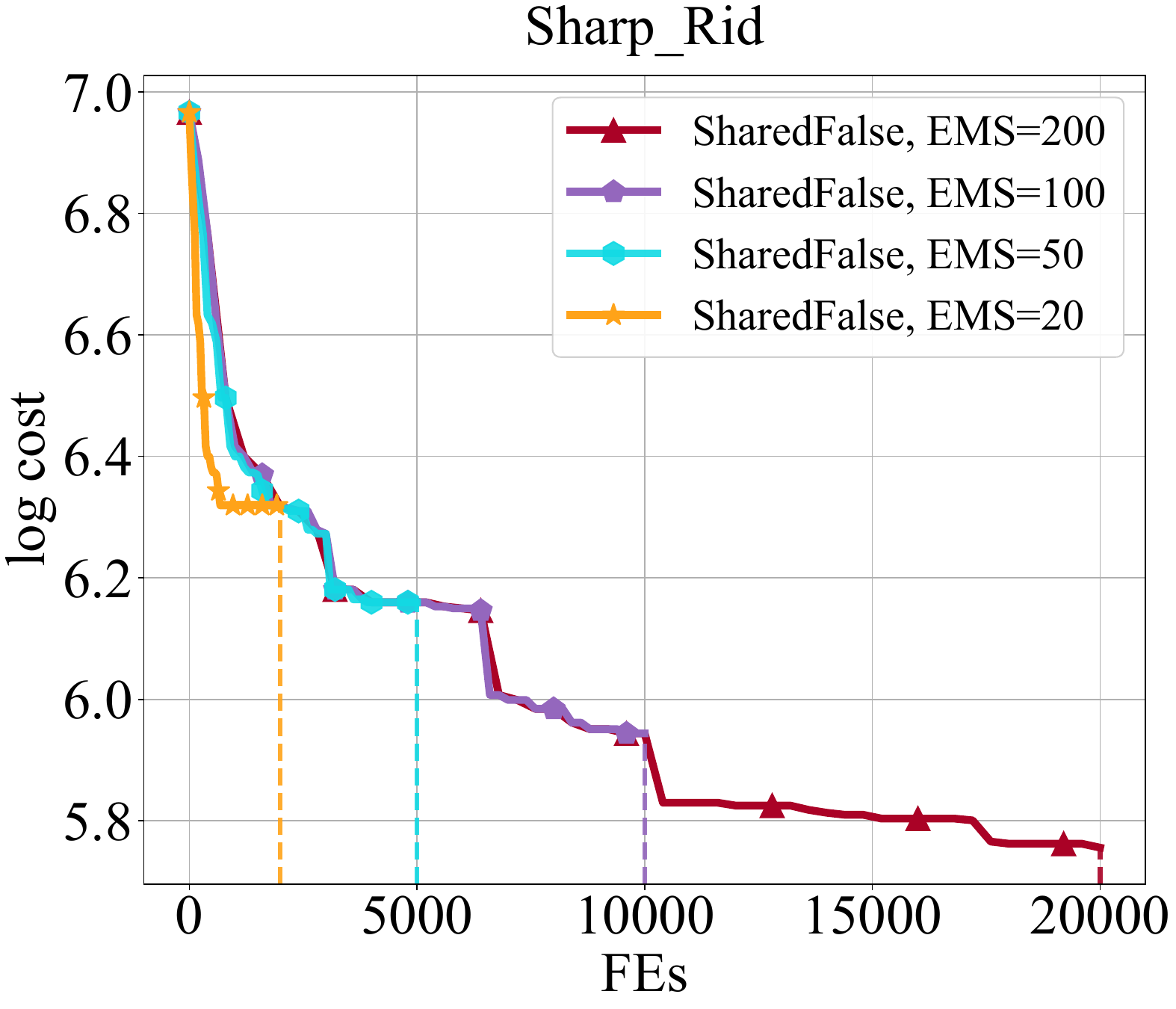}
	\end{subfigure}
	\begin{subfigure}[b]{0.24\textwidth}
		\includegraphics[width=\linewidth,
		trim=0 0 0 0, clip]{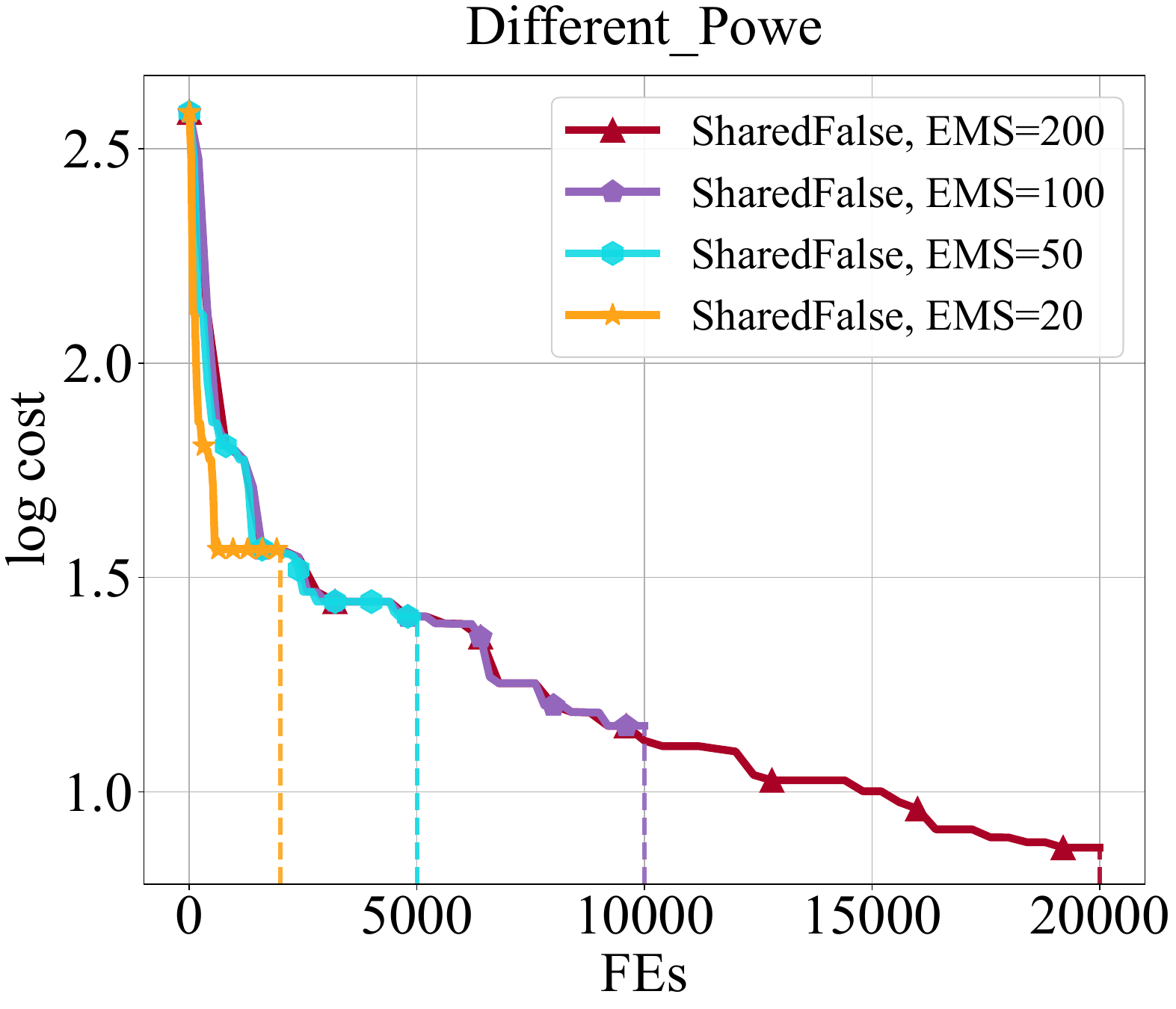}
	\end{subfigure} 
	\begin{subfigure}[b]{0.24\textwidth}
		\includegraphics[width=\linewidth,
		trim=0 0 0 0, clip]{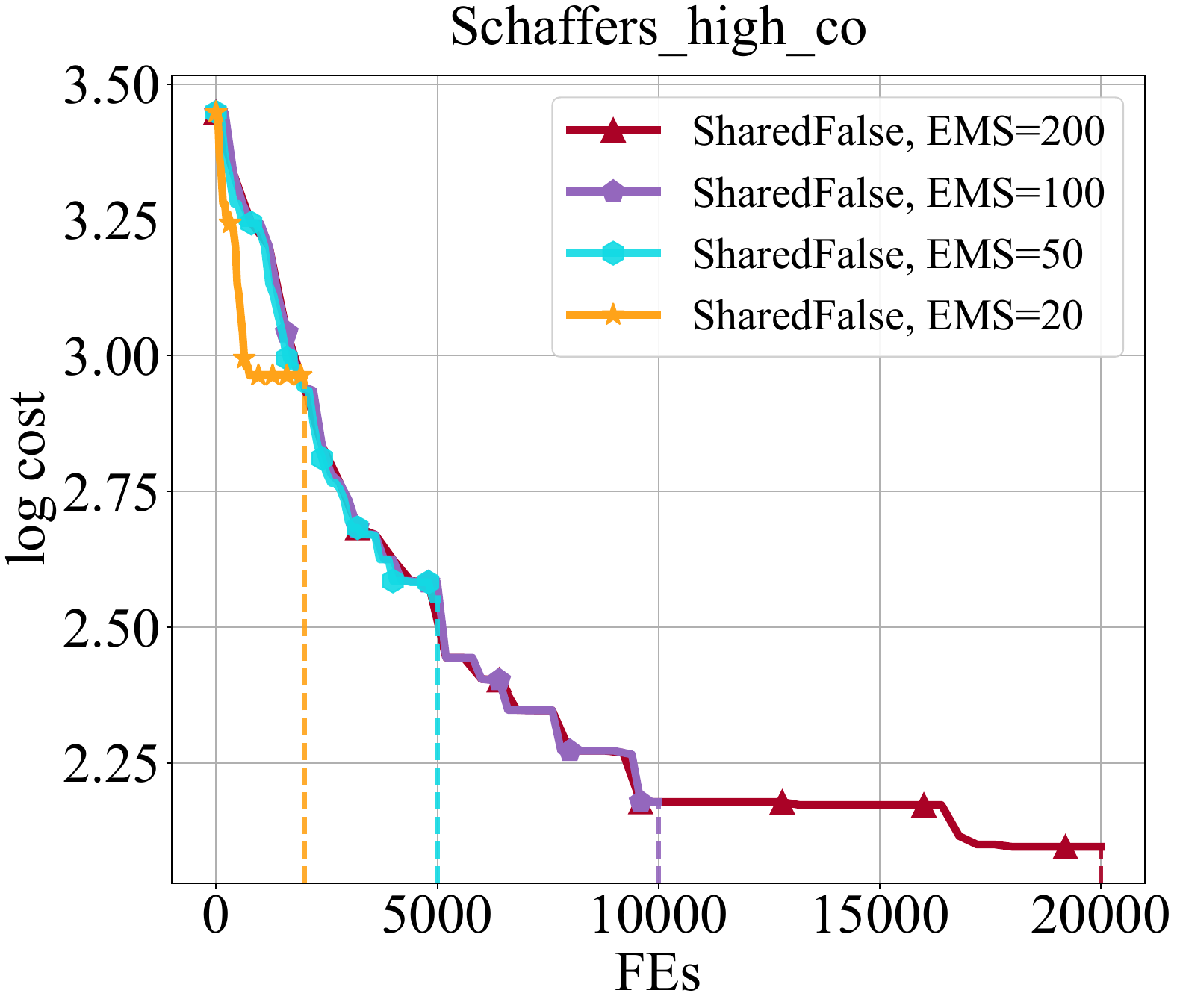}
	\end{subfigure}
	\begin{subfigure}[b]{0.24\textwidth}
		\includegraphics[width=\linewidth,
		trim=0 0 0 0, clip]{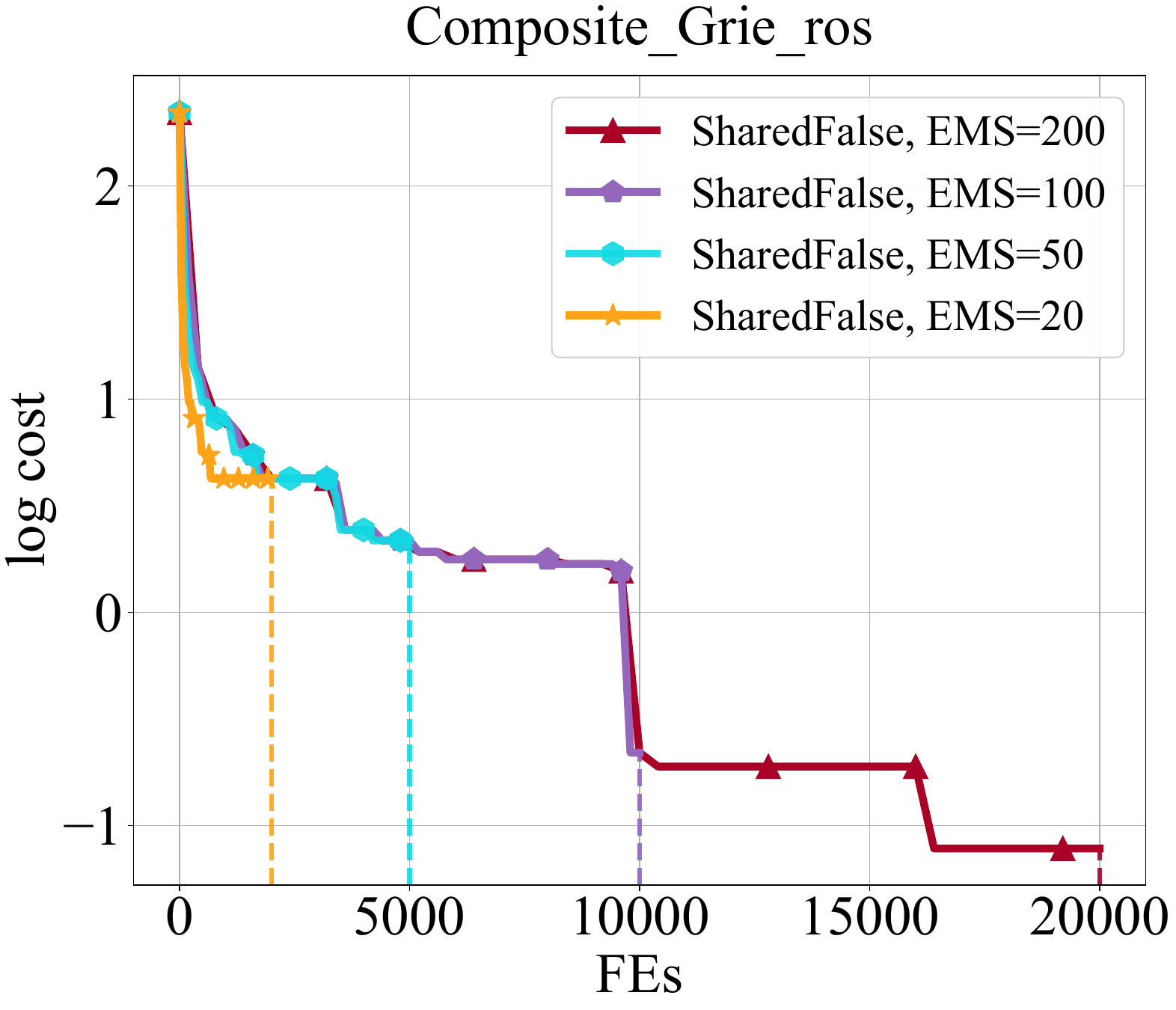}
	\end{subfigure}
	\begin{subfigure}[b]{0.24\textwidth}
		\includegraphics[width=\linewidth,
		trim=0 0 0 0, clip]{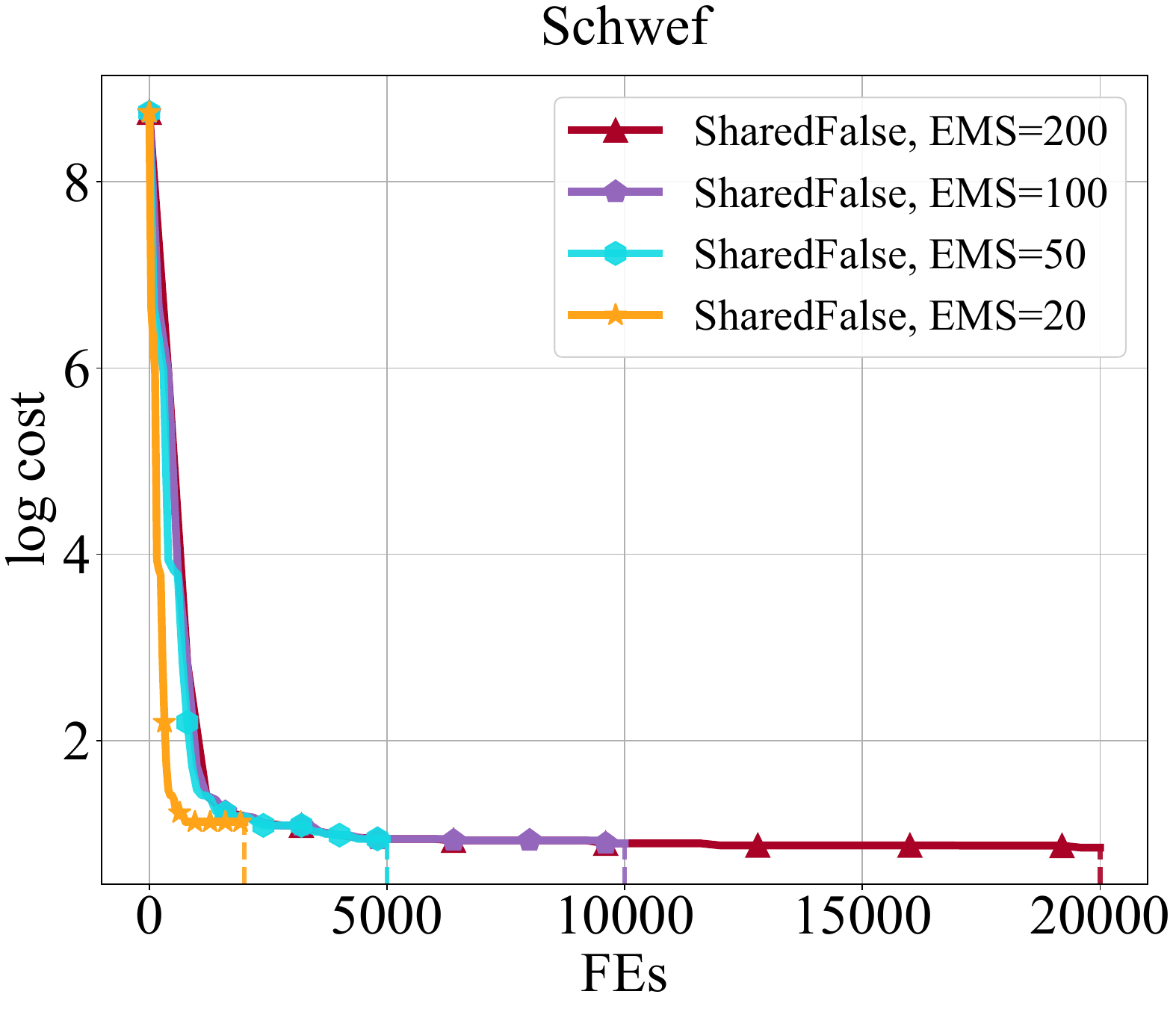}
	\end{subfigure}
	\begin{subfigure}[b]{0.24\textwidth}
		\includegraphics[width=\linewidth,
		trim=0 0 0 0, clip]{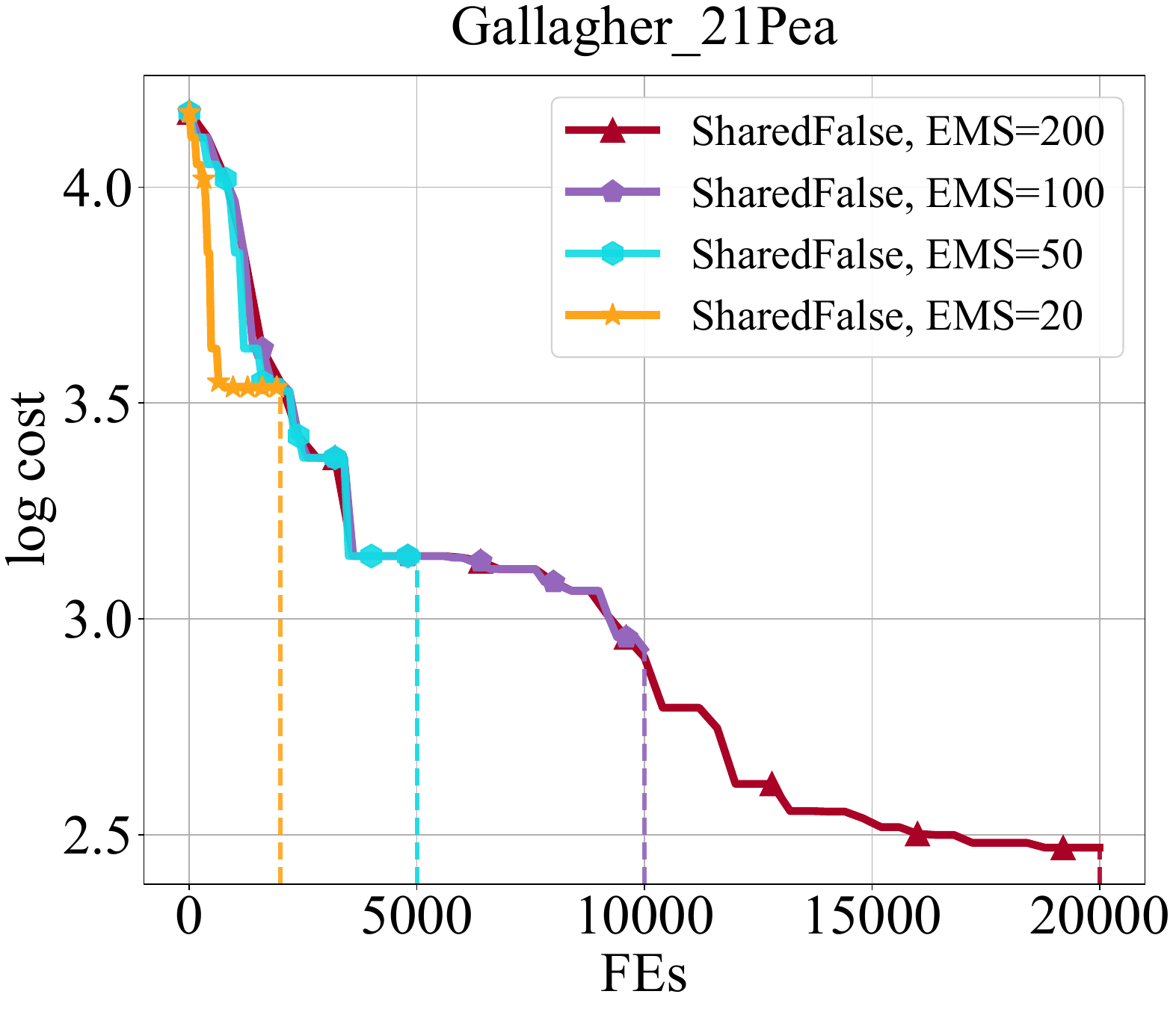}
	\end{subfigure} 
	\begin{subfigure}[b]{0.24\textwidth}
		\includegraphics[width=\linewidth,
		trim=0 0 0 0, clip]{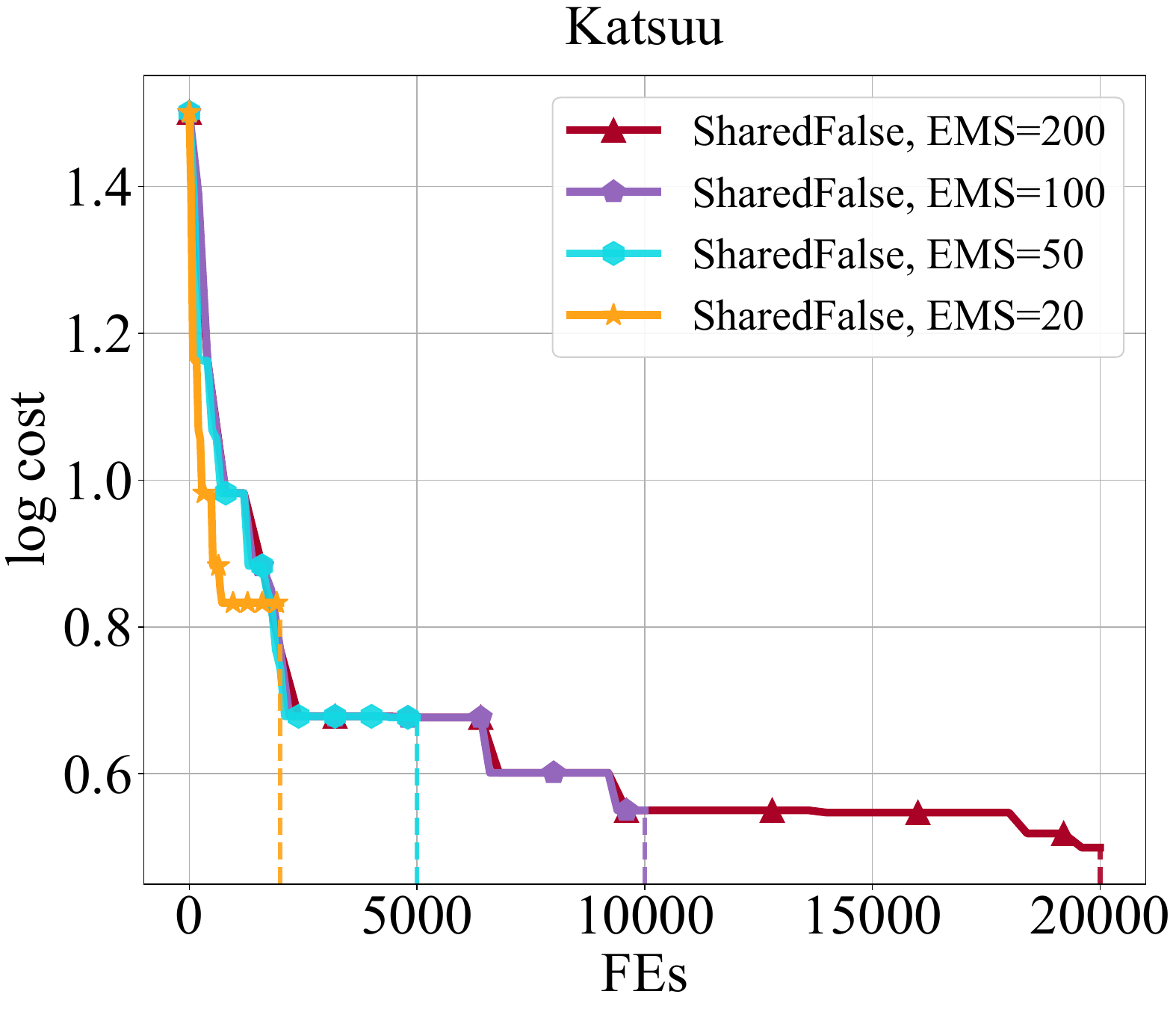}
	\end{subfigure}
	\begin{subfigure}[b]{0.24\textwidth}
		\includegraphics[width=\linewidth,
		trim=0 0 0 0, clip]{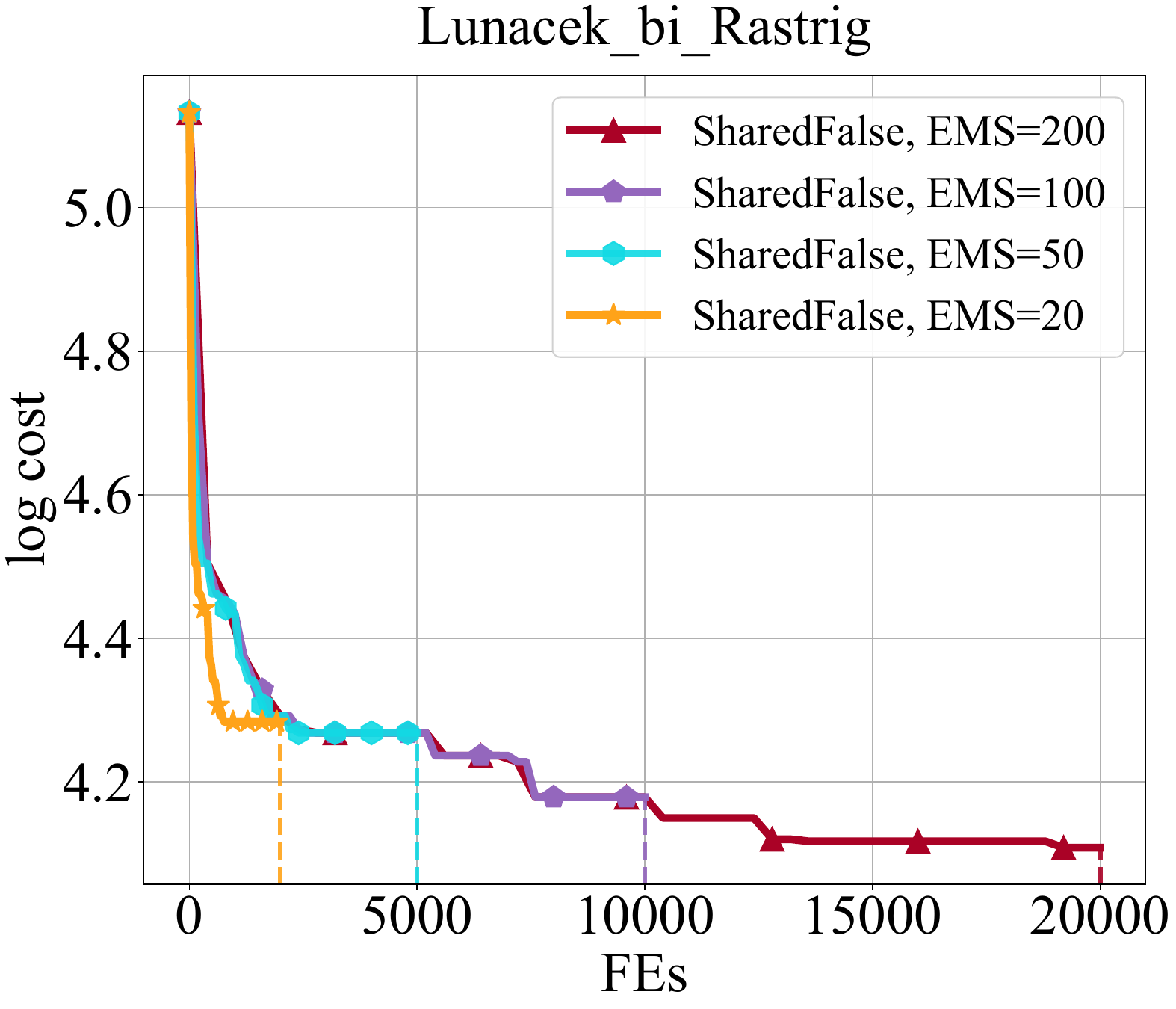}
	\end{subfigure} 
	\caption{
		Convergence behaviors under the \emph{unshared} parameterization of the 
		neural evolution operator~$\mathcal{O}_{\mathrm{evo}}$. 
		Log-cost convergence over all 16 BBOB-like test problems.  
		Vertical dashed lines denote the end of each agent’s evaluation budget.  
	}
	\label{fig:abl_unshared}
\end{figure*}
}

\newcommand{\SupFigbbobTen}{%
	\begin{table*}[!tbp]
		\centering
		\scriptsize
		\setlength{\tabcolsep}{2pt}
		\caption{Numerical performance evaluation (mean $\pm$ std) of representative algorithms on the \emph{BBOB-10D} benchmark. Each entry is calculated from 10 independent test runs. The best and second-best results are highlighted using \underline{\textbf{bold}} and \textbf{bold}, respectively.}
		\label{tab:supp-bbob-10d}
		\begin{tabular}{lcccccccc}
			\toprule
			\textbf{Method} & \texttt{Buche\_Ras} & \texttt{Attractive\_Sec} & \texttt{Step\_Ell} & \texttt{Rosenbrock\_ori} & \texttt{Rosenbrock\_rot} & \texttt{Ellipsoidal\_hc} & \texttt{Discus} & \texttt{Bent\_Cig} \\
			\midrule 
			\multirow{2}{*}{PSO} & 3.429E+02 & 1.109E+03 & 8.667E+01 & 4.003E+03 & 1.798E+03 & 3.288E+05 & 7.338E+01 & 1.768E+07 \\
			& ($\pm$6.591E+01) & ($\pm$7.005E+02) & ($\pm$1.083E+01) & ($\pm$1.471E+03) & ($\pm$7.772E+02) & ($\pm$2.522E+05) & ($\pm$3.266E+01) & ($\pm$1.077E+07) \\
			\multirow{2}{*}{DE} & \textbf{1.012E+02} & 4.934E+01 & 3.099E+01 & 3.527E+02 & 7.201E+02 & 2.859E+04 & 4.860E+01 & 2.070E+06 \\
			& ($\pm$2.988E+01) & ($\pm$6.491E+00) & ($\pm$7.657E+00) & ($\pm$1.685E+02) & ($\pm$4.851E+02) & ($\pm$2.196E+04) & ($\pm$6.461E+00) & ($\pm$5.398E+05) \\
			\multirow{2}{*}{SAHLPSO} & 1.424E+02 & 1.167E+02 & 2.402E+01 & 9.622E+02 & 1.070E+03 & 4.344E+04 & 1.141E+02 & 7.618E+06 \\
			& ($\pm$6.008E+00) & ($\pm$8.459E+01) & ($\pm$1.786E+01) & ($\pm$6.640E+02) & ($\pm$6.981E+02) & ($\pm$2.739E+04) & ($\pm$2.008E+01) & ($\pm$2.039E+06) \\
			\midrule 
			\multirow{2}{*}{RNNOPT} & 6.616E+03 & 6.720E+04 & 7.170E+01 & 1.930E+04 & 6.619E+01 & 3.805E+06 & 1.740E+05 & 6.700E+07 \\
			& ($\pm$0.000E+00) & ($\pm$0.000E+00) & ($\pm$0.000E+00) & ($\pm$0.000E+00) & ($\pm$0.000E+00) & ($\pm$0.000E+00) & ($\pm$0.000E+00) & ($\pm$0.000E+00) \\
			\multirow{2}{*}{DEDQN} & 3.627E+02 & 9.099E+03 & 1.018E+02 & 1.369E+04 & 1.021E+04 & 5.756E+05 & 2.154E+02 & 3.578E+07 \\
			& ($\pm$9.568E+01) & ($\pm$1.020E+04) & ($\pm$2.562E+01) & ($\pm$3.021E+03) & ($\pm$3.073E+03) & ($\pm$1.608E+05) & ($\pm$5.994E+01) & ($\pm$1.490E+07) \\
			\multirow{2}{*}{LES} & 1.473E+03 & 1.816E+03 & 3.076E+02 & 2.697E+03 & 1.334E+03 & 3.207E+05 & 9.363E+02 & 1.308E+07 \\
			& ($\pm$4.452E+01) & ($\pm$1.624E+03) & ($\pm$7.422E-01) & ($\pm$2.109E+02) & ($\pm$4.173E+00) & ($\pm$6.588E+04) & ($\pm$8.504E-01) & ($\pm$1.848E+06) \\
			\multirow{2}{*}{LGA} & 2.462E+02 & 4.033E+02 & 1.794E+02 & 5.188E+03 & 3.062E+04 & 2.578E+05 & 2.265E+02 & 5.390E+07 \\
			& ($\pm$5.814E+01) & ($\pm$0.000E+00) & ($\pm$9.467E+00) & ($\pm$2.615E+03) & ($\pm$1.402E+04) & ($\pm$8.073E+04) & ($\pm$7.462E+01) & ($\pm$1.930E+07) \\
			\multirow{2}{*}{GLHF} & 4.513E+02 & 1.689E+04 & 9.959E+01 & 1.324E+04 & 1.156E+04 & 4.462E+05 & 7.396E+02 & 3.449E+07 \\
			& ($\pm$1.747E+02) & ($\pm$1.254E+04) & ($\pm$1.196E+01) & ($\pm$3.514E+03) & ($\pm$2.001E+03) & ($\pm$1.668E+05) & ($\pm$5.473E+02) & ($\pm$3.621E+06) \\
			\multirow{2}{*}{B2OPT} & 2.734E+02 & 1.116E+02 & 1.061E+01 & 3.067E+02 & \textbf{2.422E+01} & 3.050E+04 & 4.083E+01 & 7.585E+06 \\
			& ($\pm$4.075E+01) & ($\pm$4.417E+00) & ($\pm$3.472E+00) & ($\pm$2.971E+02) & ($\pm$3.558E+00) & ($\pm$1.391E+04) & ($\pm$3.942E+00) & ($\pm$1.256E+06) \\
			\midrule
			\rowcolor{blue!2} \multirow{1}{*}{Ours} & 1.079E+02 & \textbf{4.523E+01} & \textbf{5.048E+00} & \textbf{5.148E+01} & \underline{\textbf{8.213E+00}} & \textbf{1.798E+04} & \underline{\textbf{3.001E+01}} & \textbf{4.299E+05} \\
			\rowcolor{blue!2} & ($\pm$1.640E+01) & ($\pm$1.305E+01) & ($\pm$8.704E$-$01) & ($\pm$1.390E+01) & ($\pm$1.041E+00) & ($\pm$4.549E+03) & ($\pm$3.792E+00) & ($\pm$9.796E+04) \\
			\rowcolor{blue!5} \multirow{1}{*}{Ours$^{\ddag}$} & \underline{\textbf{8.209E+01}} & \underline{\textbf{5.386E+00}} & \underline{\textbf{3.714E+00}} & \underline{\textbf{3.111E+01}} & 2.764E+01 & \underline{\textbf{5.438E+03}} & \textbf{3.869E+01} & \underline{\textbf{1.774E+05}} \\
			\rowcolor{blue!5} & ($\pm$1.394E+01) & ($\pm$1.164E+00) & ($\pm$1.130E+00) & ($\pm$5.541E+00) & ($\pm$3.051E+00) & ($\pm$8.623E+02) & ($\pm$8.844E+00) & ($\pm$2.297E+04) \\
					\midrule
					\textbf{Method} & \texttt{Sharp\_Rid} & \texttt{Different\_Pow} & \texttt{Schaffers\_hig} & \texttt{Composite\_Gri} & \texttt{Schwefel} & \texttt{Gallagher\_21P} & \texttt{Katsuura} & \texttt{Lunacek\_bi\_Ras} \\
					\midrule 
					\multirow{2}{*}{PSO} & 9.431E+02 & 7.173E+00 & 2.411E+01 & 4.266E+00 & 3.235E+02 & 6.513E+01 & \textbf{1.209E+00} & 1.166E+02 \\
					& ($\pm$1.627E+02) & ($\pm$3.197E+00) & ($\pm$9.562E-01) & ($\pm$6.784E-01) & ($\pm$3.334E+02) & ($\pm$5.339E+00) & ($\pm$2.412E-01) & ($\pm$7.371E+00) \\
					\multirow{2}{*}{DE} & \textbf{2.583E+02} & 2.088E+00 & 1.081E+01 & 4.509E+00 & 2.413E+00 & 1.420E+01 & 2.733E+00 & 8.125E+01 \\
					& ($\pm$2.861E+01) & ($\pm$1.005E+00) & ($\pm$8.182E-01) & ($\pm$2.996E-01) & ($\pm$9.219E-02) & ($\pm$1.058E+01) & ($\pm$7.799E-01) & ($\pm$3.929E+00) \\
					\multirow{2}{*}{SAHLPSO} & 3.679E+02 & \textbf{1.316E+00} & 1.229E+01 & 5.759E+00 & 3.942E+01 & 1.675E+01 & 1.577E+00 & 9.341E+01 \\
					& ($\pm$2.165E+01) & ($\pm$1.267E+00) & ($\pm$5.473E+00) & ($\pm$5.528E-01) & ($\pm$3.132E+01) & ($\pm$5.020E+00) & ($\pm$5.575E-01) & ($\pm$2.748E+00) \\
					\midrule 
					\multirow{2}{*}{RNNOPT} & 1.798E+03 & 1.069E+01 & 5.223E+01 & 2.056E+00 & 8.060E+03 & 8.486E+01 & 3.440E+00 & 1.221E+02 \\
					& ($\pm$0.000E+00) & ($\pm$0.000E+00) & ($\pm$0.000E+00) & ($\pm$0.000E+00) & ($\pm$0.000E+00) & ($\pm$0.000E+00) & ($\pm$0.000E+00) & ($\pm$0.000E+00) \\
					\multirow{2}{*}{DEDQN} & 1.057E+03 & 1.047E+01 & 3.073E+01 & 1.127E+01 & 4.618E+03 & 5.834E+01 & 3.613E+00 & 1.480E+02 \\
					& ($\pm$1.827E+02) & ($\pm$3.672E+00) & ($\pm$4.258E+00) & ($\pm$1.689E+00) & ($\pm$8.773E+02) & ($\pm$9.502E+00) & ($\pm$5.164E-01) & ($\pm$2.998E+01) \\
					\multirow{2}{*}{LES} & 3.274E+03 & 1.505E+03 & 7.211E+02 & 1.700E+03 & 2.104E+03 & 2.454E+03 & 6.017E+02 & 1.270E+03 \\
					& ($\pm$6.658E+01) & ($\pm$3.239E-01) & ($\pm$6.131E-01) & ($\pm$3.196E-02) & ($\pm$2.379E-01) & ($\pm$1.005E+01) & ($\pm$1.436E-01) & ($\pm$2.917E+00) \\
					\multirow{2}{*}{LGA} & 9.683E+02 & 1.663E+01 & 2.613E+01 & 1.322E+01 & 1.234E+04 & 4.875E+01 & 1.618E+00 & 1.778E+02 \\
					& ($\pm$2.034E+02) & ($\pm$5.850E+00) & ($\pm$4.232E+00) & ($\pm$5.755E+00) & ($\pm$3.810E+03) & ($\pm$1.666E+01) & ($\pm$0.000E+00) & ($\pm$1.977E+01) \\
					\multirow{2}{*}{GLHF} & 1.111E+03 & 1.530E+01 & 2.792E+01 & 9.223E+00 & 4.686E+03 & 4.876E+01 & 3.299E+00 & 1.499E+02 \\
					& ($\pm$2.052E+02) & ($\pm$8.705E-01) & ($\pm$3.296E+00) & ($\pm$1.221E+00) & ($\pm$1.331E+03) & ($\pm$1.684E+01) & ($\pm$4.095E-01) & ($\pm$2.572E+01) \\
					\multirow{2}{*}{B2OPT} & 4.721E+02 & 4.665E+00 & 1.176E+01 & \textbf{1.633E+00} & 2.877E+00 & 2.987E+01 & 2.293E+00 & 7.342E+01 \\
					& ($\pm$6.671E+01) & ($\pm$1.035E+00) & ($\pm$2.675E+00) & ($\pm$1.135E-01) & ($\pm$3.331E-01) & ($\pm$1.920E+01) & ($\pm$4.472E-02) & ($\pm$5.952E+00) \\
					\midrule
					\rowcolor{blue!2} \multirow{1}{*}{Ours} & 2.925E+02 & 2.580E+00 & \textbf{7.677E+00} & \underline{\textbf{3.739E$-$01}} & \textbf{2.234E+00} & \textbf{1.352E+01} & 1.663E+00 & \underline{\textbf{5.758E+01}} \\
					\rowcolor{blue!2} & ($\pm$5.516E+01) & ($\pm$7.181E$-$01) & ($\pm$7.121E$-$01) & ($\pm$5.895E$-$02) & ($\pm$8.563E$-$02) & ($\pm$7.355E+00) & ($\pm$1.946E$-$01) & ($\pm$2.787E+00) \\
					\rowcolor{blue!5} \multirow{1}{*}{Ours$^{\ddag}$} & \underline{\textbf{8.569E+01}} & \underline{\textbf{2.277E$-$01}} & \underline{\textbf{7.583E+00}} & 3.910E+00 & \underline{\textbf{2.058E+00}} & \underline{\textbf{6.827E$-$01}} & 1.896E+00 & 8.373E+01 \\
					\rowcolor{blue!5} & ($\pm$2.778E+00) & ($\pm$2.264E$-$02) & ($\pm$4.008E+00) & ($\pm$6.519E$-$01) & ($\pm$9.424E$-$02) & ($\pm$3.104E$-$01) & ($\pm$3.804E$-$02) & ($\pm$1.637E+01) \\
			\bottomrule
		\end{tabular}
		\vspace{0.3em}
		\begin{minipage}{\textwidth}
			\scriptsize
			\raggedright
			~~~~$^{\ddag}$ indicates ours variant enhanced with the proposed Mamba-based neural operator.
		\end{minipage}
	\end{table*}
}

\newcommand{\SupTabbbobTen}{%
	\begin{table*}[!tbp]
		\centering
		\scriptsize
		\setlength{\tabcolsep}{2pt}
		\caption{Numerical performance evaluation (mean $\pm$ std) of representative algorithms on the \emph{BBOB-30D} benchmark.. Each entry is calculated from 10 independent test runs. The best and second-best results are highlighted using \underline{\textbf{bold}} and \textbf{bold}, respectively.}
		\label{tab:supp-bbob-30d}
		\begin{tabular}{lcccccccc}
			\toprule
			\textbf{Method} & \texttt{Buche\_Ras} & \texttt{Attractive\_Sec} & \texttt{Step\_Ell} & \texttt{Rosenbrock\_ori} & \texttt{Rosenbrock\_rot} & \texttt{Ellipsoidal\_hc} & \texttt{Discus} & \texttt{Bent\_Cig} \\
			\midrule 
			\multirow{2}{*}{PSO} & 2.169E+03 & 1.224E+05 & 7.002E+02 & 1.109E+04 & 5.884E+02 & 1.110E+06 & \textbf{1.633E+02} & 1.159E+08 \\
			& ($\pm$5.998E+02) & ($\pm$5.208E+04) & ($\pm$4.272E+01) & ($\pm$2.664E+03) & ($\pm$2.849E+02) & ($\pm$2.838E+05) & ($\pm$2.059E+01) & ($\pm$1.495E+07) \\
			\multirow{2}{*}{DE} &  {\textbf{4.996E+02}} & \textbf{8.765E+02} & \textbf{1.224E+02} & 2.471E+03 & 2.814E+03 & \textbf{3.085E+05} & 1.820E+02 & 2.201E+07 \\
			& ($\pm$1.715E+01) & ($\pm$7.413E+02) & ($\pm$3.713E+01) & ($\pm$7.355E+02) & ($\pm$5.186E+02) & ($\pm$1.613E+05) & ($\pm$1.608E+01) & ($\pm$3.474E+06) \\
			\multirow{2}{*}{SAHLPSO} & 9.574E+02 & 5.158E+04 & 4.115E+02 & 2.687E+04 & 9.397E+03 & 1.296E+06 & 2.760E+02 & 1.124E+08 \\
			& ($\pm$6.404E+01) & ($\pm$4.482E+04) & ($\pm$1.473E+02) & ($\pm$1.173E+04) & ($\pm$2.132E+03) & ($\pm$5.424E+05) & ($\pm$7.851E+01) & ($\pm$1.062E+08) \\
			\midrule 
			\multirow{2}{*}{RNNOPT} & 1.603E+04 & 5.768E+05 & 1.511E+03 & 9.192E+04 & 1.975E+02 & 1.431E+07 & 4.950E+06 & 4.499E+08 \\
			& ($\pm$0.000E+00) & ($\pm$0.000E+00) & ($\pm$0.000E+00) & ($\pm$1.455E-11) & ($\pm$0.000E+00) & ($\pm$0.000E+00) & ($\pm$0.000E+00) & ($\pm$0.000E+00) \\
			\multirow{2}{*}{DEDQN} & 2.061E+03 & 2.325E+05 & 9.007E+02 & 1.712E+05 & 1.518E+05 & 3.772E+06 & 5.374E+02 & 2.796E+08 \\
			& ($\pm$3.744E+02) & ($\pm$4.669E+04) & ($\pm$4.917E+01) & ($\pm$1.802E+04) & ($\pm$9.146E+03) & ($\pm$4.637E+05) & ($\pm$2.222E+02) & ($\pm$3.171E+07) \\
			\multirow{2}{*}{LES} & 5.822E+03 & 3.620E+05 & 3.044E+03 & 5.335E+04 & 1.485E+03 & 3.974E+06 & 1.953E+03 & 2.468E+08 \\
			& ($\pm$3.715E+02) & ($\pm$2.230E+04) & ($\pm$1.385E+02) & ($\pm$9.407E+02) & ($\pm$5.223E-01) & ($\pm$2.115E+05) & ($\pm$3.210E+01) & ($\pm$1.006E+07) \\
			\multirow{2}{*}{LGA} & 1.082E+04 & 6.709E+05 & 1.266E+03 & 2.607E+05 & 2.086E+05 & 6.214E+06 & 1.187E+05 & 4.517E+08 \\
			& ($\pm$4.568E+03) & ($\pm$2.485E+05) & ($\pm$1.630E+02) & ($\pm$6.975E+04) & ($\pm$6.213E+04) & ($\pm$2.719E+06) & ($\pm$1.669E+05) & ($\pm$9.124E+07) \\
			\multirow{2}{*}{GLHF} & 4.510E+03 & 3.470E+05 & 1.275E+03 & 3.373E+05 & 1.993E+05 & 7.354E+06 & 1.892E+03 & 5.111E+08 \\
			& ($\pm$5.042E+02) & ($\pm$2.271E+05) & ($\pm$3.484E+02) & ($\pm$2.066E+04) & ($\pm$2.834E+04) & ($\pm$1.035E+06) & ($\pm$1.603E+03) & ($\pm$6.690E+07) \\
			\multirow{2}{*}{B2OPT} & 1.173E+03 & 1.243E+05 & 4.753E+02 & 2.698E+04 & 1.863E+02 & 1.270E+06 & 1.809E+02 & 1.656E+08 \\
			& ($\pm$1.705E+02) & ($\pm$4.855E+04) & ($\pm$5.596E+01) & ($\pm$1.223E+04) & ($\pm$9.315E-01) & ($\pm$3.203E+05) & ($\pm$1.733E+01) & ($\pm$1.558E+07) \\
			\midrule
			\rowcolor{blue!2} \multirow{1}{*}{Ours} &\underline{\textbf{4.914E+02}} & 2.497E+03 & 1.342E+02 & \textbf{1.753E+03} & \underline{\textbf{1.638E+02}} & 6.490E+05 & \underline{\textbf{1.518E+02}} & \textbf{2.096E+07} \\
			\rowcolor{blue!2} & ($\pm$3.327E+01) & ($\pm$2.761E+03) & ($\pm$1.739E+01) & ($\pm$2.956E+02) & ($\pm$9.913E-01) & ($\pm$1.787E+05) & ($\pm$2.712E+01) & ($\pm$7.699E+05) \\
			\rowcolor{blue!5} \multirow{1}{*}{Ours$^{\ddag}$} & 5.452E+02 & \underline{\textbf{1.585E+02}} & \underline{\textbf{6.737E+01}} & \underline{\textbf{4.076E+02}} & 2.378E+02 & \underline{\textbf{9.960E+04}} & 1.861E+02 & \underline{\textbf{2.684E+06}} \\
			\rowcolor{blue!5} & ($\pm$1.656E+01) & ($\pm$3.687E+01) & ($\pm$4.068E+00) & ($\pm$1.098E+02) & ($\pm$1.392E+01) & ($\pm$1.120E+04) & ($\pm$3.168E+01) & ($\pm$5.046E+05) \\
					\midrule 
					\textbf{Method} & \texttt{Sharp\_Rid} & \texttt{Different\_Pow} & \texttt{Schaffers\_hig} & \texttt{Composite\_Gri} & \texttt{Schwefel} & \texttt{Gallagher\_21P} & \texttt{Katsuura} & \texttt{Lunacek\_bi\_Ras} \\
					\midrule 
					\multirow{2}{*}{PSO} & 1.360E+03 & 1.928E+01 & 2.751E+01 &  {5.665E+00} & 2.016E+01 & 6.659E+01 & 3.208E+00 & 3.461E+02 \\
					& ($\pm$1.372E+02) & ($\pm$3.804E+00) & ($\pm$4.309E+00) & ($\pm$2.647E-01) & ($\pm$1.457E+01) & ($\pm$2.572E+00) & ($\pm$2.201E-01) & ($\pm$2.614E+01) \\
					\multirow{2}{*}{DE} & \textbf{7.012E+02} & 8.848E+00 & \underline{\textbf{1.536E+01}} & 6.595E+00 & 2.439E+02 &  {\textbf{1.083E+01}} & 3.000E+00 & 3.043E+02 \\
					& ($\pm$3.727E+01) & ($\pm$7.387E-01) & ($\pm$5.156E+00) & ($\pm$1.962E-01) & ($\pm$1.181E+02) & ($\pm$7.941E-01) & ($\pm$1.823E-01) & ($\pm$3.679E+01) \\
					\multirow{2}{*}{SAHLPSO} & 1.500E+03 & 1.707E+01 & 3.268E+01 & 8.649E+00 & 4.673E+03 & 4.442E+01 & 3.192E+00 & 4.744E+02 \\
					& ($\pm$3.082E+01) & ($\pm$8.338E+00) & ($\pm$4.166E+00) & ($\pm$1.578E+00) & ($\pm$1.368E+03) & ($\pm$8.733E+00) & ($\pm$7.748E-01) & ($\pm$2.208E+01) \\
					\midrule 
					\multirow{2}{*}{RNNOPT} & 2.314E+03 & 7.363E+01 & 7.426E+01 & 6.430E+00 & 2.606E+04 & 8.447E+01 & 7.372E+00 & 4.541E+02 \\
					& ($\pm$0.000E+00) & ($\pm$0.000E+00) & ($\pm$0.000E+00) & ($\pm$0.000E+00) & ($\pm$0.000E+00) & ($\pm$0.000E+00) & ($\pm$0.000E+00) & ($\pm$0.000E+00) \\
					\multirow{2}{*}{DEDQN} & 2.360E+03 & 4.823E+01 & 4.896E+01 & 2.184E+01 & 6.849E+04 & 7.642E+01 & 4.808E+00 & 8.033E+02 \\
					& ($\pm$1.659E+02) & ($\pm$2.662E+00) & ($\pm$5.639E+00) & ($\pm$1.482E+00) & ($\pm$1.161E+04) & ($\pm$3.011E+00) & ($\pm$3.796E-01) & ($\pm$9.052E+01) \\
					\multirow{2}{*}{LES} & 2.895E+03 & 1.310E+02 & 1.341E+03 & 1.300E+03 & 7.007E+03 & 1.180E+03 & 1.303E+03 & 1.749E+03 \\
					& ($\pm$3.334E+01) & ($\pm$1.310E+00) & ($\pm$2.615E+00) & ($\pm$4.597E-04) & ($\pm$1.338E+03) & ($\pm$1.277E+00) & ($\pm$2.324E-01) & ($\pm$1.423E+01) \\
					\multirow{2}{*}{LGA} & 2.821E+03 & 9.115E+01 & 6.749E+01 & 2.159E+01 & 9.618E+04 & 8.395E+01 & 3.231E+00 & 9.121E+02 \\
					& ($\pm$1.272E+02) & ($\pm$2.051E+01) & ($\pm$5.473E+00) & ($\pm$3.009E+00) & ($\pm$1.232E+04) & ($\pm$1.719E+00) & ($\pm$2.440E-01) & ($\pm$1.066E+02) \\
					\multirow{2}{*}{GLHF} & 2.611E+03 & 6.510E+01 & 5.337E+01 & 2.541E+01 & 7.450E+04 & 8.305E+01 & 6.515E+00 & 8.946E+02 \\
					& ($\pm$1.152E+02) & ($\pm$5.242E+00) & ($\pm$5.011E+00) & ($\pm$1.609E+00) & ($\pm$4.114E+03) & ($\pm$1.280E+00) & ($\pm$1.767E-01) & ($\pm$3.887E+01) \\
					\multirow{2}{*}{B2OPT} & 1.393E+03 & 1.921E+01 & 3.485E+01 & \textbf{2.156E+00} & 1.671E+03 & 7.715E+01 & 4.219E+00 & 3.267E+02 \\
					& ($\pm$1.978E+02) & ($\pm$3.729E-01) & ($\pm$1.291E+00) & ($\pm$2.386E-01) & ($\pm$8.537E+02) & ($\pm$3.433E+00) & ($\pm$2.024E-01) & ($\pm$5.097E+00) \\
					\midrule
					\rowcolor{blue!2} \multirow{1}{*}{Ours} & 7.296E+02 & \textbf{6.871E+00} &  {\textbf{2.039E+01}} & \underline{\textbf{2.530E$-$01}} & \textbf{3.372E+00} & 2.292E+01 & \textbf{2.821E+00} & \textbf{3.184E+02} \\
					\rowcolor{blue!2} & ($\pm$4.779E+01) & ($\pm$4.774E$-$01) & ($\pm$1.481E+00) & ($\pm$4.139E$-$03) & ($\pm$8.584E$-$02) & ($\pm$1.073E+00) & ($\pm$2.661E$-$01) & ($\pm$7.627E+00) \\
					\rowcolor{blue!5} \multirow{1}{*}{Ours$^{\ddag}$} & \underline{\textbf{2.553E+02}} & \underline{\textbf{7.204E$-$01}} & 3.633E+01 & 1.133E+01 & \underline{\textbf{3.002E+00}} & \underline{\textbf{1.765E+00}} & 2.989E+00 & 6.525E+02 \\
					\rowcolor{blue!5} & ($\pm$3.280E+01) & ($\pm$1.349E$-$01) & ($\pm$6.386E+00) & ($\pm$4.398E$-$01) & ($\pm$7.050E$-$02) & ($\pm$4.353E$-$02) & ($\pm$6.860E$-$02) & ($\pm$2.558E+01) \\
			\bottomrule
		\end{tabular}
		\vspace{0.3em}
		\begin{minipage}{\textwidth}
			\scriptsize
			\raggedright
			~~~~$^{\ddag}$ indicates ours variant enhanced with the proposed Mamba-based neural operator.
		\end{minipage}
	\end{table*}
}

\begin{document}

\title{Learning to Evolve for Optimization via Stability-Inducing Neural Unrolling} 

\author{Jiaxin Gao, Yaohua Liu, Ran Cheng,~\IEEEmembership{Senior Member, IEEE,} and Kay Chen Tan,~\IEEEmembership{Fellow, IEEE} 
	\thanks{J. Gao, R. Cheng, and K. C. Tan are with the Department of Data Science and Artificial Intelligence, The Hong Kong Polytechnic University, Hong Kong. R. Cheng and K. C. Tan are also with the Department of Computing. (e-mail: jiaxinn.gao@outlook.com; ranchengcn@gmail.com; kaychen.tan@polyu.edu.hk). Y. Liu is with the School of Computing and Data Science, The University of Hong Kong, Hong Kong (e-mail: liuyaohua.918@gmail.com).} 
}
\maketitle

\begin{abstract}
    Evolutionary algorithms serve as a powerful paradigm for tackling optimization challenges, yet their reliance on manually engineered heuristics inherently limits their adaptability across diverse landscapes. 
    However, the transition from the hand-crafted heuristics to data-driven algorithms faces a fundamental dilemma: achieving neural \emph{plasticity} without sacrificing algorithmic \emph{stability}. Although learned optimizers offer high adaptivity, their unconstrained update rules often result in unstable dynamics and brittle generalization on unseen landscapes. To address this challenge, this paper proposes \textit{Learning to Evolve (L2E)}, a bilevel meta-optimization framework that learns evolutionary search via \textit{stability-inducing neural unrolling}. First, L2E reformulates population evolution as an unrolled fixed-point iteration via a structured neural operator. In this design, the inner loop imposes a stability-biased update structure, while the outer loop meta-trains the operator to produce effective search trajectories across tasks. Second, to balance global exploration with local refinement, a gradient-derived composite solver adaptively fuses learned evolutionary proposals with proxy numerical guidance in a differentiable manner. Extensive experiments on synthetic benchmarks and real-world control tasks demonstrate that L2E achieves substantial optimization performance, scales to high-dimensional problems, and exhibits robust zero-shot transfer across diverse test distributions.
\end{abstract}


\section{Introduction}
\IEEEPARstart{E}{volutionary} Algorithms (EAs) are a core paradigm in computational intelligence for solving complex black-box optimization problems~\cite{yang2021gradient,majid2023deep,xue2023solution,de2025differential,cao2024global}. 
By evolving a population through selection, recombination, and perturbation, EAs can explore highly nonconvex, multimodal, and noisy landscapes without relying on explicit gradient information. 
As a result, they underpin a wide range of applications, including hyperparameter tuning~\cite{feurer2019hyperparameter}, aerodynamic shape design~\cite{jones1998efficient}, molecular conformation prediction~\cite{riniker2015better}, and the control of physical systems~\cite{fleming2002evolutionary}. 
In these scenarios, solvers must operate under tight evaluation budgets while navigating high-dimensional and noisy search spaces, requiring both effective exploration and reliable progress toward high-quality solutions~\cite{eiben2002parameter}.

Over the past decades, black-box optimization has been dominated by two major families: population-based evolutionary search and model-based Bayesian optimization (BO).
The former emphasizes exploration through handcrafted operators and heuristics, whereas the latter focuses on sample efficiency via surrogate modeling in expensive evaluation settings~\cite{espinosa2023surrogate}.
Despite their empirical success, a common bottleneck is that the update dynamics are largely driven by manually designed heuristics.
Consequently, their behavior is difficult to characterize and control in a unified way across varying scales, conditioning, and dimensionalities.
This issue becomes particularly pronounced in high-dimensional regimes and when the optimizer is deployed on unseen tasks, where stable progress under limited evaluations is challenging. 

A natural response is the transition from manually designing operators to {data-driven} and {learnable} evolutionary optimizers~\cite{wu2022ensemble,yu2024ngde,xue2024gradient,wu2024evae}.
Recent studies in learning to optimize (or meta-evolution) learn mutation-crossover-selection policies from a task distribution.
Specifically, these methods employ neural controllers (e.g., RNNs, Transformers) or set-based architectures to model population updates~\cite{chen2025expensive,andrychowicz2016learning,ma2023metabox,ma2025toward,maneural}.
Representative examples include learning-driven operator discovery and attention-based evolutionary strategies, such as LQD~\cite{bahlous2025dominated} and the evolution transformer~\cite{li2025b2opt}.
However, a critical gap remains: learned optimizers often struggle to reconcile \emph{plasticity} with \emph{stability} in a way that transfers reliably across tasks.
Most existing approaches learn update rules as unconstrained policies that minimize trajectory loss on training problems.
While such policies can be highly effective in-distribution, they may induce \emph{unstable dynamics} on unseen landscapes, such as oscillation, premature stagnation, or sensitivity to nuisance shifts in scale and conditioning.
In other words, the learned update rule may implicitly overfit the training landscape topology, i.e., \emph{``memorizing the map''} rather than \emph{``learning the compass''}.
This tendency results in brittle generalization under distribution shifts.

This observation motivates a different design principle: rather than learn an unconstrained update policy, the optimizer should learn an {iterative dynamical system} with an explicit {stability-inducing bias}.
In response, we propose \textbf{Learning to Evolve (L2E)}, a bilevel meta-optimization framework that learns evolutionary search through {stability-inducing neural unrolling}.
Concretely, L2E models population evolution as an unrolled iterative solver~\cite{Gregor2010LISTA,Monga2021UnrollingSurvey}.
The inner loop follows a stability-biased update structure (e.g., averaged/residual updates with bounded transformations), while the outer loop meta-trains the operator to produce effective search trajectories.
To balance global exploration with local refinement, L2E further introduces a gradient-derived composite solver.
This solver adaptively fuses learned evolutionary proposals with proxy numerical guidance in a differentiable manner, enabling controlled progress under limited evaluations.
From an architectural perspective, L2E parameterizes the learned evolutionary operator with a structured Mamba-based design to model population-wise interactions.
By combining state-space modeling with lightweight routing over population statistics, the operator provides a scalable inductive bias for meta-learned evolutionary search.
This stability-inducing formulation yields the behavioral contrast illustrated in Fig.~\ref{fig:fig0}. Specifically, while unconstrained learned heuristics may overfit training landscapes and exhibit unstable trajectories on unseen problems, L2E promotes controlled search dynamics by learning structured evolutionary updates via neural unrolling. 
This framework explicitly emphasizes stable dynamics, scalability to high-dimensional settings, and strong empirical generalization under distribution shifts. Our main contributions are:

\begin{itemize}

\item \textbf{Stability-Inducing Unrolling Formulation:}
	We propose \textit{Learning to Evolve (L2E)}, a bilevel meta-optimization framework that learns evolutionary search via \emph{stability-inducing neural unrolling}.
	The inner loop enforces structured population updates that are stability-oriented, while the outer loop meta-trains the operator to produce effective search trajectories across tasks.

\item \textbf{Structured Operator and Composite Solver:}
	We develop a \textit{Gradient-derived Evolutionary Composite} (GEC) solver that adaptively fuses learned evolutionary proposals with proxy numerical guidance in a differentiable manner.
	We further introduce a structured Mamba operator to capture population-wise interactions, providing a scalable inductive bias in high-dimensional regimes.

\item \textbf{Analysis and Empirical Validation:}
	We analyze how the proposed unrolling structure induces controlled optimization dynamics, and validate L2E on synthetic benchmarks and real-world control tasks.
	Results show strong performance, scalability to high dimensions, and robust zero-shot generalization under distribution shifts.

\end{itemize}

The remainder of the paper is organized as follows. Section~\ref{sec:background} reviews related work. Sections~\ref{sec:methodology} and \ref{sec:convergence} detail the proposed L2E framework and its theoretical analysis. Section~\ref{sec:experiments} presents the experimental results. Finally, Section~\ref{sec:conclusion} concludes the paper. 

\FigTeaser

\section{Related Work}
\label{sec:background}\subsection{Learning to Optimize}
Learning to Optimize (L2O) combines the rigor of classical optimization with the capabilities of machine learning~\cite{andrychowicz2016learning}. Specifically, rather than relying on hand-crafted heuristics, L2O seeks to parameterize update rules directly from data. Early works employed  {model-free} architectures, such as RNNs~\cite{Andrychowicz2016L2O} or MLPs~\cite{metz2019understanding}, to map gradients to updates in a coordinate-wise manner. However, these methods often struggle with scalability and interpretability, primarily due to their opaque nature.
In contrast,  {model-based} L2O approaches embed structural priors into the learner. For instance, transformer-based optimizers~\cite{gartner2023transformer} leverage self-attention to capture parameter dependencies. This methodology has further evolved into meta-learning frameworks that train optimizers to adapt across diverse tasks~\cite{ma2025toward, lange2023discovering}. Related directions include reinforcement learning controllers for scheduling~\cite{ma2023metabox} and gradient-guided mutation operators~\cite{li2025b2opt}.

Despite these advances, existing L2O approaches primarily function as unconstrained policy networks. Specifically, they lack intrinsic geometric constraints. This deficiency often leads to unstable optimization dynamics and brittle generalization under distribution shifts. Consequently, these limitations highlight the need for {structure-aware} learned optimizers that promote controlled update behaviors.

\subsection{Learnable Evolutionary Algorithms} 
Classical evolutionary algorithms and Bayesian Optimization methods have established the foundation for gradient-free optimization. Prominent examples include Evolutionary Strategies, such as CMA-ES~\cite{hansen2003reducing}, Differential Evolution~\cite{price2013differential}, and L-SHADE~\cite{tanabe2014lshade}. Within the Bayesian domain, notable approaches include SAASBO~\cite{eriksson2021high}, TuRBO~\cite{eriksson2019turbo}, and HEBO~\cite{cowen2022hebo}. These methods proved successful due to engineered covariance adaptation and population mechanisms~\cite{kennedy1995particle}. However, their reliance on hand-crafted, non-differentiable rules limits their flexibility in high-dimensional or ill-conditioned landscapes.
To enhance adaptability, recent research has progressively integrated learnability into evolutionary frameworks~\cite{zhou2021evolutionary,wang2024evolutionary,li2020boosting,wu2025learning,lange2024evolution}. Initial efforts utilized surrogates, such as Gaussian Processes, to approximate objective landscapes~\cite{lu2023surrogate}. Subsequent work has focused on embedding data-driven modules directly into the optimization loop. Examples include meta-evolved operators~\cite{chen2024symbol} and context-aware optimizers~\cite{ma2023metabox,ma2025toward}. Concurrently, researchers have employed Reinforcement Learning for dynamic operator selection~\cite{guo2024deep,guo2025reinforcement,li2024bridging}. Furthermore, Large Language Models are increasingly being explored for synthesizing heuristic code~\cite{tan2025towards,feng2025bopro,huang2025evaluation}.

Despite this progress, learnable EAs face substantial challenges. Most methods optimize isolated components, rather than the unified evolutionary loop, leading to fragmented training and limited end-to-end coordination between gradient guidance and population search. As a result, robust zero-shot generalization and scalability to high-dimensional regimes remain difficult to achieve.

\subsection{Neural Unrolling}
Within the broader inverse problem community, Neural Unrolling (NU) has established itself as a principled methodology for bridging model-based iteration and deep learning. Fundamentally, NU reinterprets an iterative optimization method as a computational graph, wherein each layer corresponds to a single update step and algorithmic parameters function as trainable weights~\cite{Yin2008Bregman}. Although originally developed for signal and image processing, this paradigm has produced influential architectures, including LISTA for sparse coding~\cite{Gregor2010LISTA}, ADMM-Net for compressive sensing~\cite{Yang2016ADMMNet}, and primal-dual networks for medical image reconstruction~\cite{Adler2018LPD}. NU embeds domain-specific iterations into differentiable architectures. Recent advances have extended NU to complex visual perception tasks, such as image dehazing and rain removal, by incorporating optimization-inspired heuristics into network design to enhance feature interpretability~\cite{liu2021retinex, liu2021value,liu2023hierarchical}. Furthermore, in the realm of AutoML, differentiable approaches for Neural Architecture Search (NAS) implicitly leverage the unrolling concept to optimize structural parameters via bilevel gradients~\cite{liudarts,liu2024learning,Monga2021UnrollingSurvey}. Consequently, it synergizes the interpretability of model-based methods with the adaptability of deep networks, thereby achieving robust performance across diverse problems.

Despite its prevalence in signal processing, NU remains underutilized within evolutionary computation. Although differentiable EAs exist, they typically rely on \emph{ad hoc} relaxations rather than systematic unrollings of fixed-point iterations. Consequently, the field lacks a framework that unifies bilevel meta-learning with {stability-oriented} population dynamics in a principled and trainable manner.

\subsection{Discussion} 
Existing methodologies navigate a trade-off between stability and adaptability. While classical EAs provide theoretical stability, they often lack the adaptability for navigating complex landscapes efficiently.
Conversely, emerging L2O methods offer data-driven flexibility; however, they frequently exhibit instability and poor generalization to out-of-distribution landscapes, primarily due to the absence of intrinsic geometric constraints.
To reconcile these diverging paradigms, we propose the L2E that reformulates the evolutionary loop as a neural unrolling process. 
This formulation integrates the expressivity of deep networks with {controlled optimization dynamics}. Consequently, it improves robustness under distribution shifts.
\FigPipeline

\section{Methodology}
\label{sec:methodology} 

The overall architecture operates as a bilevel dynamical system, as visualized in~Fig.~\ref{fig:pipeline}. To instantiate this paradigm, the methodology comprises three core components. First, the \emph{Generalized Neural Unrolling (GNU)} formulation establishes the theoretical foundation. Second, the \emph{Gradient-derived Evolutionary Composite (GEC)} solver integrates gradient-based exploitation with population-based exploration. Finally, the  \emph{structured Mamba neural operator} efficiently parameterizes the optimization process in high-dimensional spaces, thereby improving computational scalability and promoting controlled optimization dynamics.

\subsection{Generalized Neural Unrolling Paradigm} 
Conventional black-box optimizers often rely on heuristic mechanisms that can yield unstable dynamics on unseen landscapes.
To promote controlled update behaviors, we reinterpret population evolution as a structured operator iteration, adopting a fixed-point inspired unrolling view based on the KM formulation~\cite{xu2006variable}. Unlike methods that directly search for optima, the proposed approach optimizes the parameters of a learnable fixed-point operator. This operator implicitly governs the solution trajectory. Formally, let \( f: \mathcal{X} \to \mathbb{R} \) denote the black-box objective function. We introduce a learnable non-expansive-inspired operator \( \mathcal{NU}(\cdot;\bm{\omega}) \), parameterized by $\bm{\omega}\in \bm{\Omega}$, whose fixed point approximates the minimizer of  \(f \). Consequently, we formulate this framework as a constrained bilevel optimization problem~\cite{liu2021investigating}:
\begin{equation}
	\min _{\mathbf{x} \in \mathcal{X}, \boldsymbol{\omega} \in \Omega} \ell_{\mathtt{meta}}(\mathbf{x} ; \boldsymbol{\omega},\mathtt{Opt}), s.t.\; \mathbf{x} \in \operatorname{Fix}(\mathcal{NU}(\cdot; \boldsymbol{\omega})),
	\label{eq:fix-constraint}
\end{equation}
where $\ell_{\mathtt{meta}}(\mathbf{x};\boldsymbol{\omega},\mathtt{Opt})$ is the meta-objective that evaluates the population state $\mathbf{x}$ induced by the optimizer parameter $\boldsymbol{\omega}$ under a prescribed evaluation protocol $\mathtt{Opt}$. Furthermore, \( \text{Fix}(\mathcal{NU}) := \{ \bm{\mathbf{x}} \mid \bm{\mathbf{x}} = \mathcal{NU}(\bm{\mathbf{x}}; \bm{\omega}) \} \) defines the set of fixed points.  
Here, $\bm{\omega}$ are outer-level meta-parameters that parameterize the optimizer, whereas $\mathbf{x}$ denotes the inner-level population state of candidate solutions.

To introduce a stability-biased update structure, we construct \( \mathcal{NU} \) as an \(\alpha\)-averaged  (KM-style)  operator. Specifically, this formulation employs a convex combination of the identity and a learnable composite mapping $\mathcal{O}(\cdot, \boldsymbol{\omega})$, i.e., 
\begin{equation}
	\bm{\mathbf{x}}^{k+1}=\mathcal{NU}(\mathbf{x}^k; \boldsymbol{\omega}) = (1 - \alpha) \cdot  \mathbf{x}^k +  \alpha  \cdot \mathcal{O}(\mathbf{x}^k; \boldsymbol{\omega}), 
	\label{eq:km-unroll}
\end{equation} 
\begin{equation}
	\mathtt{where}\ 	\mathcal{O}(\cdot; \boldsymbol{\omega}) \triangleq \{ \underbrace{\mathcal{O}_{\mathrm{evo}}(\cdot;\boldsymbol{\omega})}_{\mathtt{Neural~operator}}
	\;\circ\;
	\underbrace{\mathcal{O}_{\mathrm{num}}(\cdot,\boldsymbol{\omega})}_{\mathtt{Numerical~operator}} \}.
\end{equation}The operator \( \mathcal{NU}(\cdot; \bm{\omega}) \) is defined as a composition of two components, modulated by an update relaxation parameter \( \alpha \in (0,1] \). $\mathcal{O}_{\mathrm{evo}}$ is a learnable evolutionary proposal operator, and $\mathcal{O}_{\mathrm{num}}$ provides proxy numerical guidance (e.g., projection or smoothing) to regulate the update behavior. 
Together, the averaging step and the composite structure impose a stability-inducing bias on the unrolled trajectory, which empirically reduces the brittle dynamics often observed in unconstrained learned optimizers.

This formulation gives rise to a semantically structured learning process: \textit{i) Inner-Level (Evolve)}: For fixed algorithm parameters  $\bm{\mathbf{\omega}}$, the inner loop performs 
K-step unrolled evolution of an initial population 
$\bm{\mathbf{x}}^0$. 
This unrolling implicitly defines the optimization trajectory, transforming initial samples into refined candidates.
\textit{ii) Outer-Level (Meta-Train):} The outer loop treats the terminal population \( \bm{\mathbf{x}}^K \) as a response to the optimizer parameters \( \bm{\omega} \). A meta-objective \(  \ell_{\mathtt{meta}}(\bm{\mathbf{x}}^K; \bm{\omega}) \) is computed to measure task performance, and gradients are backpropagated through the unrolling path to update \( \bm{\omega} \):
$
\bm{\omega}^{t+1} = \bm{\omega}^{t} - \gamma \nabla_{\bm{\omega}} \ell_{\mathtt{meta}}(\bm{\mathbf{x}}^K; \bm{\omega}^{t}),
$
where \( \gamma \) is the meta-step size. This teaches the operator to evolve more effective trajectories across tasks.

\AlgLTwoE

\subsection{Gradient-derived Evolutionary Composite Solver}
To instantiate the proposed operator \( \mathcal{NU}(\cdot; \bm{\omega}) \) introduced above, we propose a hybrid mechanism that fuses the global exploration capability of evolutionary algorithms with the local refinement power of gradient descent. This  GEC solver operates in a dual-mode fashion. 
\subsubsection{Learned evolutionary update} At each inner iteration \( k \), given the current population stat \( \bm{\mathbf{x}}^{k-1} \), the solver computes two candidate update directions:
\begin{equation}
	\bm{d}_{IL}^k(\bm{\omega}) = \mathcal{NU}(\bm{\mathbf{x}}^{k-1}; \bm{\omega}),
\end{equation}where \( \mathcal{NU} \) follows the KM-style fixed-point update defined in Eq.~\eqref{eq:km-unroll}, which is composed of the learnable operator \( \mathcal{O}_{\bm{\omega}} \).
\subsubsection{Proxy Gradient Update}
\begin{equation}
	\bm{d}_{OL}^k(\bm{\omega}) = \bm{\mathbf{x}}^{k-1} - s_k\, \bm{P}_{\bm{\omega}}^{-1} \nabla_{\bm{\mathbf{x}}}\,  \ell_{\mathtt{meta}}(\bm{\mathbf{x}}^{k-1}, \bm{\omega}),
\end{equation}where \( s_k \) denotes the inner step size, and \( \bm{P}_{\bm{\omega}} \) represents a preconditioning matrix, such as an identity matrix, a diagonal Hessian, or a low-rank approximation.
\subsubsection{Composite update with soft-gated fusion}  
A dimension-wise gating mask $\mathbf{M}^k$ is computed based on the relative fitness improvement $\Delta f^k = f(\bm{d}_{OL}^k) - f(\bm{d}_{IL}^k)$, and apply a smooth gating function to produce a soft selection mask
\begin{equation}
	\bm{M}^k = \sigma(-\Delta f^k / \tau), \quad \bm{M}^k \in (0, 1)^{B \times N \times 1},
\end{equation}where \( \sigma(\cdot) \) is the element-wise sigmoid function and \( \tau \) is a temperature parameter that controls the gating sharpness. The final update is computed as a soft interpolation between the two candidates:
\begin{equation}
	\bm{\mathbf{x}}^k = \text{Proj}_{\mathcal{X}, \bm{P}_{\bm{\omega}}} \left( \bm{M}^k \odot \bm{d}_{OL}^k + (1 - \bm{M}^k) \odot \bm{d}_{IL}^k \right),
\end{equation}where \( \text{Proj}_{\mathcal{X}, \bm{P}_{\bm{\omega}}} \) denotes a projection back onto the feasible set under the \( \bm{P}_{\bm{\omega}} \)-induced metric.\footnote{The subscript $(\mathcal{X},\mathbf{P}_{\bm{\omega}})$ indicates a metric projection onto the feasible set $\mathcal{X}$ under the $\mathbf{P}_{\bm{\omega}}$-induced norm:
	\[
	\mathrm{Proj}_{\mathcal{X},\mathbf{P}_{\bm{\omega}}}(\mathbf{z})
	:= \arg\min_{\mathbf{x}\in\mathcal{X}} \ \|\mathbf{x}-\mathbf{z}\|_{\mathbf{P}_{\bm{\omega}}}^2,
	\qquad
	\|\mathbf{v}\|_{\mathbf{P}_{\bm{\omega}}}^2 := \mathbf{v}^\top \mathbf{P}_{\bm{\omega}}\mathbf{v}.
	\]
}
This fusion mechanism balances exploitation and exploration. It operates in a manner that is both differentiable and stability-preserving.

\subsubsection{Meta-objective}
Let $\bar{f}(\mathbf{x})$ denote the mean objective value of a population $\mathbf{x}$. 
We define the meta-objective as the negative expected \emph{normalized improvement} over the task distribution $p(f)$:
\begin{equation}
	\ell_{\mathtt{meta}}(\bm{\omega}) = - \mathbb{E}_{f \sim p(f)} \left[ \frac{\bar{f}(\mathbf{x}^0) - \bar{f}(\mathbf{x}^K(\bm{\omega}))}{|\bar{f}(\mathbf{x}^0)| + \epsilon} \right],
	\label{eq:meta_loss}
\end{equation}
where $\mathbf{x}^K(\bm{\omega})$ is the terminal population after $K$ unrolled updates on function $f$, and $\epsilon$ is a small constant for numerical stability. 
This normalization makes the meta-loss less sensitive to the absolute scale of objective values, enabling consistent training across tasks with different magnitudes. 
In Algorithm~\ref{alg:l2e_advanced_dual}, we minimize $\ell_{\mathtt{meta}}$ via BPTT to update $\bm{\omega}$, thereby shaping the learned evolutionary dynamics to improve terminal performance under a fixed evaluation budget.

\subsection{Structured Mamba Neural Operator}
\label{sec:method_c} 
To instantiate the evolution operator $\mathcal{O}_{\mathrm{evo}}(\cdot;\bm{\omega})$, we adopt a structured Mamba-based operator implemented in a parallel selective form to respect the permutation-invariant nature of populations (Fig.~\ref{fig:fig3}).

\subsubsection{Parallel Mamba Block}
The operator takes the population $\mathbf{x} \in \mathbb{R}^{B \times N \times D}$ and fitness values $\mathbf{f}$ as input and maps them to an embedding $\mathbf{E}$. 
We then generate input-dependent parameters, including the discretization timescale $\boldsymbol{\Delta}$ and state transformation matrices $\mathbf{B}, \mathbf{C}$, and compute two parallel streams: a state-expanded representation and a passthrough signal. Specifically,
\begin{equation}
	M_{\mathbf{s}} = (\boldsymbol{\Delta} \mathbf{B}) \odot \mathbf{E}, \,  \,  [\boldsymbol{\Delta}, \mathbf{B}, \mathbf{C}] = \text{Linear}_{proj}(\mathbf{E}).
	\label{eq:streams}
\end{equation}The output is computed via a gated combination:
\begin{equation} \small
	\mathbf{H}_{mamba} = \text{LN}\left( \mathbf{z} \odot M_{\mathbf{s}} + (1-\mathbf{z}) \odot \mathbf{u} \right), \mathbf{z} = \sigma(\text{Linear}_{z}(\mathbf{E})).
	\label{eq:mamba_out}
\end{equation}

\subsubsection{Hybrid Adaptive Fusion}
To incorporate population-level coordination, we add a parallel multi-head self-attention path
$
\mathbf{H}_{attn} = \text{MHSA}(\mathbf{E}) + \mathbf{E}.
$
The final evolutionary proposal is obtained by routing between the two paths using population statistics $\mathbf{s}_{pop}$: 
\begin{equation}
	[\lambda_{ssm}, \lambda_{attn}] = \text{Softmax}(\text{MLP}_{router}(\mathbf{s}_{pop})).
	\label{eq:router}
\end{equation} 
\begin{equation} \small
	\Delta_{evo} = \lambda_{ssm} \cdot \text{Tanh}(\Psi_{m}(\mathbf{H}_{mamba})) + \lambda_{attn} \cdot \text{Tanh}(\Psi_{a}(\mathbf{H}_{attn})).
	\label{eq:hybrid_fusion}
\end{equation}Here, $\Psi_m$ and $\Psi_a$ denote projection heads. The \texttt{Tanh} activation bounds the update magnitude, serving as a stability-oriented design choice. Here we apply {Spectral Normalization}~\cite{miyato2018spectral} to the learnable linear transformations in both paths to control update sensitivity. 
Together, spectral normalization and bounded updates act as stability-oriented regularization, making the learned updates less sensitive to input perturbations in practice.

\FigMamba
\section{Theoretical Analysis}
\label{sec:convergence}

This section provides a {stability-oriented} theoretical analysis for L2E. We do not claim universal guarantees for arbitrary neural components. Instead, we clarify the \emph{sufficient conditions} under which the unrolled operator iteration exhibits controlled dynamics and admits meaningful optimization bounds. Due to space constraints, detailed proofs are deferred to the \textit{Supplementary Document}.

\subsection{Convergence Rate Analysis}
To characterize the convergence rate, we first adopt standard regularity assumptions gritounded in KM theory.
\begin{assumption}  
	\begin{enumerate}

\item[(A1)] ({Operator regularity}) 
For any fixed $\bm{\omega} \in \Omega$, the learned composite operator $\mathcal{O}(\cdot; \bm{\omega})$ is non-expansive with respect to the population state $\mathbf{x}$:
$ \| \mathcal{O}(\mathbf{x}; \bm{\omega}) - \mathcal{O}(\mathbf{x}'; \bm{\omega}) \| \leq \| \mathbf{x} - \mathbf{x}' \|, \ \forall \mathbf{x}, \mathbf{x}' \in \mathcal{X}. $

\item[(A2)] (Smoothness) $\ell_{\mathtt{meta}}(\mathbf{x}, \bm{\omega})$ is $L_{\mathbf{x}}$-smooth with respect to the population $\mathbf{x}$ and $L_{\bm{\omega}}$-smooth with respect to the parameters $\bm{\omega}$.

\item[(A3)] (Lipschitz Jacobian) The unrolled mapping $\mathbf{x}^K(\bm{\omega})$ admits a bounded Jacobian, i.e., $\| \nabla_{\bm{\omega}} \mathbf{x}^K(\bm{\omega}) \| \leq C_K$.

\item[(A4)] (Bounded domain) 
		The initial population $\mathbf{x}^0$ and $\mathbf{x}^*(\bm{\omega}) \in \text{Fix}(\mathcal{NU}(\cdot; \bm{\omega}))$ are contained within a bounded region of diameter $D_0$: $\| \mathbf{x}^0 - \mathbf{x}^*(\bm{\omega}) \| \leq D_0$.
	\end{enumerate} 
\end{assumption}

Based on these, we derive the approximation error of theinner loop and the convergence rate of the outer loop.

\begin{proposition}[Approximation Error]
	\label{prop:inner_error}
	Under Assumption (A1), the $K$-step unrolled KM iteration satisfies:
\begin{equation}
		\|\mathbf{x}^{K}(\bm{\omega}) - \mathbf{x}^{*}(\bm{\omega})\| \leq (1-\alpha)^K D_0,
	\end{equation}	where $\mathbf{x}^{*}(\bm{\omega})$ is the fixed point that satisfies $\mathbf{x}^{*} = \mathcal{O}(\mathbf{x}^{*}; \bm{\omega})$.
\end{proposition}

\begin{theorem}[Convergence Rate]
	\label{thm:joint_rate}
	Under Assumptions (A1)-(A4), let the meta-parameters be updated via gradient descent: $\bm{\omega}^{t+1} = \bm{\omega}^t - \gamma \nabla_{\bm{\omega}} \ell_{\mathtt{meta}}(\mathbf{x}^{K}(\bm{\omega}^t); \bm{\omega}^t)$.
	Provided that the step size is $\gamma < \frac{1}{L_{\bm{\omega}} + L_{\mathbf{x}} C_K}$, the meta-gradient norm satisfies:
\begin{equation}
		\min_{0 \leq t < T} \|\nabla \ell_{\mathtt{meta}}(\bm{\omega}^t)\|^2 \leq \mathcal{O}\left(\frac{1}{T} + (1-\alpha)^{2K}\right).
	\end{equation}\end{theorem}

\textit{Remark:} The bound decomposes into an optimization term $\mathcal{O}(1/T)$ and an approximation term $\mathcal{O}((1-\alpha)^{2K})$. This decomposition explicitly characterizes the influence of the unrolling depth $K$ and the averaging factor $\alpha$ on the meta-update quality under Assumptions (A1)-(A4). Moreover, the analysis aligns with established bilevel optimization literature, in which guarantees for convergence and consistency are derived from operator regularity and topological conditions~\cite{sabach2017first, liu2021investigating}.

\subsection{Asymptotic Consistency under Fixed-point Constraints}  
To interpret the limiting behavior of Algorithm~\ref{alg:l2e_advanced_dual} under the fixed-point constrained formulation, we consider stronger (idealized) topological assumptions.
These assumptions facilitate an asymptotic consistency analysis. Specifically, this analysis connects the unrolled iteration to the fixed-point set of Eq.~\eqref{eq:fix-constraint} and elucidates its relationship to the induced outer objective.

\begin{assumption}
	\label{ass:advanced_topology}
	\begin{enumerate} 
		\item[(B1)] ({Non-expansiveness}) For each $\bm{\omega} \in \Omega$, the composite operator satisfies: $ \|\mathcal{O}(\mathbf{u}_1; \bm{\omega}) - \mathcal{O}(\mathbf{u}_2; \bm{\omega})\|_{\mathbf{P}_{\bm{\omega}}} \leq \|\mathbf{u}_1 - \mathbf{u}_2\|_{\mathbf{P}_{\bm{\omega}}}, \quad \forall \mathbf{u}_1, \mathbf{u}_2 \in \mathcal{X}. $
		\item[(B2)] ({Closedness}) The graph $\mathrm{gph}(\mathcal{O}(\cdot; \bm{\omega})) := \{ (\mathbf{u}, \mathbf{v}) \in \mathcal{X} \times \mathcal{X} \mid \mathbf{v} = \mathcal{O}(\mathbf{u}; \bm{\omega}) \}$ is closed.
		\item[(B3)] ({Compactness and Continuity}) The sets $\Omega$ and $\mathcal{X}$ are compact. Furthermore, for any $\bm{\omega}$, the set $\text{Fix}(\mathcal{NU}(\cdot; \bm{\omega}))$ is non-empty. Additionally, $\ell_{\mathtt{meta}}$ is continuous on $\mathcal{X} \times \Omega$. Finally, for a fixed $\bm{\omega}$, the function $\ell_{\mathtt{meta}}(\cdot; \bm{\omega})$ is $L_\ell$-smooth, convex, and lower-bounded.
	\end{enumerate}
\end{assumption}

\begin{theorem} 
	\label{thm:joint_optimality}
	Based on Assumption~\ref{ass:advanced_topology}, we establish the following convergence properties for L2E:

\noindent\textit{(i)} Let $\{ \mathbf{x}^k(\bm{\omega}) \}$ denote the sequence generated by Algorithm~\ref{alg:l2e_advanced_dual}.  For any $\bm{\omega} \in \Omega$, assume the step size satisfies $s_k = \frac{\kappa}{k+1}$, where $0 < \kappa < \frac{\lambda_{\min}(\mathbf{P}_{\bm{\omega}})}{L_{\mathbf{x}}}$. Under these conditions, the sequence converges to the fixed-point set, i.e., $\lim_{k \to \infty} \mathrm{dist}(\mathbf{x}^k(\bm{\omega}), \mathrm{Fix}(\mathcal{NU}(\cdot; \bm{\omega}))) = 0$. Consequently, the meta-objective value converges to the optimum: 
\begin{equation}
		\lim_{k \to \infty} \ell_{\mathtt{meta}}(\mathbf{x}^k(\bm{\omega}); \bm{\omega}) = \inf_{\mathbf{x} \in \mathrm{Fix}(\mathcal{NU}(\cdot; \bm{\omega}))} \ell_{\mathtt{meta}}(\mathbf{x}; \bm{\omega}).
	\end{equation}\noindent\textit{(ii)} Let $\{ (\mathbf{x}^K(\bm{\omega}^K), \bm{\omega}^K) \}$ denote the sequence of generated solutions as the unrolling depth $K \to \infty$. We define the induced outer objective (value function) as
	$
	\varphi(\bm{\omega})
	:= \inf_{\mathbf{x}\in \mathrm{Fix}(\mathcal{NU}(\cdot;\bm{\omega}))}\ \ell_{\mathtt{meta}}(\mathbf{x};\bm{\omega}),
	$~$
	\varphi_K(\bm{\omega}) := \ell_{\mathtt{meta}}(\mathbf{x}^K(\bm{\omega});\bm{\omega}).
	$
	 We prove that any limit point $(\bar{\mathbf{x}}, \bar{\bm{\omega}})$ is a valid solution to the fixed-point constrained bilevel problem (Eq.~\eqref{eq:fix-constraint}). In particular, the limit point satisfies $\bar{\bm{\omega}} \in \arg \min_{\bm{\omega} \in \Omega} \varphi(\bm{\omega})$ and $\bar{\mathbf{x}} = \mathcal{NU}(\bar{\mathbf{x}}; \bar{\bm{\omega}})$. Furthermore, the sequence of surrogate objectives converges to the optimum of the induced outer objective, i.e., $\lim_{K \to \infty} \inf_{\bm{\omega} \in \Omega} \varphi_K(\bm{\omega}) = \inf_{\bm{\omega} \in \Omega} \varphi(\bm{\omega})$.
\end{theorem}

\section{Experimental Study}\label{sec:experiments}\subsection{Experimental Setting}

\noindent\textbf{Benchmark Setup.} We evaluate the optimization efficiency and zero-shot generalization capability of L2E using a hierarchical benchmark suite. This suite encompasses tasks ranging from synthetic functions to real-world robot control. Regarding hyperparameters, we set the relaxation factor $\alpha=0.9$, the gating temperature $\tau=1$, the unrolling horizon $K=200$, and the population size $B=100$, unless otherwise specified. For spectral normalization, we use the default coefficient $1.0$ for all linear layers. We conduct all experiments on a single NVIDIA RTX A6000 GPU (PyTorch 2.0.1). Each task is evaluated over 10 independent random seeds, and we report both the mean and standard deviation.

\FigbbobTen

\begin{itemize}
	\item \textbf{Synthetic Benchmarks (In-Distribution):} 
	We utilize {\texttt{BBOB-10D}}, {\texttt{BBOB-30D}}~\cite{hansen2021coco}, and {\texttt{LSGO-1000D}}~\cite{li2013benchmark} as the primary testbeds. 
	These benchmarks represent the same function families utilized during meta-training, although we conduct the evaluation on held-out runs or instances under identical protocols. 
	Specifically, \texttt{BBOB} assesses general optimization capability with budgets of $2 \times 10^4$ (10D) and $5 \times 10^4$ (30D), whereas \texttt{LSGO} evaluates scalability in high-dimensional regimes ($D \approx 1000$) with a budget of $3 \times 10^6$.

\item \textbf{Out-of-Distribution (OOD) Evaluation:} 
	We incorporate two benchmarks to assess robustness beyond the training distribution. 
	First, we employ the {\texttt{UAV}} benchmark~\cite{shehadeh2025benchmarking}, which consists of 56 real-world path planning instances. Specifically, we directly evaluate a model trained on {\texttt{BBOB-30D}} under a constrained budget of $2.5 \times 10^2$ without fine-tuning. 
	Second, we use {\texttt{BBOB-surrogate-10D}}, a synthetic variant derived from surrogate-induced landscapes, to test robustness to landscape shifts and approximation mismatch; we apply the model trained on {\texttt{BBOB-10D}} directly.

\item \textbf{Neuroevolution for Robot Control:} 
	We additionally evaluate performance on the {\texttt{NE}} benchmark~\cite{huang2024evox}, which comprises high-dimensional policy search tasks ($D > 1000$) for robot control. 
	This suite assesses the applicability of L2E in control scenarios and benchmarks it against representative reinforcement learning methods under a maximum evaluation budget of $2{,}500$.
\end{itemize}

\TabbbobTen	

\noindent\textbf{Compared Methods.} 
We conduct a comparative analysis against representative algorithms to evaluate the effectiveness of the proposed L2E framework, covering both classical heuristics and learning-based optimizers. 
To situate our stability-inducing unrolling framework within the broader L2O landscape, we include two categories of learned optimizers. 
The first category consists of meta-learning-based neuro-evolution strategies, including DEDQN~\cite{tan2021differential}, LES~\cite{lange2023discovering}, and LGA~\cite{lange2023discovering2}. 
The second category includes control-based solution generation models, including RNNOPT~\cite{tv2019meta}, GLHF~\cite{li2024pretrained}, and B2OPT~\cite{li2025b2opt}. 
In addition, we include representative classical population-based baselines, including DE~\cite{storn1997differential}, PSO~\cite{kennedy1995particle}, and SAHLPSO~\cite{tao2021self}, to provide reference performance under the same evaluation budgets. 
We re-evaluate all competing methods under a unified experimental protocol and identical computational budgets. 


\FigbbobThirty

\subsection{Performance on Numerical Benchmarks}
\label{sec:synthetic_results}
We first examine the optimization capability of L2E on numerical landscapes where the meta-training and testing distributions align. 
This assessment focuses on two aspects: performance on standard numerical benchmarks ({BBOB}) and scalability to high dimensions ({LSGO}).

\subsubsection{Precision and Stability on Standard Benchmarks} 
Table~\ref{tab:bbob-10d} and Table~\ref{tab:bbob-30d} report the comparative results on  {BBOB-10D} and  {BBOB-30D}, respectively. 
Overall, L2E achieves consistently strong performance across a wide range of test instances:

\begin{itemize}
	\item \textbf{Performance and Architectural Efficacy:} 
	L2E attains the lowest mean error on {31 out of 32} test instances. 
	Specifically, L2E outperforms established learned baselines, such as GLHF and B2OPT, on ill-conditioned and multimodal functions (e.g., \texttt{Rosenbrock\_rot}, \texttt{Lunacek\_bi\_Ras}). 
	We attribute these gains to the \emph{stability-inducing} unrolling design and the composite update structure, which promote controlled optimization dynamics in high-curvature regions. 
	Furthermore, the Mamba-based variant (\textit{Ours$^{\ddag}$}) consistently improves upon the base model (\textit{Ours}), particularly on non-separable landscapes (e.g., \texttt{Ellipsoidal\_hc}). 
	This finding suggests that the structured state-space operator provides an effective inductive bias for capturing population-wise variable couplings.

	\item \textbf{Trajectory and Stability Analysis:} 
	L2E demonstrates robust convergence behaviors, as visualized in Figure~\ref{fig:bbob-10d-diff-16} and Figure~\ref{fig:bbob-30d-diff-16}. 
	Specifically, the method exhibits faster early-stage progress with tighter confidence intervals. 
	On unimodal functions (e.g., \texttt{Bent Cigar}), L2E quickly reaches a low-error regime, which indicates effective exploitation when the landscape is smooth. 
	In contrast, on multimodal tasks (e.g., \texttt{Gallagher 21Peaks}), L2E maintains steady improvement without early stagnation. 
	This sustained progress reflects the benefit of the \textit{Elite-Mask Fusion}, which adaptively balances learned evolutionary proposals and proxy numerical guidance.
\end{itemize}

\subsubsection{Scalability to High-Dimensional Regimes}
Results on the 1000D  {LSGO} benchmark (Table~\ref{tab:lsgo}) further demonstrate the scalability of L2E. 
L2E ranks first across all six representative high-dimensional functions. 
Figure~\ref{fig:ecdf-lsgo-1000D} (ECDF) and Fig.~\ref{fig:boxplot-lsgo-1000D} (boxplots) illustrate the scalability of L2E in high-dimensional regimes. 
In contrast to existing learned optimizers, L2E exhibits consistent progress and reduced performance variance in the 1000D regime. 
This robustness stems from the structured operator design and the regularization strategies, i.e., bounded updates and soft-gated fusion. 
Consequently, these mechanisms promote controlled dynamics during optimization in high-dimensional spaces. 

\Figlsgo

\TabbbobThirty
\Figuav

\subsection{Generalization to Real-World UAV Planning}
\label{sec:ood_generalization} 
We evaluate the robustness of L2E under distribution shifts using the UAV Path Planning benchmark. 
Although L2E is meta-trained exclusively on synthetic  {BBOB-30D} functions, it is directly applied to 56 UAV planning instances (30D)  {without fine-tuning}. 
As shown by the radar plots in Fig.~\ref{fig:terrain-radar-all}, L2E achieves the best normalized performance across all seven terrain groups, with relative gains over the second-best baseline ranging from  {3.4\%} to  {14.2\%}.
We attribute this empirical transferability to the stability-inducing unrolling formulation and the structured composite updates in L2E. 
Compared with policy-style learned optimizers that may become brittle when the constraint patterns and noise characteristics differ from training, L2E relies on structured population updates with stability-oriented regularization (e.g., bounded proposals and soft-gated numerical guidance). 
These design choices promote controlled dynamics under the strict evaluation budget, enabling more reliable progress on irregular UAV landscapes.

\subsection{Neuroevolution for Robot Control}
\label{sec:ne_results}

\FigNE	

\Figablshare

Finally, we evaluate the applicability of L2E to high-dimensional policy search in the Neuroevolution (NE) benchmark~\cite{huang2024evox}. 
This setting introduces a distinct distribution shift from synthetic function optimization to control-oriented policy search. 
The benchmark includes six robotic tasks across two domains, \textit{ant} and \textit{humanoid}, where each domain comprises three complexity variants (ant-3/4/6 and humanoid-3/4/5). 
We compare L2E against {GLEET}~\cite{ma2024auto}, a specialized baseline that frames neuroevolution within an actor--critic reinforcement learning pipeline. 
This comparison tests whether a learned evolutionary optimizer can serve as a drop-in alternative to handcrafted population update heuristics in control scenarios.

Figure~\ref{fig:ne} reports normalized learning curves to enable a fair visual comparison across tasks with different reward scales. 
The normalization is defined as
$
y_{\text{plot}} = \frac{y - \text{mean}}{10^{\text{exp}}}.
$
On representative mid-difficulty tasks (e.g., \textit{ant-4} and \textit{humanoid-4}), L2E exhibits faster progress and improved final performance compared with the baseline. 
Moreover, the aggregated boxplots (Right) indicate that L2E achieves better terminal objective values with consistently reduced variance across all six tasks. 
These results suggest that the stability-inducing unrolling formulation and structured composite updates in L2E promote controlled policy search dynamics under noisy and high-variance feedback. 
Consequently, L2E improves robustness in neuroevolution-based control.

\section{Ablation Analysis and Mechanistic Insights}
\label{sec:ablation}
To quantify the contribution of each component and motivate the design choices, we conduct systematic ablations on  {BBOB-10D}. 
We focus on two questions: 
(1) how different parameterization strategies for the unrolled evolutionary operator affect optimization dynamics (shared vs.\ unshared weights across steps), and 
(2) how the core modules(i.e.,  {ProxyGrad},  {SoftGate}, and  {Mamba}) influence performance and stability.

\subsection{Shared vs.\ Unshared Unrolling Parameters}

A pivotal design choice in neural unrolling determines whether to share the meta-parameters of the evolution operator $\mathcal{O}_{\mathrm{evo}}(\cdot;\bm{\omega})$ across unrolling steps $k=1,\dots,K$. 
This distinction establishes two primary configurations:
\begin{itemize}
	\item \textbf{Shared ($\bm{\omega}_{\text{shared}}$):} A single set of weights is utilized across all unrolled blocks, which enforces a time-invariant update rule.
	\item \textbf{Unshared ($\bm{\omega}^k$):} Independent weights are assigned to distinct blocks, thereby enabling step-dependent behaviors that capture different phases of the optimization trajectory.
\end{itemize}

\Figablvariant
\Tabablvariant
\Figablcurve

\subsubsection{Statistical Advantage of the Unshared Design}
Across evaluation budgets, the unshared parameterization achieves consistently better performance than the shared-weight baseline. 
Table~\ref{tab:ablation-bbob10d-obj-agg-std-14x7} summarizes the aggregated results under different evaluation budgets, where the performance gap becomes more pronounced as the unrolling horizon increases. 
In particular, at $\mathrm{MaxFEs}=20{,}000$ ($\mathrm{EMS}=200$), the unshared model reduces the mean error by  {approximately 96\%} relative to the shared variant. 
These results suggest that a time-invariant update rule can be insufficient to accommodate the evolving needs of the optimization trajectory. 
By assigning step-specific meta-parameters $\bm{\omega}^k$, the unshared design enables stage-dependent update behaviors (e.g., exploration-oriented early steps and refinement-oriented later steps), which becomes increasingly beneficial as $\mathrm{EMS}$ grows. 
Therefore, we adopt separate per-block meta-parameters for $\mathcal{O}_{\mathrm{evo}}$ in the final model. 
Although this choice increases training cost moderately, it yields better accuracy and more stable behavior under practical evaluation budgets.

\subsubsection{Stage-wise Specialization}
Figure~\ref{fig:abl3} provides qualitative evidence that unshared parameterization enables \emph{stage-dependent} update behaviors along the unrolled trajectory. 
In the early phase (exploration), both the shared and unshared designs make rapid progress. 
However, in the late phase (refinement), the shared-weight variant (brown curves) tends to plateau earlier, indicating limited flexibility to adapt its update behavior as the search proceeds. 
In contrast, the unshared variant (red/purple curves) continues to refine the solutions and achieves better late-stage improvements. 
Overall, this visualization suggests that step-specific meta-parameters $\bm{\omega}^k$ allow different unrolling blocks to specialize in different roles with earlier blocks favoring broader exploration and later blocks emphasizing local refinement, which is beneficial for achieving higher final precision in black-box optimization.

\subsection{Deconstructing Component Synergy}
We further isolate the contributions of three core components in L2E: 
(i) \textit{ProxyGrad}, which injects proxy numerical guidance; 
(ii) \textit{SoftGate}, which enables adaptive fusion; and 
(iii) \textit{Mamba}, which parameterizes the structured evolutionary operator. 
We compare the full model (\textit{Ours}) with variants that remove or replace each component to quantify their individual and combined effects.
\Tabablationobj

\Figablcurvefirsteight

\subsubsection{Quantitative Impact Analysis}
Table~\ref{tab:ablation-obj-16} quantifies the performance changes when individual components are removed or replaced.
\begin{itemize}
	\item \textbf{Impact of Proxy Guidance (\textit{w/o ProxyGrad}):} 
	Removing proxy gradient guidance leads to the largest degradation, increasing the error by $\sim 103\times$. 
	This result indicates that, under the considered evaluation budgets, purely evolutionary updates can be substantially less sample-efficient, whereas proxy guidance provides an effective directional signal for accelerating progress.

\item \textbf{Impact of Adaptive Fusion (\textit{w/o SoftGate}):} 
	Disabling the gating mechanism (i.e., using a fixed combination) degrades performance by $\sim 161\times$. 
	This result underscores the necessity of selective fusion. 
	Specifically, the solver benefits from dynamically balancing proxy guidance for refinement and learned evolutionary proposals for exploration. 
	In contrast, a rigid mixing rule lacks this essential flexibility.
	
\FigablcurveEllipsoidal

\item \textbf{Impact of Operator Parameterization (\textit{w/o Mamba}):} 
	Replacing the Mamba-based operator with a standard MLP/attention block increases the error by $\sim 6.8\times$. 
	Although this degradation is smaller than that caused by removing guidance or gating, it indicates that the structured operator parameterization enhances the modeling of population-wise interactions and variable couplings. 
	This capability is particularly beneficial on complex landscapes.
\end{itemize}

\subsubsection{Convergence Dynamics}
Figure~\ref{fig:abl} and Fig.~\ref{fig:abl2} further visualize convergence behavior and run-to-run variability. 
On ill-conditioned functions such as \texttt{Ellipsoidal\_hc}, the full model (red) achieves better terminal performance with tighter variance (Fig.~\ref{fig:abl2}, Right). 
In contrast, the \textit{w/o SoftGate} variant exhibits noticeably higher variance, consistent with the intuition that rigid update rules can be brittle across diverse instances.
Overall, the three components act in a complementary manner. 
ProxyGrad provides effective numerical guidance, Mamba offers a structured parameterization for population updates, and SoftGate adaptively mediates between the two update sources. 
Removing any single component weakens this complementarity and often leads to slower progress or earlier stagnation. 
Together with unshared unrolling parameters, these design choices promote stage-dependent behaviors along the trajectory and enable stable, effective optimization across diverse numerical landscapes.

\section{Conclusion}
\label{sec:conclusion} 
In this paper, we proposed  {L2E}, a unified framework that approaches evolutionary optimization via  {neural unrolling}. 
Instead of learning unconstrained update rules, L2E adopts a \emph{stability-inducing} unrolling formulation within a bilevel meta-learning scheme, which promotes controlled optimization dynamics in practice. 
Empirical results on high-dimensional numerical benchmarks and robot control tasks demonstrate the scalability of L2E and its robust performance under distribution shifts.
\textit{Limitations:}  
The framework introduces additional training-time overhead due to unrolling and structured operator parameterization. 
This cost may limit applicability in resource-constrained or low-latency settings.
\textit{Future Work:}  
We will develop  {lightweight} operator parameterizations and more efficient training strategies to reduce computation and memory usage. 
We also plan to extend L2E to broader settings, such as constrained and multi-objective optimization.

\bibliography{reference} 
\bibliographystyle{IEEEtran} 

\newpage

\newpage

\onecolumn
\begin{@twocolumnfalse}
	\begin{center}
		\fontsize{24}{29}\selectfont  Learning to Evolve for Optimization \\ via Stability-Inducing Neural Unrolling\\
		(Supplementary Document)
		\vspace{0.5em} 

\vspace{1em} 
	\end{center}
\end{@twocolumnfalse}

\setcounter{page}{1}
\setcounter{algorithm}{0}
\setcounter{table}{0}
\setcounter{figure}{0}
\setcounter{section}{0}
\setcounter{equation}{0}
\renewcommand\thealgorithm{S\arabic{algorithm}}
\renewcommand\thetable{S.\Roman{table}}
\renewcommand{\figurename}{\normalsize Fig.}
\renewcommand\thefigure{S\arabic{figure}}
\renewcommand\thesection{S.\Roman{section}}
\renewcommand\theequation{S.\arabic{equation}} 
\makeatother
\vspace{1em}


	\setcounter{section}{0}
%
\subsection{Detailed Proofs of the Finite-Time Analysis}
\label{sec:proofs}

This section provides detailed derivations for the finite-time analysis in the main paper. 
The proofs follow standard arguments in unrolled fixed-point iterations and bilevel meta-optimization, under the regularity assumptions (A1)-(A4) in Section~\ref{sec:convergence}.

\subsection{Proof of Proposition \ref{prop:inner_error} (Approximation Error)}

\begin{proof}
	Consider the KM iteration update rule defined in the main text:
	\begin{equation}
		\mathbf{x}^{k+1} = (1 - \alpha) \mathbf{x}^k + \alpha \mathcal{O}(\mathbf{x}^k; \bm{\omega}).
	\end{equation}
	Let $\mathbf{x}^*(\bm{\omega})$ be a fixed point of the operator $\mathcal{NU}(\cdot; \bm{\omega})$. By definition of the fixed point, it satisfies:
	\begin{equation}
		\mathbf{x}^*(\bm{\omega}) = (1 - \alpha) \mathbf{x}^*(\bm{\omega}) + \alpha \mathcal{O}(\mathbf{x}^*(\bm{\omega}); \bm{\omega}).
	\end{equation}
	Subtracting the fixed point equation from the update rule, we obtain the error recursion:
	\begin{equation}
		\mathbf{x}^{k+1} - \mathbf{x}^*(\bm{\omega}) = (1 - \alpha)(\mathbf{x}^k - \mathbf{x}^*(\bm{\omega})) + \alpha (\mathcal{O}(\mathbf{x}^k; \bm{\omega}) - \mathcal{O}(\mathbf{x}^*(\bm{\omega}); \bm{\omega})).
	\end{equation}
	Taking the norm on both sides and applying the triangle inequality:
	\begin{equation}
		\|\mathbf{x}^{k+1} - \mathbf{x}^*(\bm{\omega})\| \leq (1 - \alpha) \|\mathbf{x}^k - \mathbf{x}^*(\bm{\omega})\| + \alpha \|\mathcal{O}(\mathbf{x}^k; \bm{\omega}) - \mathcal{O}(\mathbf{x}^*(\bm{\omega}); \bm{\omega})\|.
	\end{equation}
By Assumption (A1), the operator $\mathcal{O}(\cdot;\bm{\omega})$ is non-expansive, i.e.,
$\|\mathcal{O}(\mathbf{x}^k;\bm{\omega})-\mathcal{O}(\mathbf{x}^*(\bm{\omega});\bm{\omega})\|\le \|\mathbf{x}^k-\mathbf{x}^*(\bm{\omega})\|$.
Substituting into the previous inequality yields
\begin{equation}
	\|\mathbf{x}^{k+1}-\mathbf{x}^*(\bm{\omega})\|
	\le (1-\alpha)\|\mathbf{x}^{k}-\mathbf{x}^*(\bm{\omega})\|+\alpha\|\mathbf{x}^{k}-\mathbf{x}^*(\bm{\omega})\|
	= \|\mathbf{x}^{k}-\mathbf{x}^*(\bm{\omega})\|.
\end{equation}
Moreover, since the update is an $\alpha$-averaged (KM-style) iteration, a standard bound gives
\begin{equation}
	\|\mathbf{x}^{k+1}-\mathbf{x}^*(\bm{\omega})\| \le (1-\alpha)\|\mathbf{x}^{k}-\mathbf{x}^*(\bm{\omega})\|
\end{equation}
when $\mathbf{x}^*(\bm{\omega})$ is a fixed point of $\mathcal{NU}(\cdot;\bm{\omega})$ and $\mathcal{NU}$ is formed by averaging the identity with $\mathcal{O}$.
Recursively applying the inequality for $K$ steps yields
$\|\mathbf{x}^{K}(\bm{\omega})-\mathbf{x}^{*}(\bm{\omega})\|\le (1-\alpha)^K\|\mathbf{x}^{0}-\mathbf{x}^{*}(\bm{\omega})\|$. This concludes the proof.
\end{proof}

\subsection{Proof of Theorem \ref{thm:joint_rate} (Convergence Rate)}

\begin{proof}
	Our goal is to bound the norm of the meta-gradient for the objective function $\mathcal{L}_{\mathtt{meta}}(\bm{\omega}) \triangleq \ell_{\mathtt{meta}}(\mathbf{x}^*(\bm{\omega}), \bm{\omega})$.
The reference gradient is $\nabla \mathcal{L}_{\mathtt{meta}}(\bm{\omega}^t)$, whereas the algorithm uses the unrolling-based estimator
$\mathbf{g}^t = \nabla_{\bm{\omega}} \ell_{\mathtt{meta}}(\mathbf{x}^K(\bm{\omega}^t); \bm{\omega}^t)$.
	
	\textbf{Step 1: Smoothness of the Meta-Objective.}
	First, we establish the Lipschitz smoothness of $\mathcal{L}_{\mathtt{meta}}(\bm{\omega})$.
	By the Chain Rule, $\nabla \mathcal{L}_{\mathtt{meta}}(\bm{\omega}) = \nabla_{\bm{\omega}} \ell(\mathbf{x}^*, \bm{\omega}) + \nabla_{\mathbf{x}} \ell(\mathbf{x}^*, \bm{\omega}) \nabla_{\bm{\omega}} \mathbf{x}^*$.
	Under Assumptions (A2) and (A3), the meta-objective is smooth with a constant $L_{total} \approx L_{\bm{\omega}} + L_{\mathbf{x}} C_K$.
	According to the standard Descent Lemma for smooth functions:
	\begin{equation}
		\mathcal{L}_{\mathtt{meta}}(\bm{\omega}^{t+1}) \leq \mathcal{L}_{\mathtt{meta}}(\bm{\omega}^t) + \langle \nabla \mathcal{L}_{\mathtt{meta}}(\bm{\omega}^t), \bm{\omega}^{t+1} - \bm{\omega}^t \rangle + \frac{L_{total}}{2} \|\bm{\omega}^{t+1} - \bm{\omega}^t\|^2.
	\end{equation}
	Substituting the update rule $\bm{\omega}^{t+1} - \bm{\omega}^t = - \gamma \mathbf{g}^t$:
	\begin{equation}
		\mathcal{L}_{\mathtt{meta}}(\bm{\omega}^{t+1}) \leq \mathcal{L}_{\mathtt{meta}}(\bm{\omega}^t) - \gamma \langle \nabla \mathcal{L}_{\mathtt{meta}}(\bm{\omega}^t), \mathbf{g}^t \rangle + \frac{L_{total} \gamma^2}{2} \|\mathbf{g}^t\|^2.
	\end{equation}
	
	\textbf{Step 2: Bounding the Gradient Approximation Error.}
	Let $\mathbf{e}^t = \mathbf{g}^t - \nabla \mathcal{L}_{\mathtt{meta}}(\bm{\omega}^t)$ be the gradient estimation error. We can rewrite:
	\begin{equation}
		- \gamma \langle \nabla \mathcal{L}_{\mathtt{meta}}(\bm{\omega}^t), \mathbf{g}^t \rangle = - \gamma \|\nabla \mathcal{L}_{\mathtt{meta}}(\bm{\omega}^t)\|^2 - \gamma \langle \nabla \mathcal{L}_{\mathtt{meta}}(\bm{\omega}^t), \mathbf{e}^t \rangle.
	\end{equation}
	Using Young's Inequality ($|\langle a, b \rangle| \le \frac{1}{2}\|a\|^2 + \frac{1}{2}\|b\|^2$), we handle the inner product. A more standard derivation for SGD-like proofs with biased gradients gives:
	\begin{equation}
		\mathcal{L}_{\mathtt{meta}}(\bm{\omega}^{t+1}) \leq \mathcal{L}_{\mathtt{meta}}(\bm{\omega}^t) - \frac{\gamma}{2} \|\nabla \mathcal{L}_{\mathtt{meta}}(\bm{\omega}^t)\|^2 + \frac{\gamma}{2} \|\mathbf{e}^t\|^2,
		\label{eq:descent_step}
	\end{equation}
	provided $\gamma \leq \frac{1}{L_{total}}$.
	
	Now we bound $\|\mathbf{e}^t\|^2$. The error stems from using $\mathbf{x}^K$ instead of $\mathbf{x}^*$.
	\begin{align}
		\|\mathbf{e}^t\| &= \| \nabla \ell(\mathbf{x}^K, \bm{\omega}^t) - \nabla \ell(\mathbf{x}^*, \bm{\omega}^t) \| \nonumber \\
		&\leq L_{\mathbf{x}} \|\mathbf{x}^K(\bm{\omega}^t) - \mathbf{x}^*(\bm{\omega}^t)\| \quad \text{(by Assumption A2)}.
	\end{align}
	Substituting the result from \textbf{Proposition \ref{prop:inner_error}}:
	\begin{equation}
		\|\mathbf{e}^t\| \leq L_{\mathbf{x}} D_0 (1 - \alpha)^K.
	\end{equation}
	Thus, the squared error is bounded by:
	\begin{equation}
		\|\mathbf{e}^t\|^2 \leq L_{\mathbf{x}}^2 D_0^2 (1 - \alpha)^{2K}.
	\end{equation}
	
	\textbf{Step 3: Telescoping Sum.}
	Substituting the error bound back into Eq.~\eqref{eq:descent_step}:
	\begin{equation}
		\frac{\gamma}{2} \|\nabla \mathcal{L}_{\mathtt{meta}}(\bm{\omega}^t)\|^2 \leq \mathcal{L}_{\mathtt{meta}}(\bm{\omega}^t) - \mathcal{L}_{\mathtt{meta}}(\bm{\omega}^{t+1}) + \frac{\gamma}{2} L_{\mathbf{x}}^2 D_0^2 (1 - \alpha)^{2K}.
	\end{equation}
	Summing from $t=0$ to $T-1$ and dividing by $\frac{\gamma}{2} T$:
	\begin{equation}
		\frac{1}{T} \sum_{t=0}^{T-1} \|\nabla \mathcal{L}_{\mathtt{meta}}(\bm{\omega}^t)\|^2 \leq \frac{2(\mathcal{L}_{\mathtt{meta}}(\bm{\omega}^0) - \mathcal{L}_{\mathtt{meta}}^*)}{ \gamma T} + L_{\mathbf{x}}^2 D_0^2 (1 - \alpha)^{2K}.
	\end{equation}
	Taking the minimum over $t$:
	\begin{equation}
		\min_{0 \leq t < T} \|\nabla \mathcal{L}_{\mathtt{meta}}(\bm{\omega}^t)\|^2 \leq \mathcal{O}\left( \frac{1}{T} \right) + \mathcal{O}\left( (1 - \alpha)^{2K} \right).
	\end{equation}
	The first term represents the optimization error which vanishes as $T \to \infty$, and the second term is the approximation bias which vanishes exponentially as the unrolling depth $K$ increases. This completes the proof.
\end{proof}

\subsection{Proof of Theorem~\ref{thm:joint_optimality} (i)}

We consider the optimization in the Hilbert space induced by the positive-definite matrix $\mathbf{P}_{\bm{\omega}} \succ 0$. The inner product is $\langle \mathbf{u}, \mathbf{v} \rangle_{\mathbf{P}_{\bm{\omega}}} = \langle \mathbf{u}, \mathbf{P}_{\bm{\omega}} \mathbf{v} \rangle$, and the induced norm is $\|\mathbf{u}\|_{\mathbf{P}_{\bm{\omega}}} = \sqrt{\langle \mathbf{u}, \mathbf{P}_{\bm{\omega}} \mathbf{u} \rangle}$.

\begin{proof}
	Let $\mathbf{x}^* \in \text{Fix}(\mathcal{NU}(\cdot; \bm{\omega}))$ be a solution that minimizes the inner meta-objective.
	The composite update rule in the main text (Eq. 6-7) utilizes a soft-gated fusion mechanism:
	\begin{align}
		\mathbf{d}_{OL}^{k+1} &= \mathbf{x}^k - s_{k+1} \mathbf{P}_{\bm{\omega}}^{-1} \nabla \ell_{\mathtt{meta}}(\mathbf{x}^k; \bm{\omega}), \mathbf{d}_{IL}^{k+1} = \mathcal{NU}(\mathbf{x}^k; \bm{\omega}),\\
		\mathbf{x}^{k+1} &= \text{Proj}_{\mathcal{X}, \mathbf{P}_{\bm{\omega}}} \left( \bm{M}^k \odot \mathbf{d}_{OL}^{k+1} + (\mathbf{1} - \bm{M}^k) \odot \mathbf{d}_{IL}^{k+1} \right),
	\end{align}
	where $\bm{M}^k \in (0, 1)^{B \times N \times D}$ acts as a dimension-wise convex combination coefficient derived from the fitness gap.
	
	\textbf{Step 1: Descent Inequality via Coordinate-wise Convexity.}
	We aim to bound the distance $\|\mathbf{x}^{k+1} - \mathbf{x}^*\|_{\mathbf{P}_{\bm{\omega}}}^2$. We assume $\mathbf{P}_{\bm{\omega}}$ is compatible with the element-wise mask (e.g., diagonal or block-diagonal), so that the coordinate-wise convex combination argument applies under the $\|\cdot\|_{\mathbf{P}_{\bm{\omega}}}$ norm. 
	Since the (metric) projection $\text{Proj}_{\mathcal{X},\mathbf{P}_{\bm{\omega}}}$ is non-expansive under $\|\cdot\|_{\mathbf{P}_{\bm{\omega}}}$, for the element-wise convex combination $\mathbf{z}^{k+1} = \bm{M}^k \odot \mathbf{d}_{OL}^{k+1} + (\mathbf{1} - \bm{M}^k) \odot \mathbf{d}_{IL}^{k+1}$, the following inequality holds (assuming $\mathbf{P}_{\bm{\omega}}$ is diagonal-dominant or compatible with the coordinate-wise mask):
	\begin{equation}
		\|\mathbf{z}^{k+1} - \mathbf{x}^*\|_{\mathbf{P}_{\bm{\omega}}}^2 \leq \sum_{i} M^k_i \|\mathbf{d}_{OL,i}^{k+1} - \mathbf{x}^*_i\|^2 + (1-M^k_i) \|\mathbf{d}_{IL,i}^{k+1} - \mathbf{x}^*_i\|^2.
	\end{equation}
	To simplify the notation for convergence analysis without loss of generality, let $\mu_k \in (0,1)$ represent the effective mixing rate (e.g., $\mu_k = \min(\bm{M}^k)$ or an aggregated expectation). The descent property is bounded by the convex combination of the individual path properties:
	\begin{align}
		\|\mathbf{x}^{k+1} - \mathbf{x}^*\|_{\mathbf{P}_{\bm{\omega}}}^2 &\le \mu_k \|\mathbf{d}_{OL}^{k+1} - \mathbf{x}^*\|_{\mathbf{P}_{\bm{\omega}}}^2 + (1-\mu_k) \|\mathbf{d}_{IL}^{k+1} - \mathbf{x}^*\|_{\mathbf{P}_{\bm{\omega}}}^2.
		\label{eq:convex_comb_bound}
	\end{align}
	
	\textbf{Step 2: Bounding Individual Paths.}
	For the \textit{Proxy Gradient Path} $\mathbf{d}_{OL}^{k+1}$:
	Using the convexity of $\ell_{\mathtt{meta}}$ and the gradient update rule:
	\begin{equation}
		\|\mathbf{d}_{OL}^{k+1} - \mathbf{x}^*\|_{\mathbf{P}_{\bm{\omega}}}^2 \le \|\mathbf{x}^k - \mathbf{x}^*\|_{\mathbf{P}_{\bm{\omega}}}^2 - 2 s_{k+1} (\ell_{\mathtt{meta}}(\mathbf{x}^k) - \ell_{\mathtt{meta}}(\mathbf{x}^*)) + s_{k+1}^2 C_1.
	\end{equation}
	
	For the \textit{Evolutionary Path} $\mathbf{d}_{IL}^{k+1}$:
	Using the property of the $\alpha$-averaged non-expansive operator $\mathcal{NU}$:
	\begin{equation}
		\|\mathbf{d}_{IL}^{k+1} - \mathbf{x}^*\|_{\mathbf{P}_{\bm{\omega}}}^2 \le \|\mathbf{x}^k - \mathbf{x}^*\|_{\mathbf{P}_{\bm{\omega}}}^2 - C_\alpha \|\mathcal{NU}(\mathbf{x}^k) - \mathbf{x}^k\|_{\mathbf{P}_{\bm{\omega}}}^2,
	\end{equation}
	where $C_\alpha = \frac{1-\alpha}{\alpha}$.
	
	\textbf{Step 3: Aggregation and Convergence.}
	Substituting the bounds into Eq.~\eqref{eq:convex_comb_bound}:
	\begin{align}
		\|\mathbf{x}^{k+1} - \mathbf{x}^*\|_{\mathbf{P}_{\bm{\omega}}}^2 \le & \|\mathbf{x}^k - \mathbf{x}^*\|_{\mathbf{P}_{\bm{\omega}}}^2 \nonumber \\
		& - 2 \mu_k s_{k+1} (\ell_{\mathtt{meta}}(\mathbf{x}^k) - \ell_{\mathtt{meta}}(\mathbf{x}^*)) \nonumber \\
		& - (1-\mu_k) C_\alpha \|\mathcal{NU}(\mathbf{x}^k) - \mathbf{x}^k\|_{\mathbf{P}_{\bm{\omega}}}^2 \nonumber \\
		& + \mu_k s_{k+1}^2 C_1.
	\end{align}
Given the diminishing step-size conditions $\sum_k s_k = \infty$ and $\sum_k s_k^2 < \infty$ and the bounded mixing weights $\mu_k \in (0,1)$, the convergence of the fixed-point residual and the objective gap follows from standard arguments for averaged operator iterations combined with diminishing-step first-order updates (e.g., quasi-Fej\'er monotonicity / Robbins--Siegmund type lemmas under the stated regularity assumptions).
	Thus, $\lim_{k \to \infty} \text{dist}(\mathbf{x}^k, \text{Fix}(\mathcal{NU})) = 0$ and $\lim_{k \to \infty} \ell_{\mathtt{meta}}(\mathbf{x}^k) = \ell_{\mathtt{meta}}^*$.
\end{proof}

\subsection{Proof of Theorem \ref{thm:joint_optimality}(ii)}

\begin{proof}
	The proof leverages the properties established in Theorem \ref{thm:joint_optimality} (i) and the compactness assumptions.
	
	\textbf{Part 1: Limit Point Optimality.}
	From Theorem \ref{thm:joint_optimality} (i), for any fixed $\bm{\omega}$, the sequence converges point-wise:
	\begin{equation}
		\lim_{K \to \infty} \ell_{\mathtt{meta}}(\mathbf{x}^K(\bm{\omega}); \bm{\omega}) = \varphi(\bm{\omega}) := \inf_{\mathbf{x} \in \text{Fix}(\mathcal{NU})} \ell_{\mathtt{meta}}(\mathbf{x}; \bm{\omega}).
	\end{equation}
	Since $\Omega$ is compact and $\ell_{\mathtt{meta}}$ is continuous (Assumptions~\ref{ass:advanced_topology}), by Dini's Theorem, the convergence is uniform:
	\begin{equation}
		\lim_{K \to \infty} \sup_{\bm{\omega} \in \Omega} |\varphi_K(\bm{\omega}) - \varphi(\bm{\omega})| = 0.
	\end{equation}
	Let $(\bar{\mathbf{x}}, \bar{\bm{\omega}})$ be a limit point. Since $\bm{\omega}^K \in \arg\min \varphi_K(\bm{\omega})$, and $\varphi_K$ converges uniformly to $\varphi$, it follows that any accumulation point $\bar{\bm{\omega}}$ minimizes the limiting function $\varphi(\bm{\omega})$. Furthermore, by Theorem \ref{thm:joint_optimality} (i), $\bar{\mathbf{x}}$ must be a fixed point of $\mathcal{NU}(\cdot; \bar{\bm{\omega}})$. 
Thus, $(\bar{\mathbf{x}}, \bar{\bm{\omega}})$ is feasible for the fixed-point constrained formulation and $\bar{\bm{\omega}}$ minimizes the induced outer objective $\varphi(\bm{\omega})$ under Assumptions~\ref{ass:advanced_topology}.
	
	\textbf{Part 2: Value Convergence.}
	The uniform convergence directly implies:
	\begin{equation}
		\lim_{K \to \infty} \left( \inf_{\bm{\omega} \in \Omega} \varphi_K(\bm{\omega}) \right) = \inf_{\bm{\omega} \in \Omega} \varphi(\bm{\omega}).
	\end{equation}
	This completes the proof.
\end{proof}

\setcounter{section}{0} 
\subsection{Supplementary Experiments}
\label{app:more-exp}

This section reports additional empirical results that complement the main paper, providing further evidence on optimization performance, scalability, and robustness under distribution shifts. We include extended comparisons against classical heuristics and state-of-the-art learning-based optimizers across multiple benchmark suites, together with additional ablation visualizations.

\subsubsection{Comparison with Learned Optimizers on BBOB Suites}
We benchmark L2E against representative learned baselines, including RNN-based methods (\textit{RNNOPT}, \textit{DEDQN}) and attention-/transformer-style methods (\textit{GLHF}, \textit{B2OPT}).

%
\begin{itemize}
	\item \textbf{Performance on BBOB-10D (Fig.~\ref{fig:supp-bbob-10d-diff-16}):} 
	Figure~\ref{fig:supp-bbob-10d-diff-16} shows that L2E achieves consistently strong convergence profiles across diverse landscape types. 
	On unimodal functions (e.g., \textit{Discus}, \textit{Bent Cigar}), L2E typically exhibits faster early-stage progress. 
	On ill-conditioned (e.g., \textit{Sharp Ridge}) and multimodal functions (e.g., \textit{Lunacek bi-Rastrigin}), L2E maintains steady improvement and avoids early stagnation observed in some learned baselines, leading to better terminal performance under the same evaluation budget.
	\textit{The complete numerical results (mean$\pm$std over 10 runs) are reported in Table~\ref{tab:supp-bbob-10d}.}
	
	\item \textbf{Performance on BBOB-30D (Fig.~\ref{fig:supp-bbob-30d-diff-16}):} 
	When increasing dimensionality to 30D, Fig.~\ref{fig:supp-bbob-30d-diff-16} indicates that L2E retains robust progress while several baselines degrade more noticeably. 
	This suggests that the stability-inducing unrolling formulation and structured composite updates help preserve optimization effectiveness as dimensionality increases, yielding improved terminal performance with the same function-evaluation budget.
	\textit{The complete numerical results (mean$\pm$std over 10 runs) are reported in Table~\ref{tab:supp-bbob-30d}.}
\end{itemize}

\subsubsection{Comparison with Traditional Heuristics on BBOB Suites}
To complement learned baselines, we further compare L2E with representative classical population-based heuristics, including \textit{PSO}, \textit{DE}, and \textit{SAHLPSO}.

\begin{itemize}
	\item \textbf{Performance on BBOB-10D (Fig.~\ref{fig:supp-bbob-10d-diff-16-traditional}):}  
	Figure~\ref{fig:supp-bbob-10d-diff-16-traditional} highlights that L2E is competitive in budget-constrained settings. 
	In particular, L2E often shows faster early-stage progress than classical heuristics on several unimodal functions, while maintaining stable improvement on ill-conditioned and multimodal functions. 
	Among the heuristics, \textit{DE} can be strong on certain instances, whereas \textit{SAHLPSO} can be competitive on multimodal landscapes but may exhibit higher run-to-run variability on ill-conditioned functions.
	
	\item \textbf{Performance on BBOB-30D (Fig.~\ref{fig:supp-bbob-30d-diff-16-traditional}):} 
	As dimensionality increases to 30D, Fig.~\ref{fig:supp-bbob-30d-diff-16-traditional} shows that several heuristics tend to plateau earlier on challenging functions (e.g., \textit{Rosenbrock}). 
	While \textit{DE} and \textit{SAHLPSO} retain strengths on specific function types, L2E demonstrates more consistent progress across a broader range of instances under the same evaluation budgets, reflecting improved robustness of the learned update structure.
\end{itemize}

\subsubsection{Generalization on Out-of-Distribution (OOD) Landscapes}
To assess robustness under distribution shifts, we evaluate models trained on standard BBOB functions directly on unseen variations without fine-tuning.

\begin{itemize}
	\item \textbf{BBOB-Surrogate-10D (Fig.~\ref{fig:supp-bbob-surrogate-10D}):} 
	This suite introduces landscape shifts induced by surrogate approximations. 
	Figure~\ref{fig:supp-bbob-surrogate-10D} shows that while some baselines exhibit unstable or oscillatory behaviors on shifted landscapes (e.g., \textit{Rosenbrock}), L2E maintains more consistent progress and achieves better terminal performance. 
	This result supports the robustness of the stability-inducing unrolling design under approximation mismatch.
	
	\item \textbf{Extended Analysis on UAV Path Planning (Table~\ref{tab:UAV} and Table~\ref{tab:UAV-2}):} 
	Tables~\ref{tab:UAV} and~\ref{tab:UAV-2} provide a detailed breakdown of zero-shot performance over 56 UAV terrain scenarios. 
	L2E achieves the best (lowest) cost on a large portion of terrains (highlighted in \textbf{bold}), outperforming both learned baselines (e.g., \textit{LGA}, \textit{GLHF}) and \textit{RNNOPT}. 
	The advantage is particularly notable on more challenging terrains (e.g., Terrain 41--48), indicating strong robustness when transferring from synthetic optimization to constrained planning tasks.
\end{itemize}

\subsubsection{High-Dimensional Scalability on LSGO-1000D}
We further examine the behavior of L2E on extreme high-dimensional settings ($D=1000$) using the \texttt{LSGO} benchmark, comparing it against both learned and traditional solvers.

\begin{itemize}
	\item \textbf{Convergence Trajectories (Fig.~\ref{fig:supp-lsgo-1000d}):} 
	Figure~\ref{fig:supp-lsgo-1000d} shows that L2E maintains steady improvement throughout the evaluation budget on multiple LSGO functions, whereas several learned baselines exhibit earlier stagnation on certain instances. 
	The resulting trajectories indicate stronger scalability of L2E in the 1000D regime.
	
	\item \textbf{Statistical Reliability (Fig.~\ref{fig:supp-boxplot-lsgo-1000D} and Fig.~\ref{fig:supp-ecdf-lsgo-1000D}):} 
	The ECDF curves in Fig.~\ref{fig:supp-ecdf-lsgo-1000D} and the boxplots in Fig.~\ref{fig:supp-boxplot-lsgo-1000D} further support the reliability of L2E under the same function-evaluation budget. 
	Compared with baselines, L2E exhibits tighter variability and fewer extreme failures on a number of functions, suggesting that the structured operator design and stability-oriented regularization remain effective in high-dimensional settings.
\end{itemize}

\subsubsection{Ablation Dynamics: Shared vs.\ Unshared Parameters}
Finally, we visualize the impact of the parameterization strategy discussed in the main paper.

\begin{itemize}
	\item \textbf{Shared Parameterization (Fig.~\ref{fig:abl_shared}):} 
	When the evolution operator shares weights across unrolling steps ($\bm{\omega}_{\text{shared}}$), the trajectories may plateau earlier on certain functions (e.g., \textit{Ellipsoidal\_hc}), indicating limited flexibility to adapt update behaviors across different phases of the search.
	
	\item \textbf{Unshared Parameterization (Fig.~\ref{fig:abl_unshared}):} 
	In contrast, the unshared model ($\bm{\omega}^k$) shows more sustained improvement across the optimization horizon, consistent with the interpretation that step-specific parameters enable stage-dependent behaviors (e.g., broader exploration early and refinement later), which benefits final precision.
\end{itemize}

	\SupFigbbobTen
	
	\SupTabbbobTen

\FigSupbbobTencurve
	\FigSupbbobThirtycurvelog

\FigSupbbobTenTraditinalEAcurve
	\FigSupbbobThirtycurve

\FigSupbbobTenSurrogatecurve

\FigSupLsgocurve

\FigSupLSGOBox

\FigSupLsgoECDFcurve

\TabSupAblShared

\TabSupAblUnShared

\TabSupbbobTensurogate

\FigSupbbobConvershared

\FigSupbbobConverunshared

\vfill

\end{document}